\documentclass{article} 
\usepackage{iclr2027_conference,times}

\usepackage{amsmath,amsfonts,bm}

\def\eqref#1{equation~\ref{#1}}

\def\1{\bm{1}}

\DeclareMathAlphabet{\mathsfit}{\encodingdefault}{\sfdefault}{m}{sl}
\SetMathAlphabet{\mathsfit}{bold}{\encodingdefault}{\sfdefault}{bx}{n}

\usepackage{hyperref}
\usepackage{url}

\usepackage{dsfont}
\usepackage{nicefrac}
\usepackage{xfrac}
\usepackage{pifont}

\usepackage{enumitem}
\usepackage{rotating}
\usepackage{tabularx}

\usepackage{booktabs}
\usepackage{multirow}
\usepackage{tcolorbox}
\usepackage{subcaption}

\usepackage{arydshln}
\usepackage{bbding}
\usepackage{svg}
\usepackage{amsthm}
\usepackage{amsmath}
\usepackage{amssymb}
\usepackage{xspace}

\newcommand{\lm}[1]{\texttt{#1}}

\newcommand{\data}[1]{\textsf{#1}}

\usepackage{wrapfig}
\usepackage{xcolor,colortbl}
\usepackage[dvipsnames]{xcolor}

\usepackage{soul}
\definecolor{lightblue}{RGB}{239,247,250}
\definecolor{blue}{RGB}{224,230,255}
\definecolor{lighpurple}{RGB}{226, 218, 246}
\definecolor{purple}{RGB}{193, 175, 236}
\definecolor{deepred}{RGB}{183,26,26}
\definecolor{deepgreen}{RGB}{4,98,10}
\definecolor{lightred}{RGB}{242,207, 194}
\definecolor{red}{RGB}{232,116,124}
\definecolor{grey}{RGB}{192,192,192}
\definecolor{lightyellow}{RGB}{255,176,0}

\usepackage[T1]{fontenc}
\usepackage[utf8]{inputenc}
\usepackage{fontawesome5}
\usepackage{pifont}
\definecolor{gold}{RGB}{255,215,0}

\newcommand{\okmark}{\textcolor{deepgreen}{\ding{51}}}   
\newcommand{\xmark}{\textcolor{red}{\ding{55}}} 
\newcommand{\goldicon}{\textcolor{gold}{\faStar}}

\usepackage{verbatim}

\newcommand{\boxc}[2]{{\setlength{\fboxsep}{1pt}\setlength{\fboxrule}{1pt}\fcolorbox{#1}{white}{#2}}}

\title{From Concept Alignment to Causal Grounding: An Intervention Test of Chain-of-Thought Faithfulness}

\newcommand{\affilsup}[1]{\rlap{\textsuperscript{\normalfont#1}}}

\author{Qianli Wang\affilsup{1,5}
    \qquad
    Yilong Wang\affilsup{1,\footnotemark[2]}
    \qquad
    Dennis Wei\affilsup{2,\footnotemark[2]}
    \\
    \textbf{Jingyi Sun\affilsup{3}}
    \qquad
    \textbf{Simon Ostermann\affilsup{5,6,7}}
    \qquad
    \textbf{Isabelle Augenstein\affilsup{3}}
    \\
    \textbf{Pepa Atanasova\affilsup{3,\footnotemark[3]}}
    \qquad
    \textbf{Nils Feldhus\affilsup{4,5,\footnotemark[3]}}
    \\
    \AND
    $^1${\normalfont Technische Universit\"at Berlin}
    \qquad
    $^2${\normalfont IBM Research}
    \qquad
    $^3${\normalfont University of Copenhagen}
    \\
    $^4${\normalfont University of Groningen}
    \quad
    $^5$German Research Center for Artificial Intelligence (DFKI)
    \\
    $^6$Saarland Informatics Campus
    \quad
    $^7$Centre for European Research in Trusted AI (CERTAIN)\\
    \\
    \textbf{Correspondence}: 
  {\normalfont\href{mailto:qianli.wang@tu-berlin.de}{\mbox{\texttt{qianli.wang@tu-berlin.de}}}} \\
  \footnotemark[2] \ Equal contribution.
  \footnotemark[3] \ These authors contributed equally to this work as joint last author.
}

\iclrfinalcopy 

\begin{document}

\maketitle

\lhead{}

\begin{abstract}
Chain-of-thought (CoT) can sound plausible yet be unfaithful to the model's underlying reasoning. Most prior work probes CoT faithfulness through input--output behavior or input attributions, leaving internal computation largely underexplored. We instead cast faithfulness as an internal concept grounding: \textit{Does a large language model's (LLM) CoT reasoning engage the same internal concepts that support the LLM's direct prediction, and do the shared concepts causally drive its answer?} Concretely, we encode the prediction and CoT passes with a single shared sparse autoencoder (SAE), treating its latent features as concepts and thereby mapping both passes into a shared concept space. 
Within this space, we design three correlational metrics of concept-level alignment and a causal metric, $\Delta p$, which measures the drop in answer probability when the shared concepts are ablated.
Across five LLMs and four datasets, concept alignment is generally high, as indicated by the correlational metrics; yet these only identify which concepts are shared, not how much they causally contribute. $\Delta p$ fills this gap: causal faithfulness varies substantially with model depth, peaking at mid-to-late layers, and model scale reshapes the layer-wise profile. Moreover, causally important shared concepts are not always verbalized in the CoT. These dissociations suggest that faithfulness cannot be reliably assessed from surface-level or representational correspondence alone; assessing it requires causal tests of whether the internal concepts underlying a CoT actually drive the model's prediction.
\end{abstract}

\section{Introduction}
\label{sec:intro}

Chain-of-thought (CoT) is a promising technique that seemingly exposes an LLM's decision-making on various critical tasks~\citep{wei2022chain, kojima2022large}, such as safety monitoring \citep{korbak2025chainthoughtmonitorabilitynew}. 
This apparent transparency, however, hinges on CoT faithfulness, i.e., 
the degree to which it accurately represents or contributes to the underlying reasoning of the model, rather than a plausible post-hoc rationalization \citep{jacovi-goldberg-2020-towards,agarwal2024faithfulnessvsplausibilityunreliability}. The distinction matters most in high-stakes settings, where unfaithful CoTs can lull users into over-relying on systems whose stated reasoning masks their true decision basis~\citep{paul-etal-2024-making}. 

Prior work on measuring CoT faithfulness falls along complementary axes. 
One line of work intervenes on the reasoning chain, or the parametric knowledge that the reasoning invokes, and treats the CoT as faithful only if the answer changes accordingly \citep{lanham2023measuring,zaman-srivastava-2025-causal,tutek-etal-2025-measuring}; another injects answer hints and treats the CoT as unfaithful if a hint is used silently without acknowledgment \citep{turpin2023language, chen2025reasoning, chua2025deepseekr1reasoningmodels}. Both infer internal alignment only indirectly, from input--output behavior. Attribution-based methods \citep{parcalabescu-frank-2024-measuring,admoni-etal-2026-aligning} assess the alignment between input attributions for prediction and those for CoT, but they capture only dependence on the input and miss what the model computes internally. These evaluations largely treat the model as a black box, \textit{without testing which internal computations causally drive the answer.} 

A more direct line pries open this black box: it uses mechanistic interpretability (MI) to inspect internal computation as signals for CoT faithfulness \citep{occhipinti2026probing, shen2026detectingunfaithfulchainofthoughtcircuitguided}. 
Closest to our work, \citet{chen2026does} train two SAEs and patch CoT features into a no-CoT run, 
reading the gain in answer probability to test whether reasoning concepts can steer a direct answer. However, these approaches do not examine (1) whether and to what extent an LLM's CoT engages the same internal concepts that also support the LLM's direct prediction, (2) whether these shared concepts causally drive the prediction, nor (3) how these relations are distributed across model depth. \looseness=-1

\begin{figure*}[!t]
\centering
\resizebox{\textwidth}{!}{
\begin{minipage}{\textwidth}
\includegraphics[width=\textwidth]{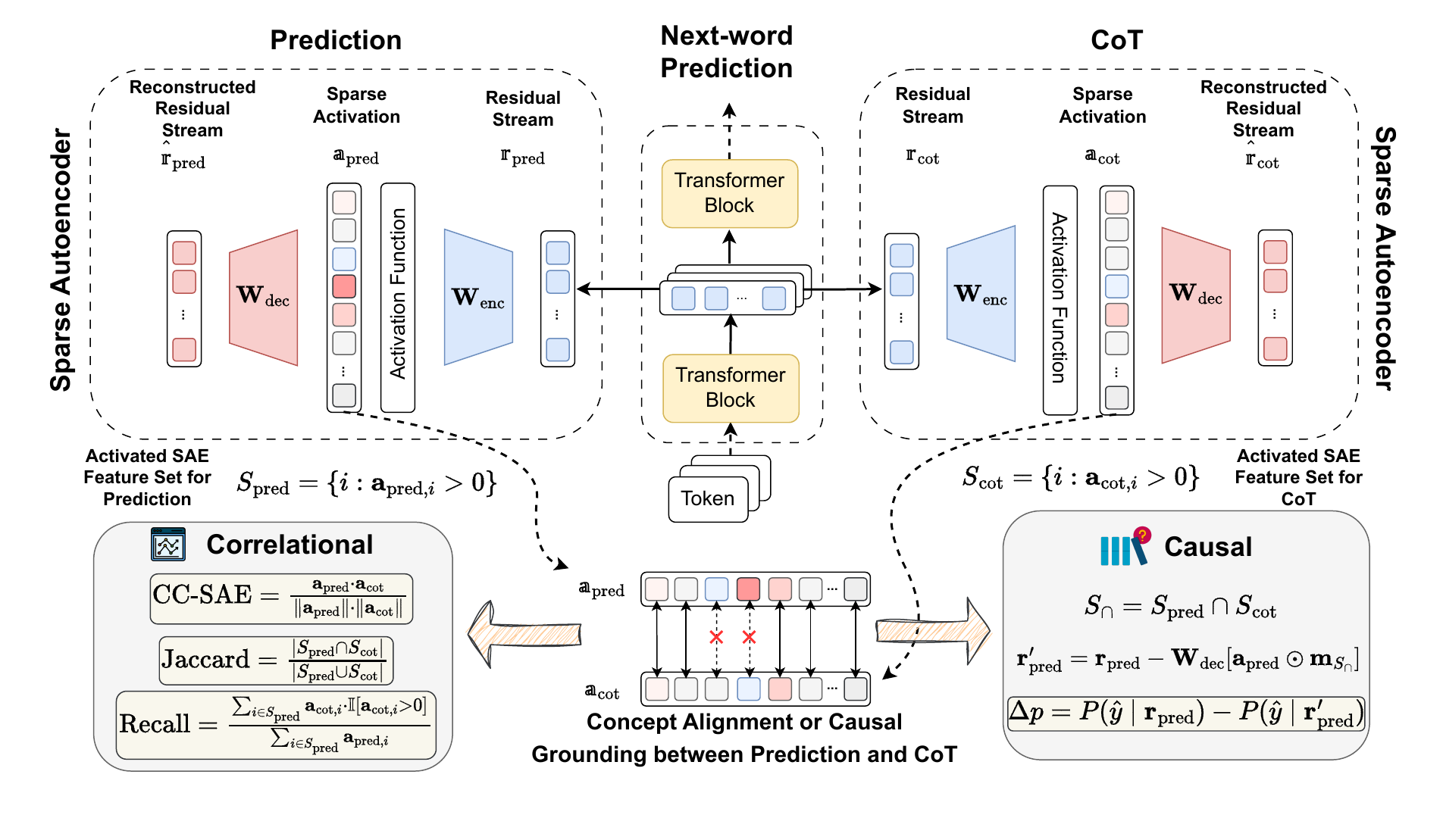}
\end{minipage}
}
\vspace{-1em}
\caption{SAE-based CoT faithfulness evaluation pipeline. We apply a shared SAE to the residual stream, extracting sparse activations from prediction and CoT-derived prediction passes. The two activations are then compared using two notions of faithfulness indicators: correlational metrics (CC-SAE, Jaccard similarity, and Recall), which quantify the overlap and alignment between the prediction and CoT activations; and a causal metric $\Delta p$, which ablates the shared features from the reconstructed prediction stream and measures the resulting change in answer probability.\looseness=-1}
\label{fig:pipeline}
\vspace{-0.5em}
\end{figure*}

We fill this gap by casting CoT faithfulness as internal concept grounding: \textit{does an LLM's CoT engage the same internal concepts that also support the LLM's direct prediction, and do the shared concepts causally drive its answer?} Concretely, we encode the answer-position residuals of a prediction pass and a CoT-derived prediction pass with a shared SAE (\S\ref{sec:sae}). We treat active SAE latent features as concepts, making their overlap across passes well-defined (Figure~\ref{fig:pipeline}). On this shared concept space, we propose metrics that capture two notions of faithfulness: 
(i) \textbf{correlational metrics} (CC-SAE, Jaccard similarity, and prediction-grounded recall) quantifying alignment in concept usage, as a form of \textit{self-consistency}\footnote{Following \citet{parcalabescu-frank-2024-measuring}, \textit{self-consistency} is defined as the alignment between a model's contribution to answer prediction and its generated CoT -- a necessary but insufficient condition for \textit{faithfulness}.}; and (ii) a \textbf{causal metric} $\Delta p$ that ablates the shared concepts $S_\cap$ from the prediction or CoT pass and measures the drop in answer probability, as a measure of \textit{faithfulness}. 
\looseness=-1

Applying this framework to five LLMs and four benchmarks yields six main findings. 
\textbf{First}, correlational metrics suggest decent concept alignment overall, which provides a solid basis for using $\Delta p$. Yet such metrics only identify which features are shared across passes. 
Precisely quantifying their contribution requires causal metrics like $\Delta p$. \textbf{Second}, we show that $\Delta p$ captures genuine causal grounding rather than artifacts of ablation count, feature selection, or the SAE itself, through random and norm-matched controls and SAE-configuration ablations, respectively. 
\textbf{Third}, our notion of faithfulness is not a single number: it varies substantially across depth, generally peaking at mid-to-late rather than final layers, with model-specific dips, underscoring the need for layer-aware reporting. \textbf{Fourth}, architectures differ qualitatively rather than merely in magnitude: \lm{Qwen3} models exhibit \textit{distributed} faithfulness that remains high across most layers, whereas \lm{Llama-3.1-8B} shows \textit{localized} faithfulness concentrated in a few layers. \textbf{Fifth}, within a model family, larger models are more faithful overall, and scaling reshapes the layer-wise profile, concentrating low-faithfulness computation into fewer, earlier layers while stabilizing alignment across the remaining depth.
\textbf{Lastly}, a case study of layer-wise concept transition described in natural language reveals that causally important shared concepts are not always verbalized in the CoT. \looseness=-1


\section{Related Work}
\textbf{CoT Faithfulness Evaluation.} 
One line of work intervenes on input or CoT and examines whether the intervention is reflected in the answer \citep{lanham2023measuring,zaman-srivastava-2025-causal,ye2026mechanistic}. The second line of work injects answer hints and verifies whether they are verbalized in CoT \citep{turpin2023language,arcuschin2025chainofthought, chua2025deepseekr1reasoningmodels}. Furthermore, \citet{tutek-etal-2025-measuring, sun2026investigatinginterplaycontextualparametric} unlearn knowledge used in CoT steps from model parameters and assess whether the answer changes after unlearning. \citet{paul-etal-2024-making} apply causal mediation and counterfactual reasoning-chain interventions to test whether intermediate CoT steps influence final answers. \citet{parcalabescu-frank-2024-measuring} measure the alignment between input contribution to answer prediction and CoT reasoning. Our work is close to theirs but measures alignment and grounding in the SAE concept space (\S\ref{sec:sae}), which addresses limitations of input-level alignment: SAE concepts match the granularity of internal computation and are near-monosemantic where tokens are not.

\textbf{Mechanistic Interpretability for CoT Faithfulness Evaluation.}
Recent advancements in MI enable the inspection of internal representations to assess faithfulness directly. 
\citet{occhipinti2026probing} train a linear probe to distinguish faithful and unfaithful CoTs.
\citet{shen2026detectingunfaithfulchainofthoughtcircuitguided} apply circuit tracing and a supervised detector to measure CoT unfaithfulness as the discrepancy between sentence-level circuits and the displayed CoT. \citet{chen2026does}, the closest to us, train \textit{two separate} SAEs and patch CoT features into a non-CoT run, reading the causal effect from the gain in answer probability, which is a magnitude-based test of whether injected reasoning features can steer a direct answer. Our intervention runs in the opposite direction (Figure~\ref{fig:compare}): with a \textit{single shared} SAE that makes cross-pass concept overlap well-defined, we ablate the shared concepts $S_\cap$ between answer prediction and CoT (\S\ref{subsec:causal}), yielding a test of whether the direct answer depends on, not merely is steerable by, the concepts that the CoT instates. \looseness=-1

\section{Methodology}
\label{sec:method}

\subsection{Problem Setup}
\label{sec:setup}

We examine faithfulness directly within the model's internal representations. Sparse autoencoders (SAEs) have been shown to effectively recover latent concepts (\S\ref{sec:sae}) that can influence LLM outputs when manipulated \citep{bricken2023monosemanticity,templeton2026scalingmonosemanticityextractinginterpretable}. We measure CoT faithfulness by checking whether \textit{the LLM's CoT-based prediction relies on the same internal representations used in a direct prediction}. Specifically, given an input $x$, we compare model behavior under two conditions: \ding{192} \textbf{Prediction Pass:} The model receives the input question $x$ and is asked to predict the answer $\hat{y}$. We extract the residual stream activation at the answer token position, denoted $\mathbf{r}_{\text{pred}} \in \mathbb{R}^{d_{\text{model}}}$, where $d_{\text{model}}$ denotes the model's hidden dimension (Table~\ref{tab:hidden_dim}). \ding{193} \textbf{CoT Pass:} Following prior work \citep{parcalabescu-frank-2024-measuring,tutek-etal-2025-measuring}, we adopt a two-step procedure: the model first generates a reasoning chain $c$ based on input $x$ \citep{kojima2022large}.  We then form the sequence ($x$; $c$), run a forward pass, and extract the residual stream activation $\mathbf{r}_{\text{cot}}  \in \mathbb{R}^{d_{\text{model}}}$ at the answer position (\S\ref{sec:extraction}), following \citet{dura2026mechanisticinterpretabilitychainofthoughtreasoning, chen2026does}.\looseness=-1

\subsection{Sparse Autoencoders}
\label{sec:sae}
To identify which internal representations are genuinely used by the model during prediction and during CoT, we apply sparse autoencoders (SAEs), which reconstruct residual-stream activations through a sparse, overcomplete bottleneck \citep{cunningham2023sparseautoencodershighlyinterpretable, zhang-etal-2026-locate} (Figure~\ref{fig:pipeline}). Given a residual stream vector $\mathbf{r} \in \mathbb{R}^{d_{\text{model}}}$, the SAE encoder produces a sparse activation vector $\mathbf{a}$: \looseness=-1
\begin{equation}
    \mathbf{a} = \sigma\left(\mathbf{W}_{\text{enc}} \mathbf{r} + \mathbf{b}_{\text{enc}}\right) \in \mathbb{R}^{d_{\text{sae}}}
\end{equation}
where $\sigma$ denotes the SAE activation function and $d_{\text{sae}}$ represents the width of the SAE (typically $d_{\text{sae}} \gg d_{\text{model}}$; an ablation study of the SAE setup is performed in \S\ref{subsec:ablation_study}). The decoder reconstructs:
\begin{equation}
    \hat{\mathbf{r}} = \mathbf{W}_{\text{dec}} \mathbf{a} + \mathbf{b}_{\text{dec}}
\end{equation}
Each dimension of $\mathbf{a}$ corresponds to a learned (SAE) concept; sparsity pushes these concepts toward monosemanticity rather than superposition \citep{bricken2023monosemanticity, shu-etal-2025-survey}.
 
\textbf{Why SAEs?} 
SAE concepts instantiate the three desiderata (\S\ref{sec:intro}): encoding both passes with a single shared SAE makes concept usage directly \textit{comparable}; sparsity promotes \textit{monosemanticity}, so overlap is meaningful rather than an artifact of entangled directions; and 
individual features can be selectively ablated, making concepts \textit{intervenable} for causal tests (\S\ref{subsec:causal}).

\subsection{Activation Extraction}
\label{sec:extraction}


In both passes, we encode the residual at the answer position, ensuring that the same functional position is compared across the prediction and CoT passes:
\noindent\begin{minipage}{.5\linewidth}
\begin{equation}
    \mathbf{a}_{\text{pred}} = \text{SAE}_{\text{enc}}(\mathbf{r}_{\text{pred}}^{(\text{ans})}) \in \mathbb{R}^{d_{\text{sae}}}
\end{equation}
\end{minipage}%
\begin{minipage}{.5\linewidth}
\begin{equation}
    \mathbf{a}_{\text{cot}} = \text{SAE}_{\text{enc}}(\mathbf{r}_{\text{cot}}^{(\text{ans})}) \in \mathbb{R}^{d_{\text{sae}}}
\end{equation}
\end{minipage}

Following \citet{chen2026does,dura2026mechanisticinterpretabilitychainofthoughtreasoning}, we extract the residual $\mathbf{r}_{\text{cot}}^{(\text{ans})}$ at the final answer position rather than aggregating across CoT tokens, which would be naturally noisy: by the point of answer commitment, relevant concepts surfaced along the reasoning path are already aggregated into this position. This mirrors the prediction pass, so $\mathbf{a}_{\text{pred}}$ and $\mathbf{a}_{\text{cot}}$ are extracted at the same functional (answer-commitment) position. The intuition is that if the CoT is faithfully grounded in the model's prediction process, it should engage answer-relevant concepts.\footnote{Following \citet{tutek-etal-2025-measuring}, we evaluate faithfulness in cases where CoT and no-CoT predictions of models agree. This avoids confounding the CoT-induced performance gain with the CoT faithfulness evaluation.\looseness=-1}

\subsection{Correlational Metrics}
\label{subsec:correlation}

Given SAE activation vectors $\mathbf{a}_{\text{pred}}$ and $\mathbf{a}_{\text{cot}}$, we first propose three new correlational metrics that quantify the overlap and alignment in concept usage between the two passes, as a form of concept-level \boxc{grey}{\textit{self-consistency}} \citep{parcalabescu-frank-2024-measuring, admoni-etal-2026-aligning}:

\textbf{CC-SAE.} 
CC-SAE measures the strength-weighted alignment of activations in concept space:
\begin{equation}
    \text{CC-SAE} = {\Large \sfrac{\mathbf{a}_{\text{pred}} \cdot \mathbf{a}_{\text{cot}}}{\|\mathbf{a}_{\text{pred}}\| \cdot \|\mathbf{a}_{\text{cot}}\|}}
\end{equation}

\textbf{Jaccard Similarity.} A set-based measure ignoring activation magnitudes, where $S_{\text{pred}} = \{i : \mathbf{a}_{\text{pred},i} > 0\}$ and $S_{\text{cot}} = \{i : \mathbf{a}_{\text{cot},i} > 0\}$ are the sets of active concept indices:
\begin{equation}
    \text{Jaccard} = {\Large \sfrac{|S_{\text{pred}} \cap S_{\text{cot}}|}{|S_{\text{pred}} \cup S_{\text{cot}}|}}
\end{equation}

\textbf{Prediction-Grounded Recall.} A directional measure of how much of the prediction's concept usage appears in the CoT:
\begin{equation}
    \text{Recall} = {\Large \sfrac{\sum_{i \in S_{\text{pred}}} \mathbf{a}_{\text{cot},i} \cdot \mathbb{I}[\mathbf{a}_{\text{cot},i} > 0]}{\sum_{i \in S_{\text{pred}}} \mathbf{a}_{\text{pred},i}}}
\end{equation}

\subsection{Causal Metric}
\label{subsec:causal}

The correlational metrics above measure whether the same concepts appear in both passes, but not whether those concepts causally drive the answer. To address this, we further introduce a new causal metric $\Delta p$. Let $S_{\cap} = S_{\text{pred}} \cap S_{\text{cot}}$ be the set of shared active concepts. We ablate these concepts from the prediction pass\footnote{Ablation results on the CoT pass are presented and discussed in \S\ref{subsubsec:cot_pass_ablation} and Appendix~\ref{app:cot_pass_ablation}.} by subtracting their contribution from the residual:
\begin{equation}
    \mathbf{r}'_{\text{pred}} = \mathbf{r}_{\text{pred}} - \mathbf{W}_{\text{dec}}[\mathbf{a}_{\text{pred}} \odot \mathbf{m}_{S_{\cap}}]
\end{equation}
where $\mathbf{m}_{S_{\cap}}$ is a binary mask that selects only the shared concepts. We then measure the change in answer probability, as a measure of \boxc{grey}{\textit{faithfulness}}:
\begin{equation}
    \Delta p = P(\hat{y} \mid \mathbf{r}_{\text{pred}}) - P(\hat{y} \mid \mathbf{r}'_{\text{pred}})
\end{equation}
We interpret larger positive values of $\Delta p$ as stronger evidence that the concepts shared between the CoT and prediction causally support the answer, and thus as indicating greater CoT faithfulness. Conversely, values near zero indicate that the answer probability is insensitive to their ablation, providing little evidence of causal support and potentially signaling concept-level post-hoc rationalization. Appendix~\ref{app:posthoc_rationalization} illustrates such low-$\Delta p$ cases.





\section{Experimental Setup}

\textbf{Models.}
We employ five open-source LLMs with varying sizes: \lm{Llama-3.1-8B} \citep{grattafiori2024llama3herdmodels}, \lm{Gemma-2-\{2B,9B\}} \citep{gemmateam2024gemma2improvingopen} and \lm{Qwen3-\{1.7B,8B\}} \citep{yang2025qwen3technicalreport}, with off-the-shelf Llama-Scope SAEs \citep{he2024llamascopeextractingmillions}, Gemma-Scope SAEs \citep{lieberum-etal-2024-gemma} and Qwen-Scope SAEs \citep{deng2026qwenscopeturningsparsefeatures}, respectively, pre-trained on the residual streams.\looseness=-1

\noindent\textbf{Datasets.} We adopt four diverse datasets: 
\data{GSM8K} \citep{cobbe2021trainingverifierssolvemath}, \data{LogiQA} \citep{liu-etal-2021-logiqa}, \data{OpenbookQA} \citep{mihaylov-etal-2018-suit}, and \data{ARC-Easy} \citep{clark2018thinksolvedquestionanswering}. 

\section{Results}

In this section, we analyze the correlational metrics and examine the relationship between correlational and causal metrics (\S\ref{subsec:corr_causal_comparison}), validate the causal metric $\Delta p$ under controlled conditions (\S\ref{subsec:sanity_check}), analyze causal layer-wise faithfulness (\S\ref{subsec:layerwise}), ablate the SAE configuration (\S\ref{subsec:ablation_study}), and characterize SAE concept transitions across layers through natural-language descriptions (\S\ref{subsec:layer_transition}).

\begin{figure*}[t]
    \centering
    \resizebox{\textwidth}{!}{%
    \begin{minipage}{\textwidth}
        \centering
        \begin{subfigure}[b]{0.32\textwidth}
            \includegraphics[width=\textwidth]{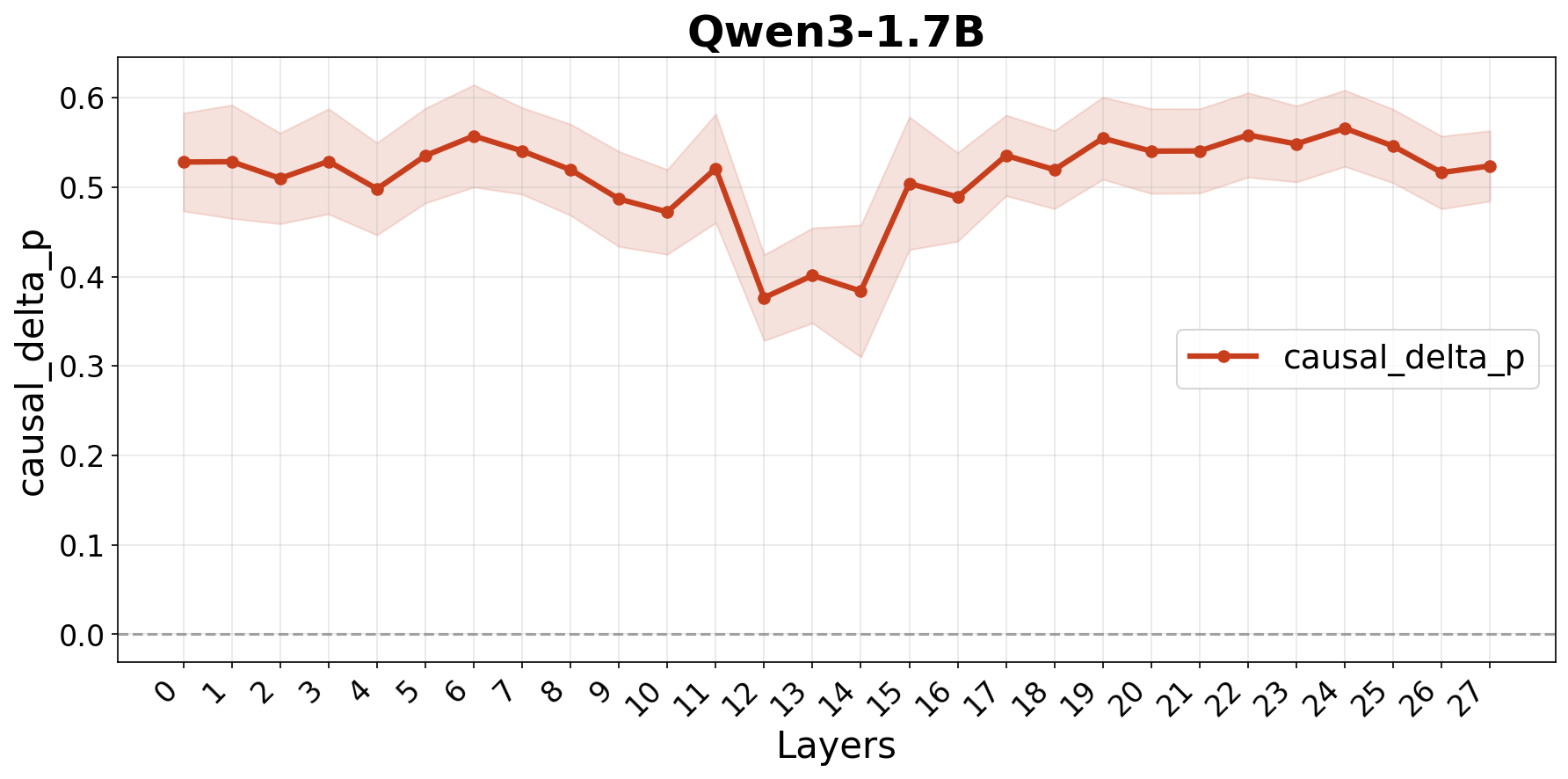}
        \end{subfigure}
        \hfill
        \begin{subfigure}[b]{0.32\textwidth}
            \includegraphics[width=\textwidth]{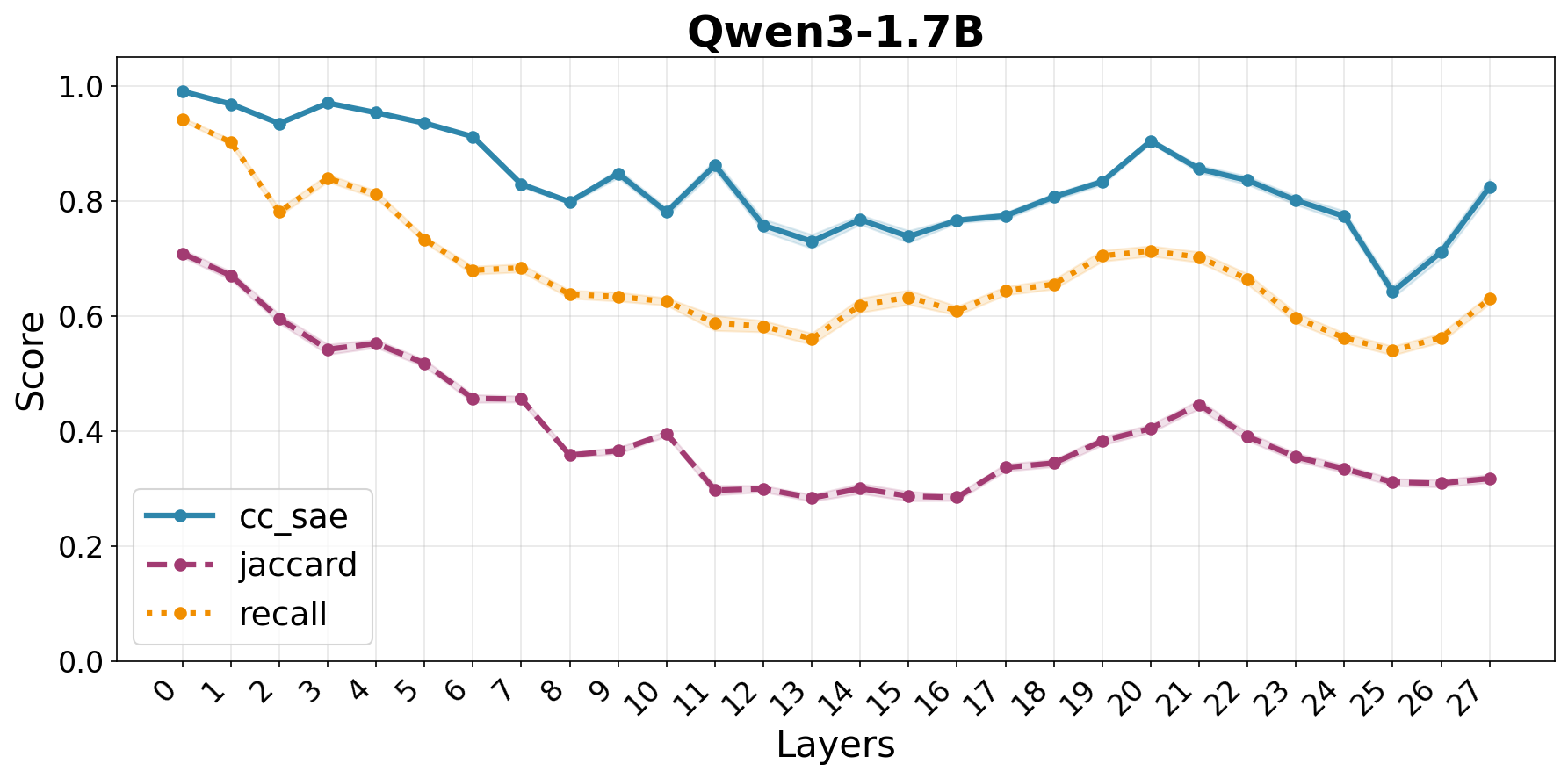}
        \end{subfigure}
        \hfill
        \begin{subfigure}[b]{0.32\textwidth}
            \includegraphics[width=\textwidth]{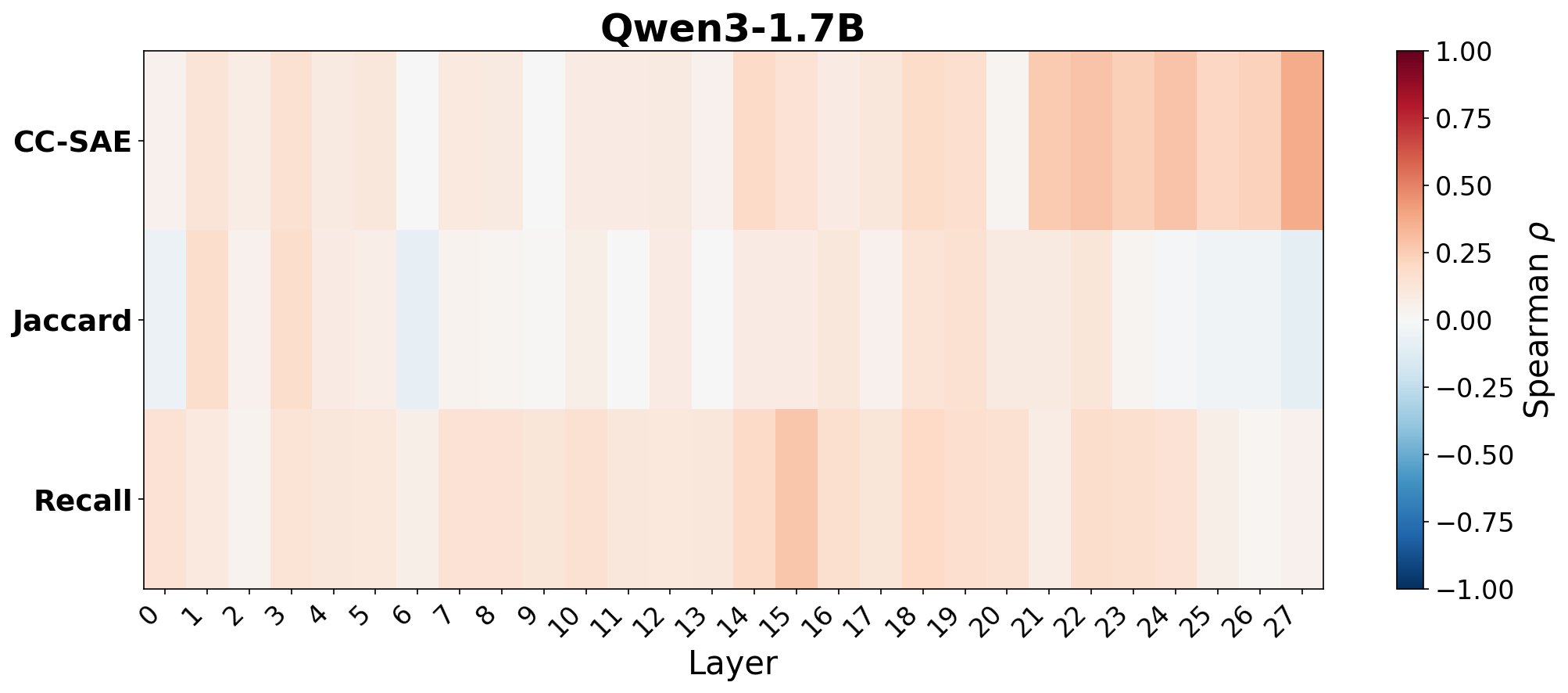}
        \end{subfigure}

        \begin{subfigure}[b]{0.32\textwidth}
            \includegraphics[width=\textwidth]{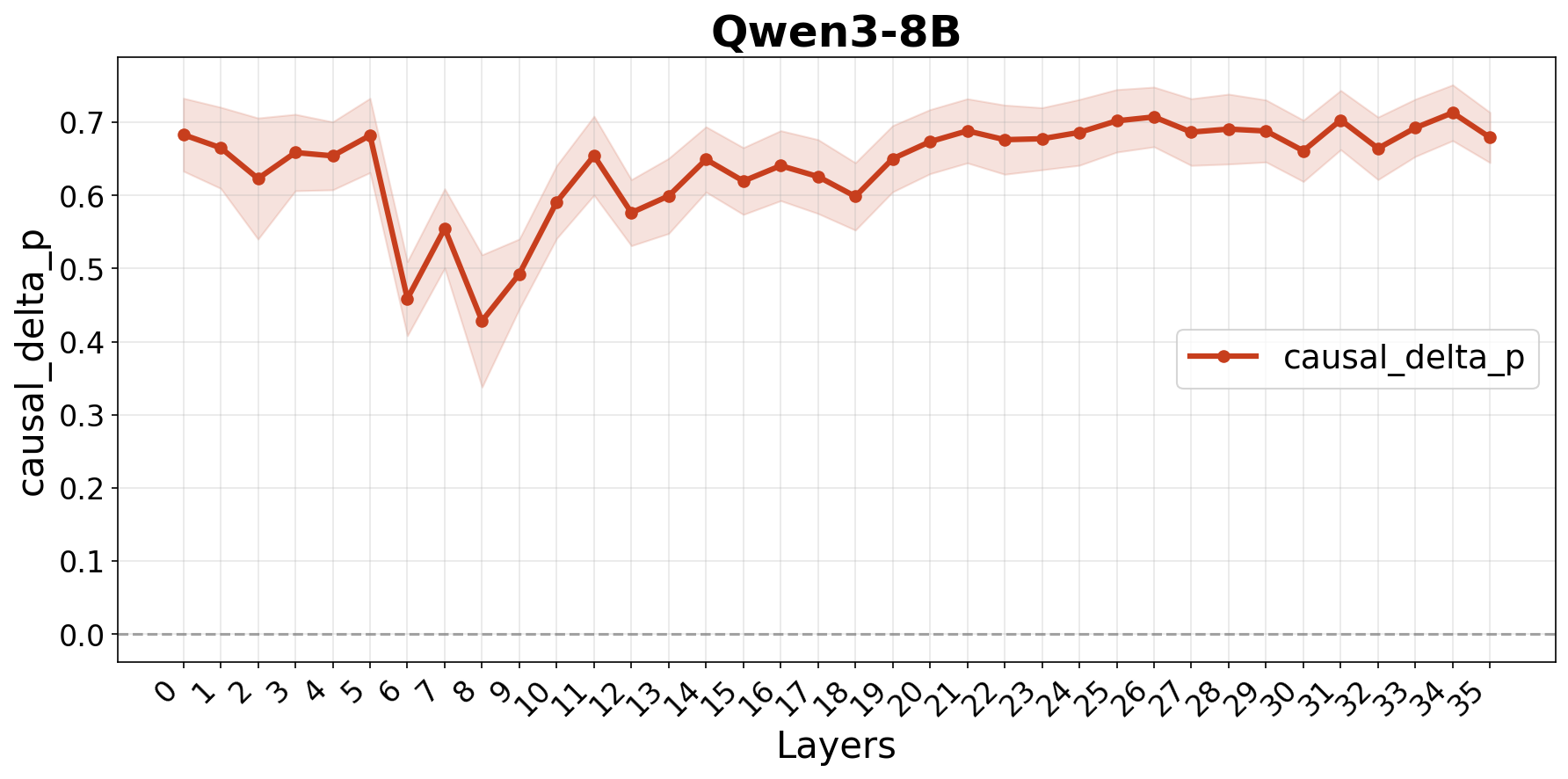}
        \end{subfigure}
        \hfill
        \begin{subfigure}[b]{0.32\textwidth}
            \includegraphics[width=\textwidth]{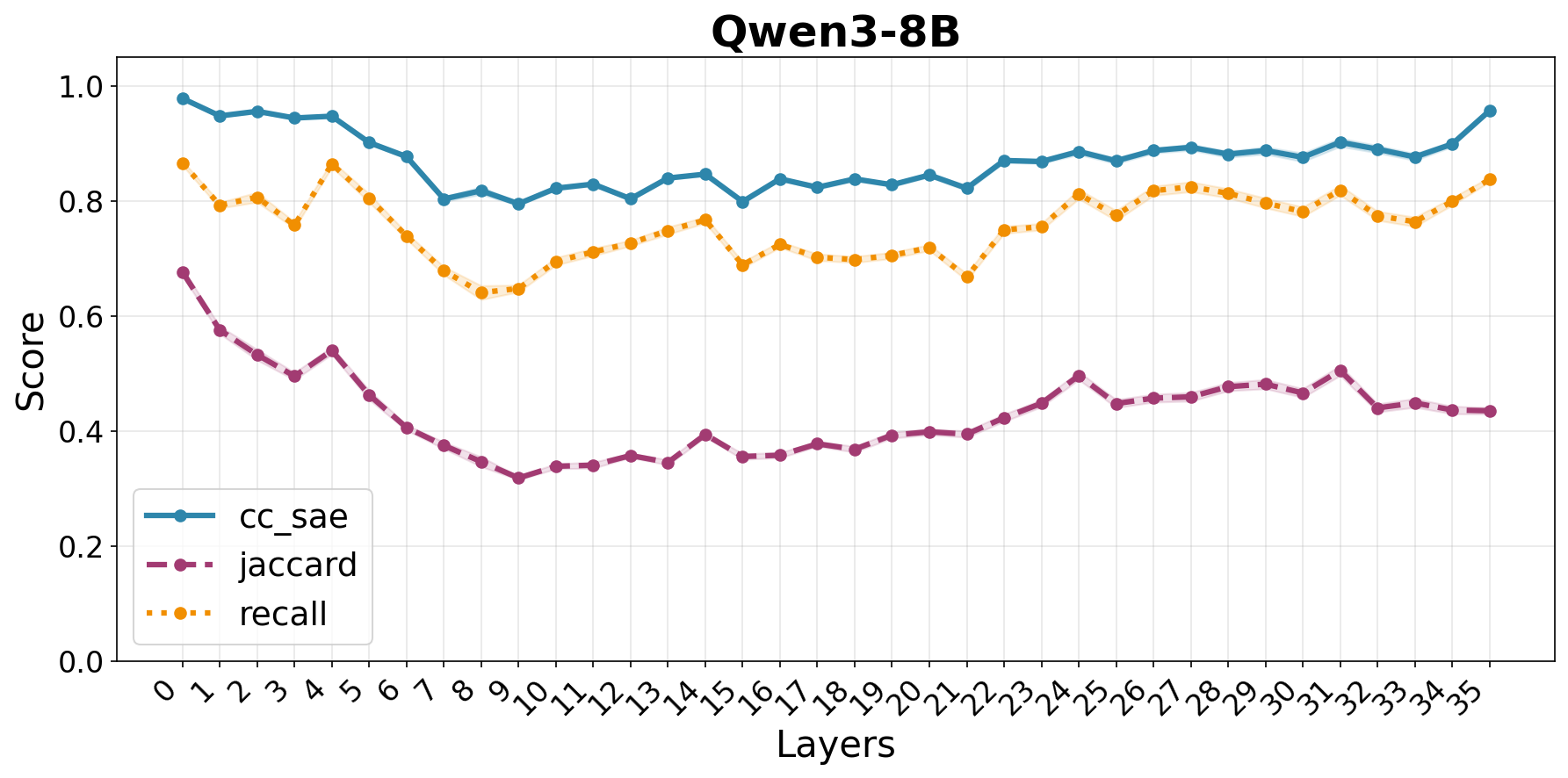}
        \end{subfigure}
        \hfill
        \begin{subfigure}[b]{0.32\textwidth}
            \includegraphics[width=\textwidth]{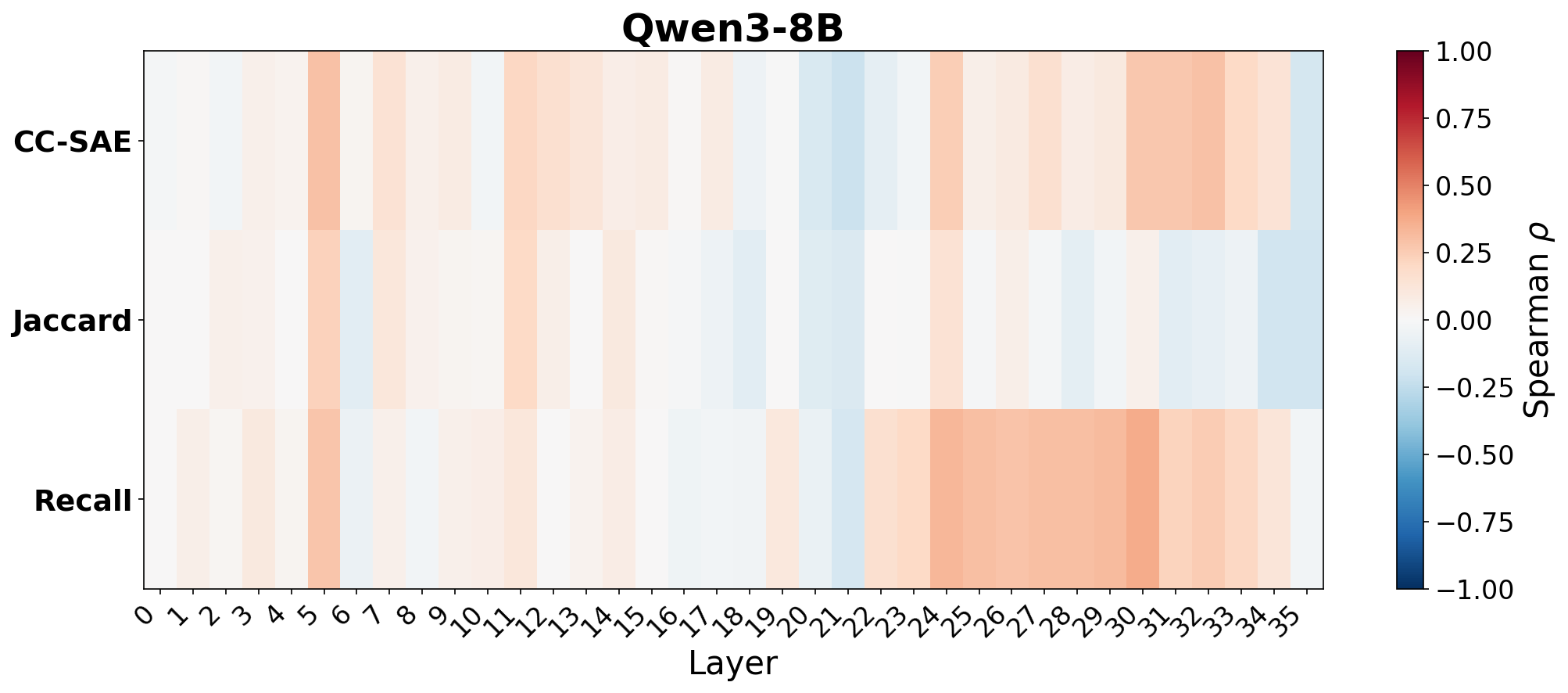}
        \end{subfigure}

        \begin{subfigure}[b]{0.32\textwidth}
            \includegraphics[width=\textwidth]{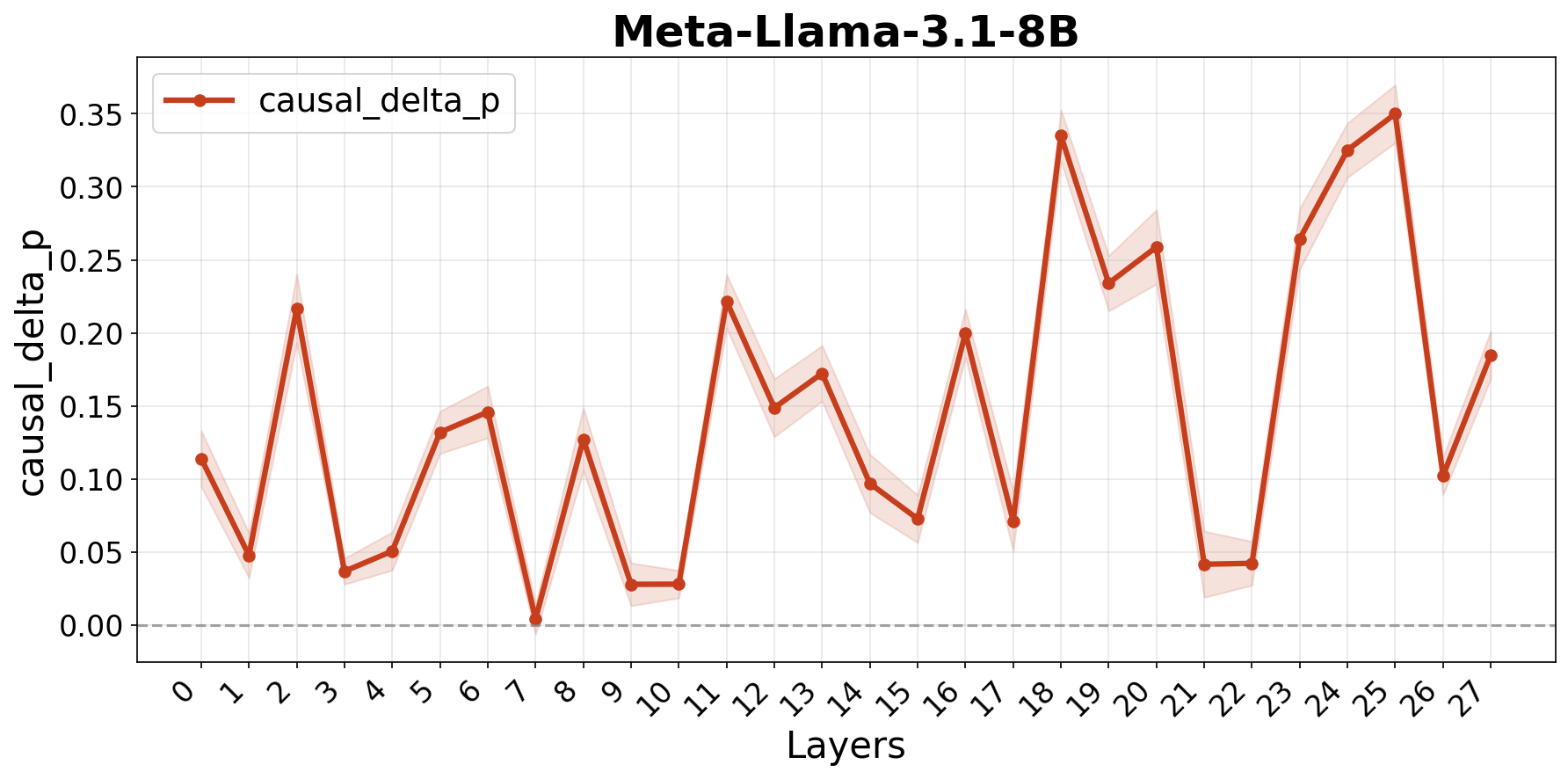}
        \end{subfigure}
        \hfill
        \begin{subfigure}[b]{0.32\textwidth}
            \includegraphics[width=\textwidth]{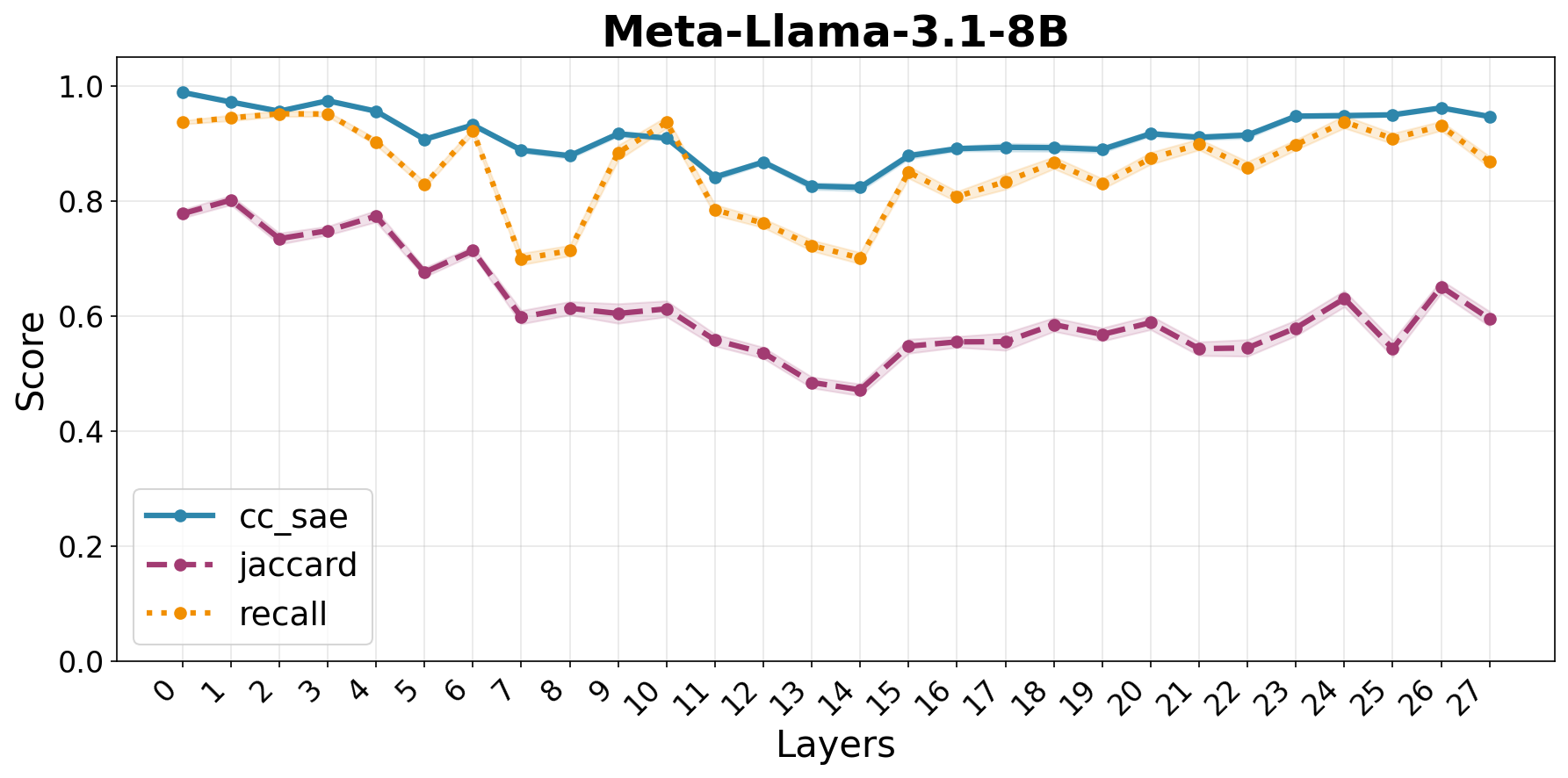}
        \end{subfigure}
        \hfill
        \begin{subfigure}[b]{0.32\textwidth}
            \includegraphics[width=\textwidth]{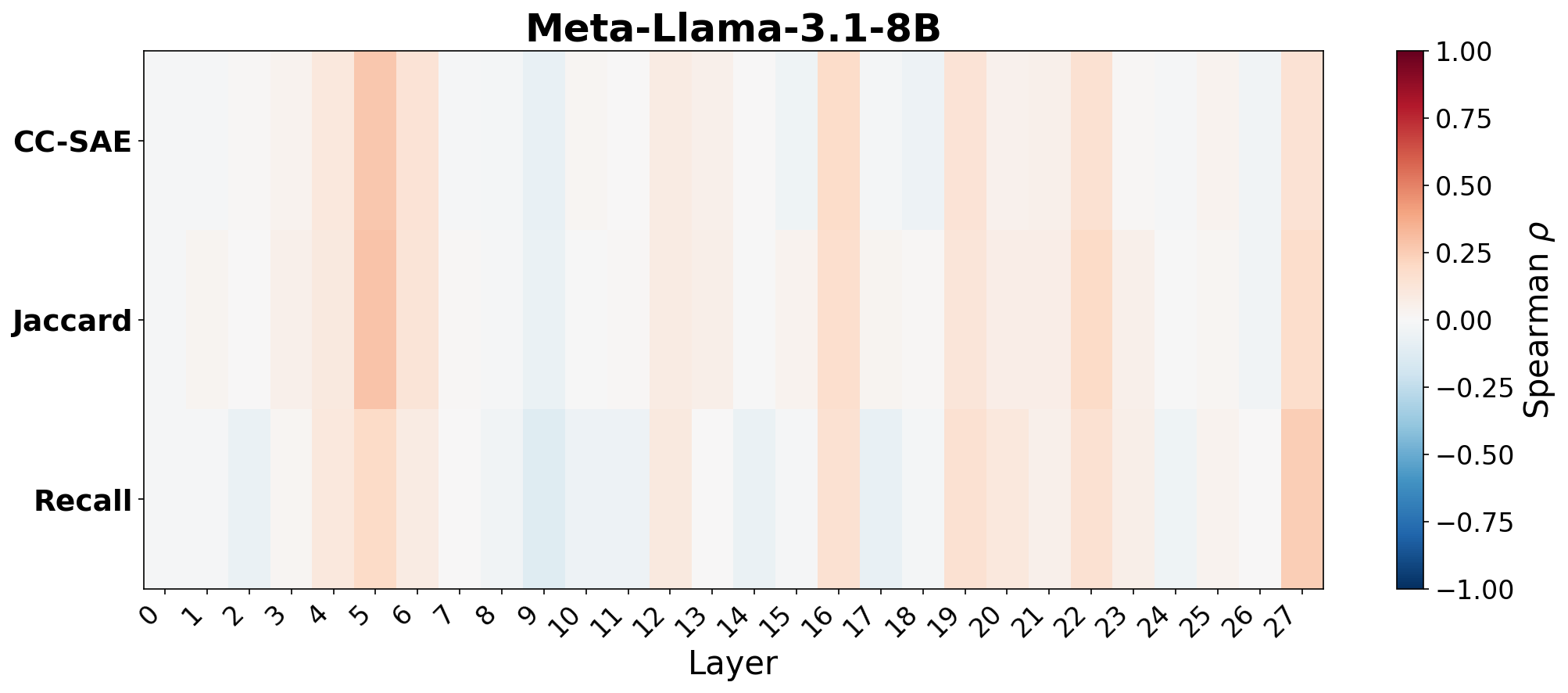}
        \end{subfigure}
    \end{minipage}%
    }
    \vspace{-0.5em}
    
    \caption{CoT faithfulness evaluation across layers for \lm{Qwen} and \lm{Llama} on \data{LogiQA} (\lm{Gemma} results in Figure~\ref{fig:ccsae-layers-app}).
    \textbf{Left column:} Causal faithfulness ($\Delta p$) measures the probability drop when ablating shared features.
    \textbf{Middle column:} Correlational metrics (CC-SAE, Jaccard, Recall) between prediction and CoT activations.
    \textbf{Right column:} Spearman correlation heatmap between each correlational metric and $\Delta p$ across layers.
    Shaded regions indicate 95\% confidence intervals (left and middle columns). Additional results on the \textit{remaining datasets} are provided in Appendix~\ref{app:faithfulness_evaluation}.}
    \label{fig:ccsae-layers}
    \vspace{-0.5em}
\end{figure*}

\subsection{Correlational Metrics and Their Relationship to The Causal Metric}
\label{subsec:corr_causal_comparison}

We first observe that the correlational metrics (\S\ref{subsec:correlation}) achieve reasonably high magnitudes (CC-SAE > Recall > Jaccard) (Figure~\ref{fig:ccsae-layers}; middle column). This indicates decent conceptual alignment between prediction and CoT passes, i.e., \boxc{grey}{\textit{self-consistency}}, which provides a solid basis for using $\Delta p$ (\S\ref{subsec:layerwise}).

We further explore the correlational metrics with respect to how and to what extent they correlate with the causal metric $\Delta p$ (\S\ref{subsec:causal}). 
Figure~\ref{fig:ccsae-layers} (right column) illustrates that the correlational metrics align more closely with $\Delta p$ in late layers (except for \lm{Llama-3.1-8B}), while in early layers the correlation can even become negative.  Additionally, Table~\ref{tab:metric-correlation} reveals that CC-SAE has a larger magnitude and correlates more strongly with $\Delta p$, outperforming Recall, followed by Jaccard.
Overall, all three Spearman correlations are negligible to weak, with substantial variance across layers. 

Table~\ref{tab:necessity_normalized} provides more evidence of weak correlation by binarizing both CC-SAE ($\theta^*$) and $\Delta p$ ($\tau^*$) using per-dataset median thresholds. The alignment–causal-effect association is weak ($A$: above-median CC-SAE; $B$: above-median $\Delta p$): $\bar{P}(A|B) = 0.53$, $\bar{P}(B|A) = 0.54$, with lift $= 1.08$. High alignment thus raises the probability of a high causal effect by only 8\% over baseline: concept overlap identifies \textit{which} features are shared, but high overlap (\boxc{grey}{\textit{self-consistency}}) neither guarantees nor precludes a strong causal effect (\boxc{grey}{\textit{faithfulness}}), consistent with \citet{parcalabescu-frank-2024-measuring}. Taken together, \textit{quantifying the contribution of overlapping features therefore requires causal metrics.}


\subsection{Validation of the Causal Role of Intersection Features}
\label{subsec:sanity_check}

\subsubsection{Prediction Pass} 
\label{subsubsec:control}
To validate the strength of the causal effect of intersection features $S_{\cap}$ with size $k$, 
we evaluate four base conditions for sanity checks (the Venn diagram of $S_\text{pred}$ and $S_\text{cot}$ is illustrated in Figure~\ref{fig:venn}):
\begin{itemize}[noitemsep,topsep=0pt,leftmargin=*]
    \item \textbf{Condition A:} Sample $k$ random features from $(S_{\text{pred}} \cup S_{\text{cot}}) \setminus S_{\cap}$. It aims to test whether intersection features influence the answer probability more than other features that are active in either pass.\looseness=-1
    \item \textbf{Condition B:} Sample $k$ random features from $S_{\text{pred}} \setminus S_{\cap}$. It aims to test whether the causal effect is specific to features reused by CoT.
    \item \textbf{Condition C:} Norm-matched sampling from $S_{\text{pred}} \setminus S_{\cap}$, controlling for the $L_2$ norm of the ablation vector (Appendix~\ref{app:condition_c}). It aims to test whether the effects are not driven by ablation magnitude.\looseness=-1
    \item \textbf{Condition D:} Ablate $S_{\cap}$ together with $k$ additional random features from $S_{\text{pred}} \setminus S_{\cap}$. If $S_{\cap}$ captures the causally important features, adding prediction-only features should not noticeably increase $\Delta p$ beyond ablating $S_{\cap}$ alone, which serves as a completeness check.
\end{itemize}
Additionally, we define and compute a \textbf{top-$k$ upper bound} (magnitude-ranked baseline), i.e., ablating the $k$ features in $S_{\text{pred}}$ with the largest activation magnitude. 

\begin{figure*}[!t]
\centering
\resizebox{\textwidth}{!}{
\begin{minipage}{\textwidth}
\includegraphics[width=\textwidth]{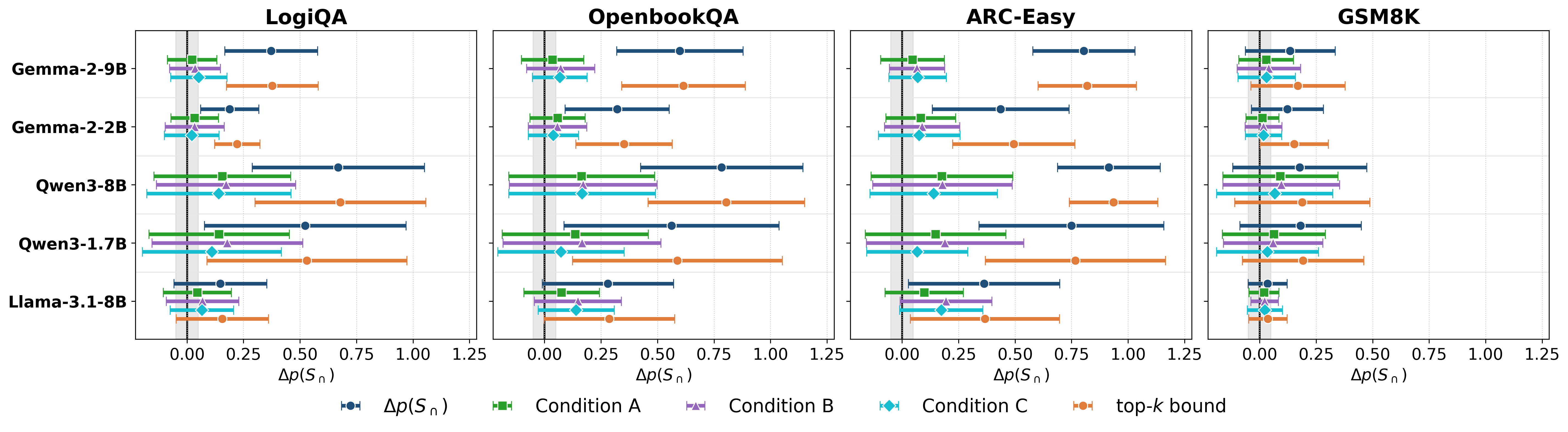}
\end{minipage}
}
\caption{Comparison of prediction-pass $\Delta p(S_{\cap})$ with three causal-ablation conditions and a top-$k$ upper bound across models and datasets. The $x$-axis shows $\Delta p$; error bars show \textit{layer-wise variance}. See Table~\ref{tab:baseline} for full raw results and Appendix~\ref{app:baseline_sanity_check} for \textbf{Condition D}.
}
\vspace{-1em}
\label{fig:baseline}
\end{figure*}

\textbf{Results.}
Figure~\ref{fig:baseline} and Table~\ref{tab:baseline} show that the intersection ablation consistently yields close to the largest $\Delta p$, while random baselines A--C yield $\Delta p$ close to 0. This reveals two findings. First, the causal effect is specific to intersection features rather than to arbitrary subsets of $S_{\text{pred}}$ or features active in one pass alone. Second, causal importance is as driven by conceptual reuse and grounding as it is by activation magnitude: the top-$k$ upper bound, which selects the $k$ highest-magnitude features in $S_{\text{pred}}$, is only marginally above $\Delta p(S_{\cap})$, so intersection features are nearly as causally potent as the best magnitude-selected subset of the same size, despite not being chosen for magnitude. For \textbf{Condition D}, Table~\ref{tab:baseline_e} illustrates a negligible incremental effect across all model-dataset combinations, indicating that $S_\cap$ alone is complete for the causal effect. Features that are active only in the prediction pass but are not reused from CoT contribute minimally to the model's final answer.



\subsubsection{CoT Pass} 

\label{subsubsec:cot_pass_ablation}
\label{subsubsec:cot_pass_ablation}
To establish that $S_{\cap}$ reflects genuine \textit{reasoning-relevant} computation, we further ablate these features on the \textbf{CoT pass} and examine their causal effect from two angles (Appendix~\ref{app:cot_pass_ablation}). First, we test cross-path consistency in terms of the causal effect of $S_\cap$. Second, analogous to the prediction pass (\S\ref{subsec:sanity_check}), we run four base conditions for sanity (Appendix~\ref{app:cot_pass_conditions}).

\textbf{Cross-pass Consistency.} For cross-pass consistency, we compare the causal effect of $S_{\cap}$ across the two inference paths. Comparing $\Delta p(S_{\cap})$ across paths, we observe similarly high values on both prediction and CoT paths (Table~\ref{tab:cot_ablation}). This rules out the interpretation that intersection features are shallow prediction heuristics that become irrelevant when explicit reasoning is available; rather, they remain causally central to both inference modes.

\textbf{Control Conditions.} Table~\ref{tab:cot_baseline} shows that \textbf{Conditions A--C} yield $\Delta p$ substantially lower than $S_{\cap}$. The top-$k$ upper bound remains marginally above $\Delta p(S_{\cap})$, confirming that intersection features, selected by conceptual reuse rather than magnitude, are nearly as causally potent as the best magnitude-selected subset. Furthermore, \textbf{Condition D} yields $\Delta p$ nearly identical to ablating all of $S_{\text{cot}}$ (Table~\ref{tab:cot_ablation}), indicating that intersection features account for essentially all causal contributions of CoT-active features to the reasoning process. In other words, CoT-only features provide no additional effect on the CoT pass. All three findings mirror those observed in the prediction pass (\S\ref{subsubsec:control}). \looseness=-1



\subsection{Layer-wise Causal CoT Faithfulness}
\label{subsec:layerwise}
Prior faithfulness metrics \citep{lanham2023measuring, turpin2023language, ye2026mechanistic} typically treat faithfulness as model-level, reporting a single number. In contrast, by operating in the internal representation space, our proposed metrics enable \textit{a layer-wise faithfulness analysis and show that \boxc{grey}{faithfulness} varies notably across layers} (Figure~\ref{fig:ccsae-layers} for \data{LogiQA}; results for \data{OpenbookQA}, \data{ARC-Easy}, and \data{GSM8K} are in Appendix~\ref{app:faithfulness_evaluation}).\footnote{$S_\text{pred}$ and $S_\text{cot}$ overlap only partially (33--75\%), indicating that $S_\cap$ captures a selective subset of prediction-relevant features that are also activated during CoT reasoning (Appendix~\ref{app:concept_composition}), rather than all features active at the answer position. Additional sanity checks rule out the potential that these are merely answer-identity features (Appendix~\ref{app:answer_identity}) or ubiquitous features (Appendix~\ref{app:ubiquitous}). Moreover, we find that $S_\cap$ is largely sufficient to recover the answer alone, compared to the random baseline (Appendix~\ref{app:sufficiency}).} Our experimental results further reveal three patterns.

\begin{wrapfigure}{r}{0.45\columnwidth}
\vspace{-1.2em}
\centering
    \includegraphics[width=\linewidth]{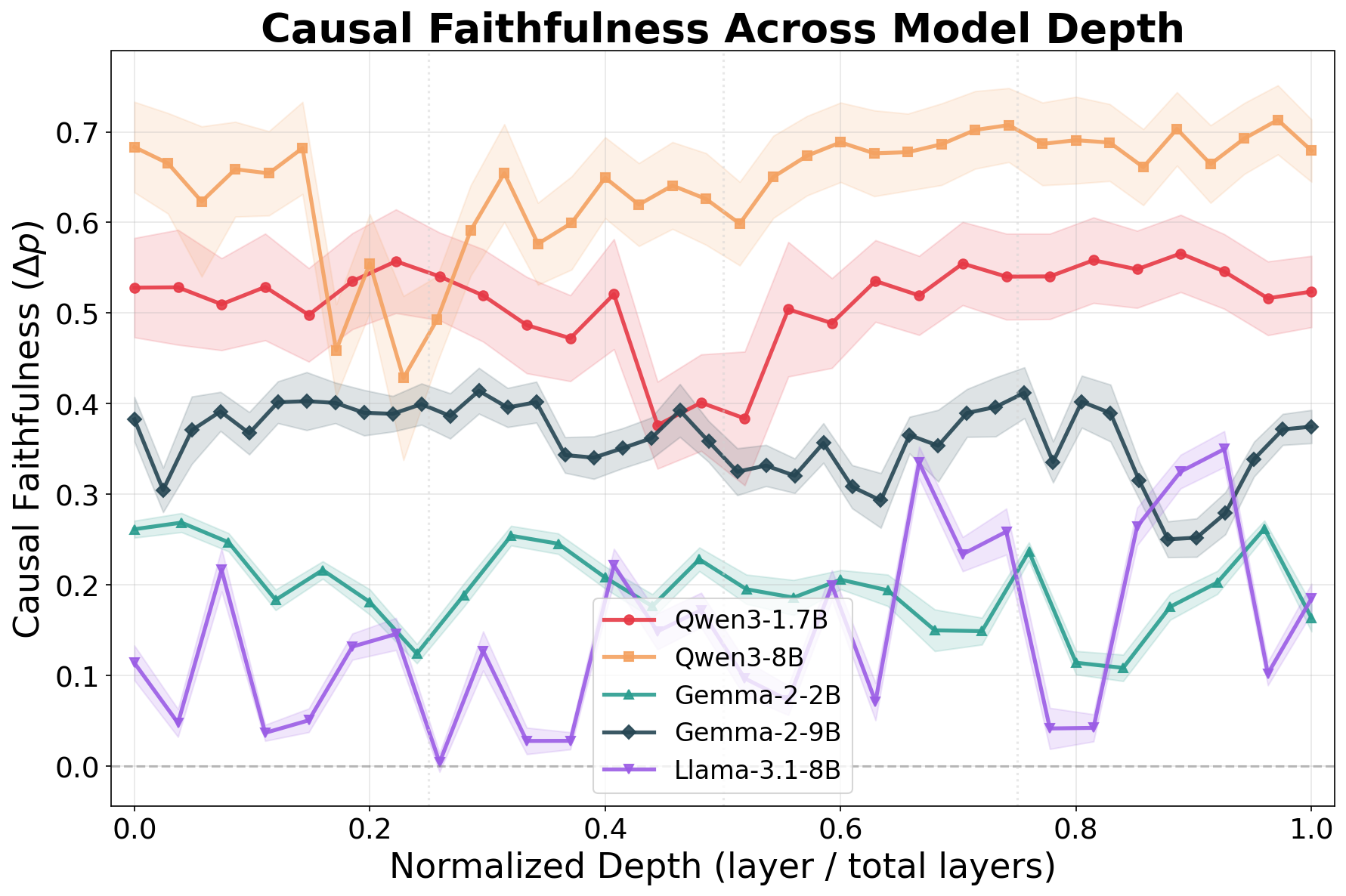}
    \caption{Causal faithfulness ($\Delta p$) variation across normalized depth on \data{LogiQA}. Full results are presented in Figure~\ref{fig:summary}.}
    \label{fig:logiqa_relative}
\vspace{-0.5em}
\end{wrapfigure}

\textbf{Both peak and dip locations are model-specific, with peaks at mid-to-late rather than final layers.}
For all models except \lm{Gemma-2-2B}, peak faithfulness often occurs at mid-to-late layers (e.g., on \data{LogiQA}, L24/28 for \lm{Qwen3-1.7B}, L31/42 for \lm{Gemma-2-9B}; Table~\ref{tab:faithfulness-peak}) and not at the final layer. A plausible explanation is that the final layers are devoted to output formatting and next-token prediction rather than manipulating reasoning concepts \citep{lad2025remarkable}. In contrast, faithfulness dips appear at model-specific layers: \lm{Qwen3} models exhibit early-mid to mid-layer dips, while \lm{Llama-3.1-8B} exhibits high layer-to-layer variance with no clear trend (Figure~\ref{fig:ccsae-layers}; left column). \lm{Gemma-2} models exhibit task-dependent trends
, with the two models showing opposite patterns: \lm{Gemma-2-9B} drops at late layers on MCQA (Figure~\ref{fig:logiqa_relative}; Figure~\ref{fig:summary_logiqa}--\ref{fig:summary_arc_easy}) but at early layers on \data{GSM8k} (Figure~\ref{fig:summary_gsm8k}), while \lm{Gemma-2-2B} exhibits a weaker and less consistent version of the reverse. \looseness=-1 


\textbf{Faithfulness manifests as distributed or localized patterns across architectures.}
Figure~\ref{fig:ccsae-layers} and Figure~\ref{fig:logiqa_relative} demonstrate that layer-wise $\Delta p$ stability varies substantially across models: \lm{Llama-3.1-8B} exhibits the highest layer-to-layer variance, with adjacent layers differing by up to $\Delta p=0.3$ (Table~\ref{tab:faithfulness-peak}); \lm{Qwen3} models are the most stable across datasets, and \lm{Gemma-2-2B} shows moderate fluctuations of consistent magnitude across layers (Figure~\ref{fig:stability}). Additionally, the contrast between \lm{Qwen3}'s \textit{distributed faithfulness}, which is consistently high across most layers with great stability, and \lm{Llama-3.1-8B}'s \textit{localized faithfulness}, concentrated at, e.g., L18--20 and L23--25 on \data{LogiQA}, suggests that \textit{CoT faithfulness may differ qualitatively across architectures, not only in magnitude}. 
Moreover, distributed faithfulness suggests a tight coupling between CoT and prediction throughout the model's computation, potentially suggesting more robust reasoning that is less susceptible to perturbations at any single layer. In contrast, localized faithfulness implies that the alignment emerges only at specific computational stages, with the CoT reasoning proceeding largely independently elsewhere (Appendix~\ref{app:repair}). 

\textbf{Scaling reshapes the layer-wise profile, not just its magnitude.}
Within the same model family, larger models generally exhibit greater faithfulness in magnitude ($\Delta p$) (Figures~\ref{fig:ccsae-layers} and \ref{fig:logiqa_relative}), aligned with \citet{siegel2026verbosity, wang2026largelanguagemodelsexplain} (Table~\ref{tab:cot_ablation} provides additional evidence: for example, \lm{Qwen3-8B} achieves a normalized $\Delta p = \frac{\Delta p (S_\cap)}{\Delta p (S_\text{pred})}$ of 0.96 on \data{LogiQA}, while \lm{Qwen3-1.7B} reaches only 0.90). Beyond this uniform shift, scaling also reshapes the layer-wise profile. For instance, within the \lm{Qwen3} family, scaling from 1.7B to 8B shifts the faithfulness dip from a normalized depth\footnote{Normalized depth is defined as the dip layer index divided by the total number of layers (Table~\ref{tab:faithfulness-peak}).} of 44\% (L12--14) to 23\% (L6--8) on \data{LogiQA} (Figure~\ref{fig:ccsae-layers}; Table~\ref{tab:faithfulness-peak}), 
suggesting that larger models may compress the early processing region. 
Within the \lm{Gemma-2} family, scaling from 2B to 9B additionally reduces the coefficient of variation across layers from 23\% to 12\%. 



\textbf{Implications for CoT faithfulness evaluation.}
The substantial layer-wise variation raises a practical question: which layer should practitioners use when reporting faithfulness? We recommend reporting either (1) the \textbf{peak-layer} $\Delta p$, which captures the model's maximum potential for faithful reasoning, or (2) the \textbf{mean} $\Delta p$ across layers, which reflects overall alignment but may underestimate faithfulness for models with localized patterns.\footnote{More actionable implications are further discussed in Appendix~\ref{app:implications}. In addition, we provide a comparison between our proposed metrics and existing CoT faithfulness metrics in Appendix~\ref{app:alternative}. Meanwhile, we present post-hoc rationalization examples identified by $\Delta p$ in Appendix~\ref{app:posthoc_rationalization}.}

\begin{figure*}[!t]
\centering

\begin{subfigure}{\textwidth}
    \centering
    \includegraphics[width=\textwidth]{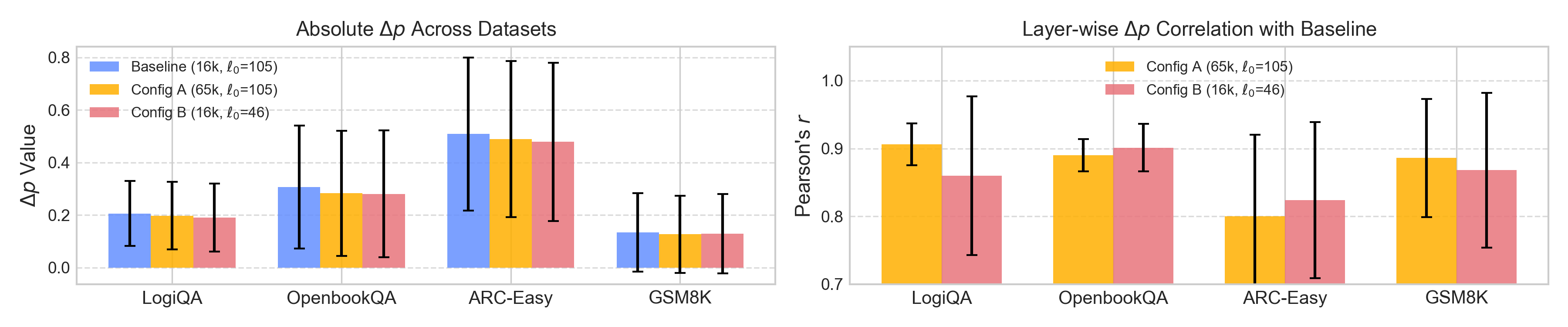}
    \caption{Comparison of layer-aggregated $\Delta p$ values and Pearson's $r$ (see Table~\ref{tab:ablation-gemma2-2b} for raw values).}
    \label{subfig:corr}
\end{subfigure}

\begin{subfigure}{\textwidth}
    \centering
    \includegraphics[width=\textwidth]{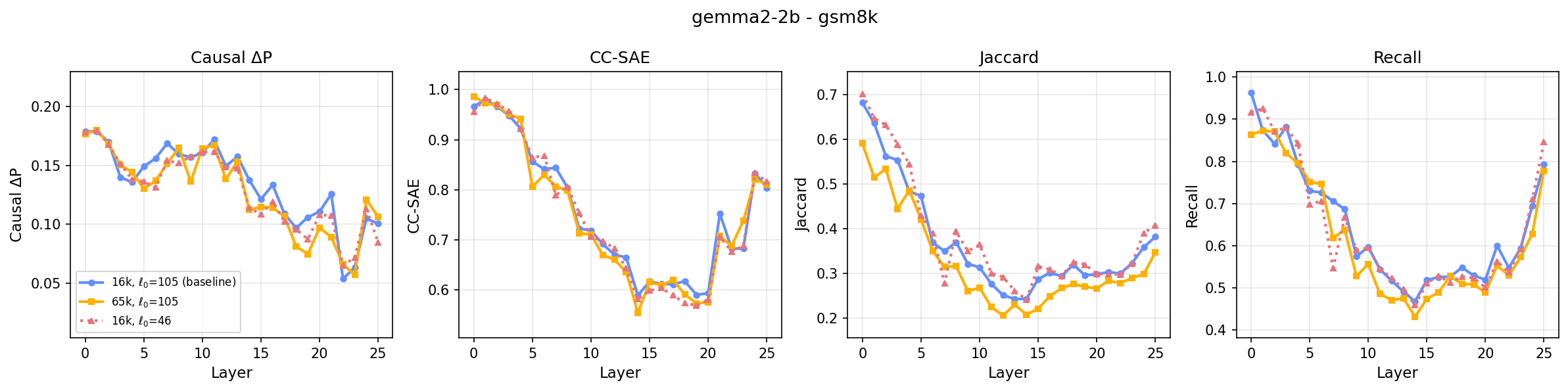}
    \caption{Layer-wise $\Delta p$, CC-SAE, Jaccard, and Recall variation on \data{GSM8K} (see full results in Figure~\ref{fig:ablation-gemma2-2b}).}
    \label{subfig:layerwise}
    \vspace{-0.5em}
\end{subfigure}
\vspace{-1em}
\caption{SAE configuration ablation study on \lm{Gemma-2-2B} with \boxc{lightyellow}{increased width} and \boxc{red}{increased sparsity}. 
Full results with other models across all datasets are provided in Appendix~\ref{app:ablation_study}.}
\label{fig:baseline_layerwise}
\end{figure*}

\subsection{SAE Configuration Ablation Study}
\label{subsec:ablation_study}
\textbf{Motivation and Setup.}
SAEs introduce hyperparameters, i.e., \textit{dictionary width} and \textit{target sparsity}, that may confound downstream analyses. To verify that our findings genuinely reflect model behavior rather than SAE-specific artifacts, we conduct ablation studies on \lm{Gemma-2-2B} (the results of other models are provided in Appendix~\ref{app:ablation_study})
. We compare the baseline (width=$16k$, $\ell_0=105$) against two ablation conditions: increased width (16$k$ $\rightarrow$ 65$k$) and increased sparsity ($\ell_0=105 \rightarrow 46)$. \looseness=-1

\textbf{Results.} Figure~\ref{subfig:corr} shows a highly correlated $\Delta p$ under different SAE configurations (Pearson $r>0.8$). This indicates that the SAE choice does not substantially alter \textit{which} examples are judged as faithful. Figure~\ref{subfig:layerwise} further demonstrates broadly consistent layer-wise $\Delta p$ profiles across configurations, with all three exhibiting noticeably similar trends in general. Some variation emerges at mid-to-late layers (L15-20), where the baseline configuration yields slightly higher $\Delta p$ than the wider or sparser alternatives.\footnote{One possible explanation is that intermediate layers encode increasingly compositional features with potentially higher superposition density, making them more sensitive to SAE hyperparameter \citep{ gurnee2026verbalizablerepresentationsformglobal}.\looseness=-1} However, the overall profile shape and peak locations remain aligned, suggesting that the layer-wise patterns reported in \S\ref{subsec:layerwise} are not artifacts of a particular SAE configuration. Moreover, CC-SAE, Jaccard, and Recall report moderate correlations across configurations, with Jaccard exhibiting the largest sensitivity to sparsity changes, which is expected given its direct dependence on feature set cardinality. Overall, these results indicate that \textit{causal faithfulness captures properties of the model's reasoning process rather than idiosyncrasies of the SAE}.

\begin{table*}[t]
\centering
\caption{Natural language descriptions of a subset of shared SAE concepts  $S_\cap$.
Full trajectory is presented in Table~\ref{tab:qual-gemma9b-logiqa} and additional case studies for the remaining datasets are provided in Appendix~\ref{app:concept_transition}.}
\label{tab:qual-gemma9b-logiqa-summary}
\scriptsize
\resizebox{0.95\textwidth}{!}{
\begin{tcolorbox}[
    width=\textwidth,
    colback=lightblue,
    colframe=purple,
    boxrule=0.7pt,
    arc=3pt,
    left=6pt, right=6pt, top=4pt, bottom=5pt,
    title={\lm{Gemma-2-9B} \ \textcolor{purple!40!black}{$\vert$} \ \data{LogiQA}},
    fonttitle=\bfseries\small,
    colbacktitle=lighpurple,
    coltitle=black,
    boxsep=2pt
]
\noindent

\begin{minipage}[t]{0.50\textwidth}
\vspace{0pt}
\raggedright
\textbf{Q.} Many elderly Beijings have never forgotten the unique skills they saw at temple fairs \ldots\ as the modernization of society accelerates, Chinese folk culture is facing unprecedented crisis \ldots\ all of which make the development of folk art face difficulties. \emph{Which of the following can be derived from this text?}
\begin{itemize}[leftmargin=1.2em, itemsep=0pt, topsep=2pt, parsep=0pt]
    \item[\textbf{A:}] Marketization is the way out for folk art
    \item[\textbf{B:}] Folk culture needs rescue protection \goldicon
    \item[\textbf{C:}] Urban construction should highlight cultural characteristics
    \item[\textbf{D:}] \ldots\ social status of folk art talents should be improved
\end{itemize}
\textbf{Pred:} B~\okmark \quad \textbf{CoT:} B~\okmark \quad \textbf{Gold:} B~\goldicon
\par\noindent\textcolor{purple!60!black}{\rule{\linewidth}{0.5pt}}
\par\noindent
\textbf{CoT:}~\textit{``\ldots\ the survival of folk art is facing unprecedented crisis \ldots\ the development of folk art is facing difficulties. So, the answer is B. The text does not say that marketization is the way out \ldots\ that urban construction should highlight cultural characteristics \ldots\ So, the answer is B.''}
\end{minipage}
\hspace{1pt}
\begin{minipage}[t]{0.50\textwidth}
\vspace{0pt}
\centering
\setlength{\tabcolsep}{1pt}
\begin{tabular}{@{}ccp{4.9cm}@{}}
\rowcolor{white}
\arrayrulecolor{purple!60!black}
\toprule
\textbf{L} & \textbf{ID ($S_\cap$)} & \textbf{Feature Description} \\
\midrule
\rowcolor{lighpurple!50}
\textbf{L5} & \colorbox{yellow!35}{\texttt{\#4649}} & Specialized compound nouns \ldots\ social sciences, economics, and formal academic assessments. \\
\arrayrulecolor{purple!30}\cmidrule(lr){1-3}
\rowcolor{lightblue!50}
\textbf{L11} & \texttt{\#10504}  & Pinyin transliterations of Chinese proper nouns \ldots\ names, locations, cultural terms. \\
\arrayrulecolor{purple!30}\cmidrule(lr){1-3}
\rowcolor{lighpurple!50}
\textbf{L21} & \texttt{\#7862}  & Formal academic or examination-style text \ldots\ Chinese-to-English translations in logical reasoning. \\
\arrayrulecolor{purple!30}\cmidrule(lr){1-3}
\rowcolor{lightblue!50}
\textbf{L41} & \colorbox{yellow!35}{\texttt{\#13986}} & Academic terminology \ldots\ cultural heritage, archaeological sites, and logical observations. \\
\arrayrulecolor{purple}\bottomrule
\end{tabular}
\vspace{4pt}
\par\raggedright\scriptsize
\textit{\colorbox{yellow!35}{\phantom{X}} marks features overlapping with concepts verbalized in the CoT.}
\end{minipage}
\end{tcolorbox}
}


\end{table*}

\subsection{Layer-to-Layer Transitions in Shared Concepts}
\label{subsec:layer_transition}
To interpret the shared concepts $S_\cap$ activated at each layer and analyze the feature transition across layers, we map the active $S_\cap$ to natural-language descriptions using an automated interpretability pipeline following \citet{pmlr-v267-paulo25a}. Specifically, we extract the top-activating shared concepts at every layer and retrieve their corresponding descriptions, allowing us to qualitatively trace how the model's encoded concepts evolve across layers (further trajectories are provided in Appendix~\ref{app:concept_transition}).


Table~\ref{tab:qual-gemma9b-logiqa-summary} illustrates a Chinese folk art case study from \data{LogiQA}, mapping its shared concept ($S_\cap$) transitions and layer-wise descriptions extracted from \lm{Gemma-2-9B}. $S_\cap$ exhibits a distinct trajectory across layers: early layers encode topic-specific concepts; middle and late layers shift toward more abstract concepts detached from specific entities or explicitly tied to evaluating candidate answers. Specifically, the trajectory reveals a cross-lingual transition: $S_\cap$ is initially anchored in Chinese topic-related entities; at a middle layer (L21), a Chinese-to-English translation feature emerges, suggesting a bridge from topic-specific Chinese concepts into English ones \citep{wendler-etal-2024-llamas}.

Moreover, we observe a high $\Delta p$ in layers associated with question content or answer evaluation; conversely, $\Delta p$ drops noticeably for layers with topically irrelevant concepts, e.g., L16 and L25 (Table~\ref{tab:qual-gemma9b-logiqa}). Additionally, certain highly activated concepts across layers mirror content explicitly verbalized in the CoT, suggesting that these concepts indeed track specific reasoning content. Nevertheless, \textit{not all concepts with high $\Delta p$ are verbalized in the CoT. Conversely, some content of the CoT verbalization may not appear among the highly activated layer-wise concepts}. \looseness=-1

\section{Conclusion}
We recast CoT faithfulness as an internal concept grounding: whether an LLM's CoT reasoning relies on the same SAE concepts that also support the LLM's direct prediction, and whether the shared concepts causally drive its answer. A single shared SAE renders the cross-pass concept overlap well-defined, supporting correlational alignment metrics and a causal metric $\Delta p$, a necessity test validated against count, arbitrary feature selection, and SAE-choice confounds. Across five LLMs and four reasoning benchmarks, although concept alignment reveals substantial feature overlap between passes, it neither establishes nor negates a strong causal link; precisely determining a feature's contribution requires causal metrics like $\Delta p$. Faithfulness generally peaks at mid-to-late depths, is distributed in some architectures yet localized in others, and, within model families, is reorganized rather than merely raised by scale, arguing for layer-aware reporting. Natural-language concept descriptions further show that causally important shared concepts may not be verbalized in the CoT. Overall, our findings suggest a future direction for faithfulness evaluations that also test whether the internal representations associated with a CoT causally support the model's prediction.\looseness=-1

\section*{AI Use Statement}
The authors used Claude (Code) to assist with language polishing (including grammar, clarity, and coherence), literature search, and code implementation support. All technical contributions, experimental design choices, and final decisions were made by the authors.

\section*{Acknowledgments}
We thank \textbf{Yifan Wang}, \textbf{Yihong Liu}, and \textbf{Ruta Binkyte} for their valuable feedback and constructive criticism, which helped improve the paper. Additionally, we thank \textbf{Pingjun Hong} and \textbf{Xu Shen} for providing CIE scores \citep{shen2026detectingunfaithfulchainofthoughtcircuitguided} to measure the correlation between our metrics and existing ones (Appendix~\ref{app:alternative}).

\bibliography{custom}
\bibliographystyle{iclr2027_conference}

\clearpage

\appendix

\section{Preliminary}

\subsection{Comparison with Prior Work}
Table~\ref{tab:comparison} illustrates the CoT faithfulness evaluation metric comparison. The evaluation dimensions are listed as follows:
\begin{itemize}
    \item \textbf{Granularity}: The level at which the CoT is treated;
    \item \textbf{Intervention}: Whether the CoT is adapted or modified;
    \item \textbf{Causal}: Whether the method employs a causal framework (e.g., activation patching);
    \item \textbf{Inference Time}: The time complexity of the given metric;
    \item \textbf{Polysemanticity}: Whether the metric can isolate decoupled features;
    \item \textbf{Training}: Whether additional training is required for the \textit{tested} model;
    \item \textbf{Per-Instance}: Whether the faithfulness evaluation is aggregate or per-instance.
\end{itemize}

For our proposed causal metric $\Delta p$ at a given layer $\ell$, an intervention can be applied, and its inference time is $O(1)$, as it only requires a constant number of forward passes, which is notably more efficient than most prior work. Moreover, it is training-free when SAEs are available, can mitigate polysemanticity, and can be used to measure the CoT faithfulness of each individual instance.

\begin{table*}[h]
\caption{CoT faithfulness evaluation metric comparison. Intervention: Whether the method perturbs the CoT or the input. $n$: number of input tokens; $L$: number of CoT steps; $m$: the number of traced tokens per step; $P$: resamples per step.}
\label{tab:comparison}

\centering
\resizebox{\textwidth}{!}{

\begin{tabular}{cccccccc}
\toprule[1.5pt]
\textbf{Method} & \textbf{Granularity} & \textbf{Intervention} & \textbf{Causal} & \textbf{Inference Time} & \textbf{Polysemanticity} & \textbf{Training} & \textbf{Per-Instance} \\
\midrule
Bias Feature \citep{turpin2023language} & CoT  & \Checkmark & \XSolidBrush & $O(1)$ & \XSolidBrush & \XSolidBrush & \Checkmark \\

CoT Corruption \citep{lanham2023measuring} & CoT  & \Checkmark & \XSolidBrush & $O(1)$ & \XSolidBrush & \XSolidBrush & \Checkmark\\

NLDD \citep{ye2026mechanistic} & Step  & \Checkmark & \XSolidBrush & $O(L)$ & \XSolidBrush & \XSolidBrush & \Checkmark \\

FUR \citep{tutek-etal-2025-measuring} & Step & \Checkmark & \XSolidBrush & $O(L)$ & \XSolidBrush & \Checkmark & \Checkmark\\

Thought Branches \citep{macar2026thought} & Step & \Checkmark  & \Checkmark & $O(L \cdot R)$ & \XSolidBrush & \XSolidBrush & \Checkmark \\

AttriCoT \citep{wei2026localcausalattributionchainofthought} & Step & \Checkmark  & \Checkmark & $O(L)$ & \XSolidBrush & \XSolidBrush & \Checkmark\\

CIE-Scorer \citep{shen2026detectingunfaithfulchainofthoughtcircuitguided} & Circuit & \Checkmark & \Checkmark &  $O(L \cdot m)$ & \XSolidBrush & \Checkmark & \Checkmark \\

CC-SHAP \citep{parcalabescu-frank-2024-measuring} & Input Token & \XSolidBrush & \XSolidBrush & $O(2^n)$ & \XSolidBrush & \XSolidBrush & \Checkmark\\

CC-LIME \citep{admoni-etal-2026-aligning} & Input Token  & \XSolidBrush & \XSolidBrush & $O(n)$ & \XSolidBrush & \XSolidBrush & \Checkmark\\

SAE Patching \citep{chen2026does} & SAE Concept & \Checkmark & \Checkmark & $O(1)$ & \Checkmark & \Checkmark & \XSolidBrush\\

\midrule

\textbf{Our} & SAE Concept & \Checkmark & \Checkmark & $O(1)$ & \Checkmark & \XSolidBrush & \Checkmark\\

\toprule[1.5pt]
\end{tabular}
}
\end{table*}

\subsection{Separate vs. Shared SAEs}
\begin{figure*}[!h]
\centering
\resizebox{\textwidth}{!}{
\begin{minipage}{\textwidth}
\includegraphics[width=\textwidth]{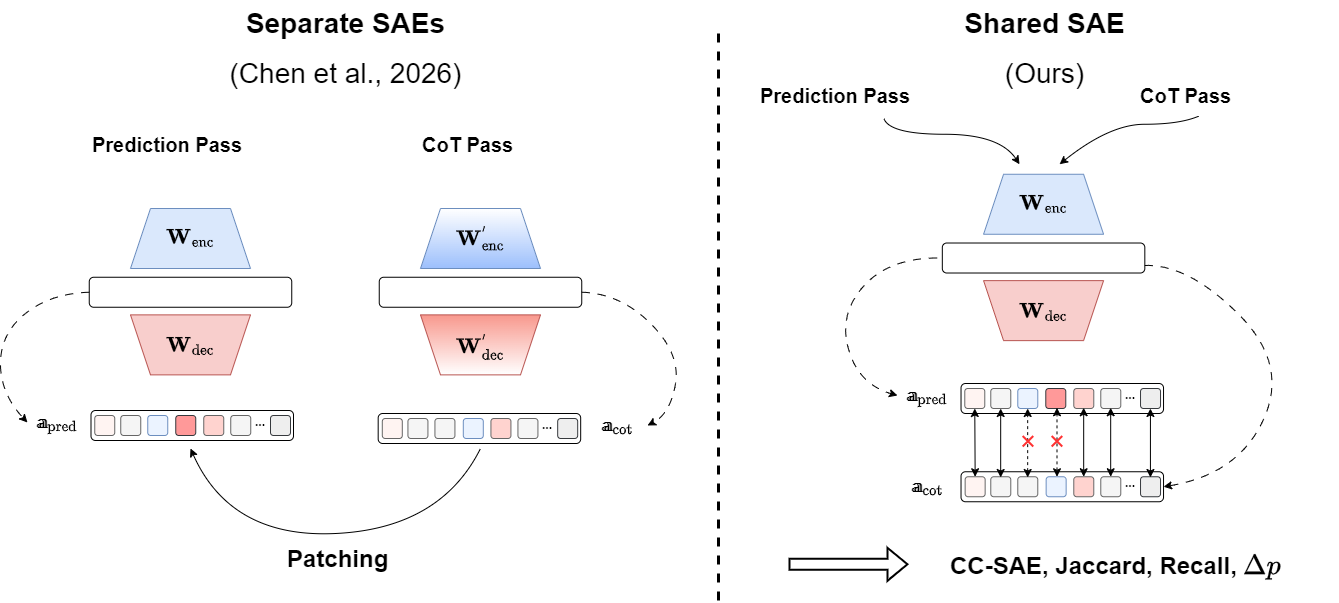}
\end{minipage}
}
\caption{Comparison between \citet{chen2026does} and our metrics.}
\label{fig:compare}
\end{figure*}

Figure~\ref{fig:compare} illustrates the differences between \citet{chen2026does} and our metrics. First, \citet{chen2026does} train and deploy two distinct SAEs on the prediction and CoT passes, respectively, which makes the two SAEs' activations not directly comparable. In contrast, we use a single SAE for both passes, which enables the measurement of concept alignment and causal grounding. Furthermore, they patch $a_\text{cot}$ into the prediction pass and measure the changes in answer log-probability. In comparison, we ablate the shared SAE features that are proven to causally drive the answer (\S\ref{subsec:sanity_check}). Additionally, we have validated that patching only the shared SAE features is sufficient to largely reproduce the original answer (Appendix~\ref{app:sufficiency}).

\subsection{Venn Diagram of SAE Features}

\begin{figure*}[!h]
\centering
\resizebox{0.5\textwidth}{!}{
\begin{minipage}{\textwidth}
\includegraphics[width=\textwidth]{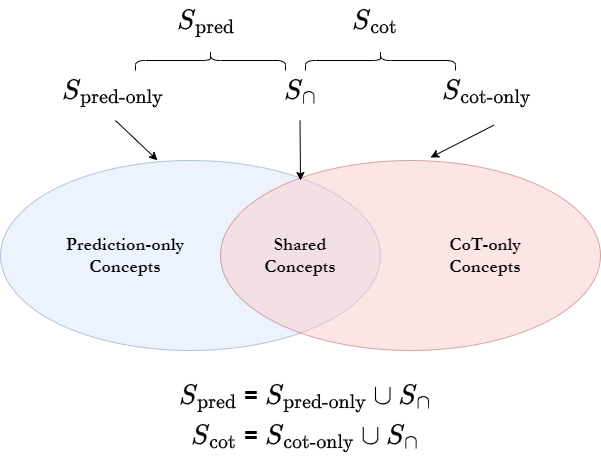}
\end{minipage}
}
\caption{Venn diagram of SAE features.}
\label{fig:venn}
\end{figure*}
Figure~\ref{fig:venn} displays an overview of the SAE features. In general, the features can be categorized into three classes: $S_\text{pred-only}$, $S_\text{cot-only}$, and $S_\cap$, where $S_\text{pred} = S_\text{pred-only} \cup S_\cap$ and $S_\text{cot} = S_\text{cot-only} \cup S_\cap$. This is enabled by employing a shared SAE, which allows us to assess the shared concepts that drive the model toward answers faithfully across the prediction and CoT passes. Appendix~\ref{app:concept_composition} further presents the SAE concept composition analysis.

\section{Experimental Detail}

\subsection{Models \& SAEs}
Table~\ref{tab:hidden_dim} illustrates the hidden state dimension, model link, and residual-stream SAE suite of each model. 

\begin{table}[htbp]
\centering
\caption{Hidden state dimension, model link, and residual-stream SAE suite of each model.}
\label{tab:hidden_dim}
\resizebox{\textwidth}{!}{
\begin{tabular}{lccc}
\toprule
\textbf{Model} & $d_\text{model}$ & \textbf{Link} & \textbf{SAE Scope} \\
\midrule
\lm{Llama-3.1-8B}~\citep{grattafiori2024llama3herdmodels} & 4096 & \href{https://huggingface.co/meta-llama/Llama-3.1-8B}{\lm{meta-llama/Llama-3.1-8B}} & \href{https://huggingface.co/OpenMOSS-Team/Llama-Scope}{Llama-Scope}~\citep{he2024llamascopeextractingmillions} \\
\lm{Gemma-2-2B}~\citep{gemmateam2024gemma2improvingopen} & 2304 & \href{https://huggingface.co/google/gemma-2-2b}{\lm{google/gemma-2-2b}} & \href{https://huggingface.co/google/gemma-scope-2b-pt-res}{Gemma-Scope}~\citep{lieberum-etal-2024-gemma} \\
\lm{Gemma-2-9B}~\citep{gemmateam2024gemma2improvingopen} & 3584 & \href{https://huggingface.co/google/gemma-2-9b}{\lm{google/gemma-2-9b}} & \href{https://huggingface.co/google/gemma-scope-9b-pt-res}{Gemma-Scope}~\citep{lieberum-etal-2024-gemma} \\
\lm{Qwen3-1.7B}~\citep{yang2025qwen3technicalreport} & 2048 & \href{https://huggingface.co/Qwen/Qwen3-1.7B}{\lm{Qwen/Qwen3-1.7B}} & \href{https://huggingface.co/Qwen/SAE-Res-Qwen3-1.7B-Base-W32K-L0_100}{Qwen-Scope}~\citep{deng2026qwenscopeturningsparsefeatures} \\
\lm{Qwen3-8B}~\citep{yang2025qwen3technicalreport} & 4096 & \href{https://huggingface.co/Qwen/Qwen3-8B}{\lm{Qwen/Qwen3-8B}} & \href{https://huggingface.co/Qwen/SAE-Res-Qwen3-8B-Base-W64K-L0_100}{Qwen-Scope}~\citep{deng2026qwenscopeturningsparsefeatures} \\
\bottomrule
\end{tabular}
}
\end{table}

We note that there is one shared SAE \textit{between passes at a given layer}, rather than a single SAE shared across layers.\footnote{Since the residual stream's distribution and semantic abstraction level change qualitatively with layer depth, a single shared SAE cannot converge to a coherent sparse mapping.} We extract activations at the output of each transformer layer (post-attention, post-MLP, post-residual connection), corresponding to \texttt{resid\_post} in TransformerLens\footnote{\url{https://github.com/TransformerLensOrg/TransformerLens}} terminology. Formally, for layer $\ell$:
\begin{equation}
    \mathbf{r}^{(\ell)} = \text{LayerNorm}\bigl(\mathbf{h}^{(\ell-1)} + \text{Attn}^{(\ell)} + \text{MLP}^{(\ell)}\bigr)
\end{equation}

\subsection{Decoding Configuration}
We evaluate CoT faithfulness using greedy decoding (\texttt{temperature}$=0$, \texttt{max\_new\_tokens}$=2048$) for CoT generation. The reasoning chain is extracted by splitting at the ``Answer:'' delimiter. 

% ...


\subsection{Inference Time}
All experiments were conducted on a single NVIDIA RTX A6000 GPU. Under this setup, the average computational cost for a single $\Delta p$ was approximately 15 seconds.


\section{CoT Faithfulness Evaluation}
\label{app:cot_faithfulness_evaluation}
\begin{figure*}[!t]
\centering
\begin{minipage}{0.49\textwidth}
  \centering
  \includegraphics[width=\linewidth]{figures/summary_relative_depth_logiqa.png}
  \subcaption{\data{LogiQA}}
  \label{fig:summary_logiqa}
\end{minipage}
\hfill
\begin{minipage}{0.49\textwidth}
  \centering
  \includegraphics[width=\linewidth]{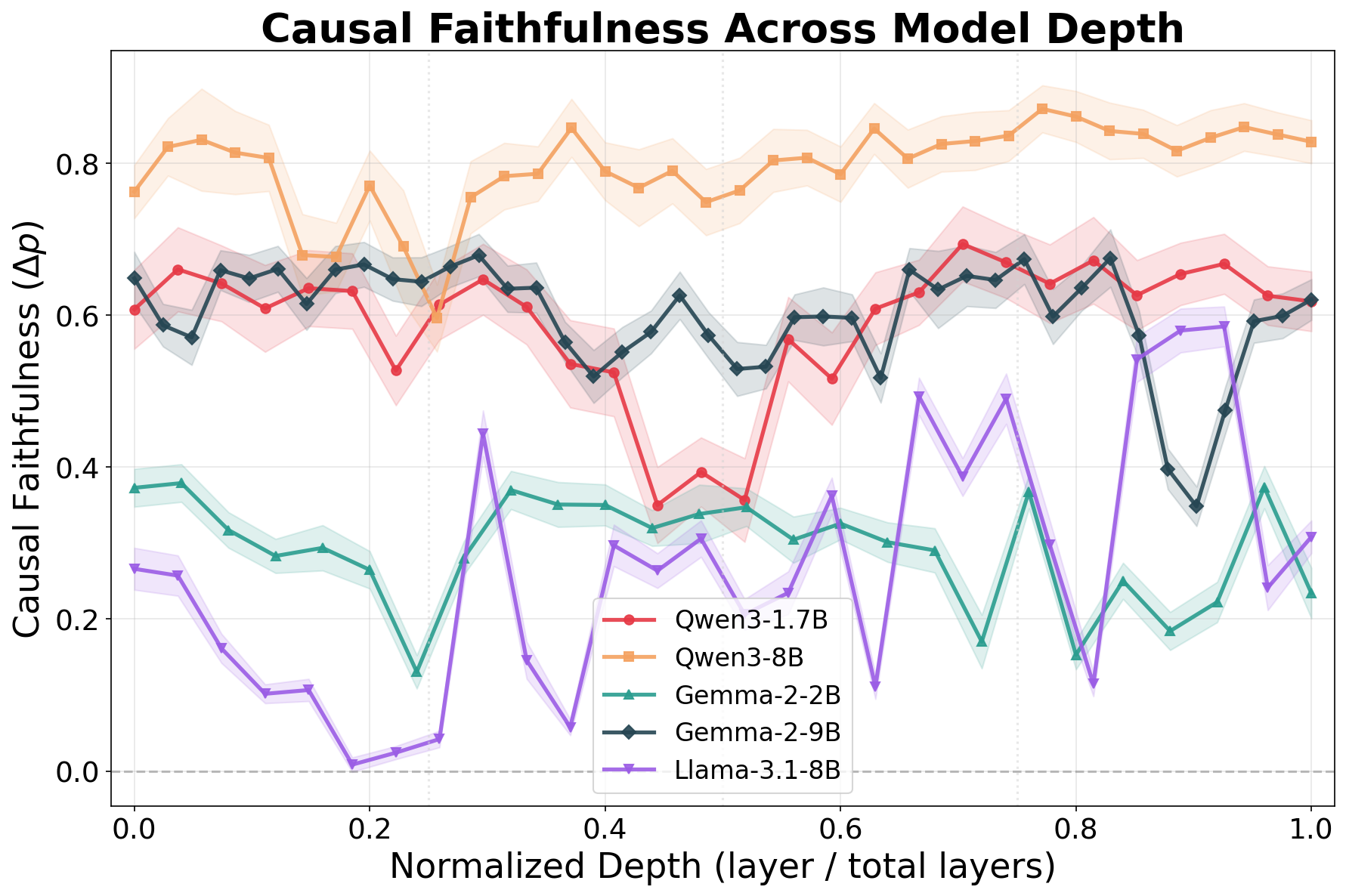}
  \subcaption{\data{OpenbookQA}}
  \label{fig:summary_openbookqa}
\end{minipage}

\vspace{0.3cm}

\begin{minipage}{0.49\textwidth}
  \centering
  \includegraphics[width=\linewidth]{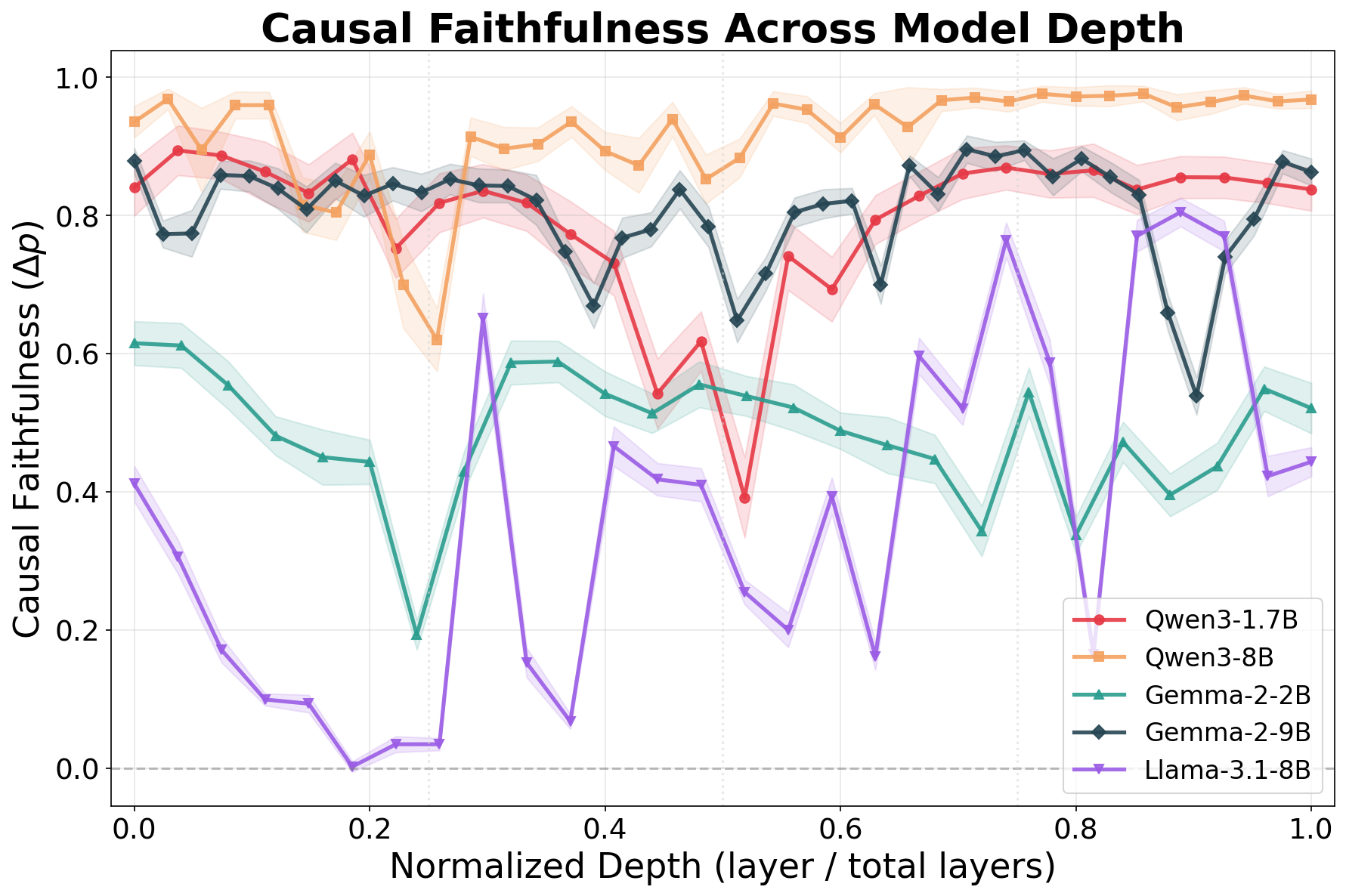}
  \subcaption{\data{ARC-Easy}}
  \label{fig:summary_arc_easy}
\end{minipage}
\hfill
\begin{minipage}{0.49\textwidth}
  \centering
  \includegraphics[width=\linewidth]{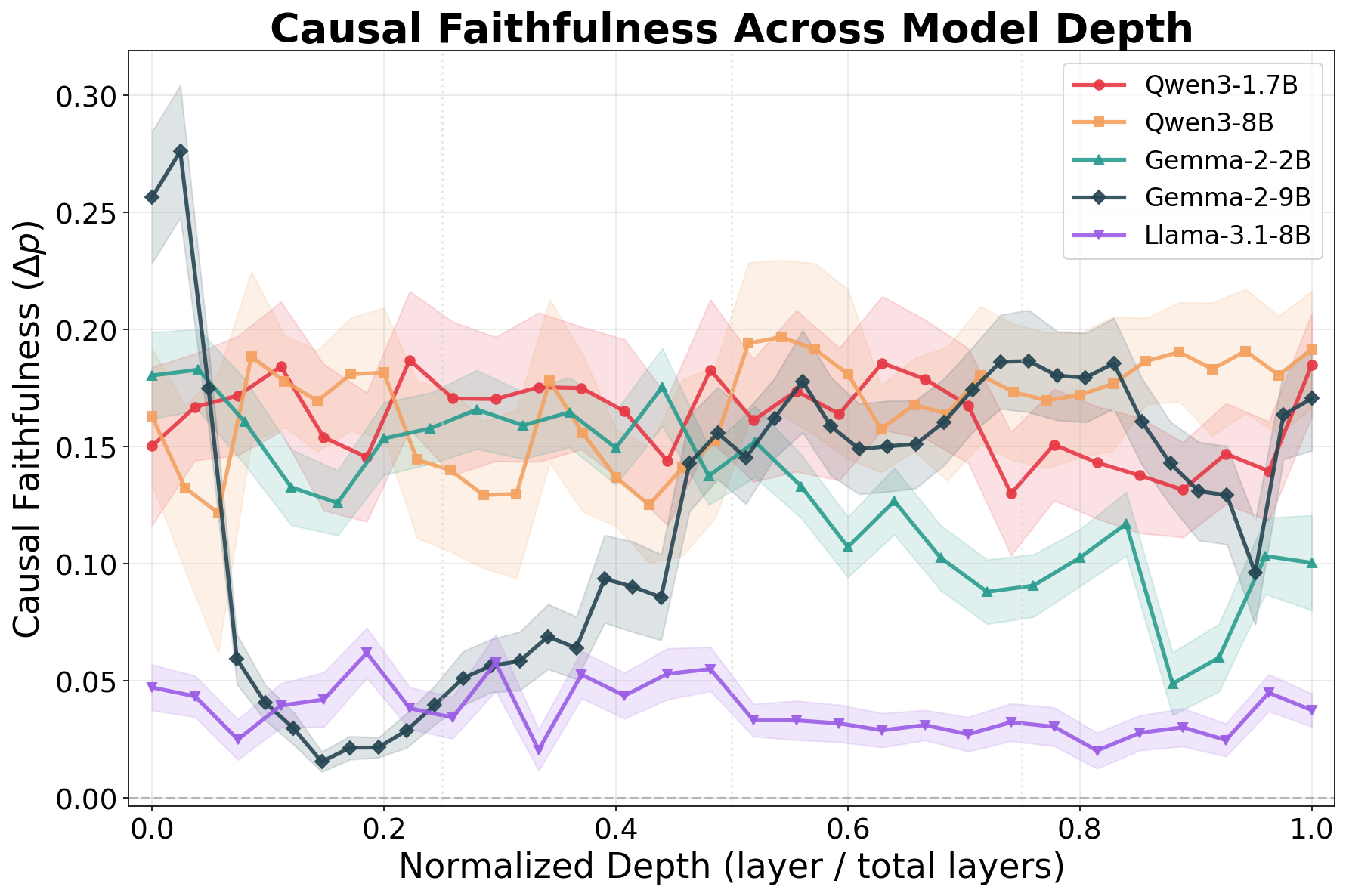}
  \subcaption{\data{GSM8K}}
  \label{fig:summary_gsm8k}
\end{minipage}
\caption{Causal faithfulness ($\Delta p$) variation across normalized model depth.}
\label{fig:summary}
\end{figure*}

\begin{figure*}[!t]
\centering
\begin{minipage}{0.49\textwidth}
  \centering
  \includegraphics[width=\linewidth]{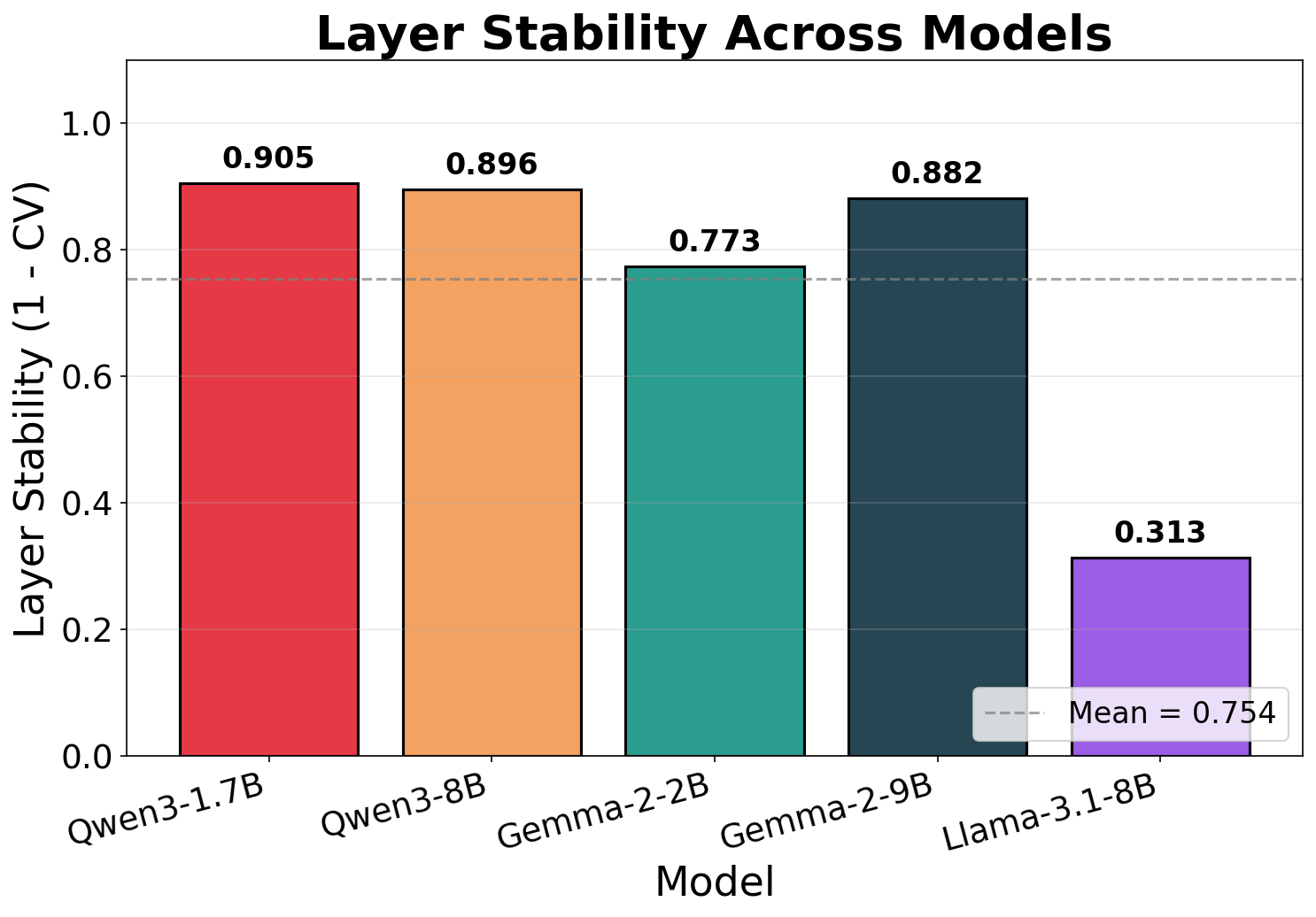}
  \subcaption{\data{LogiQA}}
  \label{fig:stability_logiqa}
\end{minipage}
\hfill
\begin{minipage}{0.49\textwidth}
  \centering
  \includegraphics[width=\linewidth]{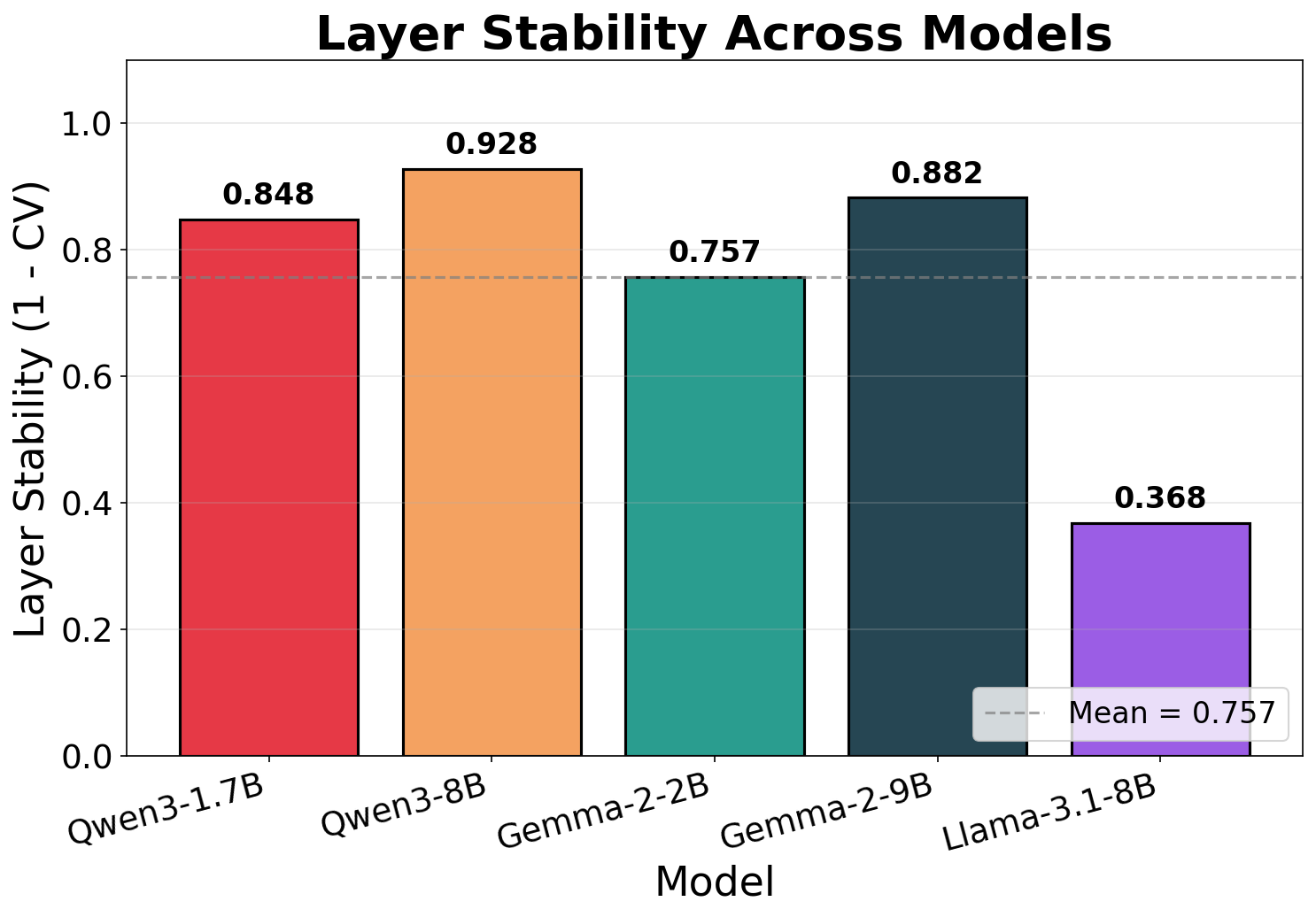}
  \subcaption{\data{OpenbookQA}}
  \label{fig:stability_openbookqa}
\end{minipage}

\vspace{0.3cm}

\begin{minipage}{0.49\textwidth}
  \centering
  \includegraphics[width=\linewidth]{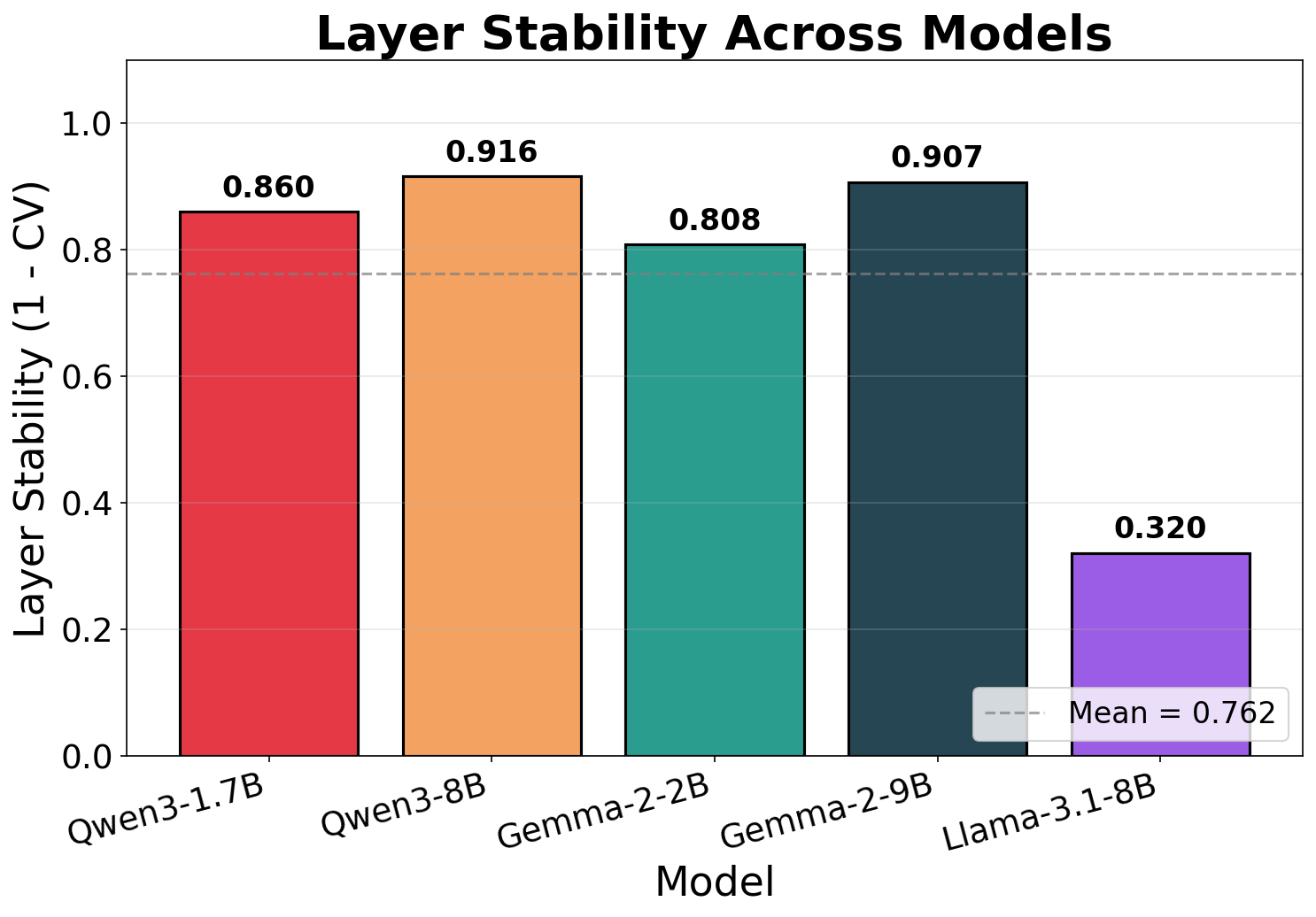}
  \subcaption{\data{ARC-Easy}}
  \label{fig:stability_arc_easy}
\end{minipage}
\hfill
\begin{minipage}{0.49\textwidth}
  \centering
  \includegraphics[width=\linewidth]{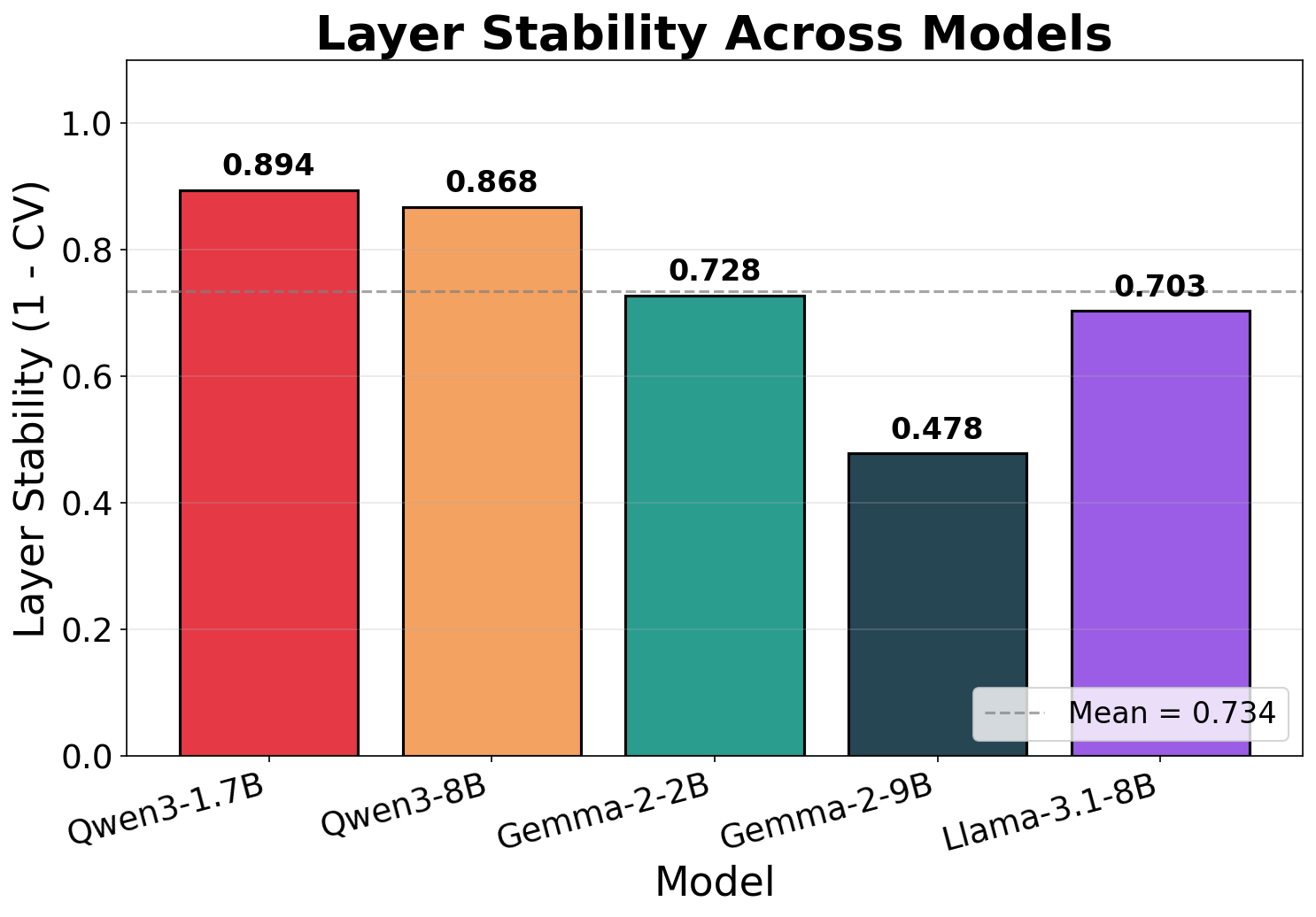}
  \subcaption{\data{GSM8K}}
  \label{fig:stability_gsm8k}
\end{minipage}
\caption{Layer stability across models.}
\label{fig:stability}
\end{figure*}

\begin{figure*}[t]
    \centering
    \resizebox{0.95\textwidth}{!}{%
    \begin{minipage}{\textwidth}
        \centering
        \begin{subfigure}[b]{0.32\textwidth}
            \includegraphics[width=\textwidth]{figures/faithfulness/logiqa/Qwen3-1.7B_causal_delta_p.png}
        \end{subfigure}
        \hfill
        \begin{subfigure}[b]{0.32\textwidth}
            \includegraphics[width=\textwidth]{figures/faithfulness/logiqa/Qwen3-1.7B_correlation_metrics.png}
        \end{subfigure}
        \hfill
        \begin{subfigure}[b]{0.32\textwidth}
            \includegraphics[width=\textwidth]{figures/faithfulness/logiqa/Qwen3-1.7B_corr_heatmap.png}
        \end{subfigure}

        \begin{subfigure}[b]{0.32\textwidth}
            \includegraphics[width=\textwidth]{figures/faithfulness/logiqa/Qwen3-8B_causal_delta_p.png}
        \end{subfigure}
        \hfill
        \begin{subfigure}[b]{0.32\textwidth}
            \includegraphics[width=\textwidth]{figures/faithfulness/logiqa/Qwen3-8B_correlation_metrics.png}
        \end{subfigure}
        \hfill
        \begin{subfigure}[b]{0.32\textwidth}
            \includegraphics[width=\textwidth]{figures/faithfulness/logiqa/Qwen3-8B_corr_heatmap.png}
        \end{subfigure}

        \begin{subfigure}[b]{0.32\textwidth}
            \includegraphics[width=\textwidth]{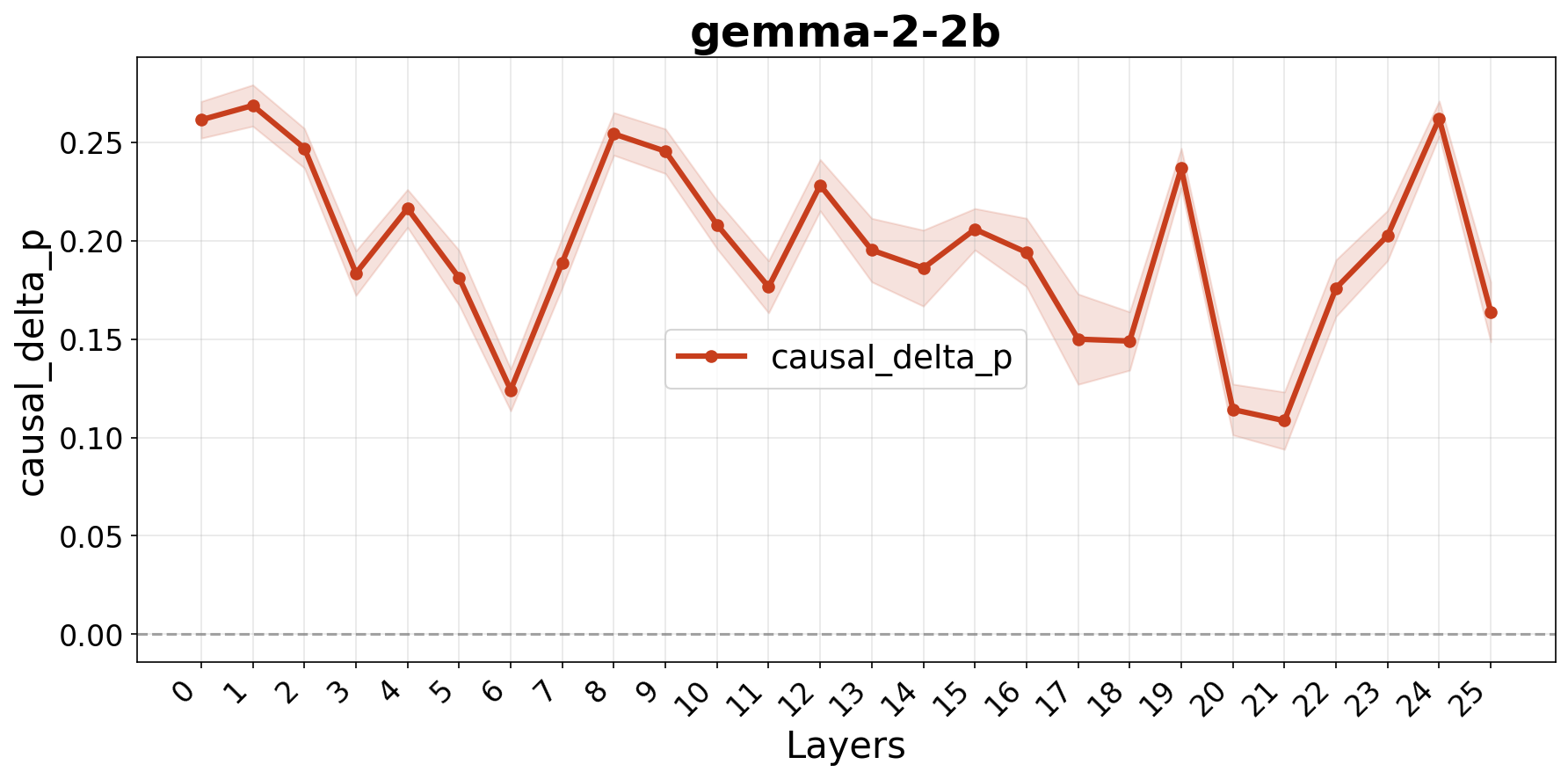}
        \end{subfigure}
        \hfill
        \begin{subfigure}[b]{0.32\textwidth}
            \includegraphics[width=\textwidth]{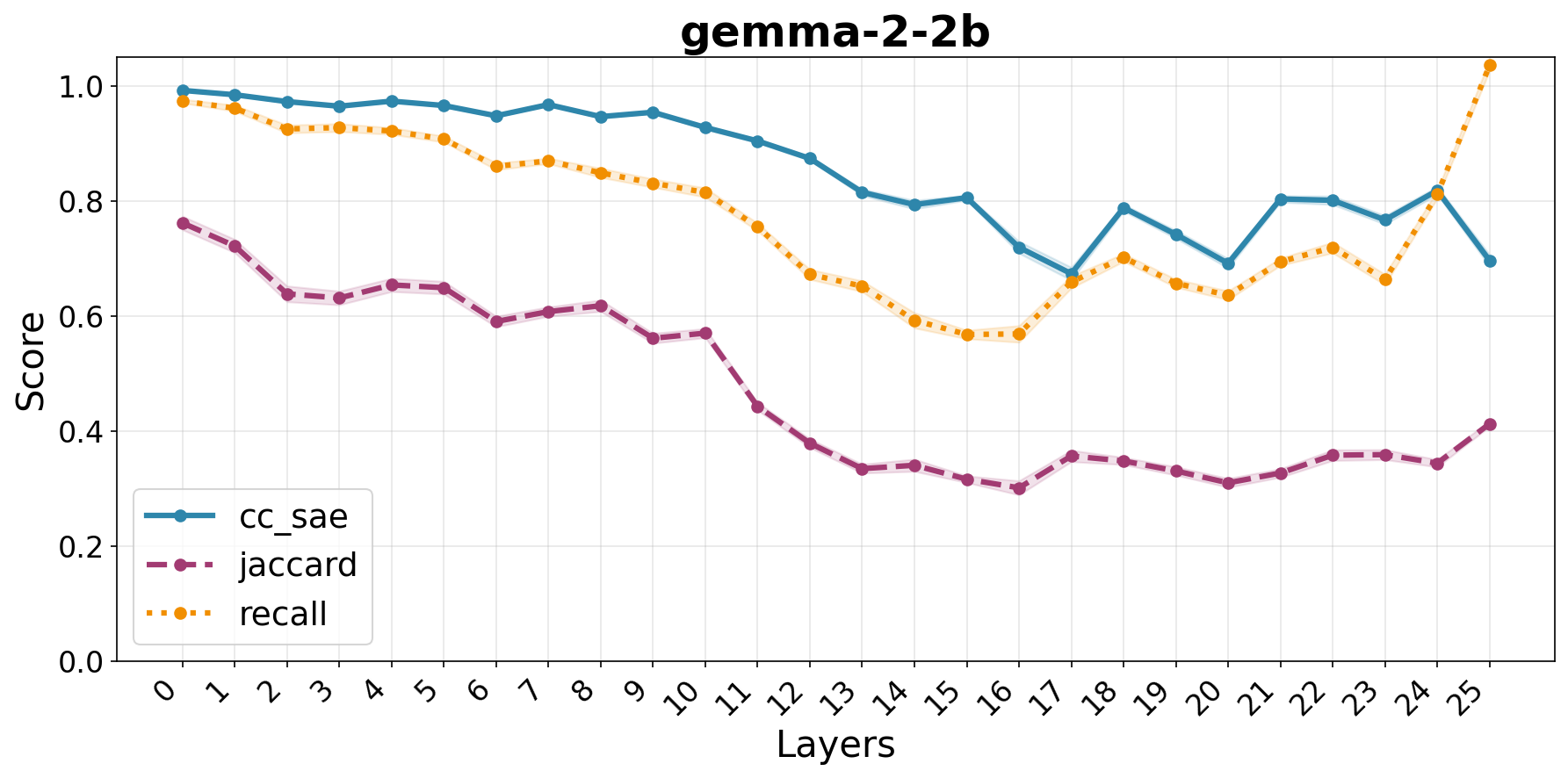}
        \end{subfigure}
        \hfill
        \begin{subfigure}[b]{0.32\textwidth}
            \includegraphics[width=\textwidth]{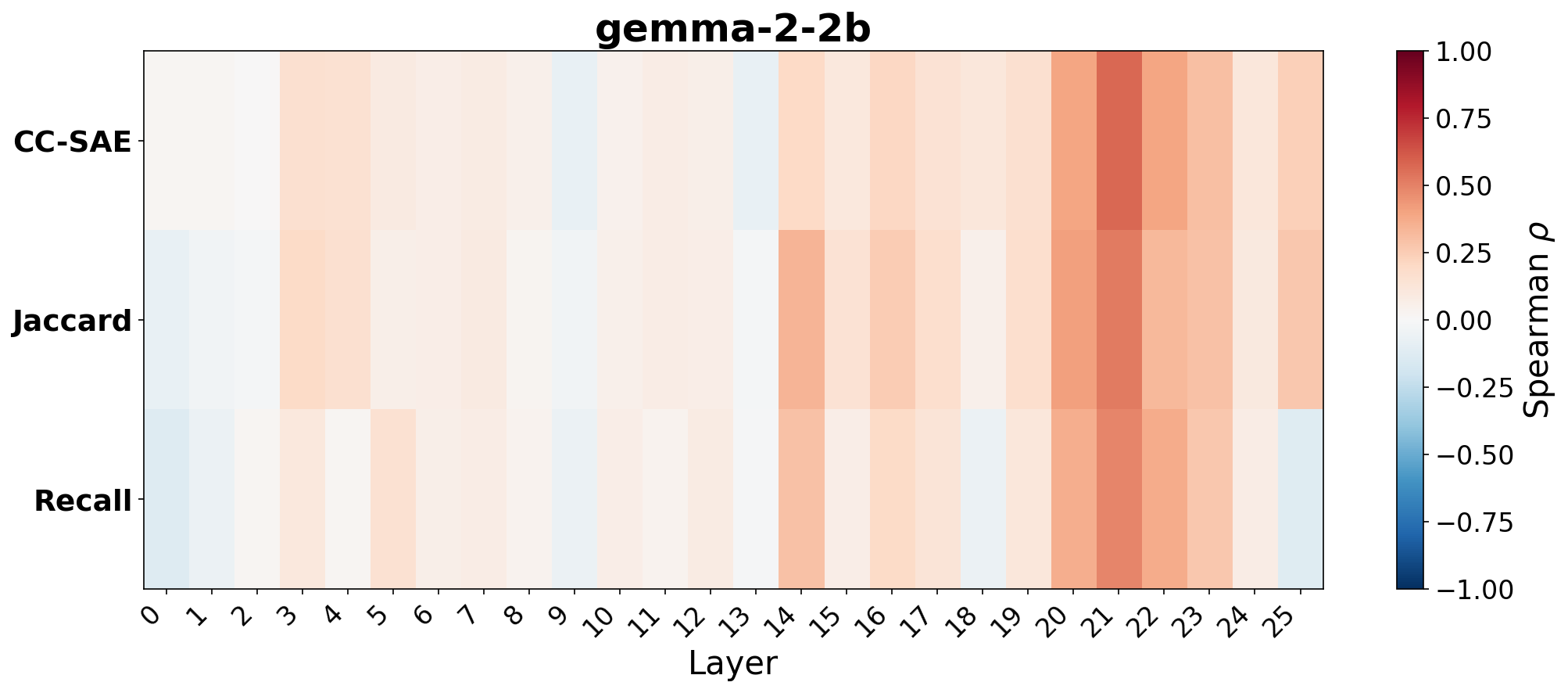}
        \end{subfigure}

        \begin{subfigure}[b]{0.32\textwidth}
            \includegraphics[width=\textwidth]{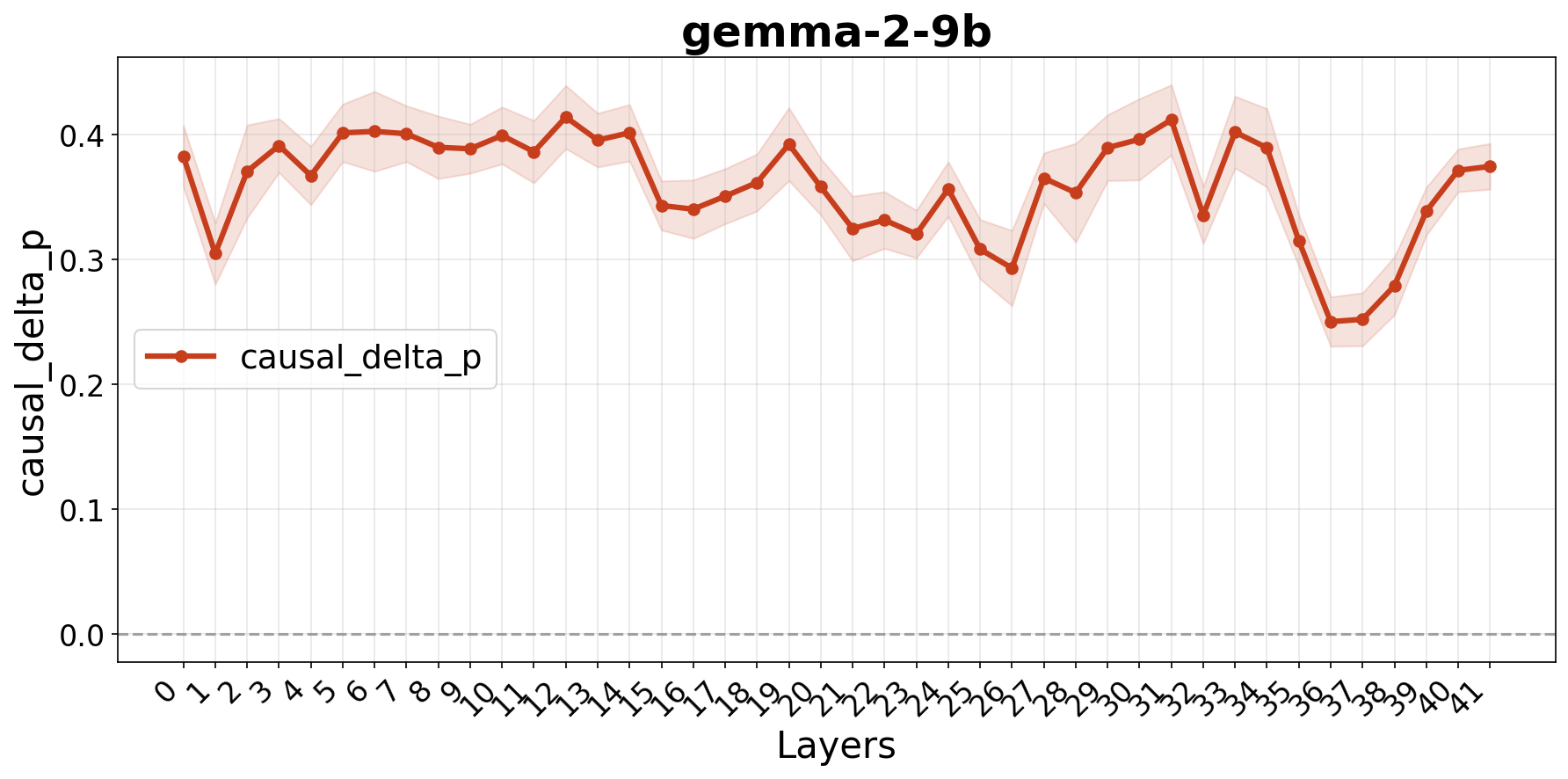}
        \end{subfigure}
        \hfill
        \begin{subfigure}[b]{0.32\textwidth}
            \includegraphics[width=\textwidth]{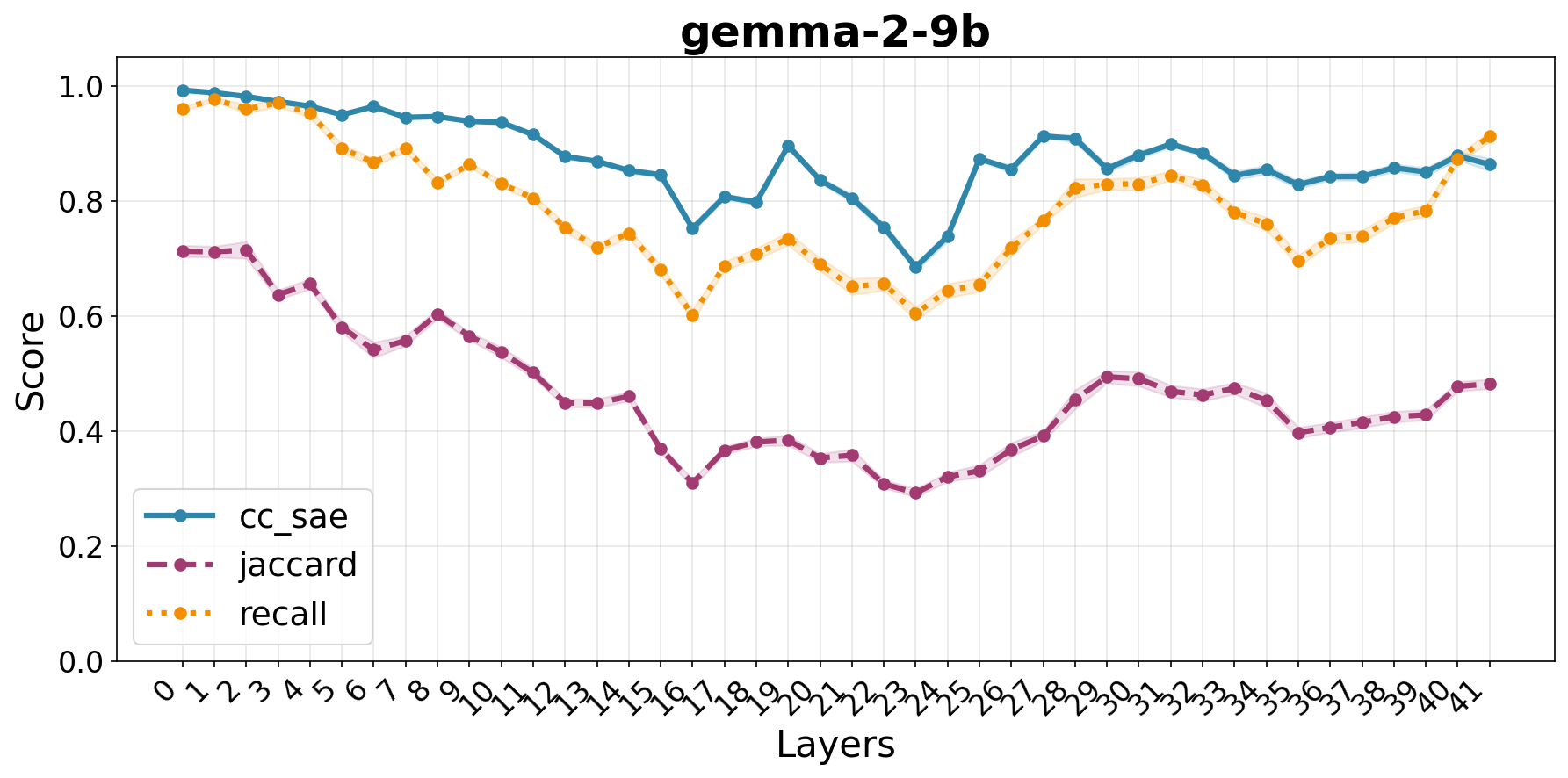}
        \end{subfigure}
        \hfill
        \begin{subfigure}[b]{0.32\textwidth}
            \includegraphics[width=\textwidth]{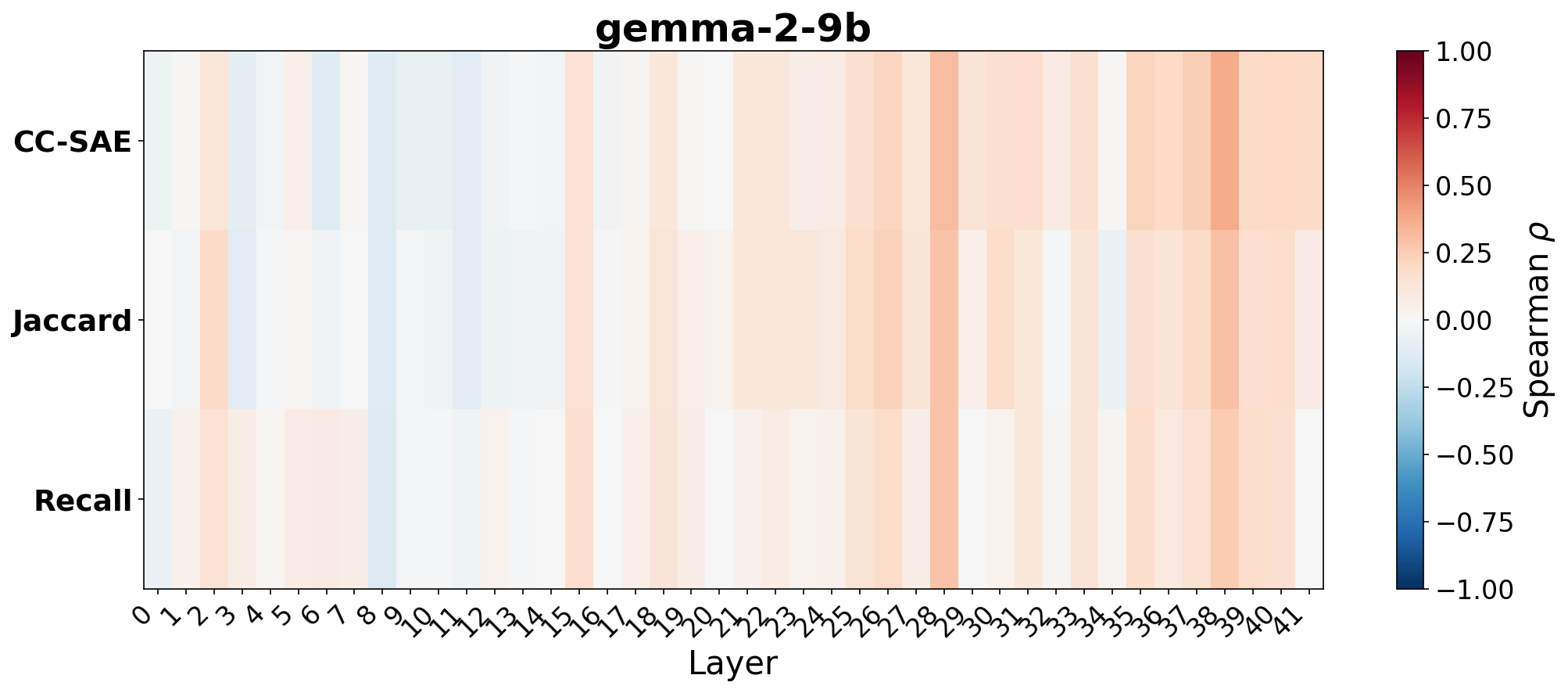}
        \end{subfigure}

        \begin{subfigure}[b]{0.32\textwidth}
            \includegraphics[width=\textwidth]{figures/faithfulness/logiqa/Meta-Llama-3.1-8B_causal_delta_p.png}
        \end{subfigure}
        \hfill
        \begin{subfigure}[b]{0.32\textwidth}
            \includegraphics[width=\textwidth]{figures/faithfulness/logiqa/Meta-Llama-3.1-8B_correlation_metrics.png}
        \end{subfigure}
        \hfill
        \begin{subfigure}[b]{0.32\textwidth}
            \includegraphics[width=\textwidth]{figures/faithfulness/logiqa/Meta-Llama-3.1-8B_corr_heatmap.png}
        \end{subfigure}
    \end{minipage}%
    }
    \caption{CoT faithfulness evaluation across layers for different models on \data{LogiQA}.
    \textbf{Left column:} Causal faithfulness ($\Delta p$) measures the probability drop when ablating shared features.
    \textbf{Middle column:} Correlational metrics (CC-SAE, Jaccard, Recall) between prediction and CoT activations.
    \textbf{Right column:} Spearman correlation heatmap between each correlational metric and $\Delta p$ across layers.
    Shaded regions indicate 95\% confidence intervals (left and middle columns). Additional results on the \textit{remaining datasets} are provided in Appendix~\ref{app:faithfulness_evaluation}.}
    \label{fig:ccsae-layers-app}
\end{figure*}
\begin{figure*}[t]
    \centering

    \begin{subfigure}[b]{0.32\textwidth}
        \includegraphics[width=\textwidth]{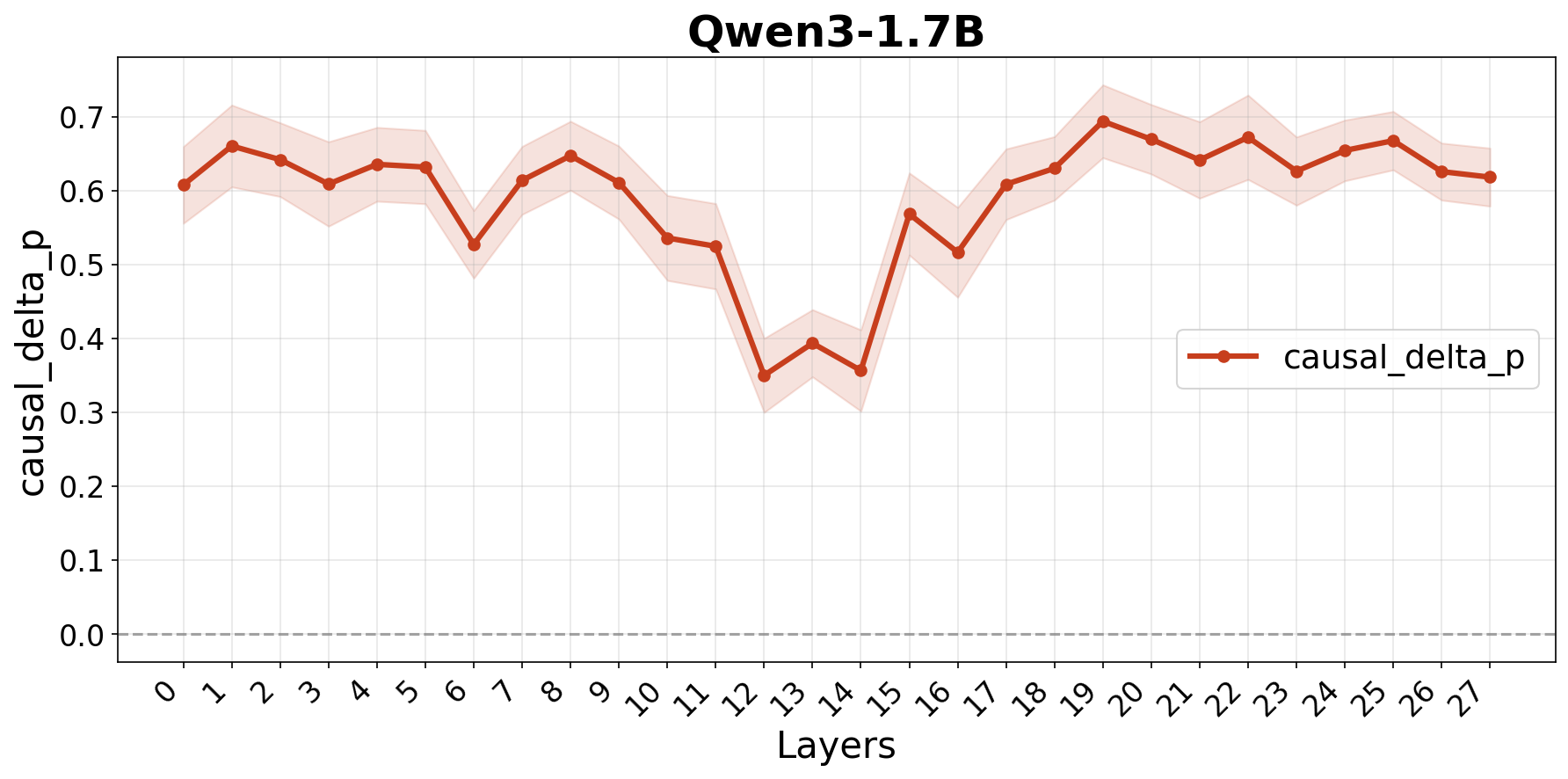}
    \end{subfigure}
    \hfill
    \begin{subfigure}[b]{0.32\textwidth}
        \includegraphics[width=\textwidth]{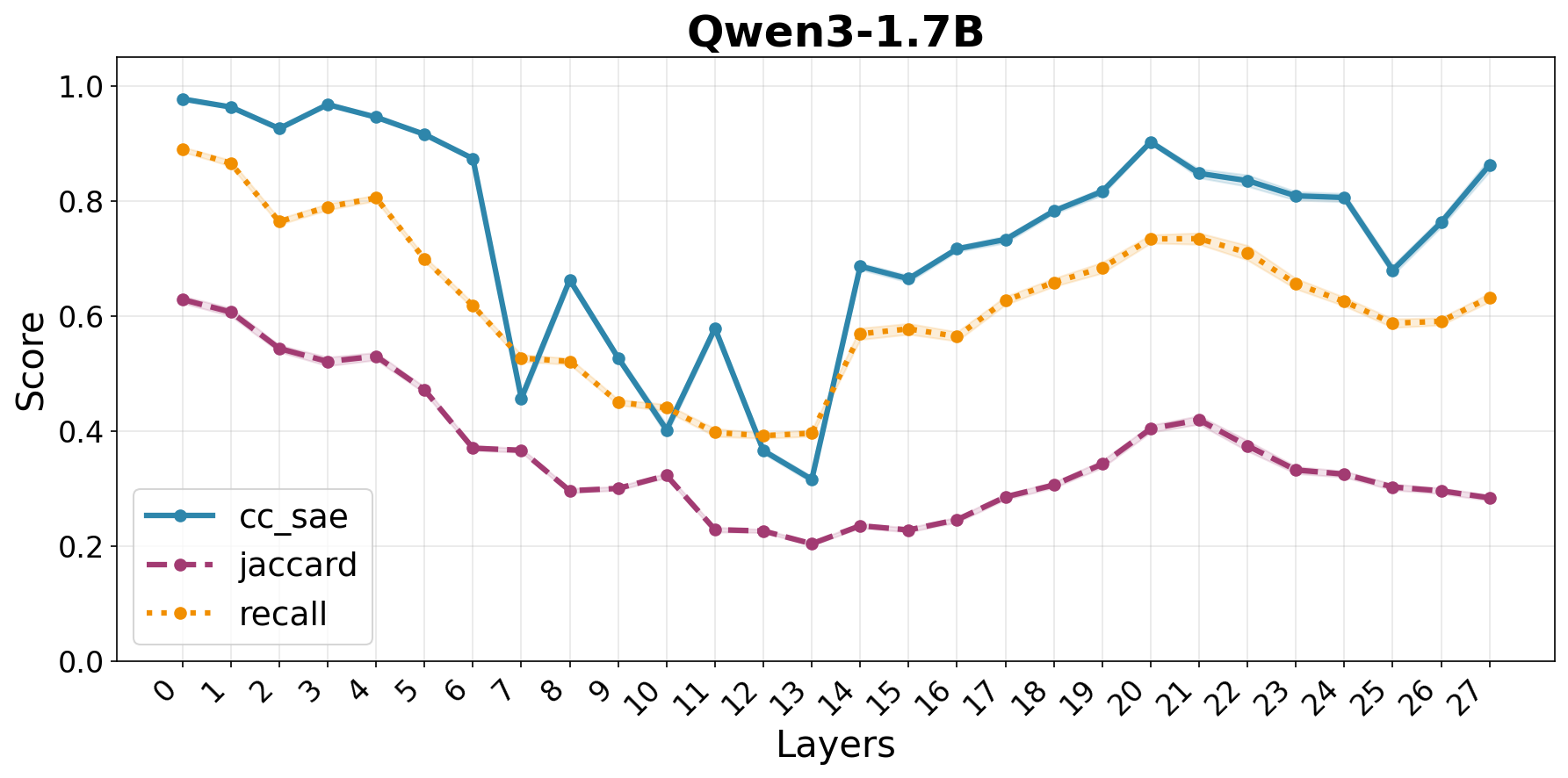}
    \end{subfigure}
    \hfill
    \begin{subfigure}[b]{0.32\textwidth}
        \includegraphics[width=\textwidth]{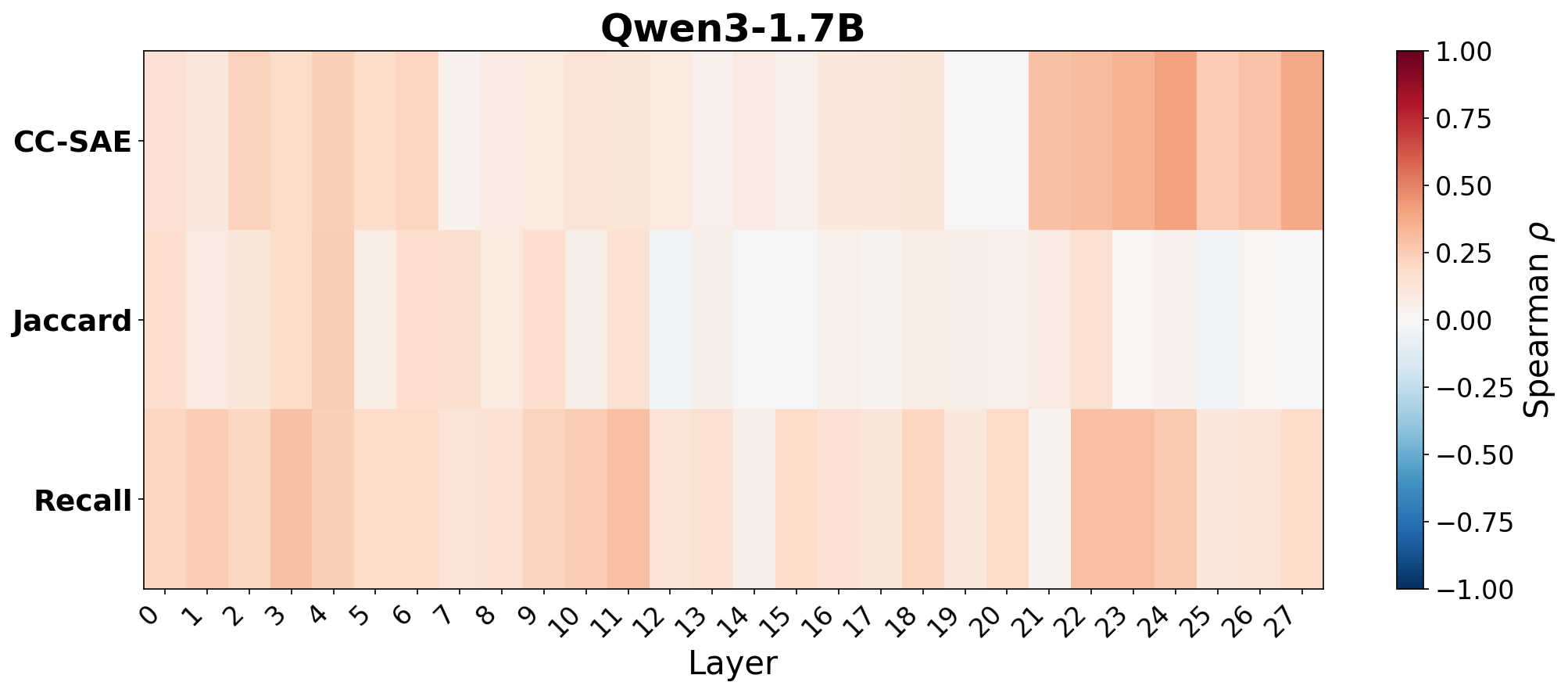}
    \end{subfigure}

    \vspace{0.5em}

    \begin{subfigure}[b]{0.32\textwidth}
        \includegraphics[width=\textwidth]{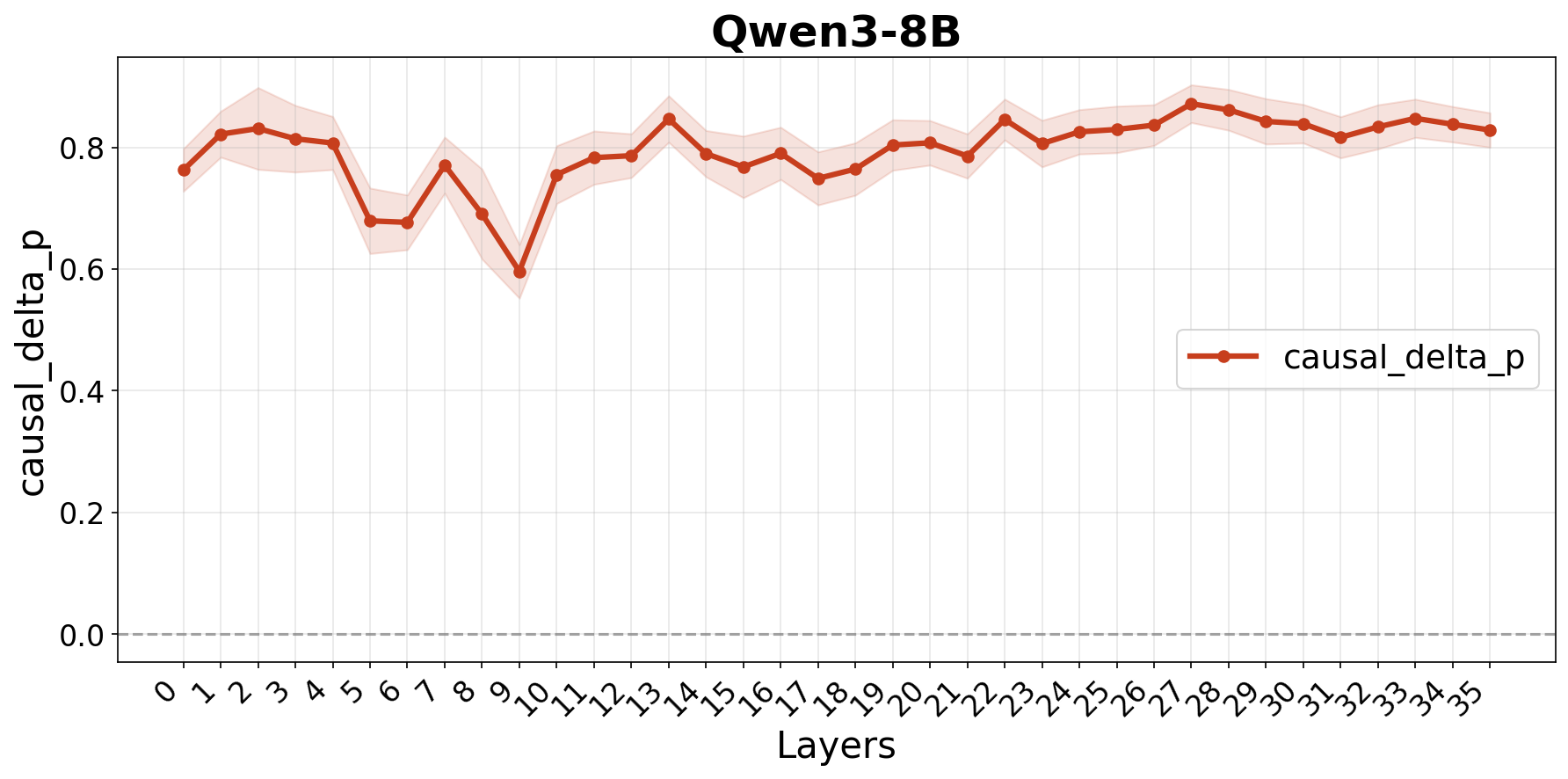}
    \end{subfigure}
    \hfill
    \begin{subfigure}[b]{0.32\textwidth}
        \includegraphics[width=\textwidth]{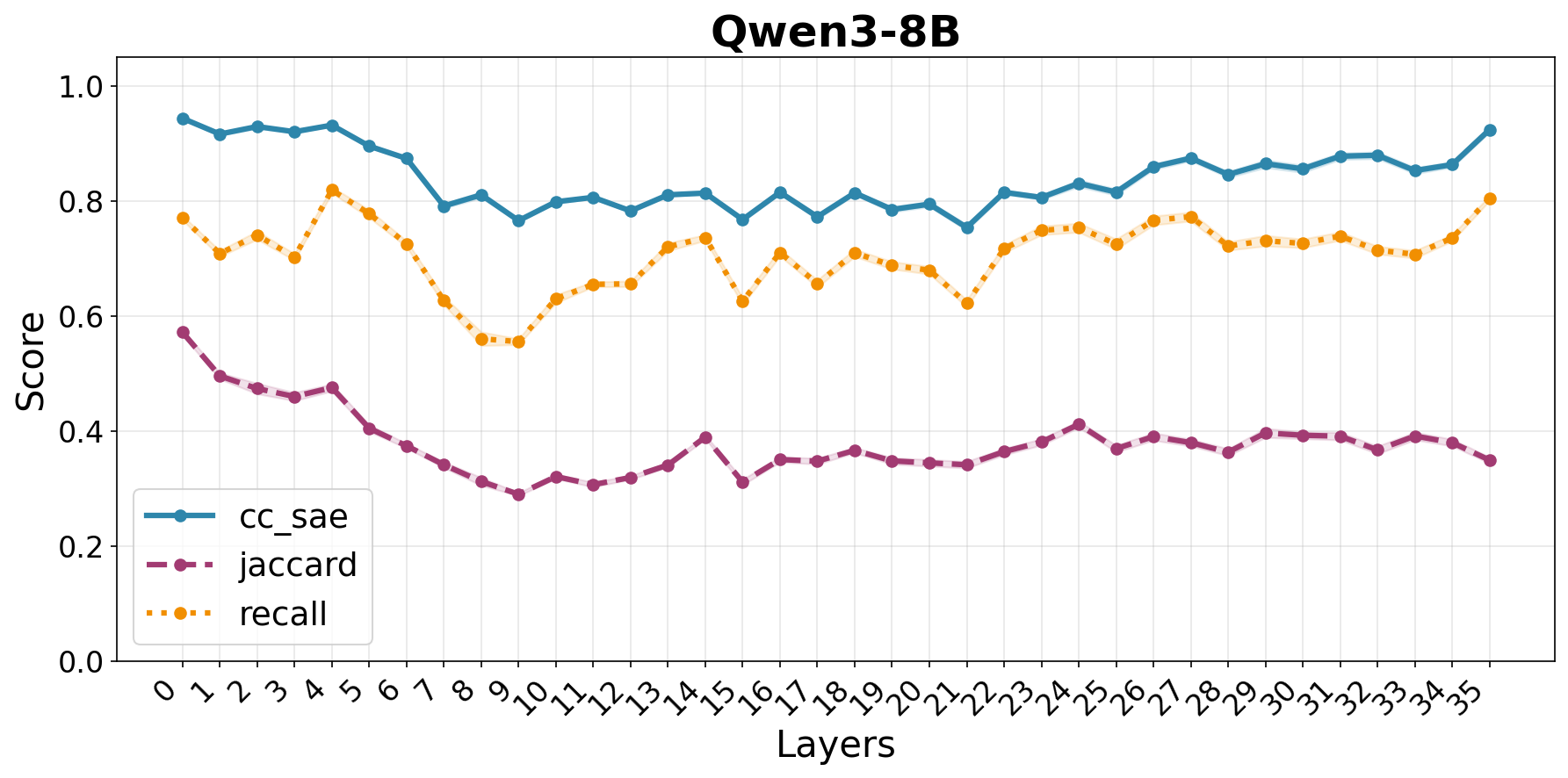}
    \end{subfigure}
    \hfill
    \begin{subfigure}[b]{0.32\textwidth}
        \includegraphics[width=\textwidth]{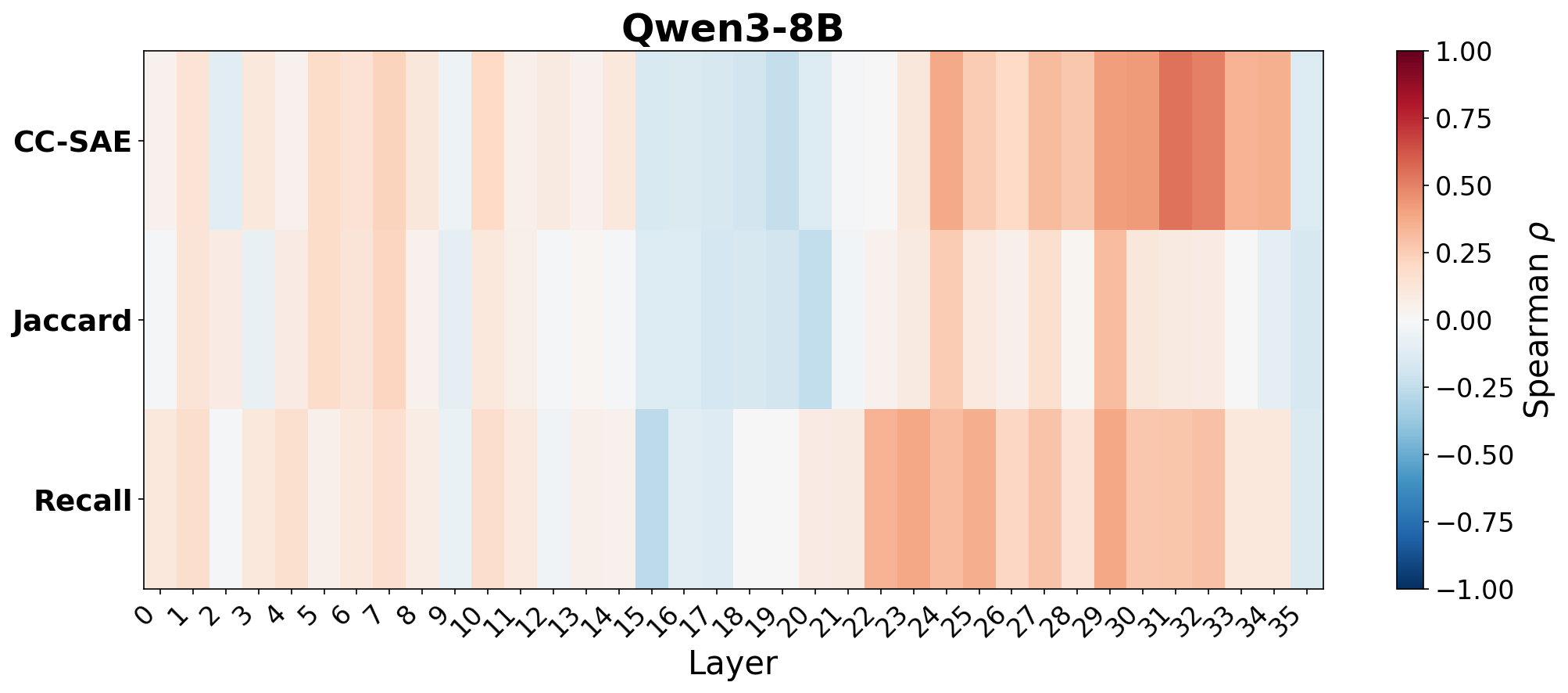}
    \end{subfigure}

    \vspace{0.5em}

    \begin{subfigure}[b]{0.32\textwidth}
        \includegraphics[width=\textwidth]{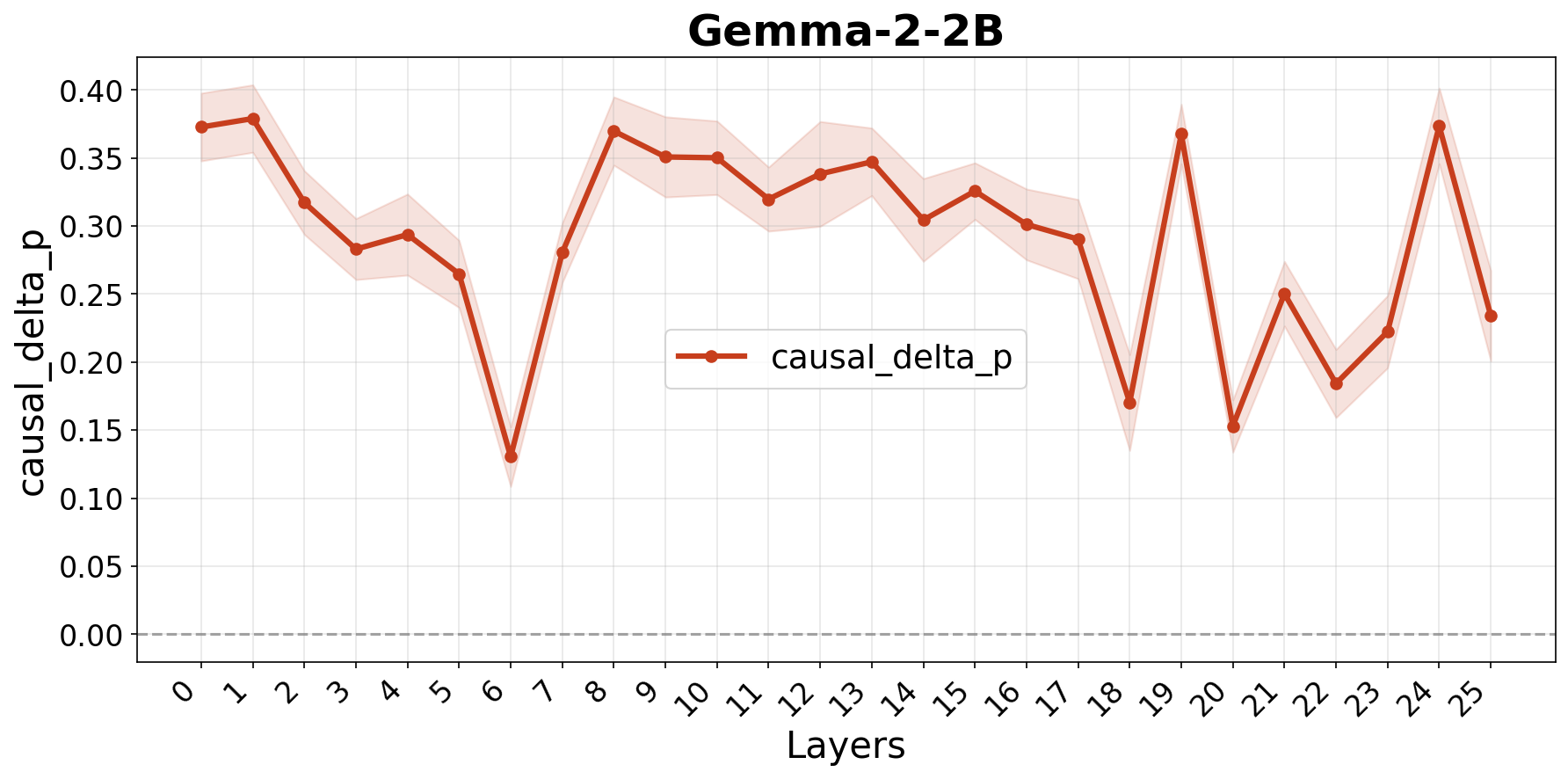}
    \end{subfigure}
    \hfill
    \begin{subfigure}[b]{0.32\textwidth}
        \includegraphics[width=\textwidth]{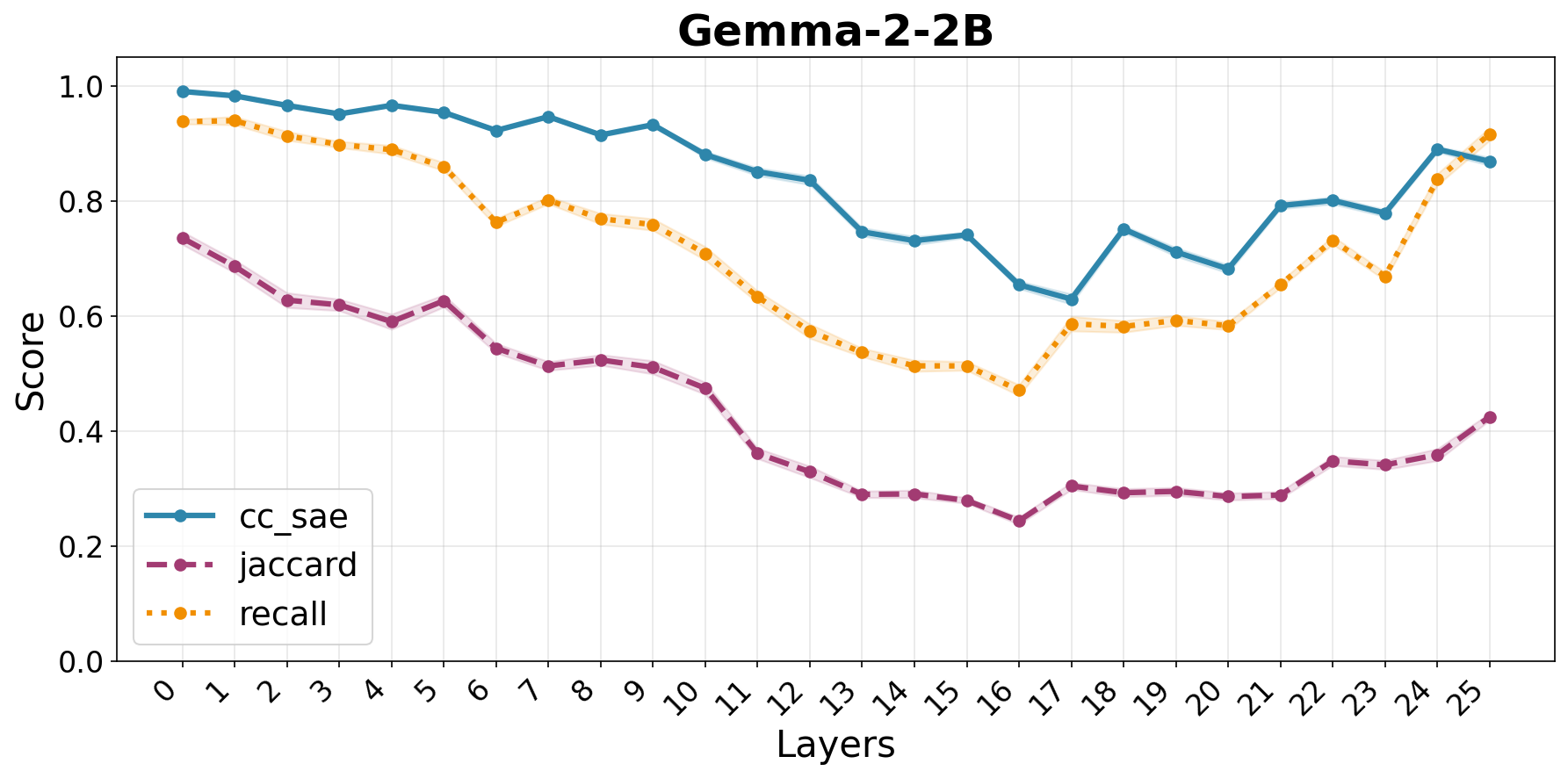}
    \end{subfigure}
    \hfill
    \begin{subfigure}[b]{0.32\textwidth}
        \includegraphics[width=\textwidth]{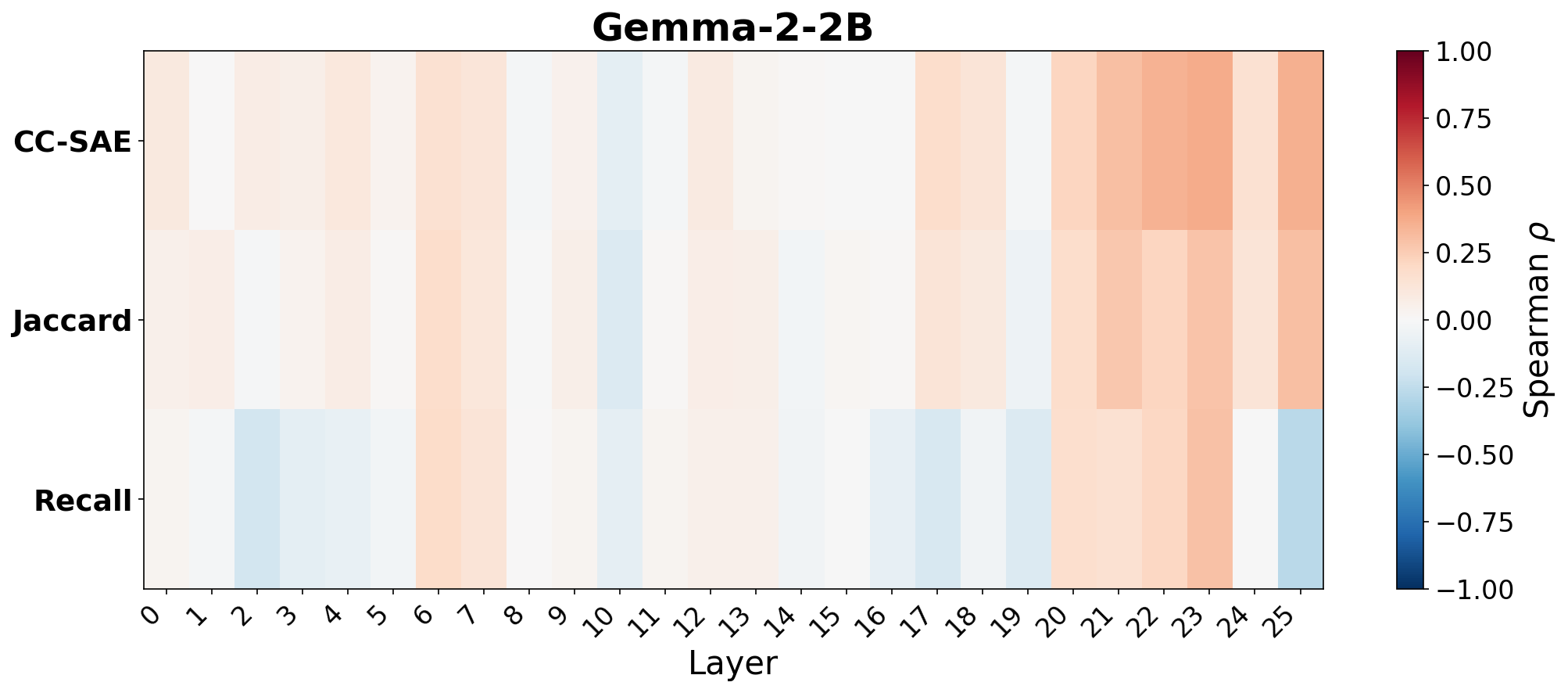}
    \end{subfigure}

    \vspace{0.5em}

    \begin{subfigure}[b]{0.32\textwidth}
        \includegraphics[width=\textwidth]{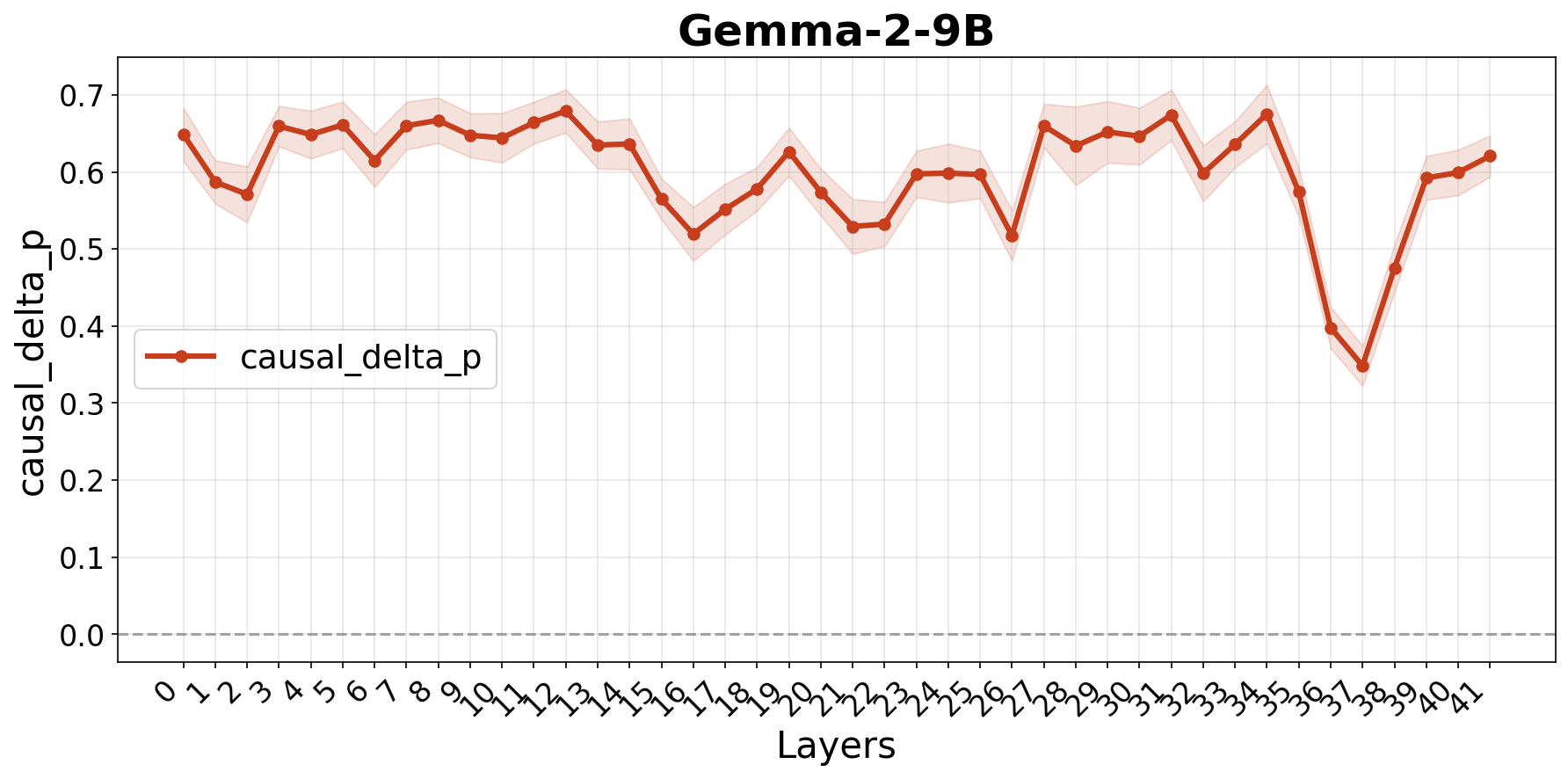}
    \end{subfigure}
    \hfill
    \begin{subfigure}[b]{0.32\textwidth}
        \includegraphics[width=\textwidth]{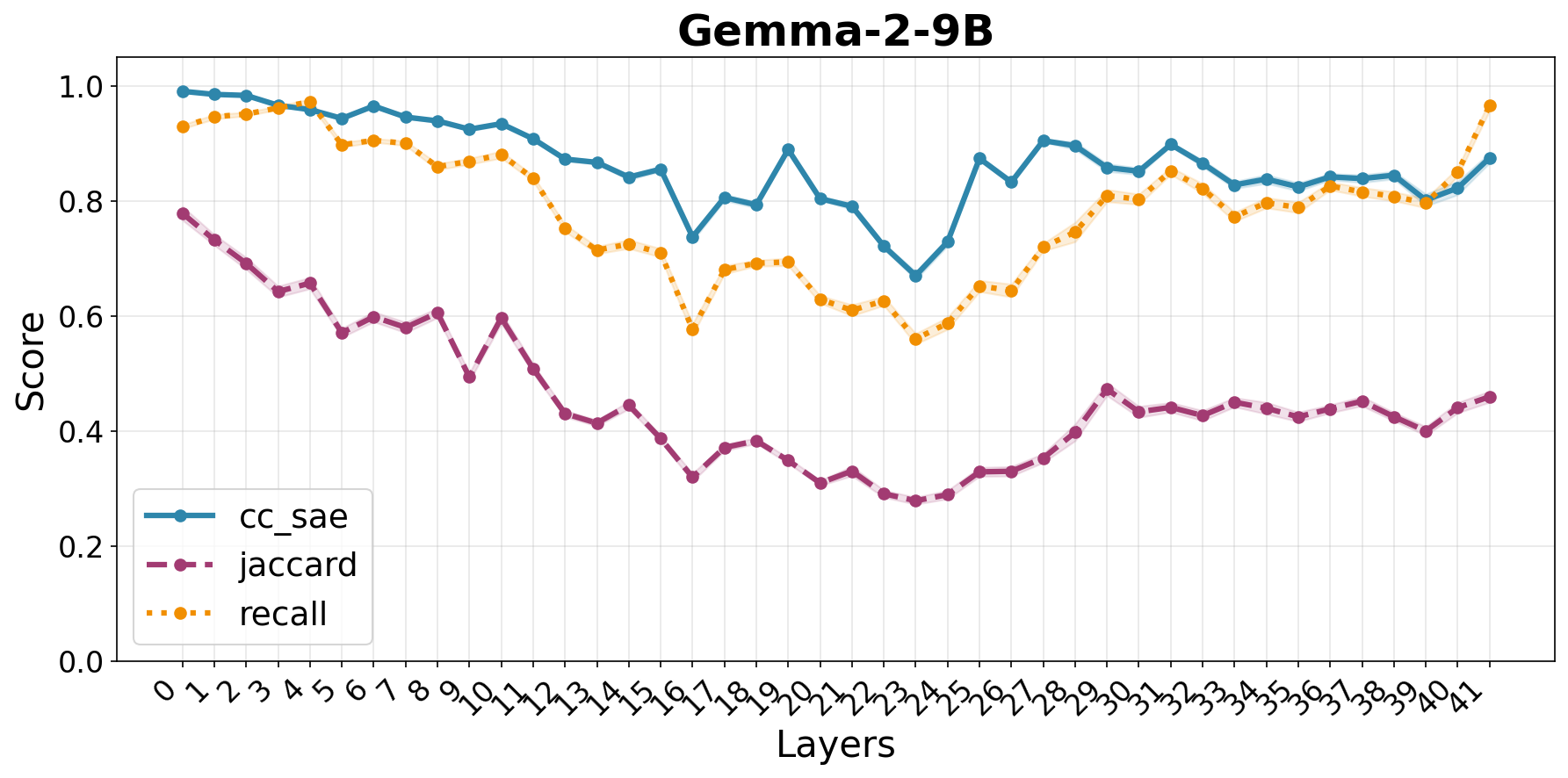}
    \end{subfigure}
    \hfill
    \begin{subfigure}[b]{0.32\textwidth}
        \includegraphics[width=\textwidth]{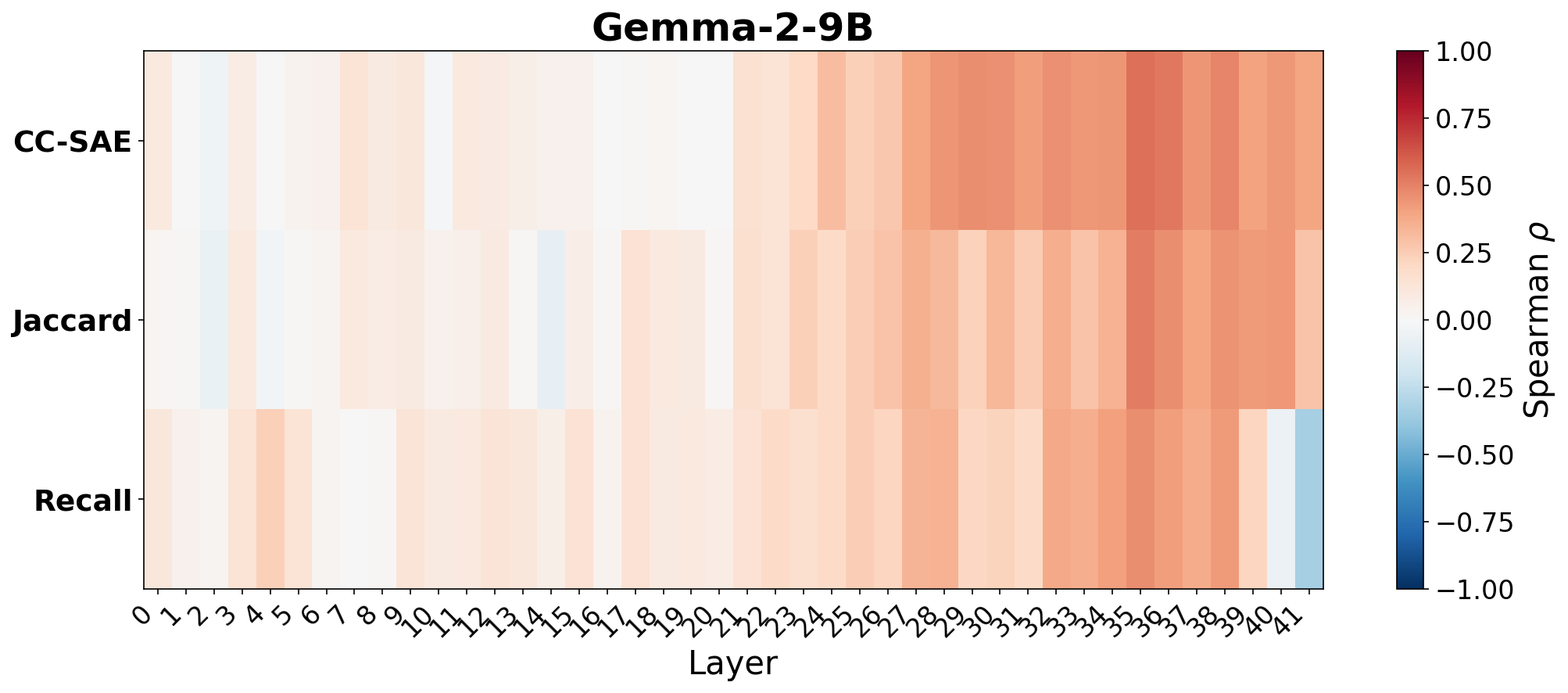}
    \end{subfigure}

    \vspace{0.5em}

    \begin{subfigure}[b]{0.32\textwidth}
        \includegraphics[width=\textwidth]{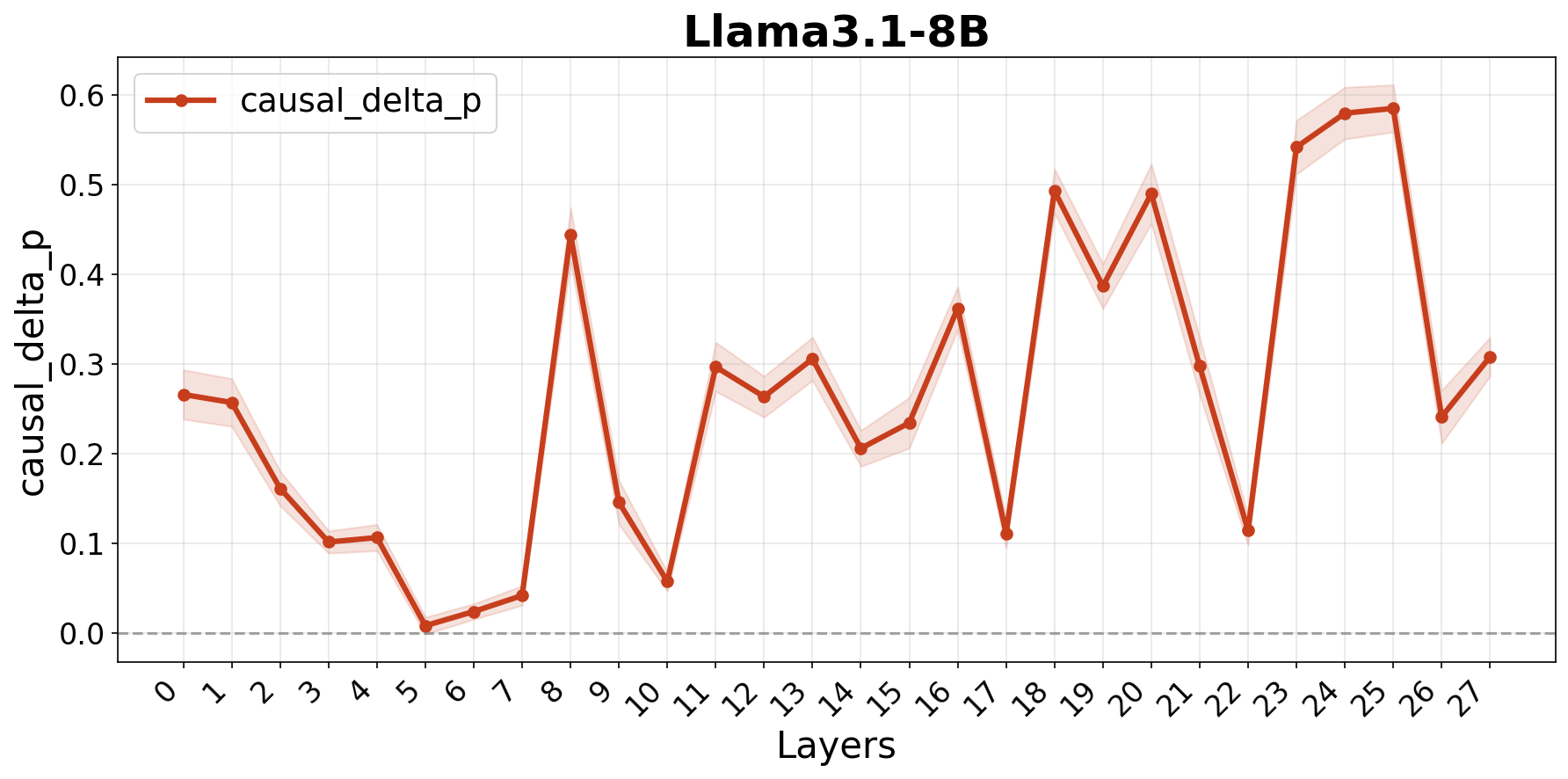}
    \end{subfigure}
    \hfill
    \begin{subfigure}[b]{0.32\textwidth}
        \includegraphics[width=\textwidth]{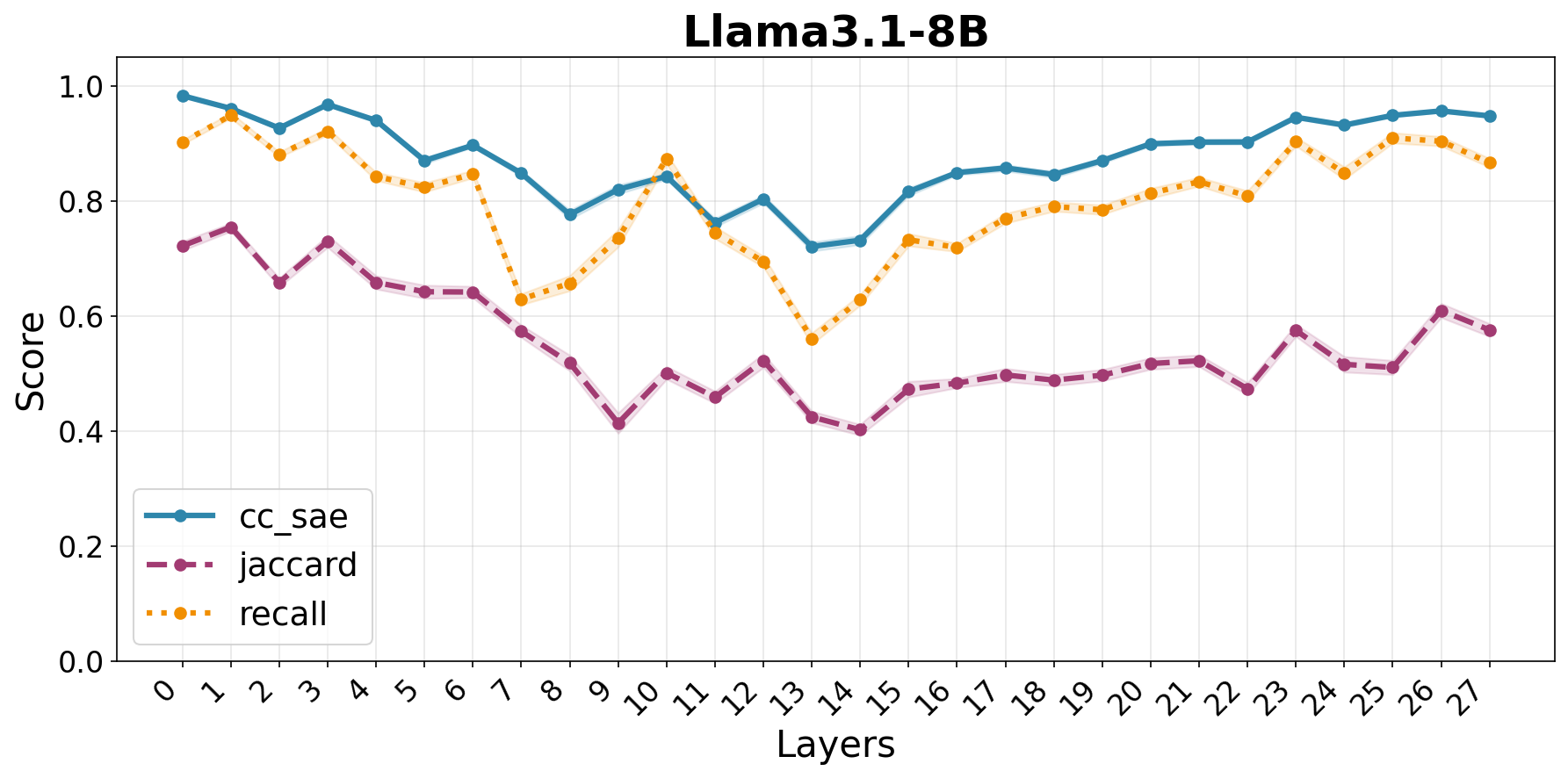}
    \end{subfigure}
    \hfill
    \begin{subfigure}[b]{0.32\textwidth}
        \includegraphics[width=\textwidth]{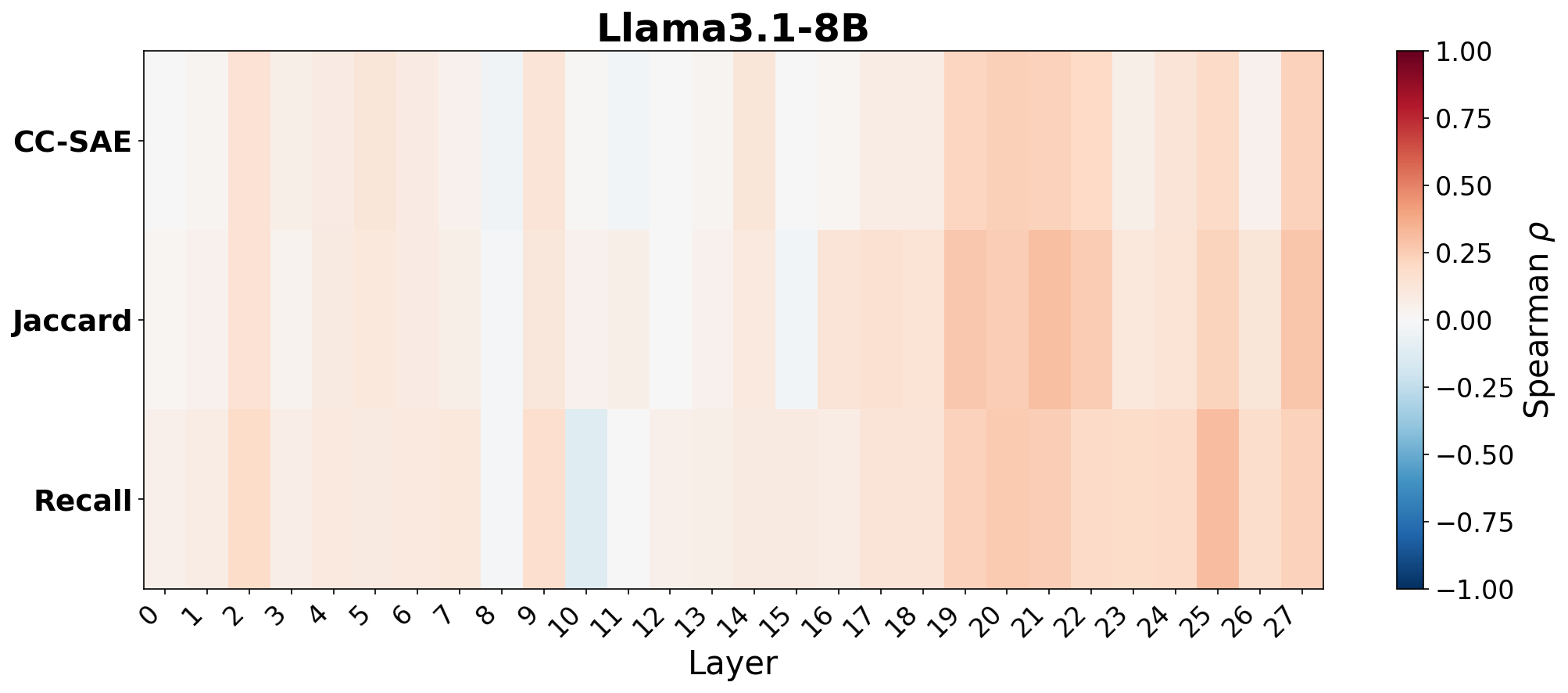}
    \end{subfigure}

    \caption{CoT faithfulness evaluation across layers for different models on \data{OpenbookQA}.
    \textbf{Left column:} Causal faithfulness ($\Delta p$) measures the probability drop when ablating shared features.
    \textbf{Middle column:} Correlational metrics (CC-SAE, Jaccard, Recall) between prediction and CoT activations.
    \textbf{Right column:} Heatmap showing Spearman correlation between each correlational metric and $\Delta p$ across layers.
    Shaded regions indicate 95\% confidence intervals.}
    \label{fig:ccsae-layers-openbookqa}
\end{figure*}

\begin{figure*}[t]
    \centering

    \begin{subfigure}[b]{0.32\textwidth}
        \includegraphics[width=\textwidth]{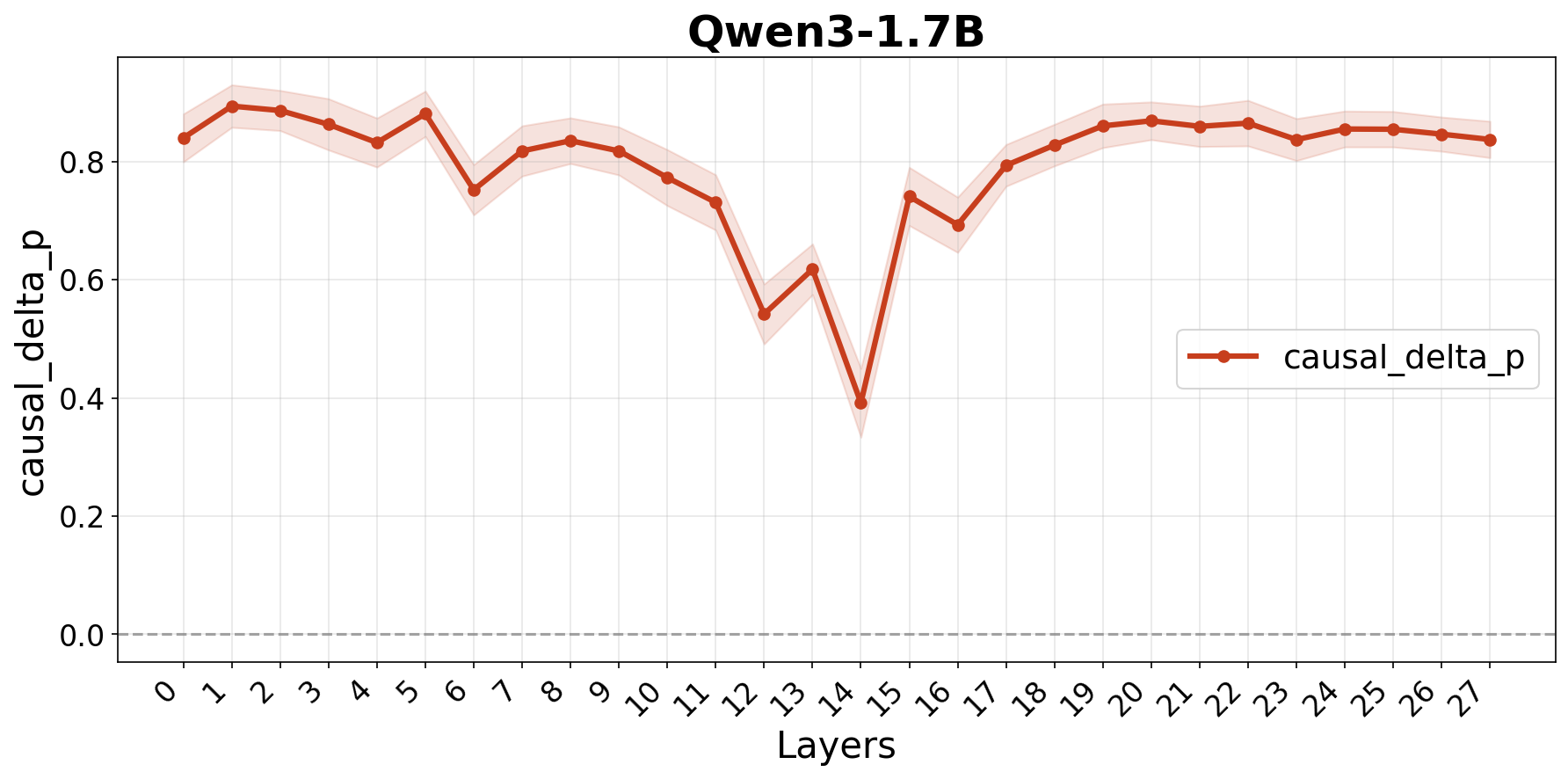}
    \end{subfigure}
    \hfill
    \begin{subfigure}[b]{0.32\textwidth}
        \includegraphics[width=\textwidth]{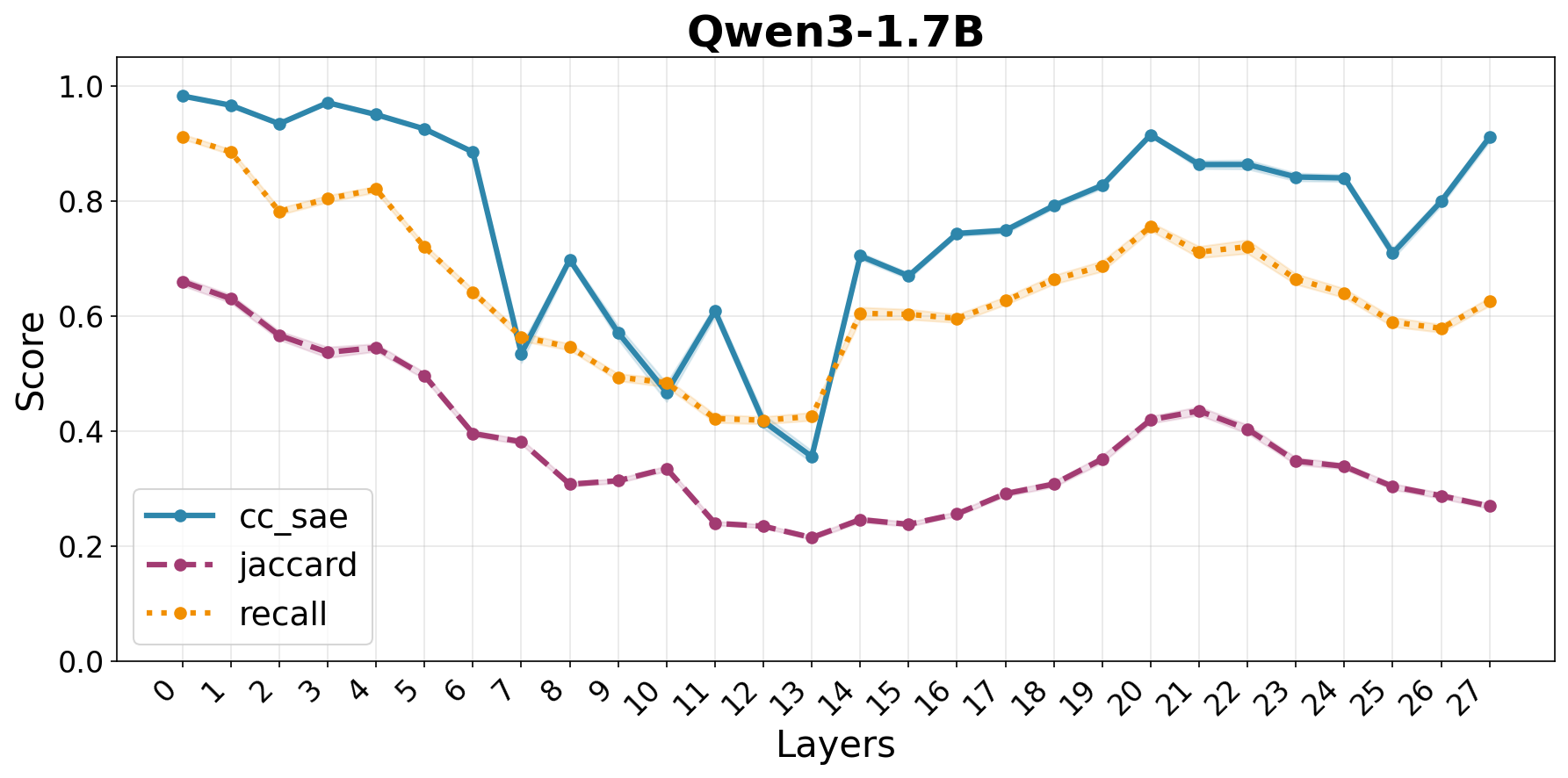}
    \end{subfigure}
    \hfill
    \begin{subfigure}[b]{0.32\textwidth}
        \includegraphics[width=\textwidth]{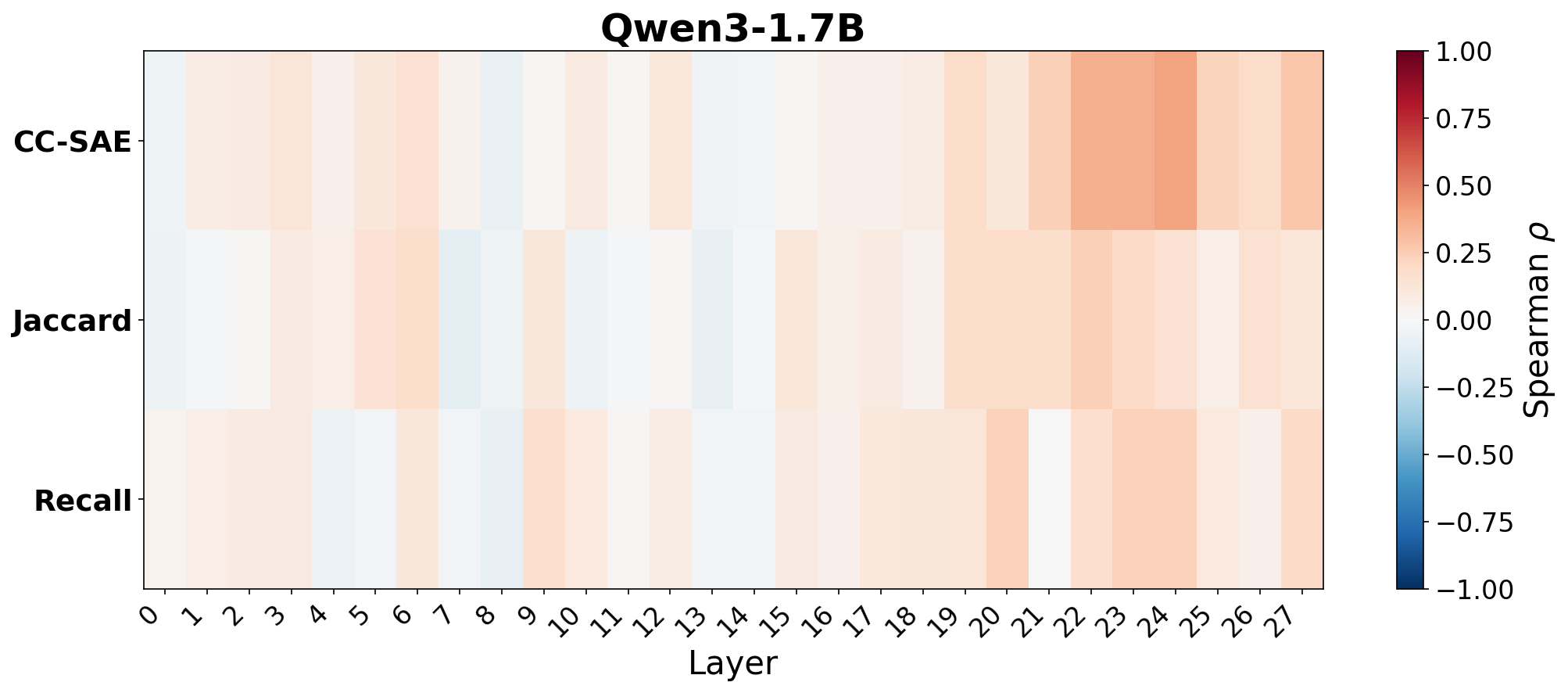}
    \end{subfigure}

    \vspace{0.5em}

    \begin{subfigure}[b]{0.32\textwidth}
        \includegraphics[width=\textwidth]{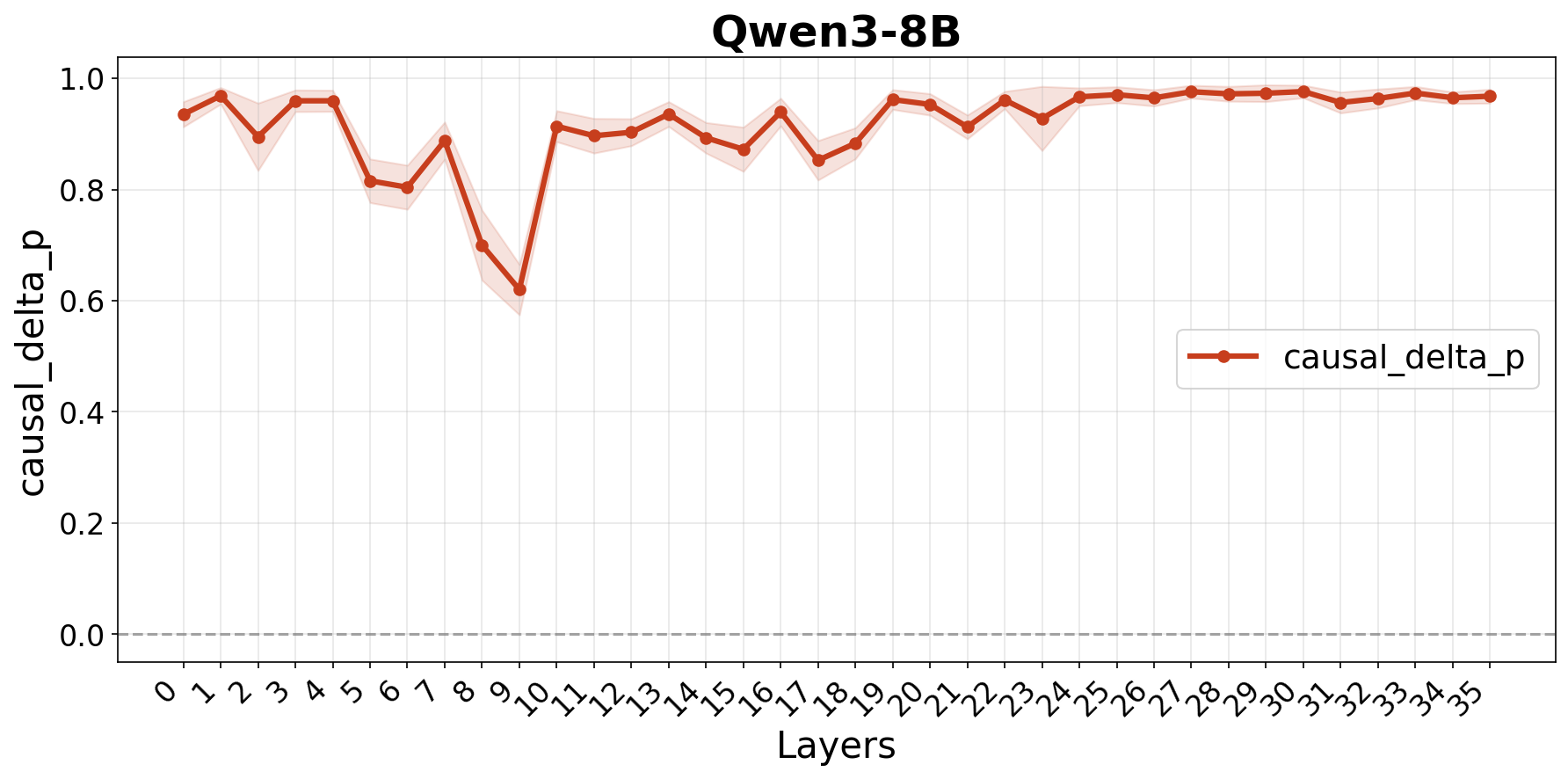}
    \end{subfigure}
    \hfill
    \begin{subfigure}[b]{0.32\textwidth}
        \includegraphics[width=\textwidth]{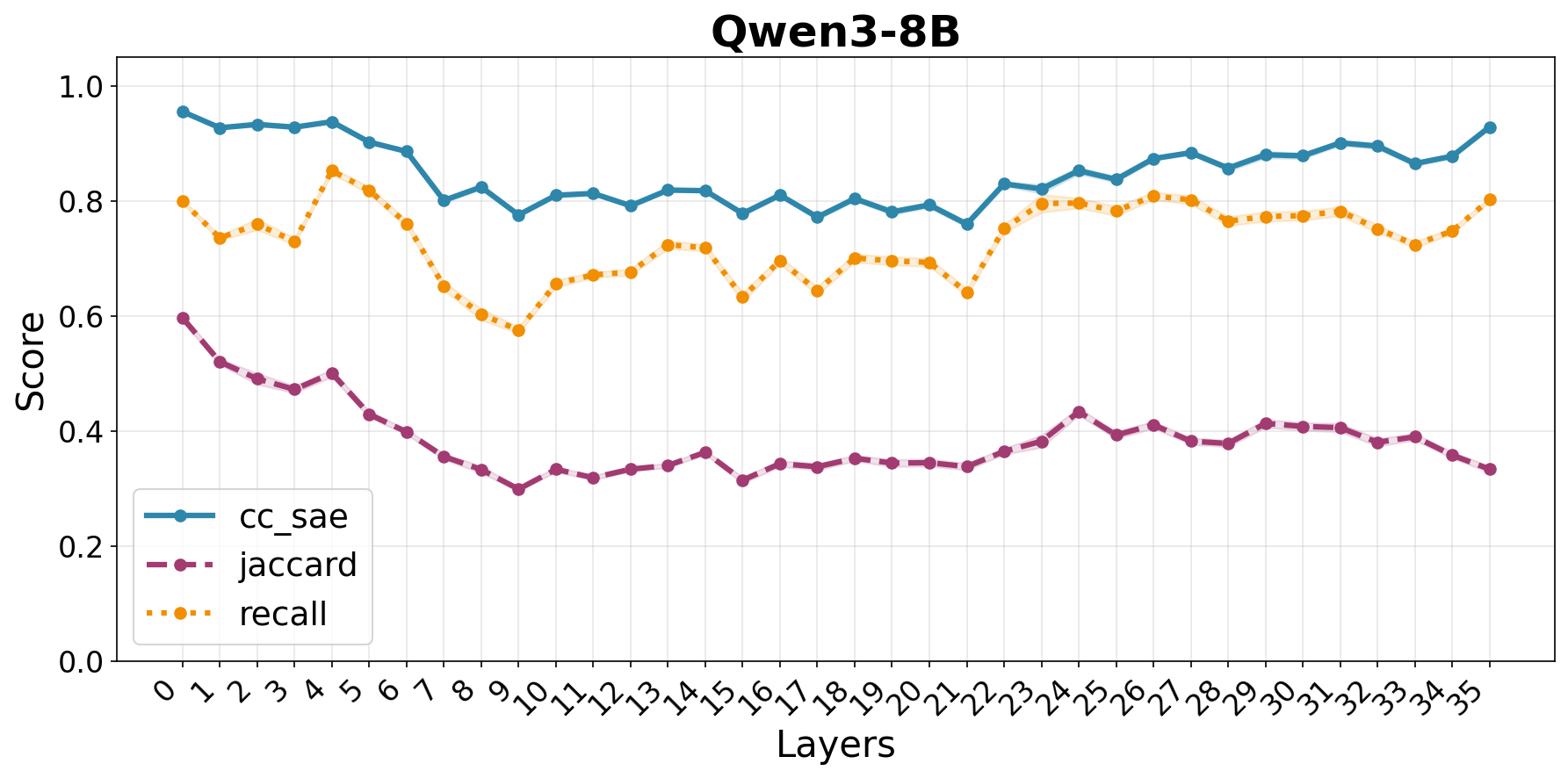}
    \end{subfigure}
    \hfill
    \begin{subfigure}[b]{0.32\textwidth}
        \includegraphics[width=\textwidth]{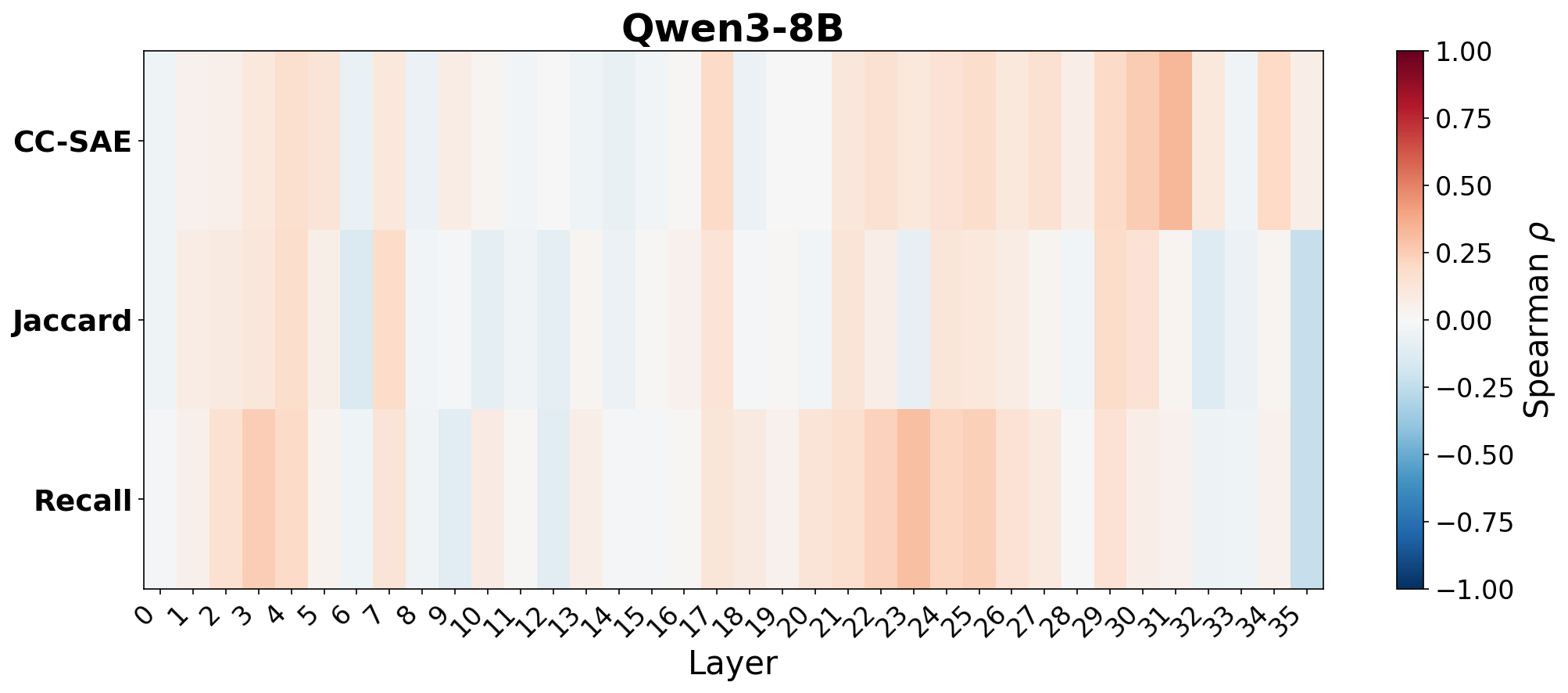}
    \end{subfigure}

    \vspace{0.5em}

    \begin{subfigure}[b]{0.32\textwidth}
        \includegraphics[width=\textwidth]{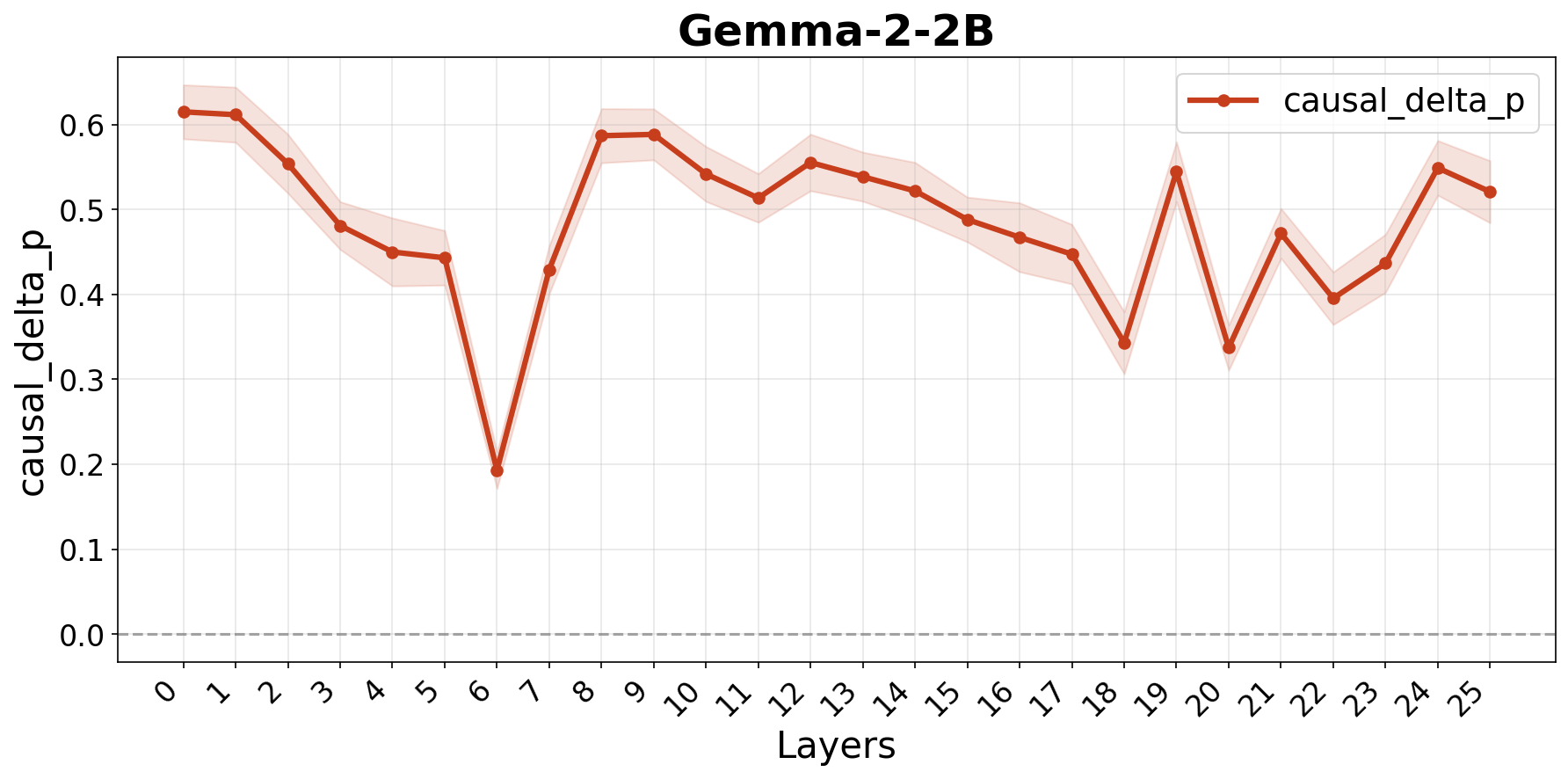}
    \end{subfigure}
    \hfill
    \begin{subfigure}[b]{0.32\textwidth}
        \includegraphics[width=\textwidth]{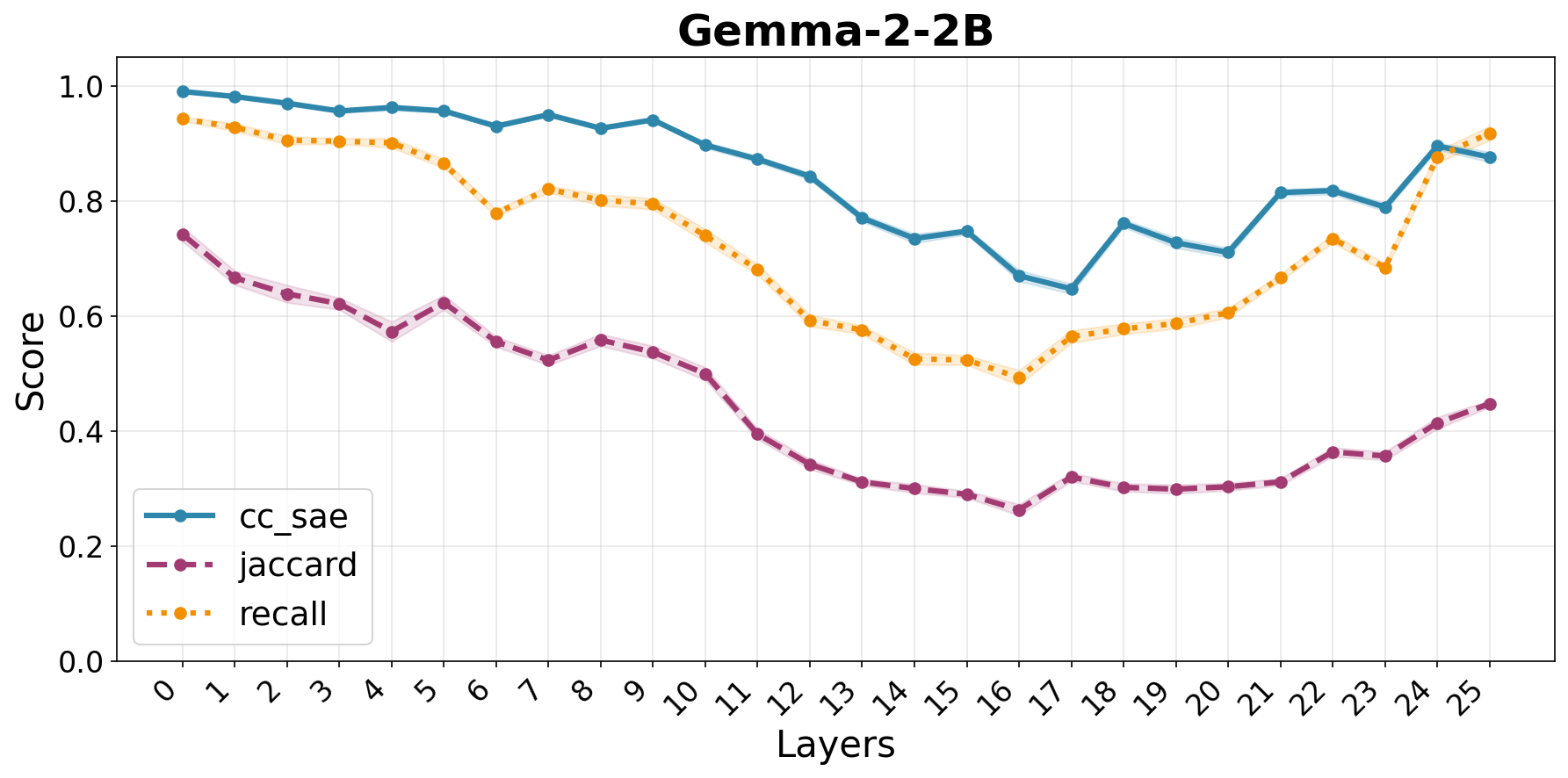}
    \end{subfigure}
    \hfill
    \begin{subfigure}[b]{0.32\textwidth}
        \includegraphics[width=\textwidth]{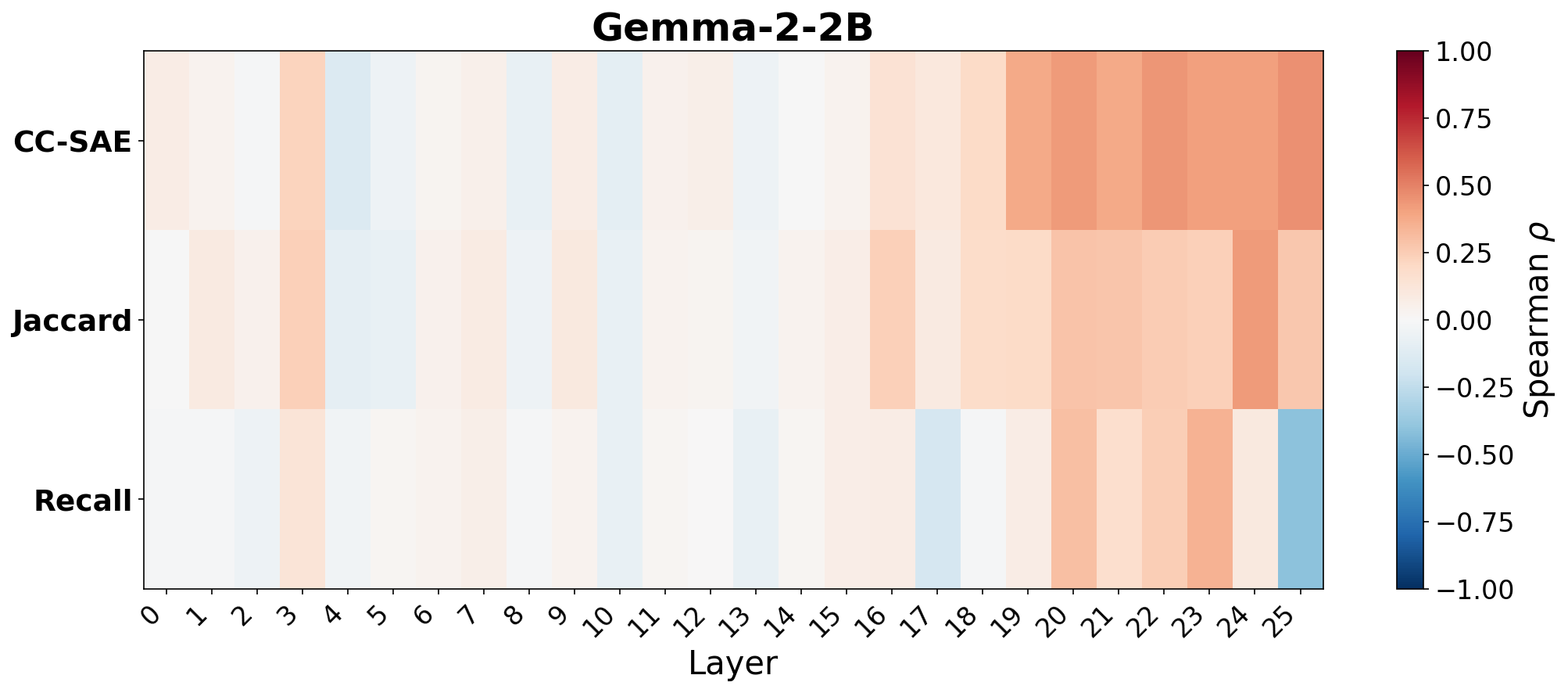}
    \end{subfigure}

    \vspace{0.5em}

    \begin{subfigure}[b]{0.32\textwidth}
        \includegraphics[width=\textwidth]{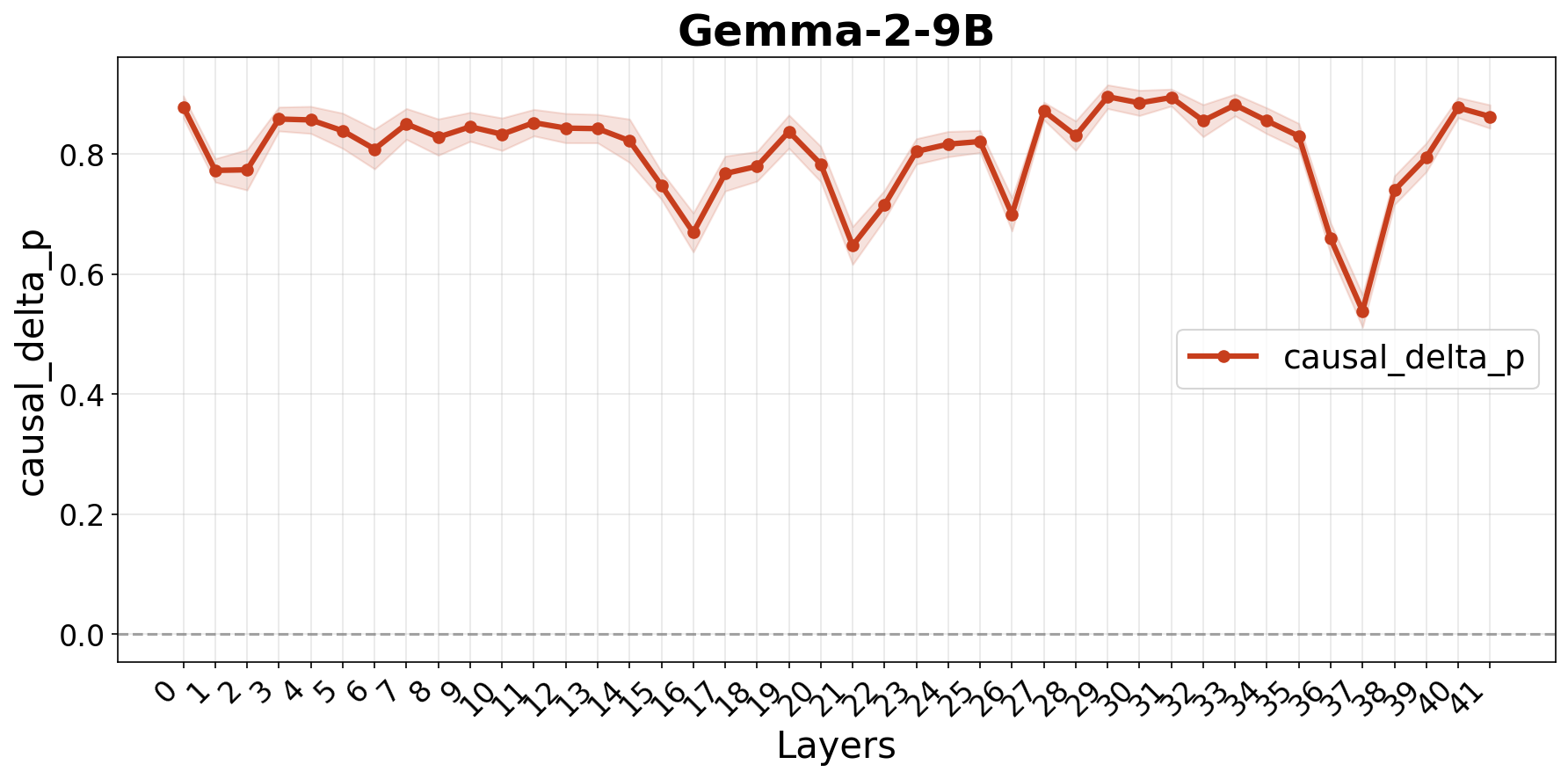}
    \end{subfigure}
    \hfill
    \begin{subfigure}[b]{0.32\textwidth}
        \includegraphics[width=\textwidth]{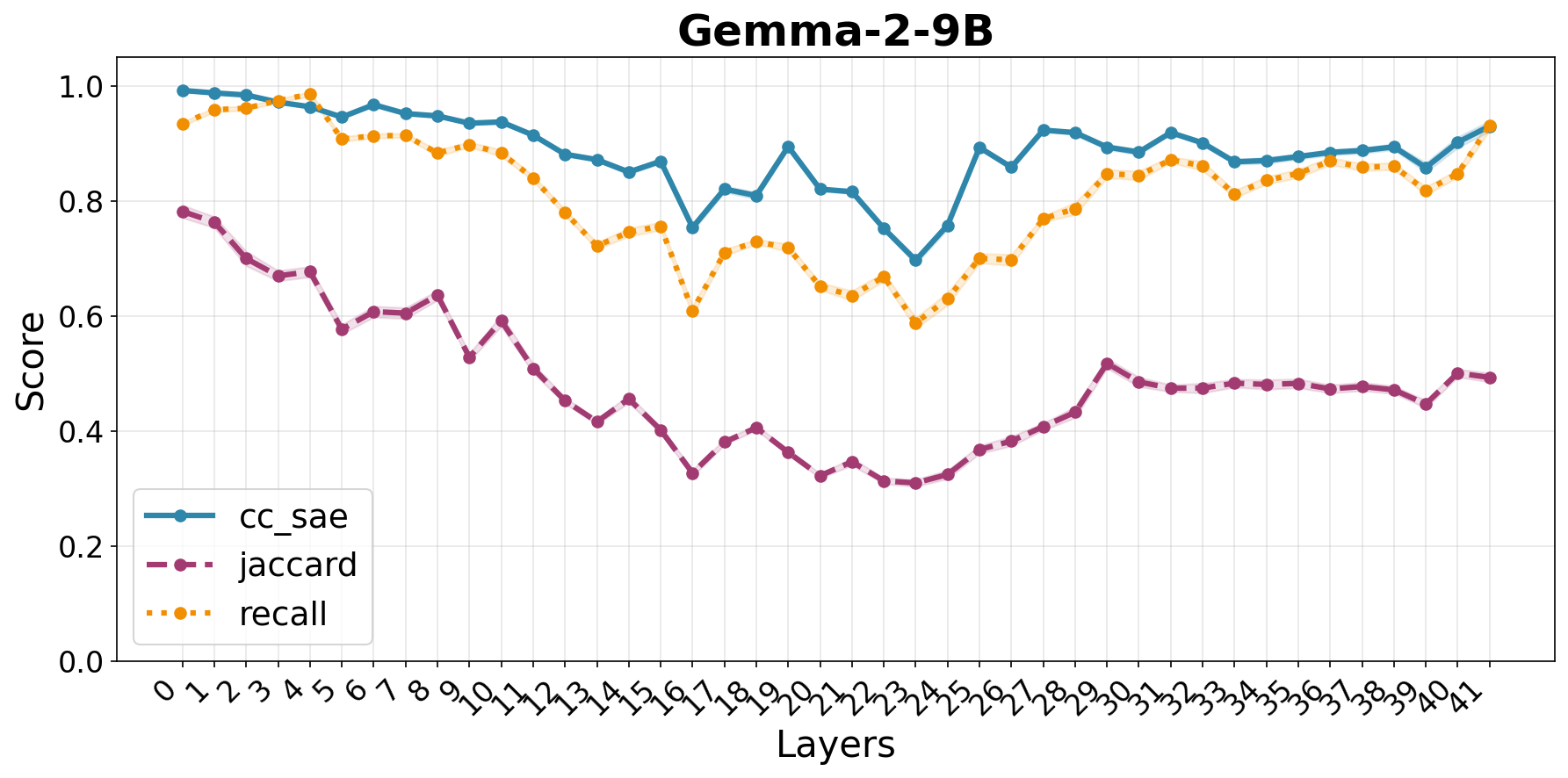}
    \end{subfigure}
    \hfill
    \begin{subfigure}[b]{0.32\textwidth}
        \includegraphics[width=\textwidth]{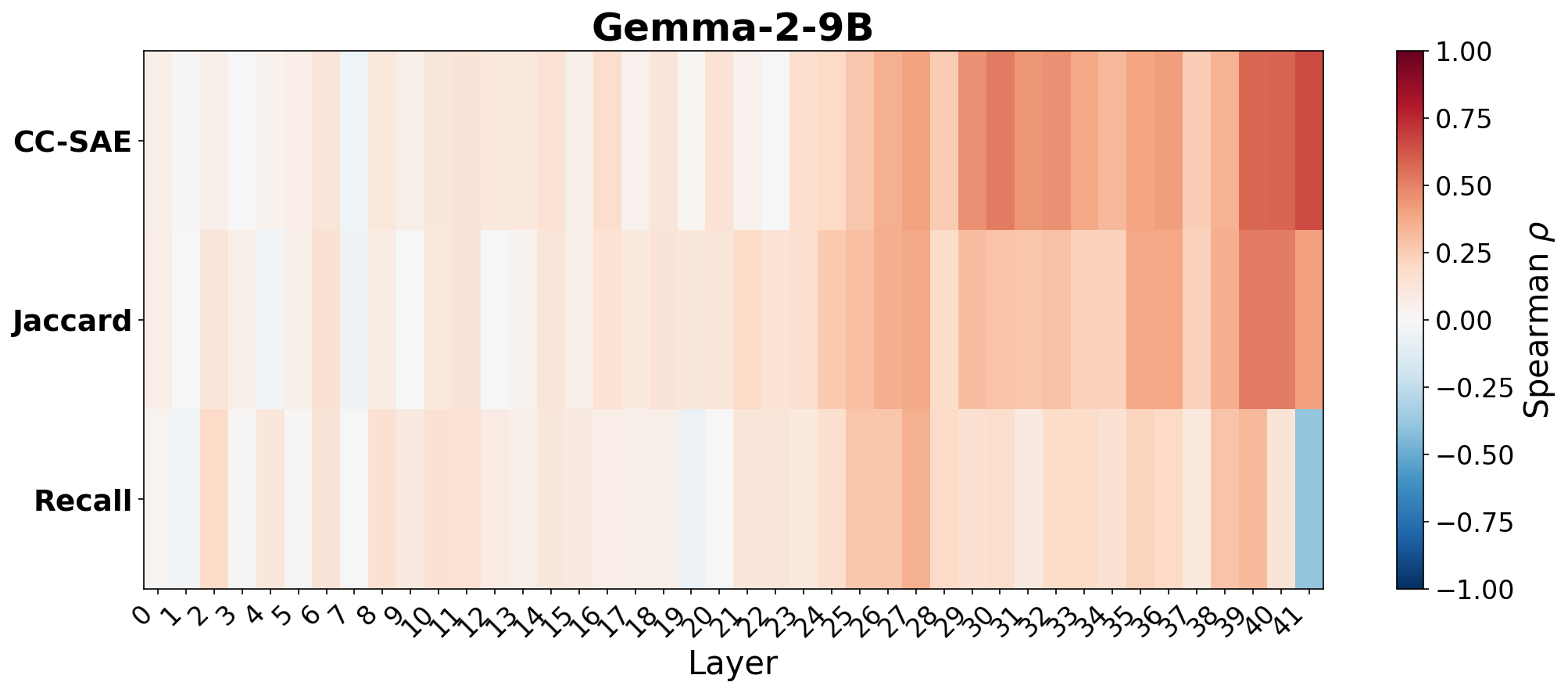}
    \end{subfigure}

    \vspace{0.5em}

    \begin{subfigure}[b]{0.32\textwidth}
        \includegraphics[width=\textwidth]{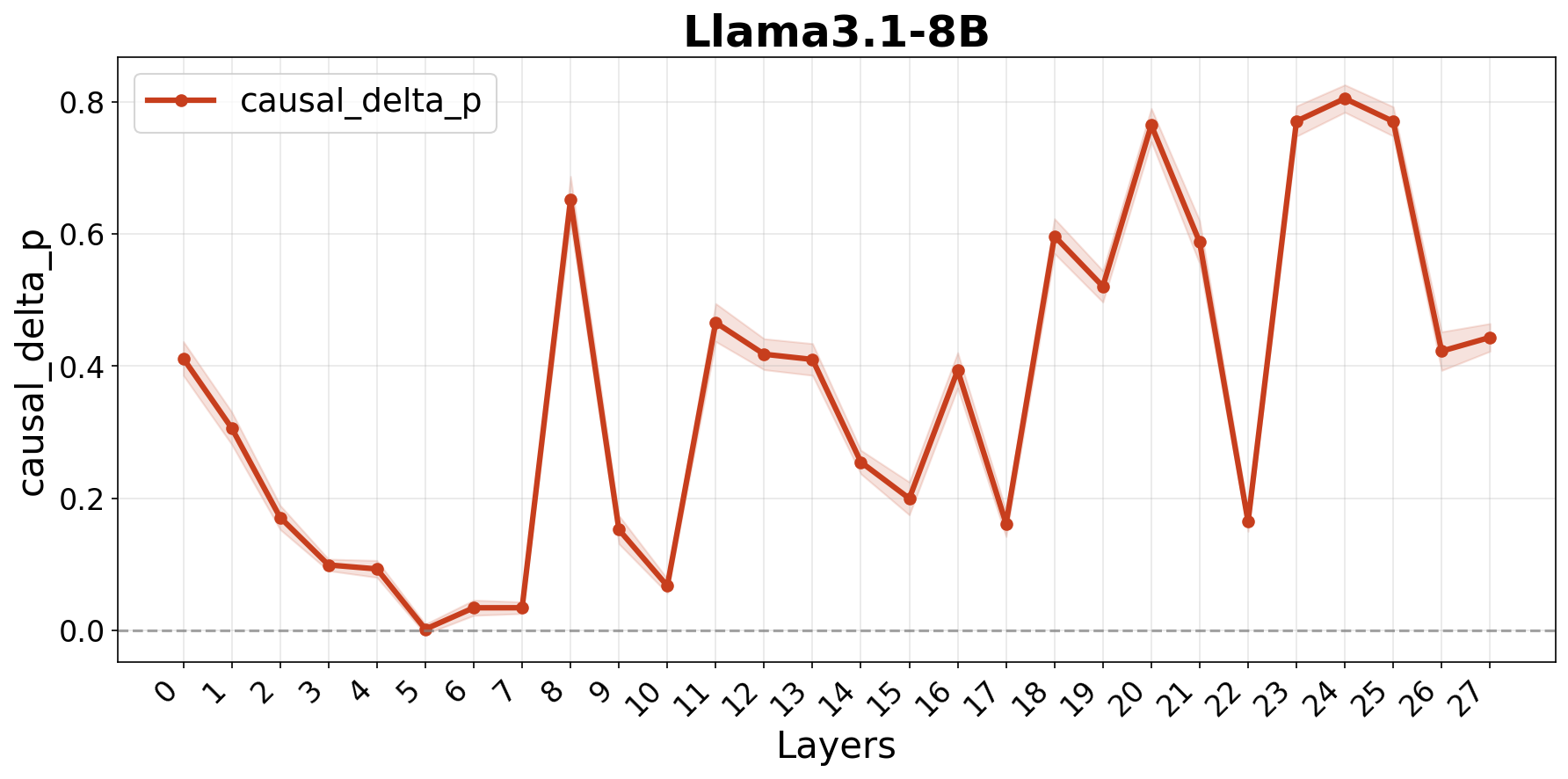}
    \end{subfigure}
    \hfill
    \begin{subfigure}[b]{0.32\textwidth}
        \includegraphics[width=\textwidth]{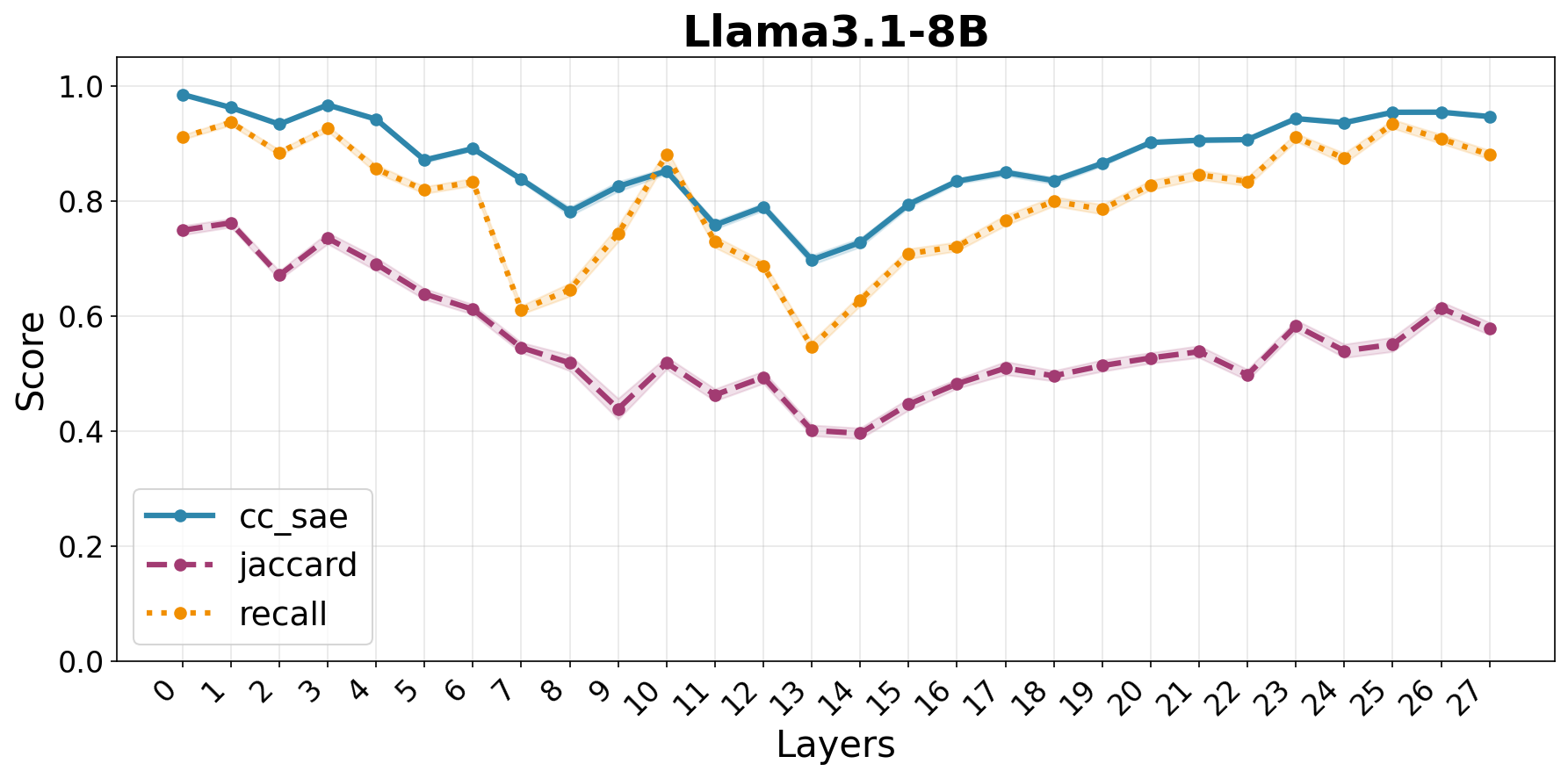}
    \end{subfigure}
    \hfill
    \begin{subfigure}[b]{0.32\textwidth}
        \includegraphics[width=\textwidth]{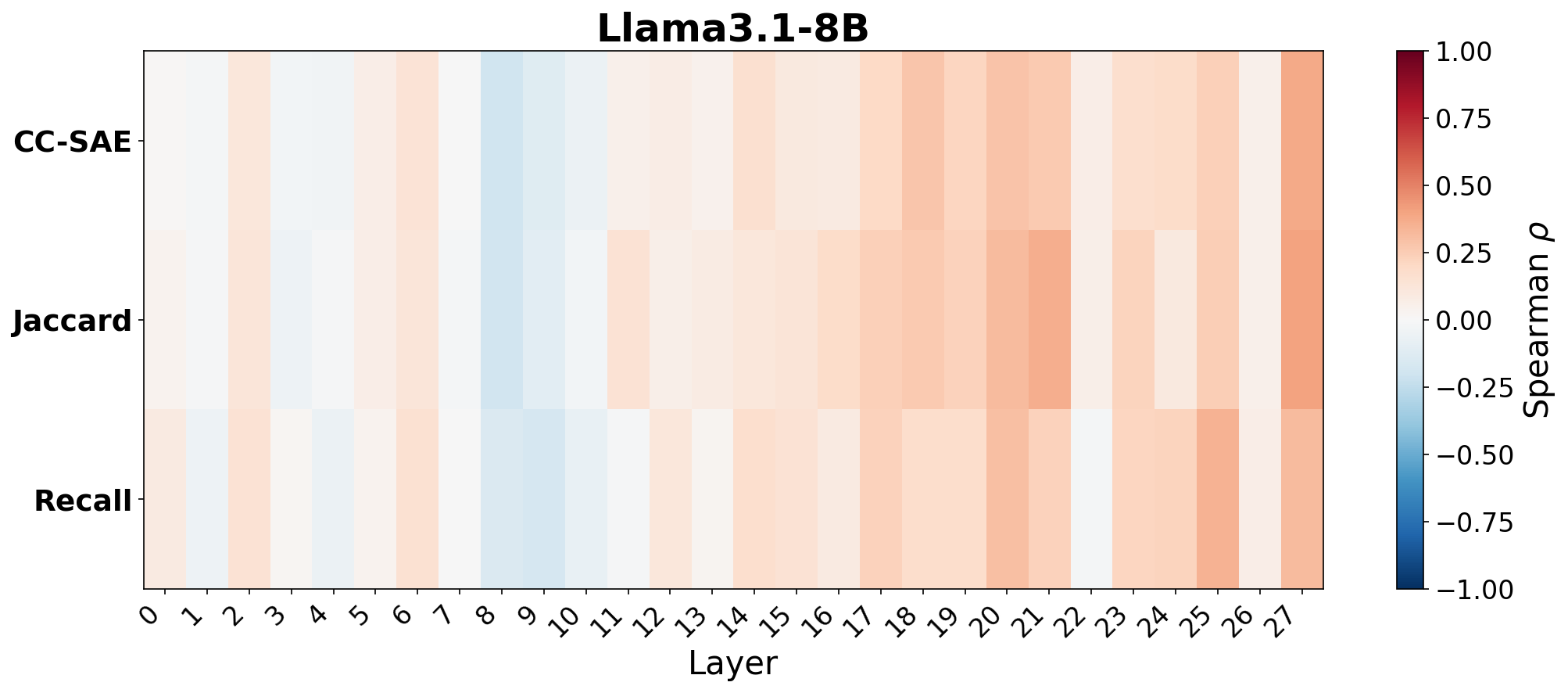}
    \end{subfigure}

    \caption{CoT faithfulness evaluation across layers for different models on \data{ARC-Easy}.
    \textbf{Left column:} Causal faithfulness ($\Delta p$) measures the probability drop when ablating shared features.
    \textbf{Middle column:} Correlational metrics (CC-SAE, Jaccard, Recall) between prediction and CoT activations.
    \textbf{Right column:} Heatmap showing Spearman correlation between each correlational metric and $\Delta p$ across layers.
    Shaded regions indicate 95\% confidence intervals.}
    \label{fig:ccsae-layers-arc_easy}
\end{figure*}

\begin{figure*}[t]
    \centering

    \begin{subfigure}[b]{0.32\textwidth}
        \includegraphics[width=\textwidth]{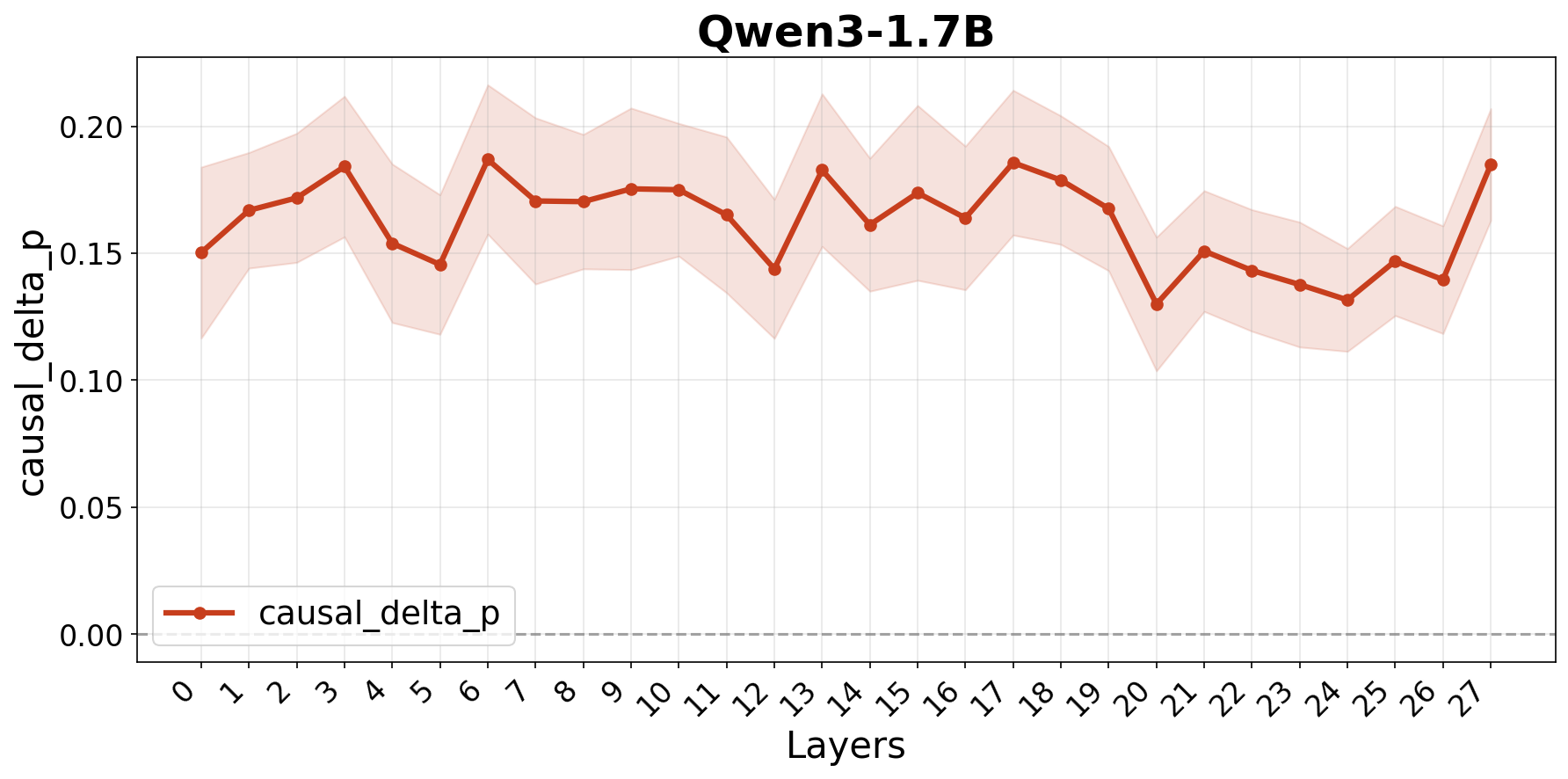}
    \end{subfigure}
    \hfill
    \begin{subfigure}[b]{0.32\textwidth}
        \includegraphics[width=\textwidth]{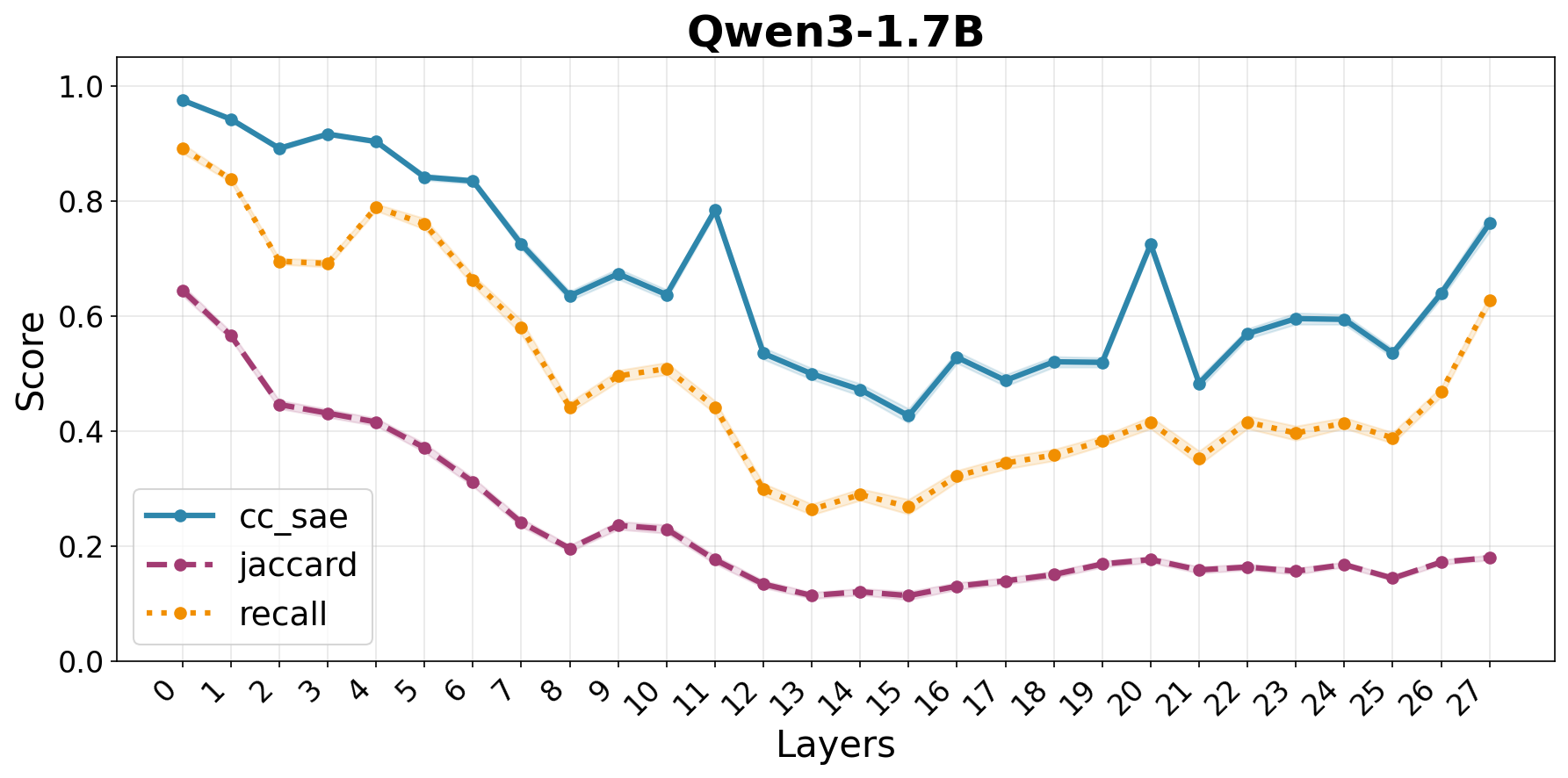}
    \end{subfigure}
    \hfill
    \begin{subfigure}[b]{0.32\textwidth}
        \includegraphics[width=\textwidth]{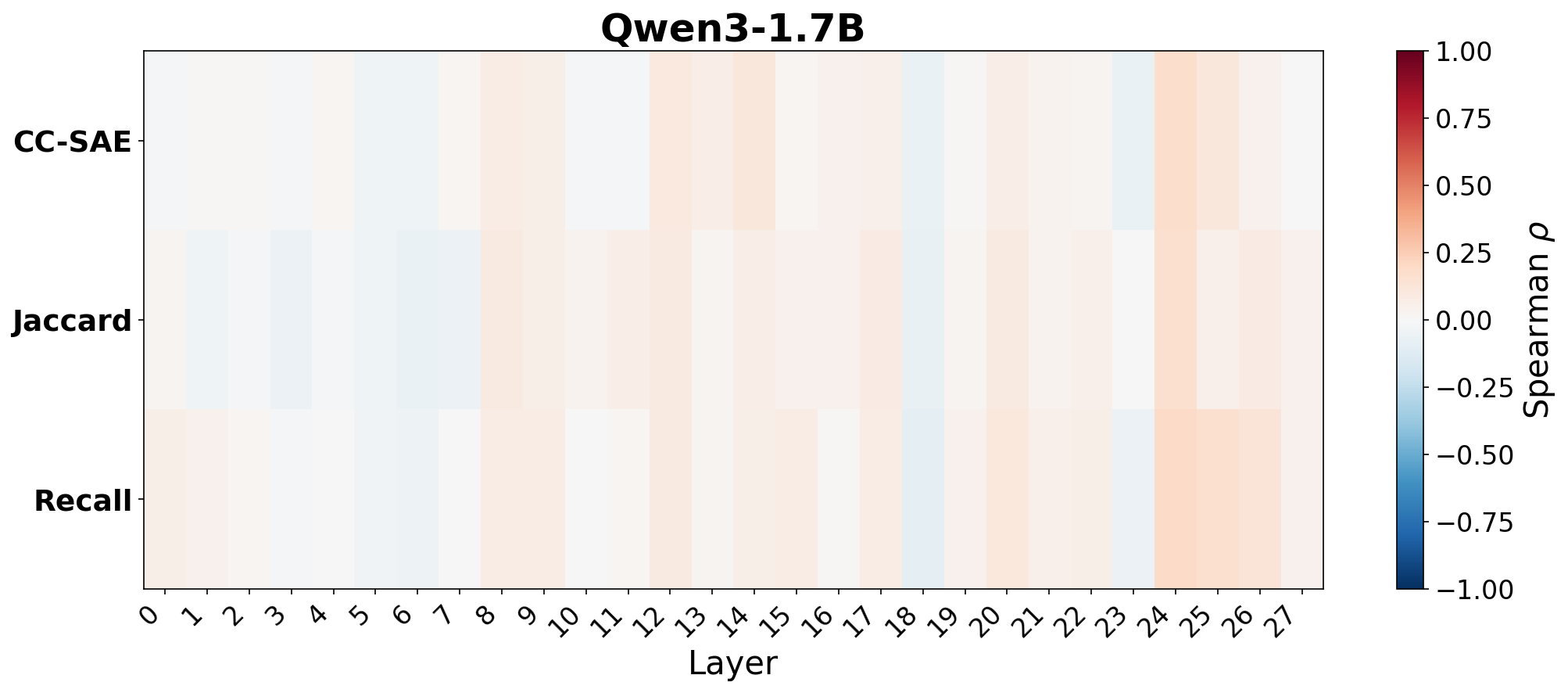}
    \end{subfigure}

    \vspace{0.5em}

    \begin{subfigure}[b]{0.32\textwidth}
        \includegraphics[width=\textwidth]{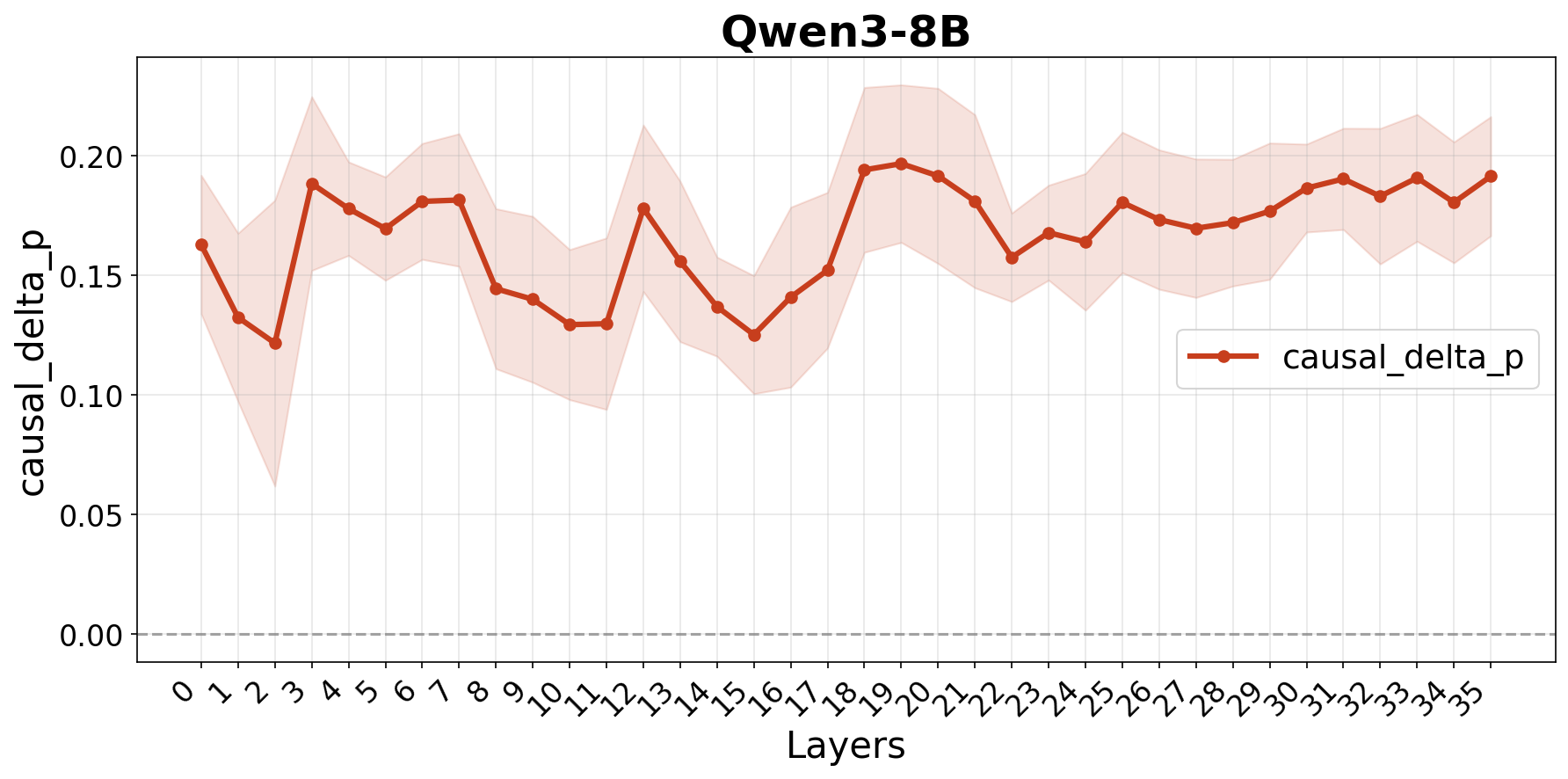}
    \end{subfigure}
    \hfill
    \begin{subfigure}[b]{0.32\textwidth}
        \includegraphics[width=\textwidth]{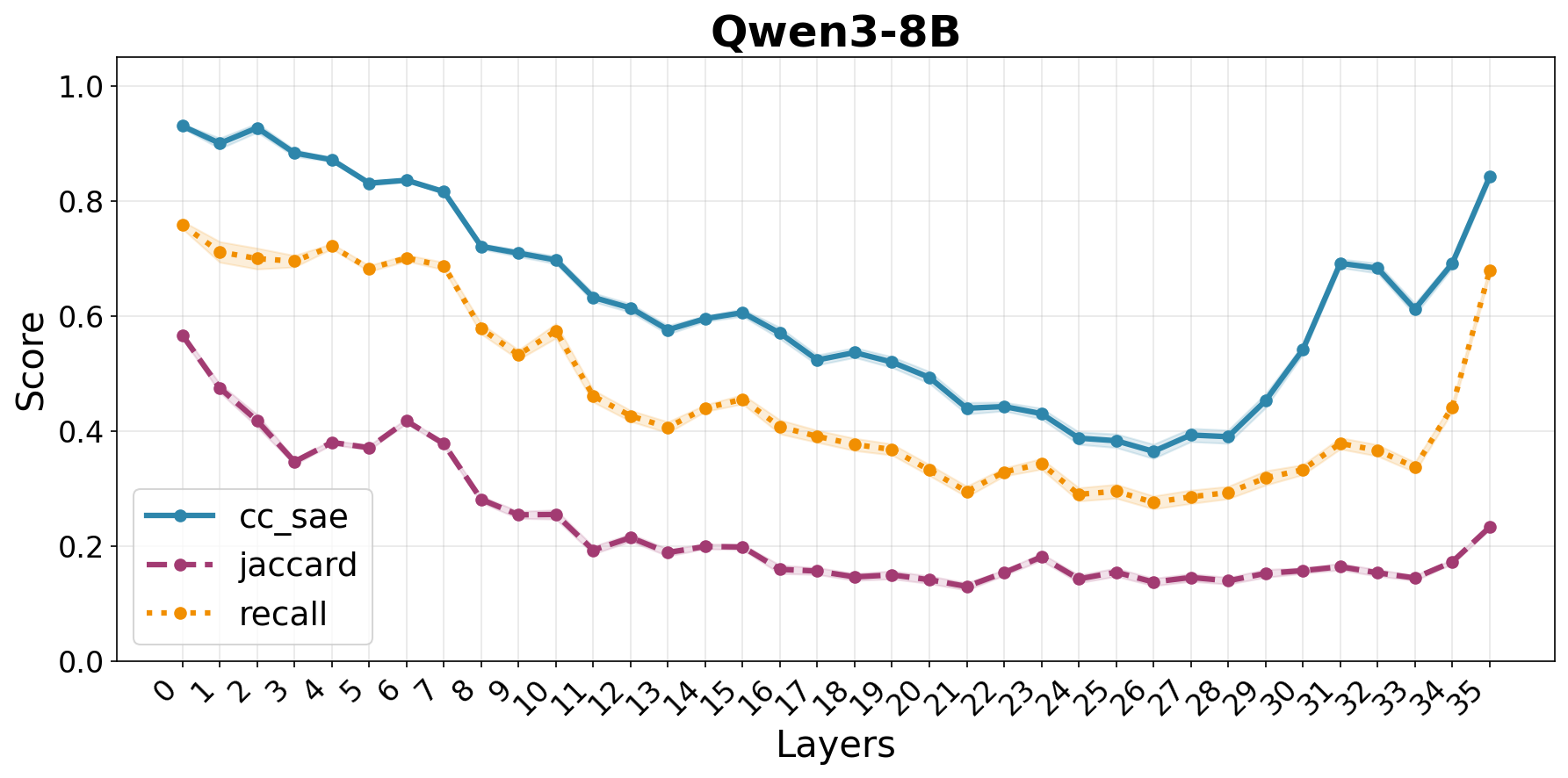}
    \end{subfigure}
    \hfill
    \begin{subfigure}[b]{0.32\textwidth}
        \includegraphics[width=\textwidth]{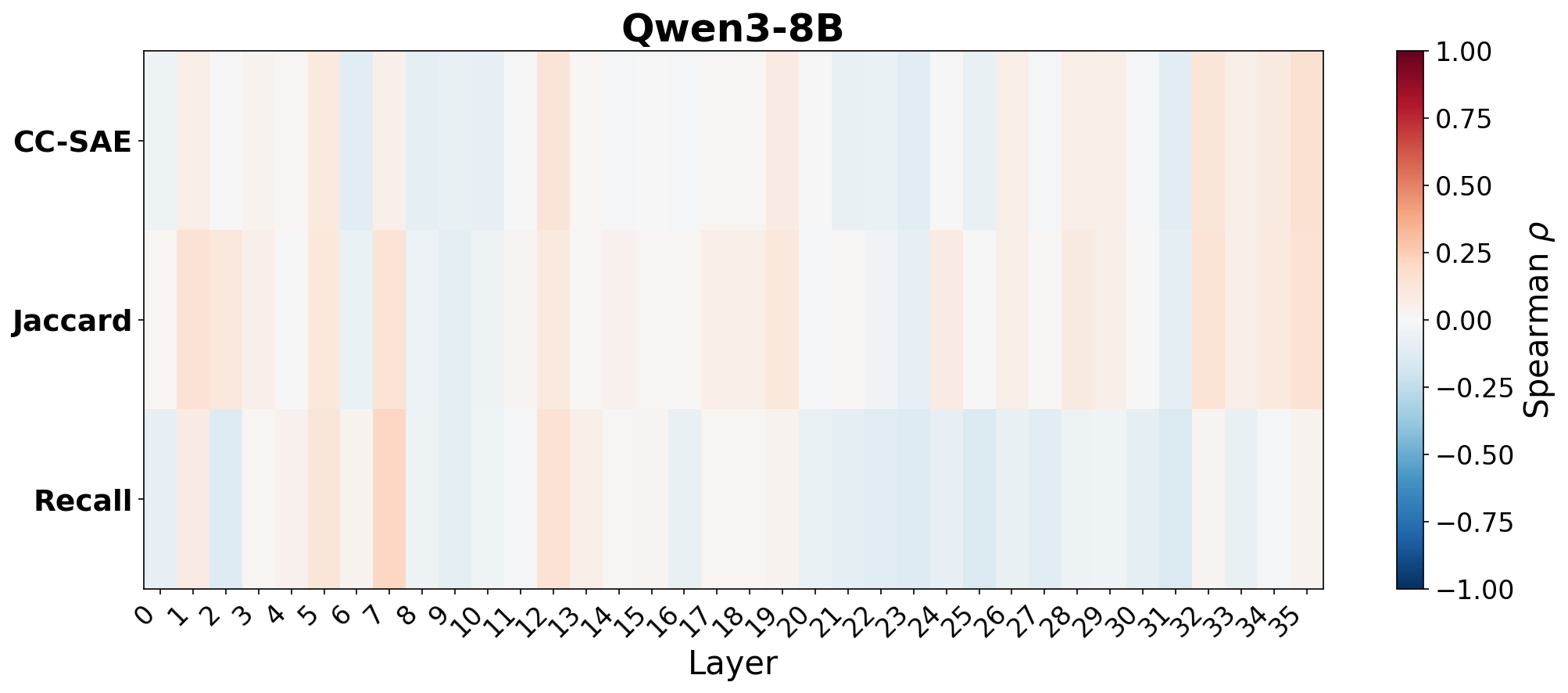}
    \end{subfigure}

    \vspace{0.5em}

    \begin{subfigure}[b]{0.32\textwidth}
        \includegraphics[width=\textwidth]{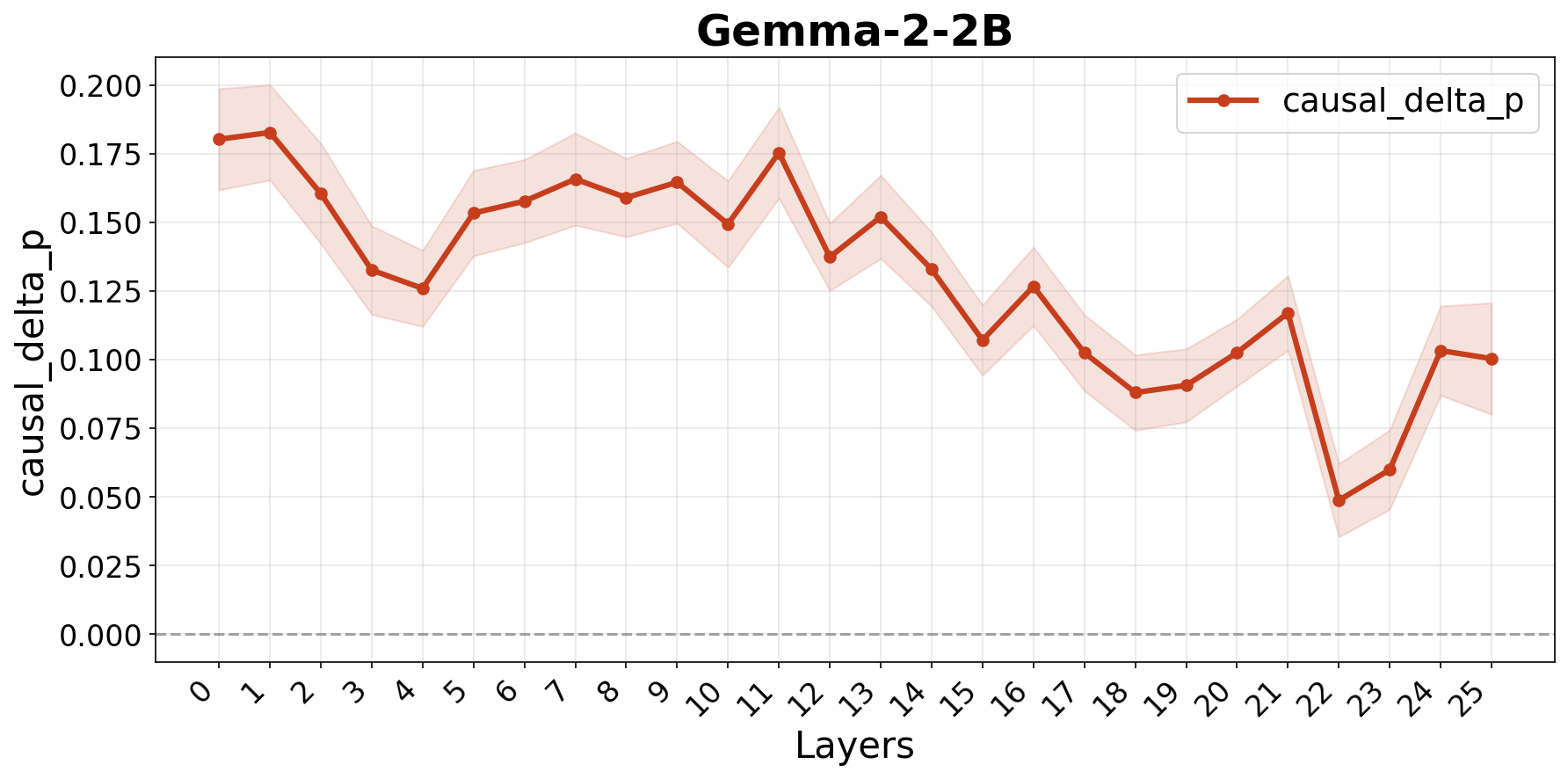}
    \end{subfigure}
    \hfill
    \begin{subfigure}[b]{0.32\textwidth}
        \includegraphics[width=\textwidth]{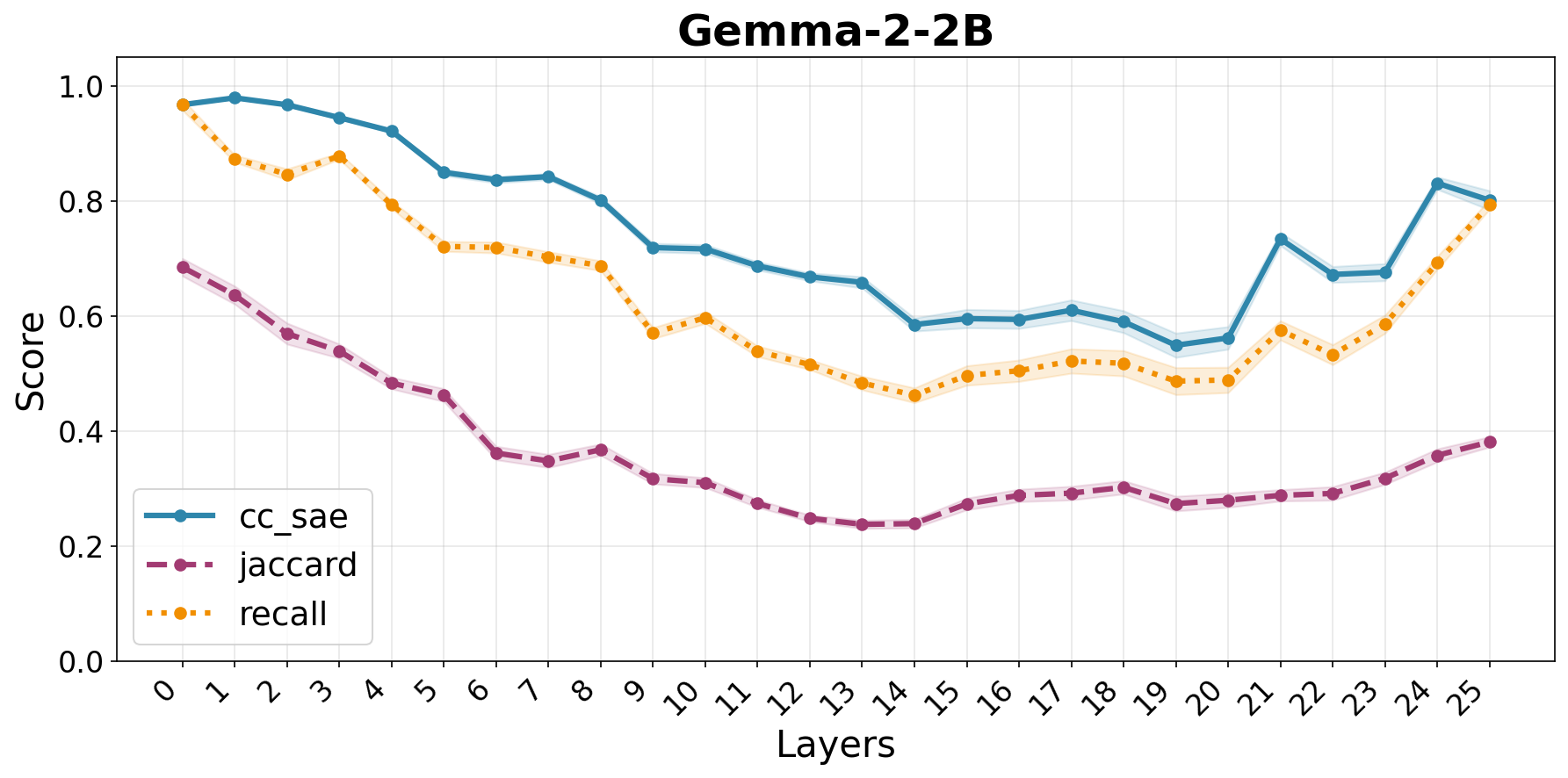}
    \end{subfigure}
    \hfill
    \begin{subfigure}[b]{0.32\textwidth}
        \includegraphics[width=\textwidth]{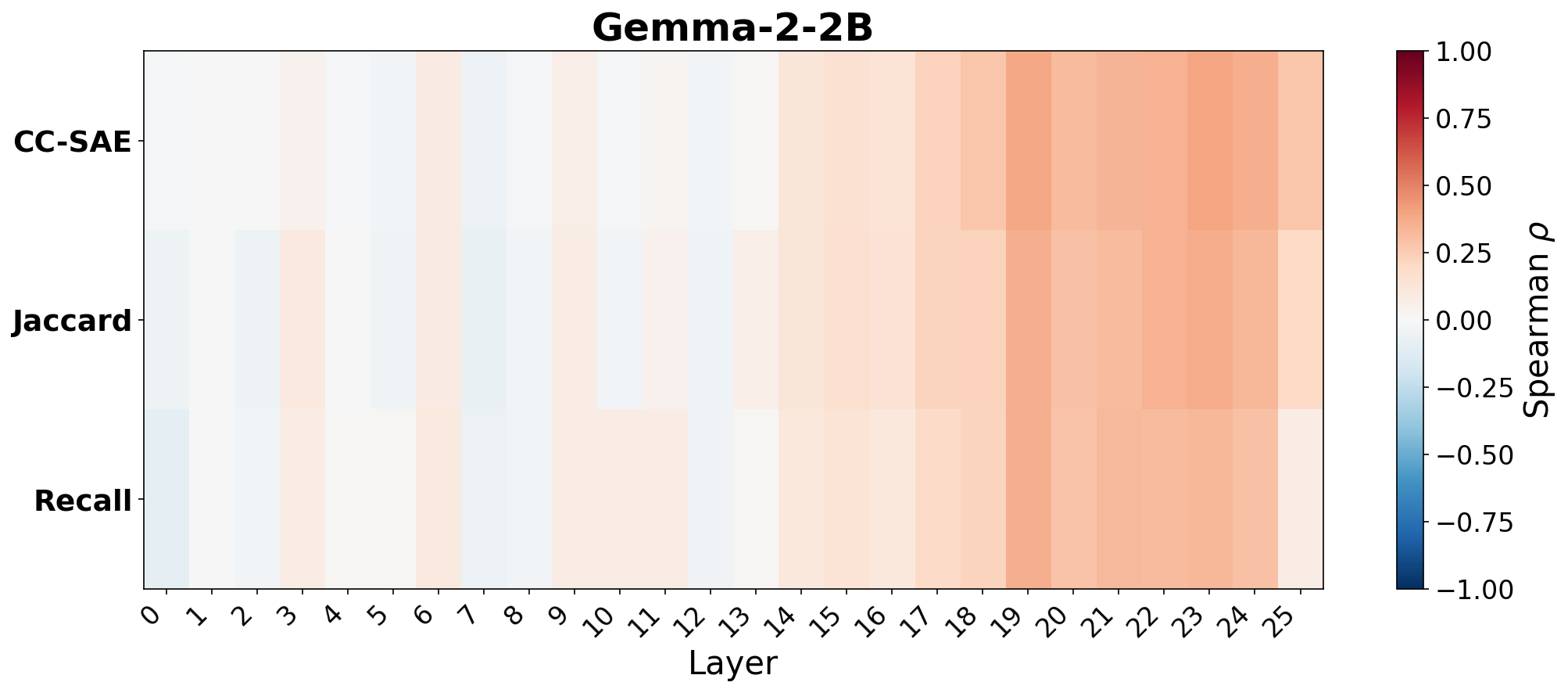}
    \end{subfigure}

    \vspace{0.5em}

    \begin{subfigure}[b]{0.32\textwidth}
        \includegraphics[width=\textwidth]{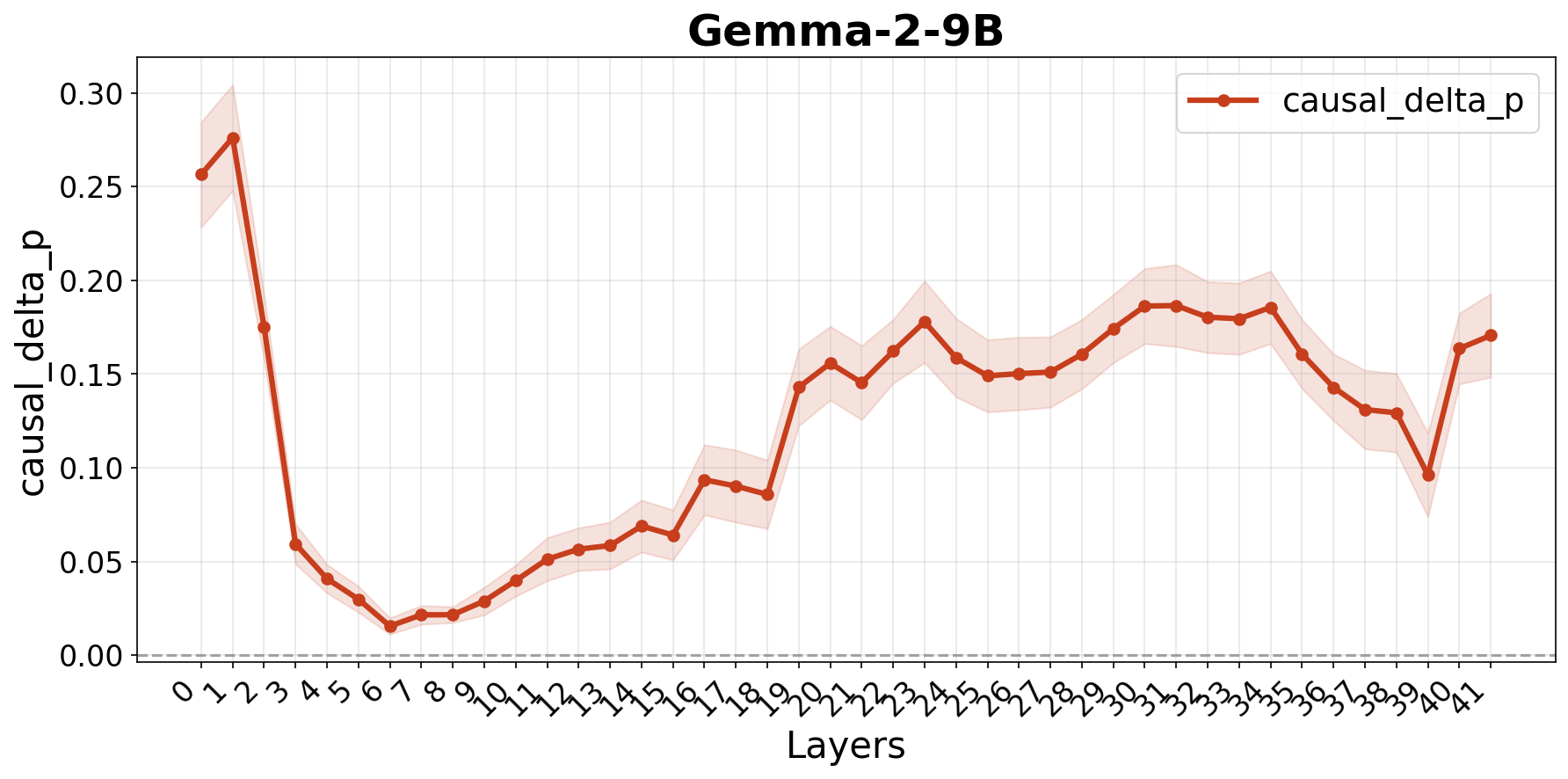}
    \end{subfigure}
    \hfill
    \begin{subfigure}[b]{0.32\textwidth}
        \includegraphics[width=\textwidth]{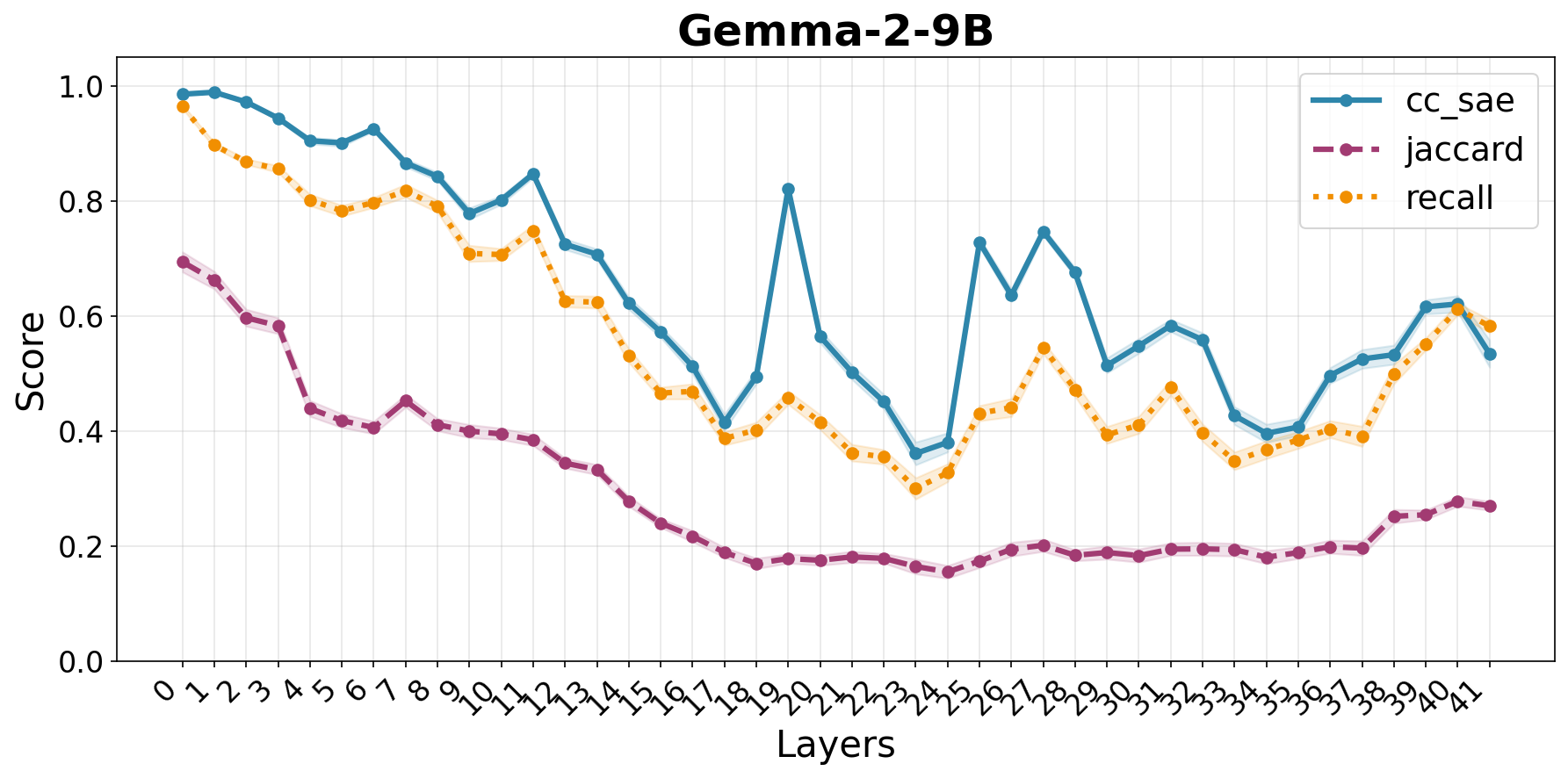}
    \end{subfigure}
    \hfill
    \begin{subfigure}[b]{0.32\textwidth}
        \includegraphics[width=\textwidth]{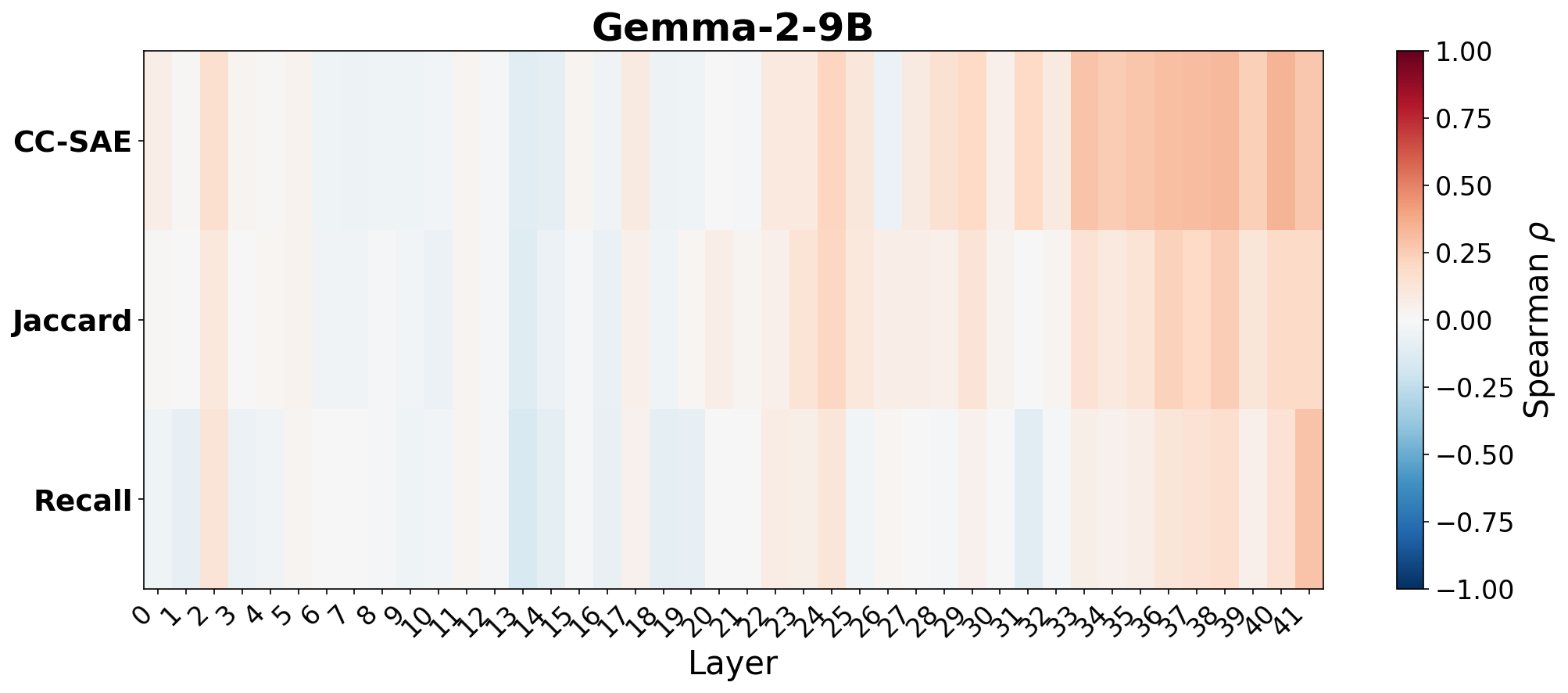}
    \end{subfigure}

    \vspace{0.5em}

    \begin{subfigure}[b]{0.32\textwidth}
        \includegraphics[width=\textwidth]{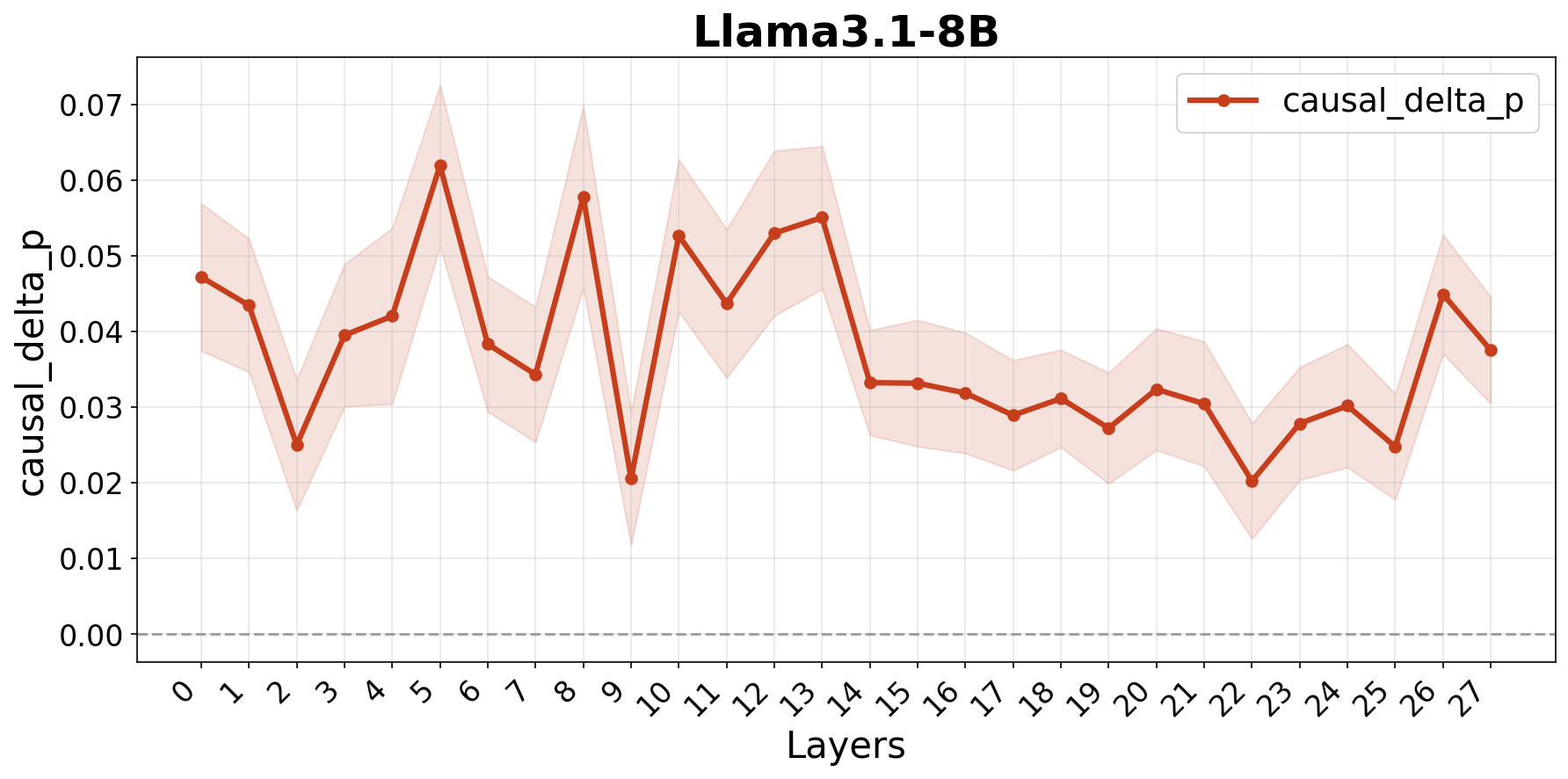}
    \end{subfigure}
    \hfill
    \begin{subfigure}[b]{0.32\textwidth}
        \includegraphics[width=\textwidth]{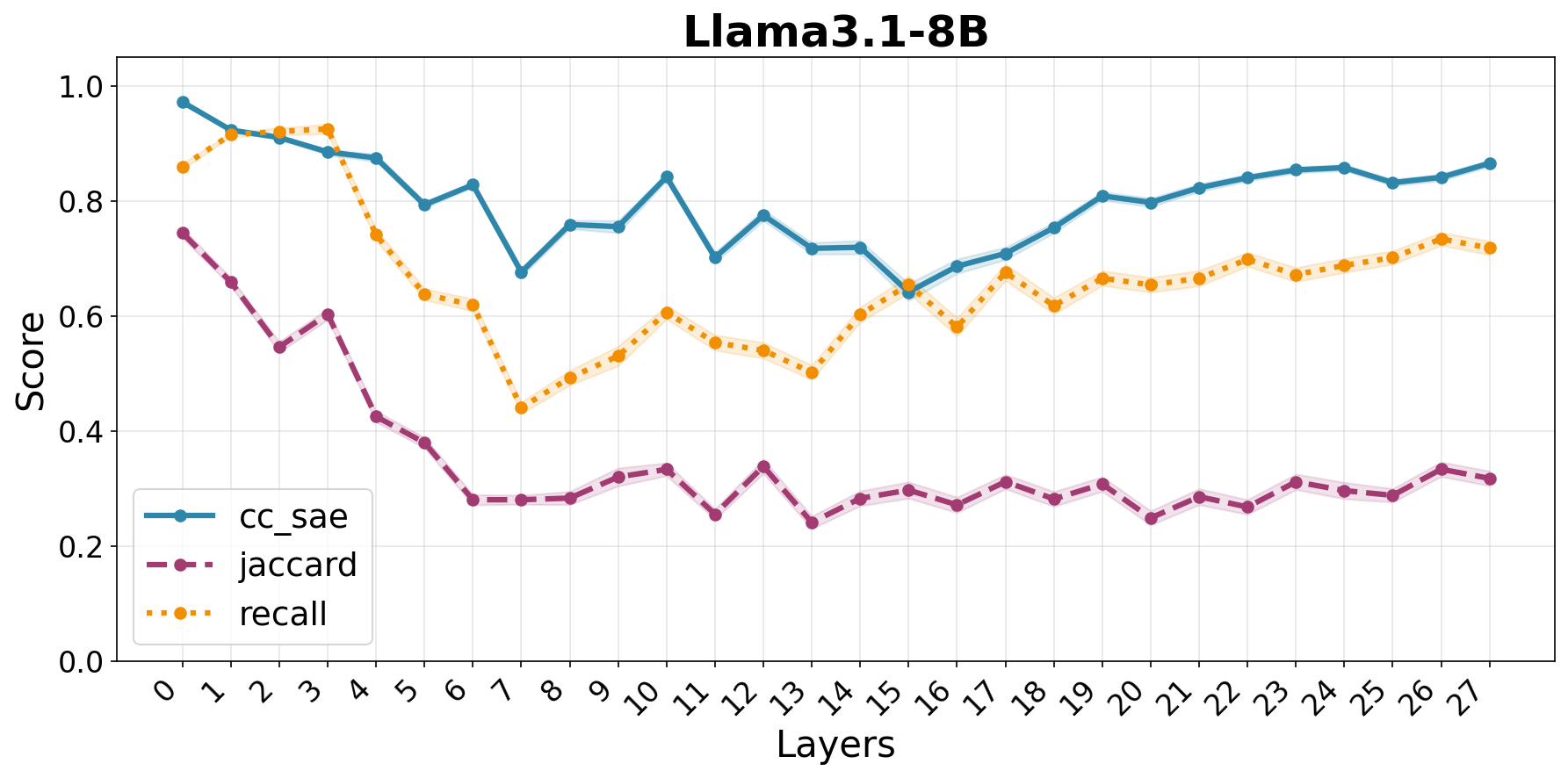}
    \end{subfigure}
    \hfill
    \begin{subfigure}[b]{0.32\textwidth}
        \includegraphics[width=\textwidth]{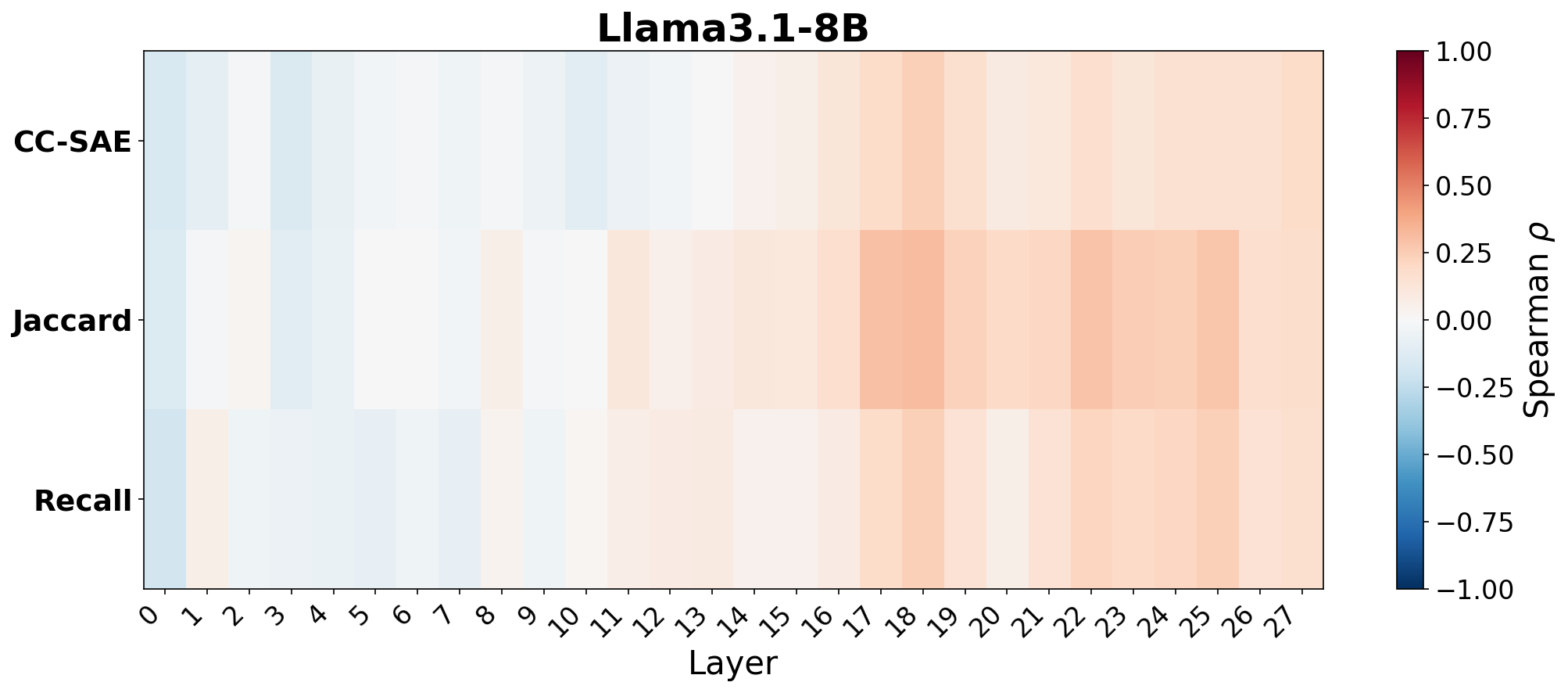}
    \end{subfigure}

    \caption{CoT faithfulness evaluation across layers for different models on \data{GSM8K}.
    \textbf{Left column:} Causal faithfulness ($\Delta p$) measures the probability drop when ablating shared features.
    \textbf{Middle column:} Correlational metrics (CC-SAE, Jaccard, Recall) between prediction and CoT activations.
    \textbf{Right column:} Heatmap showing Spearman correlation between each correlational metric and $\Delta p$ across layers.
    Shaded regions indicate 95\% confidence intervals.}
    \label{fig:ccsae-layers-gsm8k}
\end{figure*}

\begin{table*}[t]
\caption{Faithfulness peak layer, corresponding $\Delta p$ values, relative depth of the peak layer, and averaged $\Delta p$ values across models.}
\label{tab:faithfulness-peak}

\centering
    \begin{tabular}{cccccc}
      \toprule[1.5pt]
      & \textbf{Model} & \textbf{Peak Layer} & \textbf{Peak $\Delta p$} & \textbf{Rel. Dep.} & \textbf{Avg $\Delta p$} \\
    \midrule
    
    \multirow{5}{*}{\rotatebox[origin=c]{90}     {\small\data{LogiQA}}} & \lm{Llama-3.1-8B} & 25 & 0.350 & 93\% & $0.145 \pm 0.099$ \\
    & \lm{Qwen3-1.7B} & 24 & 0.566 & 89\% & $0.512 \pm 0.049$ \\
    & \lm{Qwen3-8B} & 34 & 0.713 & 97\% & $0.642 \pm 0.067$ \\
    & \lm{Gemma-2-2B} & 1 & 0.269 & 4\% & $0.197 \pm 0.045$ \\
    & \lm{Gemma-2-9B} & 31 & 0.412 & 76\% & $0.356 \pm 0.042$ \\
    
    \midrule
    
    \multirow{5}{*}{\rotatebox{90}{\small\data{OpenbookQA}}}
& \lm{Llama-3.1-8B} & 25 & 0.585 & 93\% & $0.275 \pm 0.292$ \\
& \lm{Qwen3-1.7B}   & 19 & 0.694 & 70\% & $0.589 \pm 0.458$ \\
& \lm{Qwen3-8B}    & 27 & 0.872 & 77\% & $0.798 \pm 0.339$ \\
& \lm{Gemma-2-2B}  &  1 & 0.379 & 4\% & $0.282 \pm 0.243$ \\
& \lm{Gemma-2-9B}  & 32 & 0.679 & 78\% & $0.597 \pm 0.280$ \\

\midrule

    \multirow{5}{*}{\rotatebox{90}{\small\data{ARC-Easy}}} & \lm{Llama-3.1-8B} & 24 & 0.805 & 77\% & $0.371 \pm 0.336$ \\
     & \lm{Qwen3-1.7B} & 22 & 0.894 & 81\% & $0.795 \pm 0.377$ \\
     & \lm{Qwen3-8B} & 30 & 0.976 & 86\% & $0.924 \pm 0.214$ \\
     & \lm{Gemma-2-2B} & 0 & 0.615 & 0\% & $0.479 \pm 0.298$ \\
     & \lm{Gemma-2-9B} & 29 & 0.896 & 71\% & $0.810 \pm 0.220$ \\

     \midrule
\multirow{5}{*}{\rotatebox{90}{\small\data{GSM8K}}} & \lm{Llama-3.1-8B} & 5 & 0.062 & 16\% & $0.037 \pm 0.089$ \\       
 & \lm{Qwen3-1.7B} & 6 & 0.187 & 22\% & $0.160 \pm 0.243$ \\
 & \lm{Qwen3-8B} & 19 & 0.197 & 54\% & $0.169 \pm 0.287$ \\
 & \lm{Gemma-2-2B} & 1 & 0.183 & 4\% & $0.124 \pm 0.152$ \\
 & \lm{Gemma-2-9B} & 1 & 0.276 & 2\% & $0.123 \pm 0.184$ \\
        
    \toprule[1.5pt]
    \end{tabular}
\end{table*}

\subsection{Faithfulness Metric Discussion}
We emphasize that our metric operationalizes faithfulness as \textbf{causal cross-path feature reuse}: whether the CoT-conditioned answer pass shares causally important features with the direct-answer pass. Our contribution is methodological, enabling internal comparison where prior work relied on behavioral proxies, rather than claiming to resolve the faithfulness question definitively.

\subsection{Causal Faithfulness Variation} 
Figure~\ref{fig:summary} displays $\Delta p$ variation across relative model depth, defined as the fraction of the current layer over the total model layers. Table~\ref{tab:faithfulness-peak} further reports the peak layer at which each model's internal representations exhibit maximal faithfulness, alongside the corresponding normalized depth. Across most models, faithfulness peaks in the latter half of the network, with normalized depths typically exceeding 70–90\%, suggesting that decisive, faithful representations emerge predominantly in deeper layers. In contrast, \lm{Gemma-2-2B} is a notable outlier, with its faithfulness peak occurring at or near the earliest layers, implying that its most faithful representations are formed very early in the network rather than progressively refined through depth. This divergence highlights substantial variation in how faithfulness develops across layer depth, depending on model architecture and scale.

\subsection{Layer Stability} 
Figure~\ref{fig:stability} shows the layer stability across models. \lm{Gemma-2-9B} and \lm{Qwen3} models exhibit generally high stability, whereas \lm{Llama3.1-8B} demonstrates noticeable variance across different layers, suggesting localized faithfulness. 

\subsection{CoT Faithfulness Evaluation}
\label{app:faithfulness_evaluation}

Figure~\ref{fig:ccsae-layers-app}, \ref{fig:ccsae-layers-openbookqa}, \ref{fig:ccsae-layers-arc_easy}, and \ref{fig:ccsae-layers-gsm8k} show the faithfulness evaluation with three correlational metrics and $\Delta p$ on \data{LogiQA}, \data
\data{OpenbookQA}, \data{ARC-Easy}, and \data{GSM8K}, respectively.

\subsection{Layer-wise Faithfulness Dynamics}
\label{app:repair}

A natural question arising from our layer-wise analysis is whether models exhibit ``\textit{repair}'', i.e., whether faithfulness that degrades in earlier layers can be recovered in subsequent layers. 

\subsubsection{Setup}
We track the evolution of the causal effect ($\Delta p$) across layers for each instance, using the first layer as a baseline. Specifically, for each instance $i$ and layer $\ell$, we compute the growth ratio:
\begin{equation}
    r_i^{(\ell)} = \frac{\Delta p_i^{(\ell)}}{\Delta p_i^{(0)}}
\end{equation}
We then report the median growth ratio across instances at each layer, along with the interquartile range to quantify variability. A ratio greater than 1 indicates increased faithfulness relative to the first layer, while a ratio less than 1 indicates decreased faithfulness. 

\subsubsection{Results}
\begin{figure}[htbp]
    \centering

    \subcaptionbox{\lm{Gemma-2-2B}}{\includegraphics[width=0.49\textwidth]{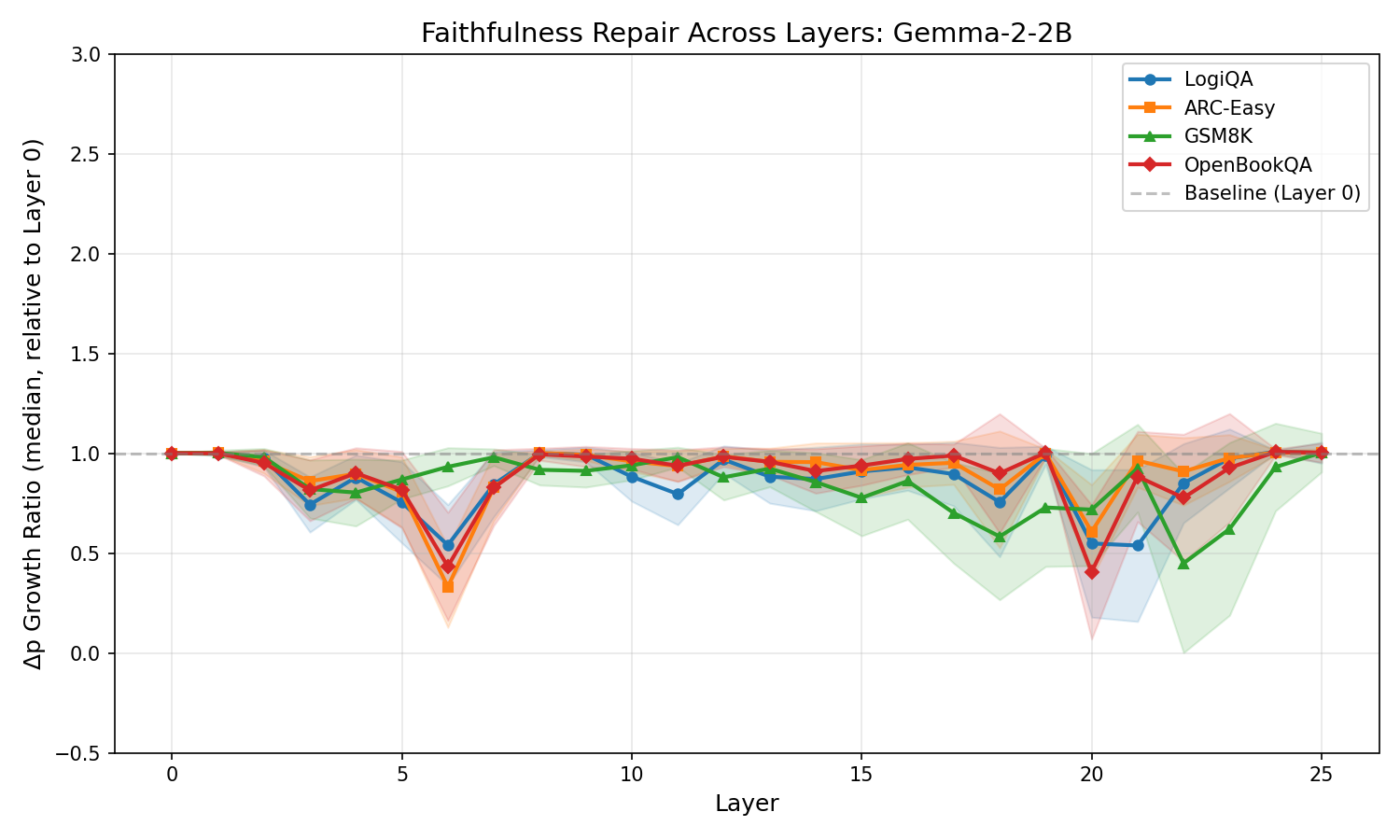}}
    \hfill
    \subcaptionbox{\lm{Gemma-2-9B}}{\includegraphics[width=0.49\textwidth]{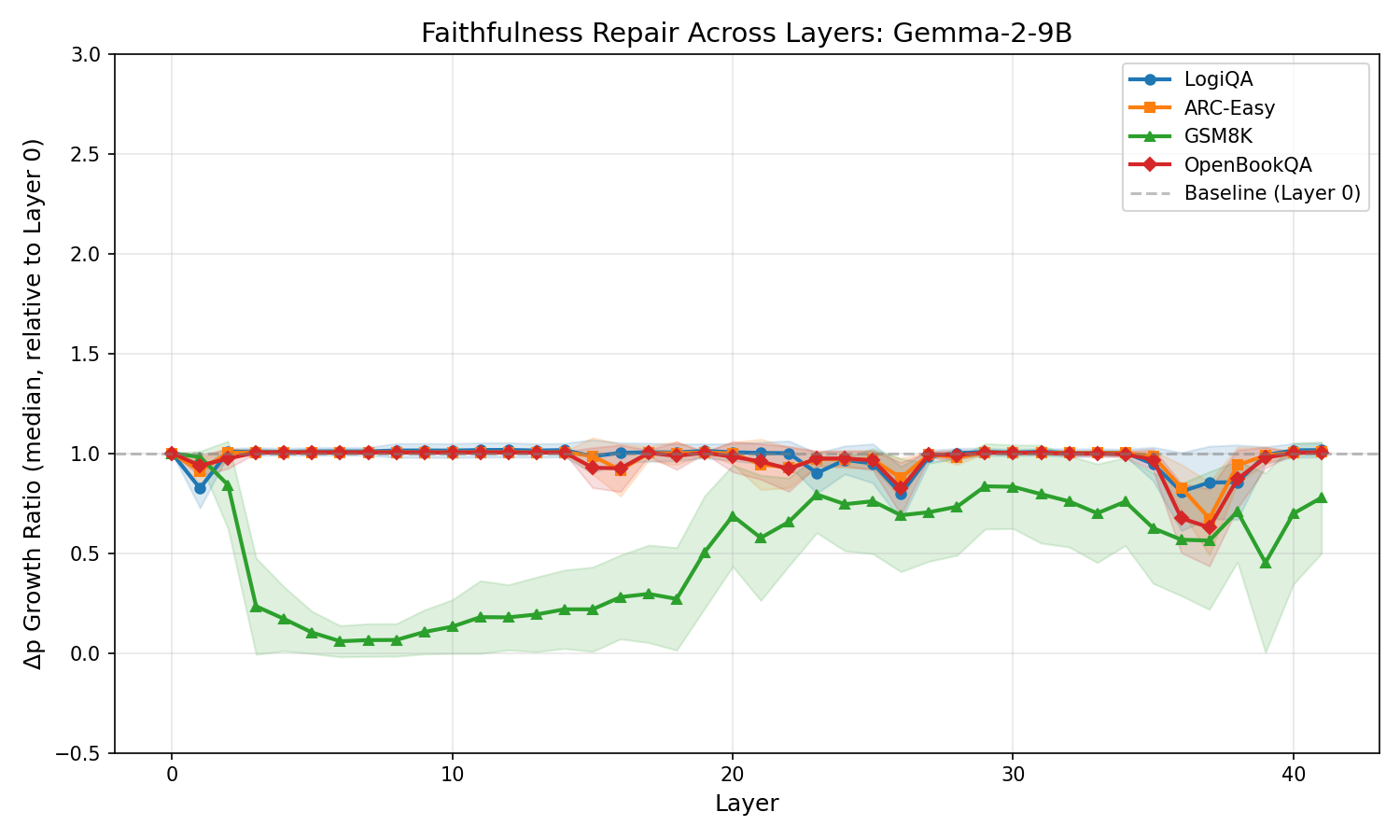}}

    \vspace{0.5cm}   

    \subcaptionbox{\lm{Qwen3-1.7B}}{\includegraphics[width=0.49\textwidth]{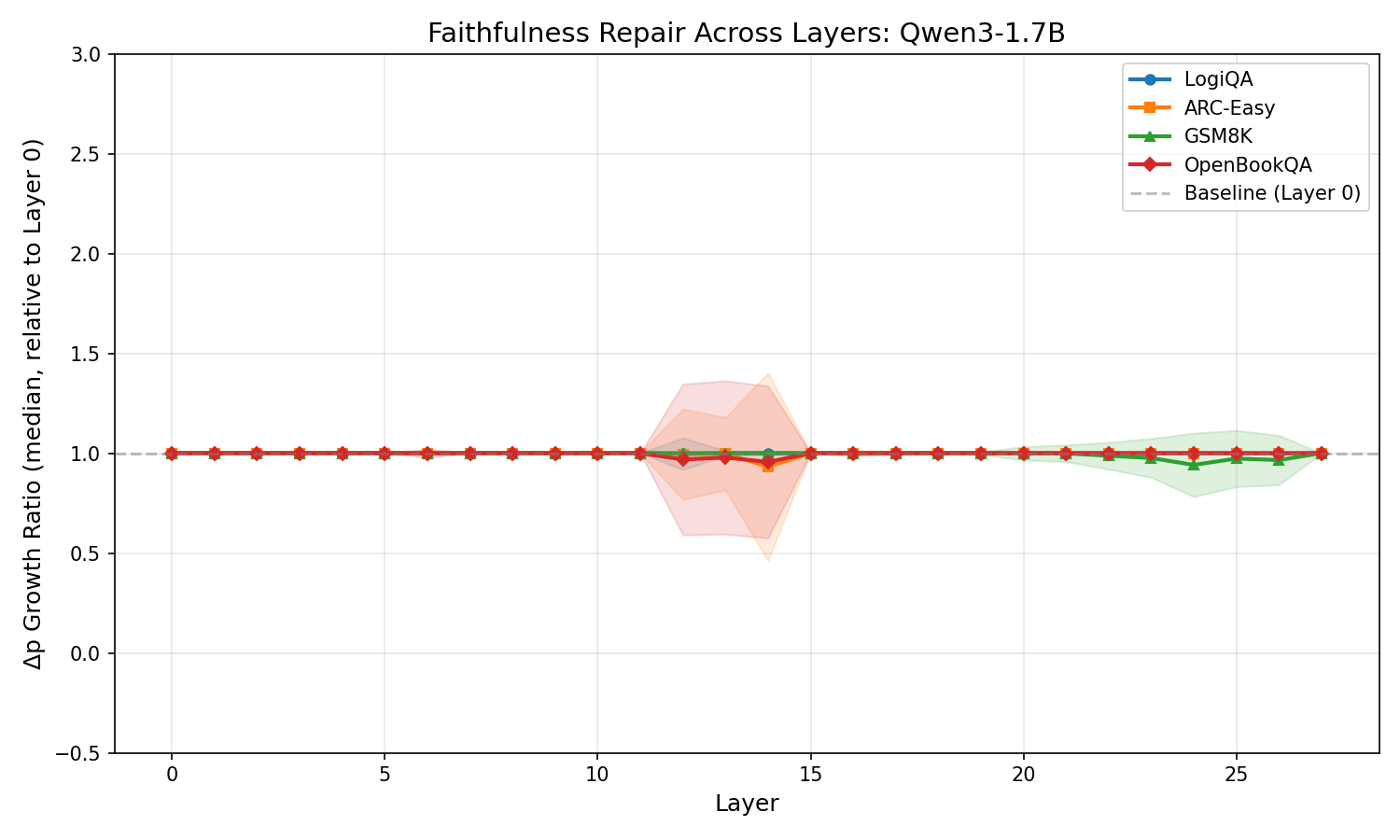}}
    \hfill
    \subcaptionbox{\lm{Qwen3-8B}}{\includegraphics[width=0.49\textwidth]{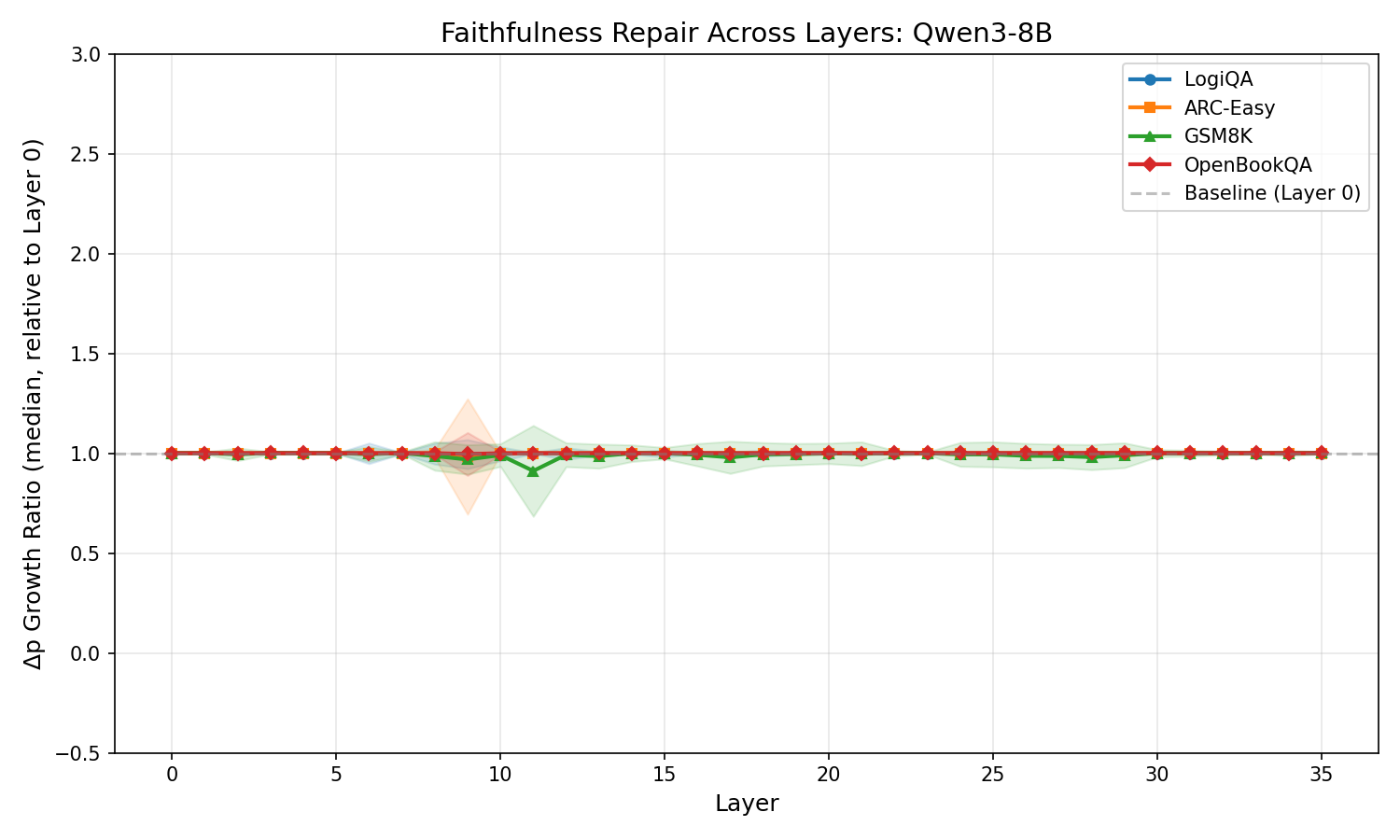}}

    \vspace{0.5cm}   

    \begin{center}
        \subcaptionbox{\lm{Llama-3.1-8B}}{\includegraphics[width=0.5\textwidth]{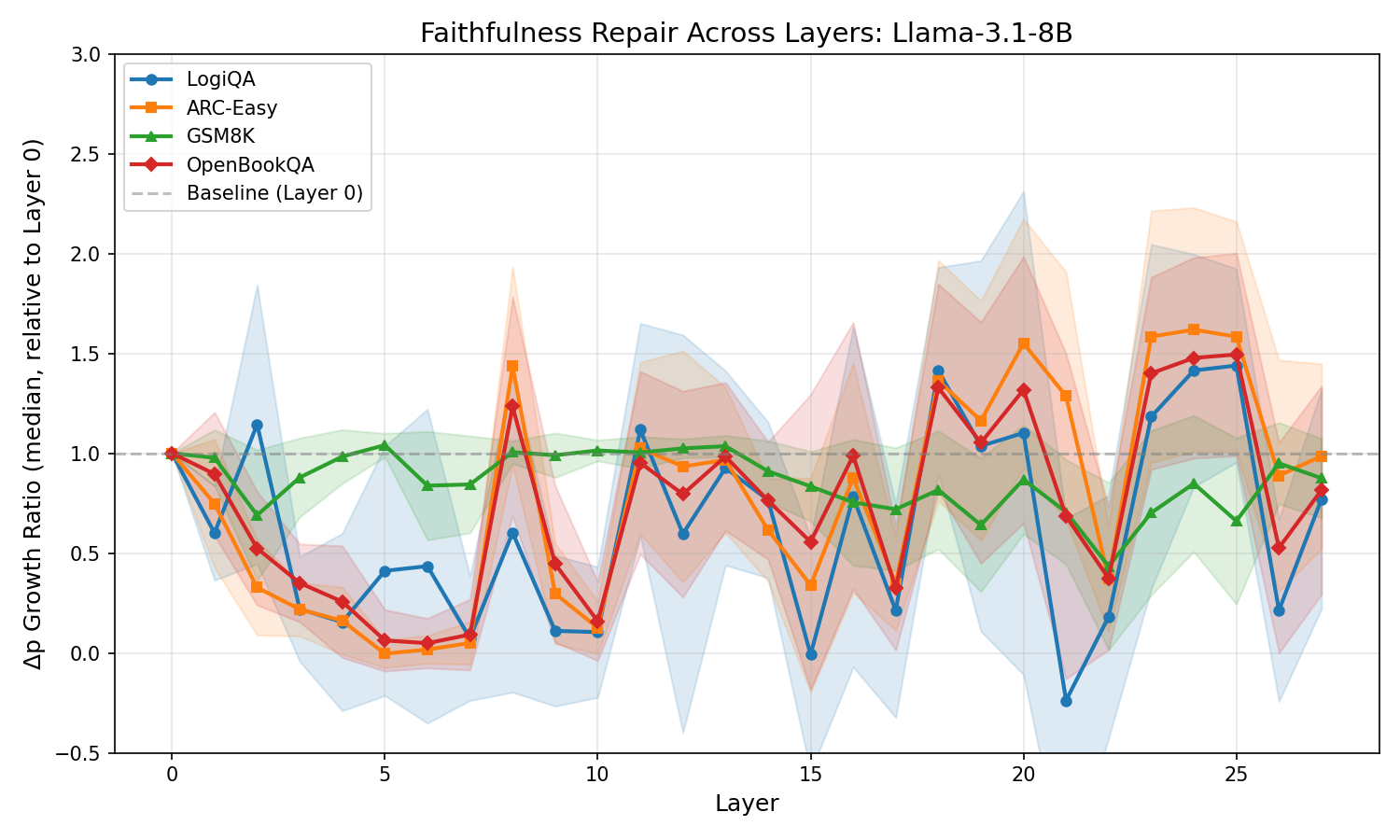}}
    \end{center}

    \caption{Layer-wise faithfulness dynamics across datasets.}
    \label{fig:repair}
\end{figure}

Figure~\ref{fig:repair} presents the layer-wise faithfulness dynamics for five models across four datasets. We observe notable differences across model families. 

\textbf{Gemma models exhibit U-shaped repair patterns.} Both \lm{Gemma-2-2B} and \lm{Gemma-2-9B} show a characteristic pattern where faithfulness drops in the early-to-middle layers before recovering toward the final layers.  For \lm{Gemma-2-2B}, the median $\Delta p$ ratio drops to approximately 0.4--0.5 around layers 5--6 across all datasets, then gradually recovers to $\approx$1.0 by the final layer. This U-shaped trajectory suggests that intermediate layers may perform transformations that temporarily reduce alignment between CoT features and answer-relevant computations, but subsequent layers repair this misalignment.

\textbf{Qwen models maintain stable faithfulness across layers.} In contrast, both \lm{Qwen3-1.7B} and \lm{Qwen3-8B} exhibit remarkably stable $\Delta p$ ratios close to 1.0 throughout all layers. This suggests that whatever causal relationship exists between CoTs and predictions is established early and preserved throughout the forward pass, with minimal perturbation or correction in later layers.

\textbf{\lm{Llama-3.1-8B} shows high layer-wise variability.} \lm{Llama-3.1-8B} exhibits substantial oscillations in the $\Delta p$ ratio across layers, with no clear monotonic trend, aligned with Figure~\ref{fig:stability}. This high variability suggests that the SAE features at different layers may capture qualitatively different aspects of the reasoning process.

\subsection{Implications}
\label{app:implications}
Beyond reporting, the layer-wise profile also carries practical guidance. As peak layers concentrate the causally important shared concepts, probing, steering, and CoT monitoring are best deployed at mid-to-late peak layers rather than at the final layers. This suggests that readouts anchored to last-layer representations may underestimate faithfulness. The distributed-localized faithfulness contrast further yields a testable prediction: models with localized faithfulness should be disproportionately brittle to pruning, quantization, or editing at their few alignment-bearing layers, while distributed models should degrade more gracefully.

\section{Sanity Check}
\subsection{SAE Concept Composition}
\label{app:concept_composition}
Table~\ref{tab:feature_sets} reports the feature set composition across models. The intersection $S_{\cap}$ comprises 33--75\% of $S_{\text{pred}}$, leaving substantial pred-only ($|S_{\text{pred}} \setminus S_{\cap}|$) and CoT-only ($|S_{\text{cot}} \setminus S_{\cap}|$) subsets. For models with top-$k$ SAEs (Qwen3 models), $|S_{\text{pred}}| = |S_{\text{cot}}| = 100$ is set by construction, yet overlap remains around 50\%. The partial overlap indicates that \textit{the two passes activate distinct feature sets despite predicting the same answer, ruling out trivial explanations where $S_{\text{pred}} \approx S_{\text{cot}}$}.

\begin{table*}[t]
\centering

\caption{Feature set statistics across models and datasets (averaged across layers). $S_{\cap}$ comprises a fraction of $S_{\text{pred}}$, with substantial pred-only and CoT-only features.}
\label{tab:feature_sets}

\small
\resizebox{\textwidth}{!}{
\begin{tabular}{llrrrrrr}
\toprule[1.5pt]
\textbf{Dataset} & \textbf{Model} & \textbf{$|S_{\text{pred}}|$} & \textbf{$|S_{\text{cot}}|$} & \textbf{$|S_{\cap}|$} & \textbf{Pred-only} & \textbf{CoT-only} & \textbf{$|S_{\cap}|/|S_{\text{pred}}|$} \\
\midrule

\data{LogiQA} & \lm{Llama-3.1-8B} & 17.6 & 18.6 & 13.4 & 4.2 & 5.2 & 75.3\% \\
 & \lm{Qwen3-1.7B} & 100.0 & 100.0 & 54.9 & 45.1 & 45.1 & 54.9\% \\
 & \lm{Qwen3-8B} & 100.0 & 100.0 & 61.0 & 39.0 & 39.0 & 61.0\% \\
 & \lm{Gemma-2-2B} & 68.2 & 66.8 & 40.6 & 27.6 & 26.2 & 62.2\% \\
 & \lm{Gemma-2-9B} & 93.1 & 93.9 & 56.7 & 36.4 & 37.2 & 63.9\% \\
\midrule
\data{OpenbookQA} & \lm{Llama-3.1-8B} & 18.3 & 18.9 & 12.7 & 5.6 & 6.3 & 68.3\% \\
 & \lm{Qwen3-1.7B} & 100.0 & 100.0 & 48.8 & 51.2 & 51.2 & 48.8\% \\
 & \lm{Qwen3-8B} & 100.0 & 100.0 & 49.5 & 50.5 & 50.5 & 49.5\% \\
 & \lm{Gemma-2-2B} & 74.6 & 69.1 & 37.3 & 37.3 & 31.8 & 53.8\% \\
 & \lm{Gemma-2-9B} & 90.7 & 93.3 & 51.9 & 38.8 & 41.4 & 61.0\% \\

\midrule

\data{ARC-Easy} & \lm{Llama-3.1-8B} & 18.7 & 19.5 & 13.5 & 5.2 & 6.0 & 70.7\% \\
 & \lm{Qwen3-1.7B} & 100.0 & 100.0 & 48.1 & 51.9 & 51.9 & 48.1\% \\
 & \lm{Qwen3-8B} & 100.0 & 100.0 & 51.1 & 48.9 & 48.9 & 51.1\% \\
 & \lm{Gemma-2-2B} & 73.1 & 67.7 & 38.2 & 34.8 & 29.5 & 56.2\% \\
 & \lm{Gemma-2-9B} & 85.3 & 88.7 & 52.8 & 32.5 & 35.9 & 65.8\% \\
\midrule

\data{GSM8K} & \lm{Llama-3.1-8B} & 18.6 & 18.7 & 10.2 & 8.4 & 8.5 & 50.4\% \\
 & \lm{Qwen3-1.7B} & 100.0 & 100.0 & 36.4 & 63.6 & 63.6 & 36.4\% \\
 & \lm{Qwen3-8B} & 100.0 & 100.0 & 33.3 & 66.7 & 66.7 & 33.3\% \\
 & \lm{Gemma-2-2B} & 56.8 & 55.3 & 25.8 & 30.9 & 29.4 & 50.1\% \\
 & \lm{Gemma-2-9B} & 72.1 & 71.1 & 24.9 & 47.2 & 46.2 & 40.4\% \\
 
\toprule[1.5pt]
\end{tabular}
}

\end{table*}

\subsection{Answer Identity}
\label{app:answer_identity}
To rule out the possibility that shared features merely encode the answer token (e.g., A/B/C/D) rather than reflecting genuine reasoning on MCQA datasets, we identify answer-identity features by encoding each letter in isolation.\footnote{\data{GSM8K} is intentionally excluded from this sanity check, as it does not use fixed options.} Specifically,
\begin{enumerate}
    \item \textbf{Isolated-token feature extraction.} For each answer letter $\ell \in \{A, B, C, D\}$, we tokenize $\ell$ in isolation, take the model's residual stream at the target layer and final token position, and SAE-encode it. The set of features with nonzero activation, $F_{\ell}$, is the answer-identity feature set for letter $\ell$; this is computed once per dataset--model pair and reused across items.
    \item \textbf{Per-item prediction pass.} For each item, we build the prediction-only prompt (no CoT), take the model's greedy answer $y = \arg\max_{\ell \in \{A,B,C,D\}} \text{logit}(\ell)$ at the final position, and extract $S_{\text{pred}}$. We use $S_{\text{pred}}$ rather than $S_{\cap}$ as the ablated set in this control, since it isolates the identity confound from the prediction pass directly, independent of whether the CoT also activates the same identity features.
    \item \textbf{Targeted ablation.} We compute the causal-faithfulness effect (\S\ref{subsec:causal}) twice: once ablating the full $S_{\text{pred}}$ ($\Delta p_{\text{full}}$), and once ablating $S_{\text{pred}} \setminus F_{y}$, i.e., $S_{\text{pred}}$ with the letter's own identity features removed ($\Delta p_{\text{no-pred}}$), i.e., we ablate $F_\ell$ -- all four letters' feature sets.
    \item \textbf{Aggregation.} We average $\Delta p_{\text{full}}$ and $\Delta p_{\text{no-pred}}$ across items per dataset--model pair and report retention as the ratio of means, $\overline{\Delta p_{\text{no-pred}}} / \overline{\Delta p_{\text{full}}}$.
\end{enumerate}

\paragraph{Results.}Table~\ref{tab:answer-identity} shows that after removing these features from $S_\cap$, the causal effect remains nearly unchanged (mean retention = $0.997 \pm 0.007$). Furthermore, the number of option-related features contained in $S_\cap$ is very limited. Together, these observations indicate that \textit{$\Delta p$ reflects reasoning-related features, rather than superficial answer encoding.}


\begin{table}[t]
\centering
\caption{Answer-identity ablation results. $\delta_X = \Delta p_{\text{full}} - \Delta p_{\text{no-}X}$ is the drop in causal effect after removing option $X$'s answer-identity features; $\#X$ is the number of option-$X$ features in $S_\cap$ that are responsible for outputting the answer corresponding to one of the predefined option letters. Avg.\ retention $\approx 1$ indicates the causal effect is driven by reasoning features, not answer-encoding.}
\label{tab:answer-identity}
\resizebox{\textwidth}{!}{

\begin{tabular}{llcccccc|ccccc}
\toprule
\textbf{Dataset} & \textbf{Model} & $\Delta p_{\text{full}}$ & $\delta_A$ & $\delta_B$ & $\delta_C$ & $\delta_D$ & \textbf{Avg.\ Ret.} & $\#A$ & $\#B$ & $\#C$ & $\#D$ & $|S_\cap|$ \\
\midrule
\multirow{5}{*}{\data{ARC-Easy}} & \lm{Gemma-2-2B} & 0.6812 & 0.0000 & 0.0000 & 0.0000 & 0.0000 & 1.0000 & 8.8 & 7.3 & 6.7 & 7.4 & 89 \\
 & \lm{Gemma-2-9B} & 0.9126 & 0.0045 & 0.0000 & 0.0000 & -0.0000 & 0.9985 & 4.0 & 2.4 & 2.3 & 4.4 & 91 \\
 & \lm{Llama-3.1-8B} & 0.8513 & 0.0001 & 0.0052 & 0.0030 & 0.0036 & 0.9963 & 4.9 & 2.1 & 5.7 & 4.0 & 37 \\
 & \lm{Qwen3-1.7B} & 0.9898 & -0.0000 & -0.0000 & -0.0000 & -0.0000 & 1.0000 & 4.5 & 5.6 & 5.6 & 5.4 & 100 \\
 & \lm{Qwen3-8B} & 0.9879 & 0.0000 & 0.0000 & 0.0000 & 0.0000 & 1.0000 & 3.2 & 4.6 & 4.7 & 4.6 & 100 \\
\midrule
\multirow{5}{*}{\data{LogiQA}} & \lm{Gemma-2-2B} & 0.3349 & 0.0000 & 0.0000 & 0.0000 & 0.0000 & 1.0000 & 8.7 & 6.3 & 5.2 & 5.5 & 84 \\
 & \lm{Gemma-2-9B} & 0.5236 & 0.0052 & 0.0000 & 0.0000 & 0.0000 & 0.9980 & 3.3 & 2.3 & 3.1 & 5.0 & 92 \\
 & \lm{Llama-3.1-8B} & 0.5177 & 0.0004 & 0.0140 & 0.0122 & 0.0116 & 0.9809 & 3.9 & 3.1 & 3.5 & 3.7 & 38 \\
 & \lm{Qwen3-1.7B} & 0.9129 & -0.0000 & -0.0000 & -0.0000 & -0.0000 & 1.0000 & 4.4 & 5.5 & 5.5 & 5.2 & 100 \\
 & \lm{Qwen3-8B} & 0.9053 & 0.0000 & 0.0000 & 0.0000 & 0.0000 & 1.0000 & 3.8 & 6.0 & 6.3 & 5.7 & 100 \\
\midrule
\multirow{5}{*}{\data{OpenBookQA}} & \lm{Gemma-2-2B} & 0.5020 & 0.0000 & 0.0000 & 0.0000 & 0.0000 & 1.0000 & 8.5 & 6.9 & 5.9 & 6.6 & 90 \\
 & \lm{Gemma-2-9B} & 0.7656 & 0.0223 & 0.0000 & 0.0000 & 0.0000 & 0.9923 & 2.9 & 2.2 & 2.4 & 3.8 & 83 \\
 & \lm{Llama-3.1-8B} & 0.7337 & 0.0002 & 0.0092 & 0.0050 & 0.0060 & 0.9924 & 4.8 & 2.1 & 5.7 & 4.1 & 38 \\
 & \lm{Qwen3-1.7B} & 0.9576 & -0.0000 & -0.0000 & -0.0000 & -0.0000 & 1.0000 & 4.3 & 5.2 & 5.2 & 5.1 & 100 \\
 & \lm{Qwen3-8B} & 0.9771 & 0.0000 & 0.0000 & 0.0000 & 0.0000 & 1.0000 & 3.2 & 4.5 & 4.6 & 4.6 & 100 \\
\bottomrule
\end{tabular}
}
\end{table}



\subsection{Robustness to Ubiquitous Features}
\label{app:ubiquitous}
$S_{\cap}$ may partly consist of \text{ubiquitous} features, i.e., generic multiple-choice artifacts that recur across nearly all items regardless of content, rather than item-specific reasoning concepts. If such features drive a substantial share of $\Delta p$, the causal effect would reflect superficial task formatting rather than genuine CoT-driven reasoning. For each dataset--model pair, we identify ubiquitous features and quantify their contribution to $\Delta p$ in two passes:
\begin{enumerate}
    \item \textbf{Pass 1: frequency collection.} For every item $i$, we extract $S_{\cap}^{(i)} = S_{\text{pred}}^{(i)} \cap S_{\text{cot}}^{(i)}$ (\S\ref{subsec:causal}) and accumulate a corpus-level frequency count $\text{count}(f) = |\{i : f \in S_{\cap}^{(i)}\}|$ for every feature $f$ observed in any item's intersection set.
    \item \textbf{Thresholding.} Given $N$ items with a valid intersection set, we define the ubiquitous set $U_{\tau} = \{f : \text{count}(f)/N > \tau\}$ for a chosen frequency threshold $\tau \in (0, 1)$.
    \item \textbf{Pass 2: ablation comparison.} For every item, we recompute the causal-faithfulness intervention (\S\ref{subsec:causal}) three times, ablating (i) the full intersection $S_{\cap}^{(i)}$, (ii) the intersection with ubiquitous features removed, $S_{\cap}^{(i)} \setminus U_{\tau}$, and (iii) the ubiquitous features alone, $S_{\cap}^{(i)} \cap U_{\tau}$, yielding $\Delta p_{\text{full}}$, $\Delta p_{\text{no-gen}}$, and $\Delta p_{\text{gen}}$ per item.
    \item \textbf{Aggregation.} We average each quantity across items within a dataset--model pair and report \textbf{retention} as
    \[
    \text{Retention} = \frac{\overline{\Delta p_{\text{no-gen}}}}{\overline{\Delta p_{\text{full}}}},
    \]
\end{enumerate}
Retention $\approx 1$ indicates the causal effect is largely content-specific, while retention $\ll 1$ indicates it is substantially driven by generic, ubiquitous features. 

\paragraph{Results.}
Table~\ref{tab:ubiquitous} reports $\Delta p_{\text{full}}$, $\Delta p_{\text{no-gen}}$, $\Delta p_{\text{gen}}$, and retention for each dataset--model pair. \lm{Qwen3-1.7B} and \lm{Qwen3-8B} retain $1.00$ of the causal effect after removing ubiquitous features across all three datasets, indicating their $S_{\cap}$-driven causal effect is content-specific. \lm{Gemma-2-2B} similarly retains $0.88$--$1.28$. \lm{Llama-3.1-8B} and \lm{Gemma-2-9B} show more moderate retention. Averaged across all 15 dataset--model pairs, $87\%$ of the causal effect measured by $\Delta p$ survives removal of ubiquitous features, indicating that \textit{the causal effect of $S_{\cap}$ predominantly reflects content-specific concept reuse rather than generic MCQA artifacts. }



\begin{table*}[t]
\centering
\small
\caption{Ubiquitous-feature ablation at threshold $\tau=0.9$. $N_{\text{ubiq}}/N_{\text{total}}$: number of features flagged ubiquitous (frequency $>\tau$ across items) out of all features observed in any $S_\cap$. Retention $= \overline{\Delta p_{\text{no-gen}}} / \overline{\Delta p_{\text{full}}}$; retention $\approx 1$ indicates the causal effect is content-specific rather than driven by ubiquitous features.}
\label{tab:ubiquitous}
\resizebox{\textwidth}{!}{

\begin{tabular}{llccccc}
\toprule
\textbf{Dataset} & \textbf{Model} & $N_{\text{ubiq}}/N_{\text{total}}$ & $\Delta p_{\text{full}}$ & $\Delta p_{\text{no-gen}}$ & $\Delta p_{\text{gen}}$ & \textbf{Retention} \\
\midrule
\data{ARC-Easy} & \lm{Gemma2-2B} & 24/296 & $0.49 \pm 0.39$ & $0.44 \pm 0.29$ & $0.11 \pm 0.44$ & $0.88$ \\
 & \lm{Gemma2-9B} & 23/476 & $0.88 \pm 0.17$ & $0.61 \pm 0.33$ & $0.83 \pm 0.30$ & $0.69$ \\
 & \lm{Llama3.1-8B} & 9/53 & $0.46 \pm 0.23$ & $0.30 \pm 0.19$ & $0.28 \pm 0.17$ & $0.64$ \\
 & \lm{Qwen3-1.7B} & 14/382 & $0.81 \pm 0.37$ & $0.81 \pm 0.35$ & $0.00 \pm 0.04$ & $1.00$ \\
 & \lm{Qwen3-8B} & 14/402 & $0.97 \pm 0.14$ & $0.97 \pm 0.14$ & $0.19 \pm 0.25$ & $1.00$ \\
\midrule
\data{LogiQA} & \lm{Gemma2-2B} & 18/206 & $0.18 \pm 0.21$ & $0.18 \pm 0.08$ & $-0.04 \pm 0.31$ & $1.00$ \\
 & \lm{Gemma2-9B} & 22/360 & $0.41 \pm 0.21$ & $0.20 \pm 0.16$ & $0.39 \pm 0.21$ & $0.49$ \\
 & \lm{Llama3.1-8B} & 10/48 & $0.20 \pm 0.15$ & $0.06 \pm 0.14$ & $0.13 \pm 0.10$ & $0.30$ \\
 & \lm{Qwen3-1.7B} & 14/277 & $0.53 \pm 0.45$ & $0.53 \pm 0.44$ & $0.02 \pm 0.09$ & $1.00$ \\
 & \lm{Qwen3-8B} & 16/317 & $0.70 \pm 0.37$ & $0.70 \pm 0.37$ & $0.19 \pm 0.28$ & $1.00$ \\
\midrule
\data{OpenbookQA} & \lm{Gemma2-2B} & 23/262 & $0.31 \pm 0.31$ & $0.31 \pm 0.23$ & $-0.07 \pm 0.35$ & $1.00$ \\
 & \lm{Gemma2-9B} & 20/495 & $0.66 \pm 0.28$ & $0.41 \pm 0.31$ & $0.48 \pm 0.38$ & $0.63$ \\
 & \lm{Llama3.1-8B} & 9/45 & $0.32 \pm 0.23$ & $0.22 \pm 0.21$ & $0.20 \pm 0.16$ & $0.68$ \\
 & \lm{Qwen3-1.7B} & 17/278 & $0.61 \pm 0.46$ & $0.61 \pm 0.44$ & $0.01 \pm 0.06$ & $1.00$ \\
 & \lm{Qwen3-8B} & 15/346 & $0.82 \pm 0.34$ & $0.82 \pm 0.34$ & $0.11 \pm 0.23$ & $1.00$ \\

\bottomrule
\end{tabular}
}
\end{table*}

\subsection{Sufficiency Test}
\label{app:sufficiency}
To complement our necessity analysis, which shows that ablating $S_{\cap}$ reduces answer probability in \S\ref{subsec:layerwise}, we additionally conduct a sufficiency test to answer the question: can $S_{\cap}$ alone largely recover the model's answer? Specifically, we define three conditions:
\begin{enumerate}
    \item \textbf{Base}: No intervention; measure $p_{\text{base}} = p(\text{answer} \mid \mathbf{r}_\text{pred})$;
    \item \textbf{Zeroed}: Replace $\mathbf{r}_\text{pred}$ with $\mathbf{0}$; measure $p_{\text{zero}}$;
    \item \textbf{Patched}: Replace $\mathbf{r}_\text{pred}$ with only the $S_{\cap}$ contribution: $\mathbf{r}_{\text{patch}} = \text{Dec}(\mathbf{a}_\text{pred} \odot \mathbf{m}_{S_{\cap}})$, 
    where $\mathbf{m}_{S_{\cap}}$ is a binary mask selecting only shared features; measure $p_{\text{patch}}$.
\end{enumerate}
Recall that $\mathbf{r}_\text{pred} \in \mathbb{R}^\text{model}$ denotes the residual stream activation and $\mathbf{a}_\text{pred} = \text{Enc}(\mathbf{r}_\text{pred})$ represents the corresponding SAE latent activation (\S\ref{subsec:causal}).

We quantify sufficiency via the \textit{recovery} metric:
\begin{equation}
    \text{Recovery} = \frac{p_{\text{patch}} - p_{\text{zero}}}{p_{\text{base}} - p_{\text{zero}}}
\end{equation}
A recovery of 1.0 indicates that $S_{\cap}$ fully restores the original answer probability
. To verify that $S_{\cap}$ is \textit{specifically} sufficient (not just any $k$ features), we compare against a random baseline: for each example, we sample $k = |S_{\cap}|$ random features from $S_{\text{pred}}$, compute recovery, and repeat 20 times. If $S_{\cap}$ captures causally important features, its recovery should noticeably exceed the random baseline.

\paragraph{Results.}
Table~\ref{tab:sufficiency} presents the sufficiency results across all model--dataset combinations. On average, patching only $S_{\cap}$ achieves a recovery of $1.05 \pm 0.09$, indicating that these shared features not only restore but sometimes \emph{exceed} the original answer probability, likely because the patched representation removes interfering non-shared features. In contrast, random features of equal size achieve only $0.45 \pm 0.35$ recovery, with a substantial gap of $0.60$.


Together, these findings suggest that \textbf{$S_{\cap}$ is both necessary and sufficient}: removing it disrupts the answer probability, and retaining it alone largely recovers the answer compared to the random baseline, indicating that the shared concepts between prediction and CoT passes capture the causally relevant reasoning.

\begin{table*}[t]
\centering
\caption{Sufficiency test results: patching only $S_{\cap}$ features into zeroed activations. Recovery measures how much of the original answer probability is restored. $S_{\cap}$ noticeably outperforms random features of the same size.}
\label{tab:sufficiency}
\small
\begin{tabular}{llrrrr}
\toprule[1.5pt]
\textbf{Dataset} & \textbf{Model} & \textbf{$S_{\cap}$ Recov.} & \textbf{Rand Recov.} & \textbf{Gap} & \textbf{$S_{\cap}$ > Rand} \\
\midrule
\multirow{5}{*}{\data{LogiQA}} & \lm{Llama-3.1-8B} & 1.018 & 0.919 & 0.099 & 63.5\% \\
 & \lm{Qwen3-1.7B} & 0.978 & 0.068 & 0.910 & 83.9\% \\
 & \lm{Qwen3-8B} & 1.147 & 0.081 & 1.066 & 97.3\% \\
 & \lm{Gemma-2-2B} & 1.000 & 0.711 & 0.289 & 65.5\% \\
 & \lm{Gemma-2-9B} & 0.950 & 0.750 & 0.200 & 60.5\% \\
\midrule
\multirow{5}{*}{\data{OpenbookQA}} & \lm{Llama-3.1-8B} & 1.079 & 0.939 & 0.140 & 75.5\% \\
 & \lm{Qwen3-1.7B} & 1.213 & 0.140 & 1.073 & 92.8\% \\
 & \lm{Qwen3-8B} & 1.226 & 0.125 & 1.101 & 97.3\% \\
 & \lm{Gemma-2-2B} & 0.954 & 0.782 & 0.172 & 58.5\% \\
 & \lm{Gemma-2-9B} & 0.992 & 0.447 & 0.545 & 90.0\% \\
\midrule
\multirow{5}{*}{\data{ARC-Easy}} & \lm{Llama-3.1-8B} & 1.075 & 0.923 & 0.152 & 88.5\% \\
 & \lm{Qwen3-1.7B} & 1.085 & 0.084 & 1.001 & 97.7\% \\
 & \lm{Qwen3-8B} & 1.045 & 0.077 & 0.967 & 100.0\% \\
 & \lm{Gemma-2-2B} & 0.996 & 0.502 & 0.494 & 78.0\% \\
 & \lm{Gemma-2-9B} & 0.950 & 0.136 & 0.814 & 96.5\% \\
\toprule[1.5pt]
\end{tabular}
\end{table*}

\section{Alternatives to Measuring Faithfulness}
\label{app:alternative}

\paragraph{Lack of Ground-truth.} We want to highlight that there is no gold metric for measuring CoT faithfulness in the absence of a ground truth for the model's underlying reasoning \citep{parcalabescu-frank-2024-measuring, sun2026investigatinginterplaycontextualparametric}. Therefore, the existing metrics are used only as proxies for faithfulness. 

\paragraph{Infeasibility of Human Evaluation.} The effectiveness of human evaluation depends on the assumption that people can meaningfully assess whether a CoT faithfully represents a model’s internal reasoning. Yet this assumption becomes questionable in the context of modern LLMs. With billions of parameters composing intricate, nonlinear interactions that map inputs to outputs, the actual computational pathways are essentially opaque to human observers. We currently lack both the tools and the theoretical understanding required to trace how these vast networks collectively produce any given prediction. As a result, human judgments of CoT faithfulness, though intuitively compelling, may be methodologically unsound given our present limited visibility into LLM internals.

\paragraph{Contextual Metrics.} We consider contextual CoT faithfulness metrics from \citep{lanham2023measuring}, which are based on context perturbation: (i) \textbf{Early Answering}, which truncates the CoT before the original answer is produced and forces an early response; (ii) \textbf{Adding Mistake}, which injects an error into one CoT step and regenerates the remaining steps; (iii) \textbf{Paraphrasing}, which restates the initial portion of the CoT while preserving its meaning and regenerates the rest; and (iv) \textbf{Filler Token}, which substitutes the CoT with meaningless symbols such as ellipses. The results of these metrics are binary, i.e., faithful or unfaithful, which is coarse-grained compared to our proposed metrics, which range from 0 to 1. Therefore, we follow \citet{zaman-srivastava-2025-causal} and instead measure the prediction score $\hat{z}_i$ changes, ranging from -1 to 1. 

\begin{align}
    \hat{z}_i = \frac{\exp(p_\theta(L_i \mid x))}{\sum_{L_j \in L} \exp(p_\theta(L_j \mid x))}
\end{align}
where $L$ denotes the set of possible labels and $x$ represents the input text.

The hint injection method \citep{turpin2023language} is intentionally excluded. It produces binary labels (faithful vs. unfaithful), whereas our metric yields continuous values. As a result, we cannot compute a meaningful rank (Spearman) correlation. 

\paragraph{Parametric Metric.} Additionally, we follow \citet{tutek-etal-2025-measuring, sun2026investigatinginterplaycontextualparametric} to assess faithfulness through interventions on the parametric knowledge of model $\mathcal{M}$. Specifically, given an input $x$, the target CoT $c$ is first decomposed into $T$ reasoning steps $c=(s_1,...,s_T)$. FUR provides a continuous answer as to whether a CoT is faithful by measuring the extent to which unlearning the knowledge used by models in CoT steps causes model to change their output prediction:
\begin{align}
    \mathrm{FUR} = \frac{\sum_{i=1}^{T} \mathbb{I}[s_i \text{ such that } y \neq y^{(i)*}]}{T}
\end{align}
where $s_i$ is the $i$-th reasoning step, $y^{(i)*}$ denotes the prediction made by $\mathcal{M}^{(i)*}$ after unlearning the knowledge involved in the $i$-th reasoning step, and $\mathbb{I}(\cdot)$ represents the indicator function. 

\paragraph{Circuit-guided Internal-External Discrepancy (CIE).}
CIE compares the model's internal computation against its displayed reasoning \citep{shen2026detectingunfaithfulchainofthoughtcircuitguided}. For each trace, an internal representation is created by tracing sentence-level attribution circuits over informative tokens, and an external representation from the hidden states of the reasoning sentences is extracted and organized into sentence graphs. Their structural discrepancy is measured via the Fused Gromov-Wasserstein distance. A larger CIE score indicates greater internal-external divergence and thus lower faithfulness.

\begin{table*}[h]
\centering
\caption{Existing faithfulness metrics results with \lm{Llama3.1-8B} on \data{LogiQA}.}
\label{tab:faithful_metric_corr}

\begin{tabular}{lccc}
\toprule[1.5pt]
\textbf{Metric} & \textbf{Mean} & \textbf{Std} & \textbf{95\% CI} \\
\midrule
Early Answering \citep{zaman-srivastava-2025-causal} & 0.6469 & 0.1119 & [0.6371, 0.6567] \\
Filler Tokens \citep{zaman-srivastava-2025-causal} & 0.6476 & 0.1118 & [0.6378, 0.6574] \\
Adding Mistakes \citep{zaman-srivastava-2025-causal} & 0.3953 & 0.1259 & [0.3843, 0.4064] \\
Paraphrasing \citep{zaman-srivastava-2025-causal} & -0.0447 & 0.0568 & [-0.0497, -0.0398]\\
FUR \citep{tutek-etal-2025-measuring} & 0.2217 & 0.1710 & [0.1876, 0.2558] \\
CIE \citep{shen2026detectingunfaithfulchainofthoughtcircuitguided} & 0.6160 & 0.0098 & [0.6139, 0.6181] \\
\midrule
CC-SAE & 0.9142 & 0.0251 & [0.9120, 0.9164] \\
Jaccard & 0.6145 & 0.0588 & [0.6093, 0.6196] \\
Recall & 0.8580 & 0.0456 & [0.8540, 0.8620] \\
$\Delta p$ & 0.1450 & 0.0990 & [0.1400, 0.1585] \\
\toprule[1.5pt]
\end{tabular}

\end{table*}

\paragraph{Relationship to Prior Metrics.} This comparison serves as a sanity check to verify that our metric captures information not already redundant with existing approaches, rather than establishing superiority over prior methods, because no ground-truth can be provided as discussed in the prior paragraph. We only conduct a small-scale comparison with \lm{Llama3.1-8B} on \data{LogiQA}, as most metrics are computationally too expensive for extensive evaluation (for example, per-instance faithfulness using FUR with \lm{Llama3.1-8B} takes approximately 25 minutes). Therefore, we defer a more systematic and comprehensive comparison to future work. Table~\ref{tab:faithful_metric_corr} reports correlations between our metrics and existing faithfulness measures.  We observe slightly positive Spearman correlations between $\Delta p$ and contextual metrics ($\rho \approx 0.10$), while noting weak-to-moderate correlations with the parametric faithfulness metric FUR ($\rho=0.25$) and the MI-based metric CIE score ($\rho=0.30$). The weak-to-moderate correlations are \textit{expected}, not a deficiency of either approach, aligned with \citet{parcalabescu-frank-2024-measuring, gurarieh2026faithfulnessmetricsdontmeasure}. Prior metrics probe different dimensions: contextual metrics test sensitivity to textual perturbations; FUR tests dependence on unlearned knowledge; CIE measures circuit-trace discrepancy. Our $\Delta p$ tests whether \textit{shared internal features} causally drive the answer. These dimensions are conceptually orthogonal, i.e., a CoT can be contextually robust yet rely on different internal features, or vice versa. The low correlations confirm that our metric captures a complementary signal rather than redundantly measuring what existing methods already measure. Together, these metric families provide a more comprehensive assessment than any single approach.

\section{Causal Validation via Conditions}

\subsection{Prediction Pass Ablation}

\subsubsection{Three Base Conditions}
Table~\ref{tab:baseline} shows the complete raw values of three base conditions (Conditions A--C) for all models and datasets. Conditions A--C yield $\Delta p \approx 0$ across all settings, while $\Delta p(S_{\cap})$ ranges from 0.1 to 0.9. This demonstrates that the causal effect is specific to intersection features rather than an artifact of set size or ablation magnitude. Even norm-matched random sampling (Condition C) fails to replicate the effect, indicating that $S_{\cap}$ captures semantically meaningful features beyond what magnitude alone would predict.

We further observe that \lm{Llama-3.1-8B} conditions are magnitude-wise notably smaller than those of other models and exhibit markedly lower variance than those of the other models, indicating that random ablations have essentially no effect on \lm{Llama-3.1-8B}.

\begin{table*}[t]
  \centering
    \caption{Prediction pass conditions and top-$k$ upper bound comparison against $\Delta p$ ($S_{\cap}$) across models.}
  \label{tab:baseline}
  \resizebox{\textwidth}{!}{

\renewcommand*{\arraystretch}{1}
  
  \begin{tabular}{lccccc}
  \toprule[1.5pt]
 & \textbf{Model} & \textbf{Condition A} & \textbf{Condition B} & \textbf{Condition C} & \textbf{top-$k$ upper bound} \\
  
  \midrule
  
  \multirow{5}{*}{\rotatebox[origin=c]{90}{\small\data{LogiQA}}} & \lm{Llama-3.1-8B} & $0.045 \pm 0.150$ & $0.068 \pm 0.160$ & $0.065 \pm 0.141$ & $0.156 \pm 0.204$ \\
  
   & \lm{Qwen3-1.7B} & $0.142 \pm 0.311$ & $0.178 \pm 0.334$ & $0.110 \pm 0.308$ & $0.530 \pm 0.442$ \\
  
   & \lm{Qwen3-8B} & $0.156 \pm 0.302$ & $0.172 \pm 0.308$ & $0.141 \pm 0.319$ & $0.678 \pm 0.378$ \\
  
   & \lm{Gemma-2-2B} & $0.034 \pm 0.105$ & $0.034 \pm 0.130$ & $0.021 \pm 0.121$ & $0.222 \pm 0.100$ \\
  
   & \lm{Gemma-2-9B} & $0.022 \pm 0.109$ & $0.035 \pm 0.112$ & $0.052 \pm 0.124$ & $0.378 \pm 0.203$ \\
  
  \midrule

  \multirow{5}{*}{\rotatebox[origin=c]{90}{\small\data{OpenbookQA}}} & \lm{Llama-3.1-8B} & $0.075 \pm 0.167$ & $0.147 \pm 0.192$ & $0.140 \pm 0.169$ & $0.286 \pm 0.289$ \\
  
   & \lm{Qwen3-1.7B} & $0.135 \pm 0.323$ & $0.165 \pm 0.349$ & $0.072 \pm 0.279$ & $0.588 \pm 0.464$ \\
  
   & \lm{Qwen3-8B} & $0.164 \pm 0.323$ & $0.170 \pm 0.327$ & $0.166 \pm 0.324$ & $0.804 \pm 0.347$ \\
  
   & \lm{Gemma-2-2B} & $0.057 \pm 0.122$ & $0.057 \pm 0.129$ & $0.039 \pm 0.111$ & $0.351 \pm 0.213$ \\
  
   & \lm{Gemma-2-9B} & $0.035 \pm 0.138$ & $0.071 \pm 0.150$ & $0.067 \pm 0.121$ & $0.614 \pm 0.273$ \\
  
  \midrule

  \multirow{5}{*}{\rotatebox[origin=c]{90}{\small\data{ARC-Easy}}} & \lm{Llama-3.1-8B} & $0.098 \pm 0.173$ & $0.195 \pm 0.203$ & $0.173 \pm 0.184$ & $0.366 \pm 0.330$ \\
  
   & \lm{Qwen3-1.7B} & $0.148 \pm 0.311$ & $0.189 \pm 0.348$ & $0.066 \pm 0.224$ & $0.766 \pm 0.399$ \\
  
   & \lm{Qwen3-8B} & $0.176 \pm 0.313$ & $0.178 \pm 0.309$ & $0.139 \pm 0.282$ & $0.935 \pm 0.195$ \\
  
   & \lm{Gemma-2-2B} & $0.082 \pm 0.154$ & $0.088 \pm 0.167$ & $0.076 \pm 0.181$ & $0.494 \pm 0.270$ \\
  
   & \lm{Gemma-2-9B} & $0.046 \pm 0.141$ & $0.066 \pm 0.122$ & $0.069 \pm 0.127$ & $0.819 \pm 0.218$ \\
  
  \midrule

  \multirow{5}{*}{\rotatebox[origin=c]{90}{\small\data{GSM8K}}} & \lm{Llama-3.1-8B} & $0.019 \pm 0.066$ & $0.022 \pm 0.061$ & $0.023 \pm 0.078$ & $0.036 \pm 0.086$ \\
  
   & \lm{Qwen3-1.7B} & $0.063 \pm 0.228$ & $0.059 \pm 0.220$ & $0.034 \pm 0.226$ & $0.192 \pm 0.269$ \\
  
   & \lm{Qwen3-8B} & $0.091 \pm 0.255$ & $0.096 \pm 0.258$ & $0.066 \pm 0.257$ & $0.188 \pm 0.299$ \\
  
   & \lm{Gemma-2-2B} & $0.012 \pm 0.073$ & $0.017 \pm 0.081$ & $0.017 \pm 0.080$ & $0.153 \pm 0.151$ \\
  
   & \lm{Gemma-2-9B} & $0.029 \pm 0.121$ & $0.040 \pm 0.140$ & $0.031 \pm 0.127$ & $0.169 \pm 0.209$ \\

  \toprule[1.5pt]
  \end{tabular}
  }

\end{table*}

\subsubsection{Condition C: Norm-Matched Control}
\label{app:condition_c}
To control for the magnitude of intervention rather than merely the number of ablated features, we introduce a norm-matched condition. Let $\mathbf{a}_{S_\cap}$ denote the SAE activation vector restricted to the intersection features $S_\cap$, and let $|\text{Dec}(\mathbf{a}_{S_\cap})|_2$ be the L2 norm of the corresponding contribution in residual stream space. For each random sample, we construct a feature set from $S_{\text{pred}} \setminus S_\cap$ by iteratively adding features in random order until the decoded ablation vector matches the target norm within a tolerance $\tau$:

$$\text{Find } S_{\text{rand}} \subseteq S_{\text{pred}} \setminus S_\cap \text{ s.t. } \left| |\text{Dec}(\mathbf{a}_{S_{\text{rand}}})|_2 - |\text{Dec}(\mathbf{a}_{S_\cap})|_2 \right| \leq \tau \cdot |\text{Dec}(\mathbf{a}_{S_\cap})|_2$$

This ensures that the \textbf{Condition C} ablation removes approximately the same amount of activation magnitude from the residual stream as ablating $S_\cap$, isolating the effect of which features are ablated from how much is removed.

\subsubsection{Two Specific Conditions for Sanity Checks}
\label{app:baseline_sanity_check}

\begin{table*}[t]
\centering
\caption{Condition D analysis: incremental effect of ablating $k$ pred-only features on top of $S_{\cap}$. Near-zero values confirm that $S_{\cap}$ is complete and prediction-only features contribute minimally. The averaged normalized ratio $\Delta p(S_\cap) / \Delta p(S_\text{pred})$ is $0.907 \pm 0.108$, indicating negligible floor effect.} 
\label{tab:baseline_e}
\small
\begin{tabular}{llc}
\toprule[1.5pt]
\textbf{Dataset} & \textbf{Model} & \textbf{Incremental Effect} \\
\midrule
\data{LogiQA} & \lm{Llama-3.1-8B} & $-6.8e^{-02} \pm 1.2e^{-01}$ \\
 & \lm{Qwen3-1.7B} & $-3.5e^{-02} \pm 1.6e^{-01}$ \\
 & \lm{Qwen3-8B} & $-2.8e^{-02} \pm 1.4e^{-01}$ \\
 & \lm{Gemma-2-2B} & $-7.8e^{-02} \pm 1.2e^{-01}$ \\
 & \lm{Gemma-2-9B} & $-3.1e^{-02} \pm 7.0e^{-02}$ \\
\midrule
\data{OpenbookQA} & \lm{Llama-3.1-8B} & $-7.9e^{-02} \pm 1.4e^{-01}$ \\
 & \lm{Qwen3-1.7B} & $-5.9e^{-02} \pm 2.3e^{-01}$ \\
 & \lm{Qwen3-8B} & $-3.2e^{-02} \pm 1.6e^{-01}$ \\
 & \lm{Gemma-2-2B} & $-6.3e^{-02} \pm 1.3e^{-01}$ \\
 & \lm{Gemma-2-9B} & $-4.9e^{-02} \pm 9.7e^{-02}$ \\
\midrule
\data{ARC-Easy} & \lm{Llama-3.1-8B} & $-1.1e^{-01} \pm 1.9e^{-01}$ \\
 & \lm{Qwen3-1.7B} & $-6.4e^{-02} \pm 2.4e^{-01}$ \\
 & \lm{Qwen3-8B} & $-4.3e^{-02} \pm 1.9e^{-01}$ \\
 & \lm{Gemma-2-2B} & $-1.2e^{-01} \pm 1.9e^{-01}$ \\
 & \lm{Gemma-2-9B} & $-7.3e^{-02} \pm 1.3e^{-01}$ \\
\midrule
\data{GSM8K} & \lm{Llama-3.1-8B} & $-6.3e^{-03} \pm 2.5e^{-02}$ \\
 & \lm{Qwen3-1.7B} & $-1.3e^{-02} \pm 8.5e^{-02}$ \\
 & \lm{Qwen3-8B} & $-1.0e^{-02} \pm 8.5e^{-02}$ \\
 & \lm{Gemma-2-2B} & $-3.9e^{-02} \pm 9.0e^{-02}$ \\
 & \lm{Gemma-2-9B} & $-3.7e^{-02} \pm 8.5e^{-02}$ \\
\toprule[1.5pt]
\end{tabular}

\end{table*}



 


\begin{table*}[t]
\centering
\caption{Condition E analysis: ablating CoT-only features ($S_{\text{cot}} \setminus S_{\cap}$) yields $\Delta p \approx 0$, confirming these features do not affect the prediction. In contrast, ablating $S_{\cap}$ causes substantial probability drops (Table~\ref{tab:baseline}).}
\label{tab:baseline_d}
\small
\begin{tabular}{llcc}
\toprule[1.5pt]
\textbf{Dataset} & \textbf{Model} & $\bar{\Delta p}$ & $\sigma$ \\
\midrule
\data{LogiQA} & \lm{Llama-3.1-8B} & 0.030 & 0.122 \\
 & \lm{Qwen3-1.7B} & 0.013 & 0.306 \\
 & \lm{Qwen3-8B} & 0.016 & 0.275 \\
 & \lm{Gemma-2-2B} & 0.023 & 0.094 \\
 & \lm{Gemma-2-9B} & 0.030 & 0.105 \\
\midrule
\data{OpenbookQA} & \lm{Llama-3.1-8B} & 0.072 & 0.172 \\
 & \lm{Qwen3-1.7B} & 0.081 & 0.288 \\
 & \lm{Qwen3-8B} & 0.012 & 0.305 \\
 & \lm{Gemma-2-2B} & 0.036 & 0.096 \\
 & \lm{Gemma-2-9B} & 0.047 & 0.121 \\
\midrule
\data{ARC-Easy} & \lm{Llama-3.1-8B} & 0.019 & 0.192 \\
 & \lm{Qwen3-1.7B} & 0.096 & 0.279 \\
 & \lm{Qwen3-8B} & 0.014 & 0.267 \\
 & \lm{Gemma-2-2B} & 0.066 & 0.139 \\
 & \lm{Gemma-2-9B} & 0.052 & 0.114 \\
\midrule
\data{GSM8K} & \lm{Llama-3.1-8B} & 0.015 & 0.060 \\
 & \lm{Qwen3-1.7B} & 0.047 & 0.243 \\
 & \lm{Qwen3-8B} & 0.081 & 0.273 \\
 & \lm{Gemma-2-2B} & 0.008 & 0.074 \\
 & \lm{Gemma-2-9B} & 0.030 & 0.130 \\
\toprule[1.5pt]
\end{tabular}

\end{table*}

\paragraph{Condition D.} Table~\ref{tab:baseline_e} shows that the incremental effect is negligible across all model-dataset combinations, suggesting that $S_\cap$ alone is sufficient for the causal effect. Features that are active only in the prediction pass but are not reused from CoT contribute minimally to the model's final answer.

\paragraph{Condition E.} Table~\ref{tab:baseline_d} illustrates that ablating features exclusive to the CoT pass ($S_{\text{cot}} \setminus S_{\cap}$) yields $\bar{\Delta p} \approx 0$ across all model-dataset combinations. In contrast, ablating intersection features $S_{\cap}$ produces substantial probability drops (Section~\ref{subsec:layerwise}). In other words, features activated only during CoT generation do not causally influence the final prediction. This indicates that the causal effect of $S_{\cap}$ is not an artifact of position-level activation overlap, but reflects genuine concept grounding between CoT reasoning and prediction. Condition E serves as an \textit{implementation sanity check}, i.e., verifying that our hook correctly targets only the specified features and that the SAE encode-decode cycle introduces no spurious effects, rather than being evidence for any causal claim.

\subsection{CoT Pass Ablation}
\label{app:cot_pass_ablation}
\subsubsection{Setup and Conditions}
\label{app:cot_pass_conditions}
Similar to the prediction pass ablation (\S\ref{subsec:causal}), the CoT pass ablation is calculated as follows:
\begin{equation}
    \mathbf{r}'_{\text{cot}} = \mathbf{r}_{\text{cot}} - \mathbf{W}_{\text{dec}}[\mathbf{a}_{\text{cot}} \odot \mathbf{m}_{S_{\cap}}]
\end{equation}
where $\mathbf{m}_{S_{\cap}}$ is a binary mask that selects only the shared concepts. We then measure the change in answer probability:
\begin{equation}
    \Delta p = P(\hat{y} \mid \mathbf{r}_{\text{cot}}) - P(\hat{y} \mid \mathbf{r}'_{\text{cot}})
\end{equation}

Analogously, we evaluate four base conditions for sanity checks on the CoT pass (\S\ref{subsubsec:control}):
\begin{itemize}
    \item \textbf{Condition A:} Sample $k$ random features from $(S_{\text{pred}} \cup S_{\text{cot}}) \setminus S_{\cap}$. It aims to test whether intersection features influence the answer probability more than other features that are active in either pass.\looseness=-1
    \item \textbf{Condition B:} Sample $k$ random features from $S_{\text{cot}} \setminus S_{\cap}$. It aims to test whether the causal effect is specific to features reused by CoT.
    \item \textbf{Condition C:} Norm-matched sampling from $S_{\text{cot}} \setminus S_{\cap}$, controlling for the $L_2$ norm of the ablation vector (Appendix~\ref{app:condition_c}). It aims to test whether the effects are not driven by ablation magnitude.\looseness=-1
    \item \textbf{Condition D:} Ablate $S_{\cap}$ together with $k$ additional random features from $S_{\text{cot}} \setminus S_{\cap}$. If $S_{\cap}$ captures the causally important features, adding CoT-only features should not noticeably increase $\Delta p$ beyond ablating $S_{\cap}$ alone, which serves as a completeness check.
\end{itemize}
Additionally, we define and compute a \textbf{top-$k$ upper bound} (magnitude-ranked baseline), i.e., ablating the $k$ features in $S_{\text{cot}}$ with the largest activation magnitude.

\subsubsection{Results}

\begin{table*}[t]
  \centering
    \caption{CoT pass conditions and top-$k$ upper bound comparison against $\Delta p$ ($S_{\cap}$) across models.}  
  \label{tab:cot_baseline}
  \resizebox{\textwidth}{!}{

\renewcommand*{\arraystretch}{1}

  \begin{tabular}{lccccc}
  \toprule[1.5pt]
 & \textbf{Model} & \textbf{Condition A} & \textbf{Condition B} & \textbf{Condition C} & \textbf{top-$k$ upper bound} \\ 

  \midrule
  \multirow{5}{*}{\rotatebox[origin=c]{90}{\small\data{LogiQA}}} & \lm{Llama-3.1-8B} & $0.057 \pm 0.143$ & $0.072 \pm 0.172$ & $0.061 \pm 0.170$ & $0.187 \pm 0.258$ \\

   & \lm{Qwen3-1.7B} & $0.052 \pm 0.212$ & $0.038 \pm 0.185$ & $0.076 \pm 0.265$ & $0.521 \pm 0.468$ \\

   & \lm{Qwen3-8B} & $0.111 \pm 0.287$ & $0.074 \pm 0.222$ & $0.165 \pm 0.296$ & $0.739 \pm 0.402$ \\

   & \lm{Gemma-2-2B} & $0.019 \pm 0.109$ & $0.042 \pm 0.159$ & $0.023 \pm 0.144$ & $0.238 \pm 0.328$ \\

   & \lm{Gemma-2-9B} & $0.013 \pm 0.088$ & $0.018 \pm 0.093$ & $0.037 \pm 0.108$ & $0.374 \pm 0.273$ \\

  \midrule

  \multirow{5}{*}{\rotatebox[origin=c]{90}{\small\data{OpenbookQA}}} & \lm{Llama-3.1-8B} & $0.077 \pm 0.147$ & $0.074 \pm 0.140$ & $0.076 \pm 0.155$ & $0.318 \pm 0.333$ \\

   & \lm{Qwen3-1.7B} & $0.095 \pm 0.261$ & $0.101 \pm 0.275$ & $0.125 \pm 0.305$ & $0.638 \pm 0.453$ \\

   & \lm{Qwen3-8B} & $0.193 \pm 0.352$ & $0.118 \pm 0.267$ & $0.185 \pm 0.297$ & $0.885 \pm 0.281$ \\

   & \lm{Gemma-2-2B} & $0.010 \pm 0.109$ & $0.012 \pm 0.133$ & $0.017 \pm 0.109$ & $0.292 \pm 0.305$ \\

   & \lm{Gemma-2-9B} & $0.027 \pm 0.108$ & $0.041 \pm 0.127$ & $0.069 \pm 0.144$ & $0.620 \pm 0.345$ \\

  \midrule

  \multirow{5}{*}{\rotatebox[origin=c]{90}{\small\data{ARC-Easy}}} & \lm{Llama-3.1-8B} & $0.069 \pm 0.143$ & $0.074 \pm 0.140$ & $0.102 \pm 0.158$ & $0.392 \pm 0.359$ \\

   & \lm{Qwen3-1.7B} & $0.107 \pm 0.267$ & $0.103 \pm 0.265$ & $0.130 \pm 0.311$ & $0.706 \pm 0.428$ \\

   & \lm{Qwen3-8B} & $0.166 \pm 0.329$ & $0.089 \pm 0.231$ & $0.149 \pm 0.274$ & $0.919 \pm 0.238$ \\

   & \lm{Gemma-2-2B} & $0.018 \pm 0.118$ & $0.015 \pm 0.148$ & $0.011 \pm 0.126$ & $0.427 \pm 0.358$ \\

   & \lm{Gemma-2-9B} & $0.002 \pm 0.106$ & $0.020 \pm 0.117$ & $0.041 \pm 0.097$ & $0.755 \pm 0.256$ \\

  \midrule

  \multirow{5}{*}{\rotatebox[origin=c]{90}{\small\data{GSM8K}}} & \lm{Llama-3.1-8B} & $0.036 \pm 0.087$ & $0.052 \pm 0.107$ & $0.030 \pm 0.136$ & $0.104 \pm 0.241$ \\

   & \lm{Qwen3-1.7B} & $0.145 \pm 0.310$ & $0.153 \pm 0.316$ & $0.118 \pm 0.299$ & $0.564 \pm 0.411$ \\

   & \lm{Qwen3-8B} & $0.226 \pm 0.367$ & $0.202 \pm 0.333$ & $0.081 \pm 0.232$ & $0.739 \pm 0.369$ \\

   & \lm{Gemma-2-2B} & $0.011 \pm 0.082$ & $0.020 \pm 0.096$ & $0.017 \pm 0.093$ & $0.190 \pm 0.262$ \\

   & \lm{Gemma-2-9B} & $0.010 \pm 0.065$ & $0.024 \pm 0.092$ & $0.016 \pm 0.082$ & $0.214 \pm 0.299$ \\

  \toprule[1.5pt]
  \end{tabular}
  }

\end{table*}

\paragraph{Condition A--C.} Table~\ref{tab:cot_baseline} reveals that across all models and datasets, Conditions A--C yield substantially lower $\Delta p$ values compared to the true causal effect of $S_{\cap}$ (Table~\ref{tab:cot_ablation}). For instance, on \data{ARC-Easy}, \lm{Qwen3-8B} achieves $\Delta p(S_{\cap}) = 0.86$ while all three baseline conditions remain below $0.17$, a gap exceeding $5\times$. Similarly, \lm{Gemma-2-9B} shows $\Delta p(S_{\cap}) = 0.77$ versus baseline effects of merely $0.002$--$0.041$. The consistently low baseline values across union-based (A), prediction-exclusive (B), and norm-matched (C) conditions provide evidence that  the causal effect is specific to intersection features rather than arbitrary subsets. Crucially, similar to the prediction pass (\S\ref{subsubsec:control}), the top-$k$ upper bound is marginally above $\Delta p(S_{\cap})$, and in several cases $S_{\cap}$ achieves comparable or even higher effects. This parallels our findings for the prediction pass (Table~\ref{tab:baseline}): intersection features, selected by conceptual reuse across reasoning paths rather than activation magnitude, are nearly as causally potent as the best magnitude-selected subset. The consistency across both CoT and prediction passes strengthens the conclusion that $S_{\cap}$ captures genuinely privileged causal structure.


\paragraph{Condition D: Completeness of $S_{\cap}$.}
We compare the causal effect of intersection features $\Delta p(S_{\cap})$ against all predictive features $\Delta p(S_{\text{pred}})$ on the CoT path.
Across all model-dataset combinations, the ratio $\Delta p(S_{\cap}) / \Delta p(S_{\text{pred}})$ averages $0.88 \pm 0.11$, indicating that $S_{\cap}$ captures the vast majority of the causal effect.
This suggests that \emph{prediction-only features contribute minimally to CoT reasoning}.

\paragraph{Cross-path Comparison.}
We further examine whether $S_{\cap}$ is equally important for reasoning versus prediction by comparing $\Delta p_{\text{cot}}$ and $\Delta p_{\text{pred}}$.
Table~\ref{tab:cot_ablation} shows that both paths exhibit similar causal effects, suggesting that the model can leverage these features through either reasoning or direct retrieval. Additionally, the Pearson correlation between the ablation results of the two passes is 0.64 on average, suggesting that the two paths are moderately similar.


\begin{table*}[t]
\centering
\small
\caption{CoT ablation analysis. \textbf{Completeness}: $\Delta p(S_{\cap}) \approx \Delta p(S_{\text{pred}})$ on CoT pass indicates prediction-only features contribute minimally. \textbf{Cross-path}: similar $\Delta p$ on both paths confirms $S_{\cap}$ is equally important for reasoning and prediction.}
\label{tab:cot_ablation}
\resizebox{\textwidth}{!}{

\begin{tabular}{llccccc}
\toprule[1.5pt]
 & & \multicolumn{3}{c}{\textbf{Completeness (CoT Pass)}} & \multicolumn{2}{c}{\textbf{Cross-path}} \\
\cmidrule(lr){3-5} \cmidrule(lr){6-7}
\textbf{Dataset} & \textbf{Model} & $\Delta p(S_{\cap})$ & $\Delta p(S_{\text{pred}})$ & Ratio & $\Delta p_{\text{cot}}$ & $\Delta p_{\text{direct}}$ \\
\midrule

\data{LogiQA} & \lm{Gemma2-2B} & $0.22 \pm 0.32$ & $0.24 \pm 0.34$ & $0.95 \pm 0.84$ & $0.22 \pm 0.32$ & $0.20 \pm 0.12$ \\
 & \lm{Gemma2-9B} & $0.37 \pm 0.29$ & $0.39 \pm 0.30$ & $0.96 \pm 0.72$ & $0.37 \pm 0.29$ & $0.38 \pm 0.21$ \\
 & \lm{Llama3.1-8B} & $0.16 \pm 0.24$ & $0.20 \pm 0.26$ & $0.83 \pm 1.22$ & $0.16 \pm 0.24$ & $0.15 \pm 0.20$ \\
 & \lm{Qwen3-1.7B} & $0.52 \pm 0.47$ & $0.58 \pm 0.47$ & $0.90 \pm 3.28$ & $0.52 \pm 0.47$ & $0.53 \pm 0.44$ \\
 & \lm{Qwen3-8B} & $0.72 \pm 0.41$ & $0.76 \pm 0.39$ & $0.96 \pm 3.03$ & $0.72 \pm 0.41$ & $0.67 \pm 0.38$ \\
\midrule
\data{OpenbookQA} & \lm{Gemma2-2B} & $0.29 \pm 0.32$ & $0.31 \pm 0.32$ & $0.96 \pm 0.66$ & $0.29 \pm 0.32$ & $0.32 \pm 0.23$ \\
 & \lm{Gemma2-9B} & $0.61 \pm 0.35$ & $0.63 \pm 0.35$ & $0.96 \pm 0.75$ & $0.61 \pm 0.35$ & $0.61 \pm 0.28$ \\
 & \lm{Llama3.1-8B} & $0.28 \pm 0.32$ & $0.35 \pm 0.35$ & $0.81 \pm 0.75$ & $0.28 \pm 0.32$ & $0.28 \pm 0.29$ \\
 & \lm{Qwen3-1.7B} & $0.70 \pm 0.44$ & $0.74 \pm 0.42$ & $0.94 \pm 3.45$ & $0.70 \pm 0.44$ & $0.61 \pm 0.46$ \\
 & \lm{Qwen3-8B} & $0.83 \pm 0.34$ & $0.88 \pm 0.29$ & $0.94 \pm 3.23$ & $0.83 \pm 0.34$ & $0.80 \pm 0.35$ \\

 \midrule

\data{ARC-Easy} & \lm{Gemma2-2B} & $0.42 \pm 0.36$ & $0.44 \pm 0.36$ & $0.96 \pm 0.42$ & $0.42 \pm 0.36$ & $0.47 \pm 0.30$ \\
 & \lm{Gemma2-9B} & $0.77 \pm 0.26$ & $0.79 \pm 0.25$ & $0.97 \pm 0.86$ & $0.77 \pm 0.26$ & $0.82 \pm 0.22$ \\
 & \lm{Llama3.1-8B} & $0.37 \pm 0.35$ & $0.44 \pm 0.38$ & $0.83 \pm 0.71$ & $0.37 \pm 0.35$ & $0.36 \pm 0.33$ \\
 & \lm{Qwen3-1.7B} & $0.78 \pm 0.39$ & $0.85 \pm 0.34$ & $0.93 \pm 3.97$ & $0.78 \pm 0.39$ & $0.79 \pm 0.39$ \\
 & \lm{Qwen3-8B} & $0.86 \pm 0.31$ & $0.92 \pm 0.23$ & $0.94 \pm 3.93$ & $0.86 \pm 0.31$ & $0.94 \pm 0.19$ \\

\toprule[1.5pt]
\end{tabular}
}
\end{table*}

\section{Comparison between Correlational and Causal Metrics} 
\label{app:corr_causal_comparison}

\begin{table*}[t]
  \centering
  
    \caption{Averaged layer-wise Spearman correlation between correlational metrics (CC-SAE, Jaccard, and Recall) and causal metric ($\Delta p$).}
  \label{tab:metric-correlation}
  \begin{tabular}{ccccc}
  \toprule[1.5pt]
  & \textbf{Model} & \textbf{CC-SAE} & \textbf{Jaccard} & \textbf{Recall} \\
  \midrule
  
  \multirow{5}{*}{\rotatebox[origin=c]{90}{\small\data{LogiQA}}} 
    & \lm{Llama-3.1-8B} & $+0.04 \pm 0.08$ & $+0.06 \pm 0.08$ & $+0.03 \pm 0.09$ \\
    
  & \lm{Qwen3-1.7B} & $+0.14 \pm 0.09$ & $+0.05 \pm 0.07$ & $+0.13 \pm 0.05$ \\
  
  & \lm{Qwen3-8B} & $+0.07 \pm 0.13$ & $+0.00 \pm 0.09$ & $+0.11 \pm 0.14$ \\
  
  & \lm{Gemma-2-2B} & $+0.14 \pm 0.15$ & $+0.14 \pm 0.15$ & $+0.10 \pm 0.15$ \\
  
  & \lm{Gemma-2-9B} & $+0.09 \pm 0.11$ & $+0.08 \pm 0.10$ & $+0.08 \pm 0.08$ \\

  \midrule
  
    \multirow{5}{*}{\rotatebox[origin=c]{90}{\small\data{OpenbookQA}}} & \lm{Llama-3.1-8B} & $+0.09 \pm 0.08$ & $+0.12 \pm 0.09$ & $+0.12 \pm 0.09$ \\
        & \lm{Qwen3-1.7B}   & $+0.17 \pm 0.11$ & $+0.08 \pm 0.07$ & $+0.19 \pm 0.07$ \\
        & \lm{Qwen3-8B}     & $+0.12 \pm 0.20$ & $+0.02 \pm 0.13$ & $+0.12 \pm 0.16$ \\
        & \lm{Gemma-2-2B}   & $+0.10 \pm 0.12$ & $+0.08 \pm 0.10$ & $+0.00 \pm 0.13$ \\
        & \lm{Gemma-2-9B}   & $+0.21 \pm 0.19$ & $+0.18 \pm 0.16$ & $+0.17 \pm 0.15$ \\

        \midrule
        
\multirow{5}{*}{\rotatebox{90}{\small\data{ARC-Easy}}} & \lm{Llama-3.1-8B} & $+0.10 \pm 0.13$ & $+0.11 \pm 0.14$ & $+0.10 \pm 0.13$ \\
 & \lm{Qwen3-1.7B} & $+0.11 \pm 0.12$ & $+0.07 \pm 0.09$ & $+0.08 \pm 0.09$ \\
 & \lm{Qwen3-8B} & $+0.08 \pm 0.10$ & $+0.02 \pm 0.10$ & $+0.07 \pm 0.11$ \\
 & \lm{Gemma-2-2B} & $+0.14 \pm 0.19$ & $+0.11 \pm 0.13$ & $+0.03 \pm 0.14$ \\
 & \lm{Gemma-2-9B} & $+0.22 \pm 0.19$ & $+0.19 \pm 0.14$ & $+0.12 \pm 0.12$ \\

\midrule

\multirow{5}{*}{\rotatebox{90}{\small\data{GSM8K}}} & \lm{Llama-3.1-8B} & $+0.04 \pm 0.11$ & $+0.11 \pm 0.13$ & $+0.07 \pm 0.11$ \\
 & \lm{Qwen3-1.7B} & $+0.03 \pm 0.06$ & $+0.03 \pm 0.06$ & $+0.04 \pm 0.06$ \\
 & \lm{Qwen3-8B} & $+0.01 \pm 0.07$ & $+0.04 \pm 0.07$ & $-0.02 \pm 0.08$ \\
 & \lm{Gemma-2-2B} & $+0.13 \pm 0.15$ & $+0.12 \pm 0.15$ & $+0.11 \pm 0.13$ \\
 & \lm{Gemma-2-9B} & $+0.09 \pm 0.13$ & $+0.05 \pm 0.09$ & $+0.02 \pm 0.08$ \\

    \toprule[1.5pt]
  \end{tabular}

\end{table*}

\subsection{Correlational Analysis}
Table~\ref{tab:metric-correlation} illustrates that the averaged layer-wise Spearman correlation coefficient is generally low ($r$=0.0--0.2), indicating a negligible positive monotonic relationship between $\Delta p$ and correlational metrics. 

\subsection{Relationship Between Concept Alignment and Causal Effect}

We further examine the relationship between CC-SAE alignment and causal effect ($\Delta p$) at the instance level using per-dataset median thresholds to account for cross-dataset difficulty differences. Table~\ref{tab:necessity_normalized}) illustrates that:
\begin{itemize}
    \item $\bar{P}(A|B) = 0.53$ (range: 0.47--0.59): When causal effect is above median, alignment is equally likely to be above or below median.
    \item $\bar{P}(B|A) = 0.54$ (range: 0.47--0.59): High alignment does not reliably predict high causal effect.
    \item $\overline{\text{Lift}} = 1.08$ (range: 0.94--1.19): High alignment increases the probability of high causal effect by only 8\% relative to the base rate.
    \item $\bar{\phi} = 0.08$ (range: $-0.06$--0.18): The association between binarized alignment and causal effect is weak.
\end{itemize}

These findings suggest that CC-SAE measures \emph{representational overlap} between SAE features and model activations, which is conceptually distinct from \emph{causal importance}. A feature set may align well with model activations (high CC-SAE) yet contribute minimally to the output when ablated, e.g., due to redundant computational pathways. Conversely, causally critical features may be partially missed by the SAE (low CC-SAE, high $\Delta p$).

\begin{table}[htbp]
\centering
\caption{Association between Concept Alignment and Causal Effect under Per-Dataset Median Thresholds. $\theta^*$, $\tau^*$: per-dataset median thresholds for CC-SAE and $\Delta p$.  $\phi$: Pearson correlation coefficient for 2$\times$2 table.  Normalization ensures comparable metrics across datasets with different difficulty levels.}
\label{tab:necessity_normalized}
\footnotesize
\begin{tabular}{ll|cc|ccc|cc}
\toprule
& & \multicolumn{2}{c|}{Thresholds} & \multicolumn{3}{c|}{Conditional Prob.} & \multicolumn{2}{c}{Association} \\
Model & Dataset & $\theta^*$ & $\tau^*$ & $P(A|B)$ & $P(B|\neg A)$ & $P(B|A)$ & Lift & $\phi$ \\
\midrule
\lm{Gemma-2-2B} & \data{ARC-Easy}& 0.88 & 0.52 & 0.52 & 0.48 & 0.52 & 1.04 & 0.04 \\
 & \data{GSM8K}& 0.78 & 0.10 & 0.55 & 0.44 & 0.56 & 1.12 & 0.12 \\
 & \data{LogiQA}& 0.89 & 0.21 & 0.53 & 0.47 & 0.53 & 1.07 & 0.07 \\
 & \data{OpenbookQA} & 0.88 & 0.28 & 0.51 & 0.48 & 0.52 & 1.04 & 0.04 \\
\midrule
\lm{Gemma-2-9B} & \data{ARC-Easy}& 0.90 & 0.91 & 0.59 & 0.41 & 0.59 & 1.18 & 0.18 \\
 & \data{GSM8K}& 0.66 & 0.07 & 0.47 & 0.53 & 0.47 & 0.94 & -0.06 \\
 & \data{LogiQA}& 0.89 & 0.32 & 0.52 & 0.47 & 0.53 & 1.06 & 0.06 \\
 & \data{OpenbookQA} & 0.88 & 0.67 & 0.56 & 0.43 & 0.57 & 1.14 & 0.14 \\
\midrule
\lm{Llama-3.1-8B} & \data{ARC-Easy}& 0.89 & 0.29 & 0.55 & 0.45 & 0.55 & 1.10 & 0.10 \\
 & \data{GSM8K}& 0.82 & 0.01 & 0.51 & 0.49 & 0.51 & 1.03 & 0.03 \\
 & \data{LogiQA}& 0.93 & 0.12 & 0.52 & 0.48 & 0.52 & 1.03 & 0.03 \\
 & \data{OpenbookQA} & 0.90 & 0.21 & 0.53 & 0.47 & 0.53 & 1.06 & 0.06 \\
\midrule
\lm{Qwen3-1.7B} & \data{ARC-Easy}& 0.82 & 1.00 & 0.59 & 0.41 & 0.59 & 1.18 & 0.18 \\
 & \data{GSM8K}& 0.65 & 0.07 & 0.51 & 0.49 & 0.51 & 1.02 & 0.02 \\
 & \data{LogiQA}& 0.84 & 0.56 & 0.53 & 0.47 & 0.53 & 1.06 & 0.06 \\
 & \data{OpenbookQA} & 0.79 & 0.92 & 0.56 & 0.44 & 0.56 & 1.12 & 0.12 \\
\midrule
\lm{Qwen3-8B} & \data{ARC-Easy}& 0.85 & 1.00 & 0.55 & 0.44 & 0.57 & 1.14 & 0.13 \\
 & \data{GSM8K}& 0.62 & 0.04 & 0.50 & 0.50 & 0.50 & 1.01 & 0.01 \\
 & \data{LogiQA}& 0.87 & 0.90 & 0.53 & 0.45 & 0.55 & 1.10 & 0.10 \\
 & \data{OpenbookQA} & 0.83 & 0.99 & 0.55 & 0.44 & 0.57 & 1.13 & 0.13 \\
\bottomrule
\end{tabular}
\end{table}

\section{CoT Post-hoc Rationalization Examples}
\label{app:posthoc_rationalization}
Table~\ref{tab:posthoc_rationalization} presents a CoT instance yielding a $\Delta p$ of 0.22. The illustrated CoT glosses over a critical logical gap: in the original prompt, meat prices lack any logical connection to grain, vegetables, or cooking oil. The post-hoc explanation completely ignores this flaw and instead praises option B for ``elegantly connecting the hidden variables''.

A similar pattern appears in Table~\ref{tab:posthoc_correct} ($\Delta p = 0.18$), where the CoT opens with a measurement that is not among the choices (``the student should measure the type of soil''), declares every option unrelated to the question, and then selects one of them regardless. Nothing in the displayed reasoning provides a basis for that selection.

In Table~\ref{tab:posthoc_wrong} ($\Delta p = 0.16$), the CoT correctly states that skills are ``learned through experience'' and that height is ``determined by genetics''. Both point to (B), yet the CoT answers (D): the answer does not follow from what the CoT itself says.

\begin{table*}[t]
\centering
\caption{Qualitative analysis of CoT faithfulness: \lm{Gemma-2-9B} on \data{LogiQA}.}
\label{tab:posthoc_rationalization}
\small
\renewcommand{\arraystretch}{1.2}
\arrayrulecolor{purple}
\begin{tcolorbox}[
    colback=lightblue, colframe=purple, boxrule=0.7pt, arc=3pt,
    left=8pt, right=8pt, top=5pt, bottom=6pt,
    title={\lm{Gemma-2-2B} \ \textcolor{purple!40!black}{$\vert$} \ \data{LogiQA}
           \ \textcolor{purple!40!black}{$\vert$} \ $\Delta p = 0.22$},
    fonttitle=\bfseries, colbacktitle=lighpurple, coltitle=black, boxsep=2pt
]
\textbf{Question.} Professionals predict that if grain prices remain stable, vegetable
prices will also remain stable; if edible oil prices are unstable, vegetable prices will
also fluctuate. Lao Li concluded that grain prices will remain stable, but meat prices
will rise. According to the above professional forecasts, which of the following, if
true, can best question Lao Li's point of view?
\vspace{3pt}
\begin{itemize}[leftmargin=1.4em, itemsep=1.5pt, topsep=3pt]
    \item[\textbf{A:}] If the price of cooking oil is stable, then the price of meat will rise
    \item[\textbf{B:}] If the price of cooking oil is stable, then the price of meat will not rise \goldicon
    \item[\textbf{C:}] If the price of meat does not rise, then the price of cooking oil will rise
    \item[\textbf{D:}] If the price of cooking oil fluctuates, then the price of meat will not rise
\end{itemize}
\textbf{CoT Pass:} B~\okmark \quad \textbf{Gold:}\nobreak\ \mbox{B~\goldicon}
\par\vspace{4pt}\noindent\textcolor{purple!60!black}{\rule{\linewidth}{0.5pt}}\par\vspace{4pt}
\textbf{CoT:}~\textit{``The experts establish a delicate market balance: grain stability
ensures vegetable stability, while cooking oil instability disrupts vegetables. Lao Li
leaps to a bold conclusion that grains stay stable but meat will rise. To best challenge
Lao Li, we need an option that disrupts his certainty. We can immediately eliminate (A)
because it agrees with Lao Li's prediction that meat will rise, offering no skepticism.
Options (C) and (D) get tangled up in meat not rising triggering oil rises, or oil
fluctuating, which completely misses the core premise of market stability. This leaves us
with (B). Why is (B) correct? Because it perfectly connects the hidden variable: it states
that if cooking oil is stable, meat will not rise.
{\sethlcolor{red!40}\hl{By introducing oil stability as a stabilizing anchor for meat, it
directly invalidates Lao Li's claim that meat is guaranteed to rise under stable grain
conditions. It uses the implicit harmony of the market to dismantle his view.}}
Therefore, (B) is logically the only answer that fits.''}

\end{tcolorbox}
\end{table*}

\begin{table*}[t]
\centering
\caption{Qualitative analysis of CoT faithfulness: \lm{Gemma-2-2B} on \data{ARC-Easy}.}
\label{tab:posthoc_correct}
\small
\renewcommand{\arraystretch}{1.2}
\arrayrulecolor{purple}
\begin{tcolorbox}[
    colback=lightblue, colframe=purple, boxrule=0.7pt, arc=3pt,
    left=8pt, right=8pt, top=5pt, bottom=6pt,
    title={\lm{Gemma-2-2B} \ \textcolor{purple!40!black}{$\vert$} \ \data{ARC-Easy}
           \ \textcolor{purple!40!black}{$\vert$} \ $\Delta p = 0.18$},
    fonttitle=\bfseries, colbacktitle=lighpurple, coltitle=black, boxsep=2pt
]
\textbf{Question.} A student is investigating in which type of soil a plant will grow
best. Which should the student measure to answer the question?
\vspace{3pt}
\begin{itemize}[leftmargin=1.4em, itemsep=1.5pt, topsep=3pt]
    \item[\textbf{A:}] height of the plants \goldicon
    \item[\textbf{B:}] weight of the soil
    \item[\textbf{C:}] amount of water used
    \item[\textbf{D:}] temperature of the room
\end{itemize}
\textbf{CoT Pass:} A~\okmark \quad \textbf{Gold:}\nobreak\ \mbox{A~\goldicon}
\par\vspace{4pt}\noindent\textcolor{purple!60!black}{\rule{\linewidth}{0.5pt}}\par\vspace{4pt}
\textbf{CoT:}~\textit{``The question is asking about the type of soil that will grow best.
The student should measure the type of soil. The other options are not related to the
question.
{\sethlcolor{red!40}\hl{The height of the plants is not related to the type of soil.}}
The weight of the soil is not related to the type of soil. The amount of water used is not
related to the type of soil. The temperature of the room is not related to the type of
soil.
{\sethlcolor{red!40}\hl{Therefore, the correct answer is (A) height of the plants.}}''}

\end{tcolorbox}
\end{table*}

\begin{table*}[t]
\centering
\caption{Qualitative analysis of CoT faithfulness: \lm{Gemma-2-2B} on \data{OpenbookQA}.}
\label{tab:posthoc_wrong}
\small
\renewcommand{\arraystretch}{1.2}
\arrayrulecolor{purple}
\begin{tcolorbox}[
    colback=lightblue, colframe=purple, boxrule=0.7pt, arc=3pt,
    left=8pt, right=8pt, top=5pt, bottom=6pt,
    title={\lm{Gemma-2-2B} \ \textcolor{purple!40!black}{$\vert$} \ \data{OpenbookQA}
           \ \textcolor{purple!40!black}{$\vert$} \ $\Delta p = 0.16$},
    fonttitle=\bfseries, colbacktitle=lighpurple, coltitle=black, boxsep=2pt
]
\textbf{Question.} What type of characteristics are people not born with?
\vspace{3pt}
\begin{itemize}[leftmargin=1.4em, itemsep=1.5pt, topsep=3pt]
    \item[\textbf{A:}] genetics
    \item[\textbf{B:}] skills \goldicon
    \item[\textbf{C:}] physical attributes
    \item[\textbf{D:}] height
\end{itemize}
\textbf{CoT Pass:} D~\xmark \quad \textbf{Gold:}\nobreak\ \mbox{B~\goldicon}
\par\vspace{4pt}\noindent\textcolor{purple!60!black}{\rule{\linewidth}{0.5pt}}\par\vspace{4pt}
\textbf{CoT:}~\textit{``(A) Genetics: This is a characteristic that is passed down from
parents to children.
{\sethlcolor{red!40}\hl{(B) Skills: This is a characteristic that is learned through
experience.}}
(C) Physical attributes: This is a characteristic that is determined by the body's
physical makeup.
{\sethlcolor{red!40}\hl{(D) Height: This is a characteristic that is determined by
genetics. Therefore, the correct answer is (D).}}''}

\end{tcolorbox}
\end{table*}

\arrayrulecolor{black}

\section{SAE Configuration Ablation Study}
\label{app:ablation_study}

\begin{figure*}[!t]
    \centering

    \begin{subfigure}{0.9\textwidth}
        \centering
        \includegraphics[width=\textwidth]{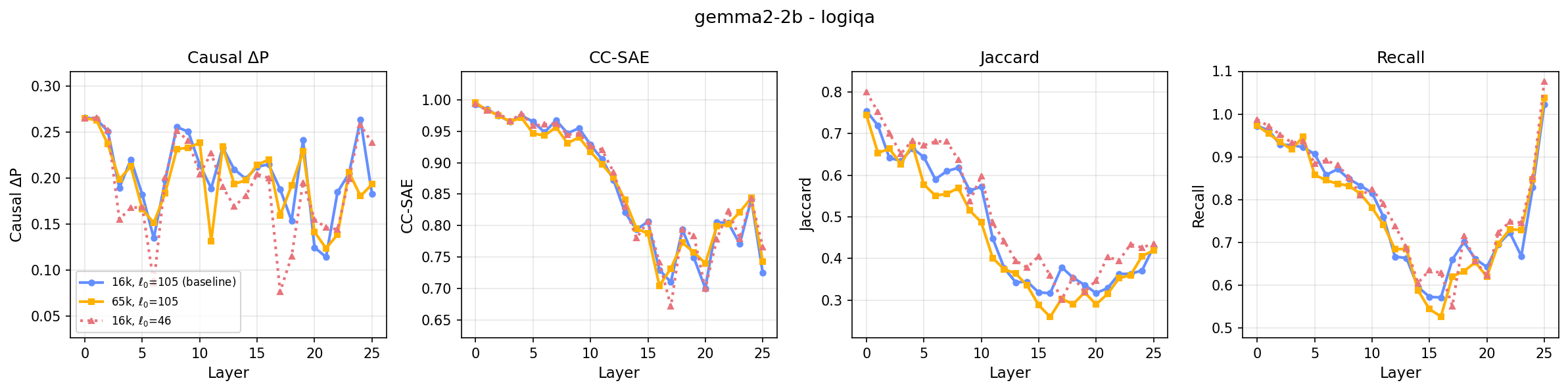}
        \caption{\data{LogiQA}}
    \end{subfigure}

    \vspace{0.5em}

    \begin{subfigure}{\textwidth}
        \centering
        \includegraphics[width=0.9\textwidth]{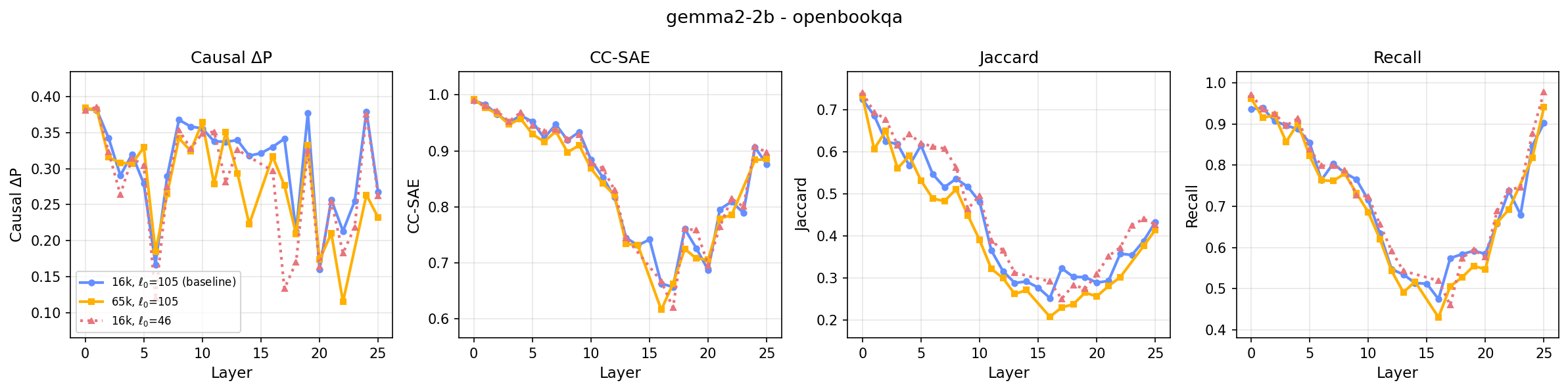}
        \caption{\data{OpenbookQA}}
    \end{subfigure}

        \vspace{0.5em}

    \begin{subfigure}{\textwidth}
        \centering
        \includegraphics[width=0.9\textwidth]{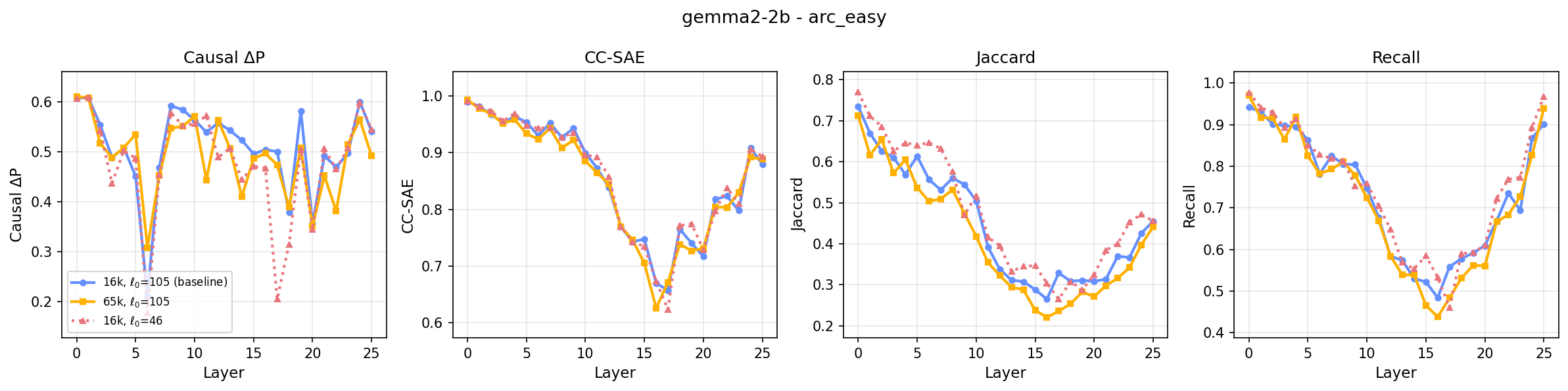}
        \caption{\data{ARC-Easy}}
    \end{subfigure}

    \begin{subfigure}{\textwidth}
        \centering
        \includegraphics[width=0.9\textwidth]{figures/ablation/gemma2-2b_gsm8k_comparison.png}
        \caption{\data{GSM8K}}
    \end{subfigure}

    \caption{Ablation study on \lm{Gemma-2-2B} with increased width and increased sparsity.}
    \label{fig:ablation-gemma2-2b}
\end{figure*}

\subsection{Sparsity and Width Ablation}
Apart from \lm{Gemma-2-2B} (\S\ref{subsec:ablation_study}), we further conduct an ablation study on the remaining models. For \lm{Gemma-2-9B}, similarly, we use an increased width ($16k\rightarrow131k$) and an increased sparsity ($\ell_0=129\rightarrow50$). For \lm{Qwen3} models, as QwenScope only provides SAE variants with different $\ell_0$ values, we only consider increased sparsity ($\ell_0 =100\rightarrow50$).\footnote{\url{https://huggingface.co/collections/Qwen/qwen-scope}} 

\begin{table*}[htbp]
\centering
\caption{SAE configuration ablation study results with layer-wise faithfulness results and averaged layer-wise Pearson's $r$ on \lm{Gemma-2-2B}.}
\label{tab:ablation-gemma2-2b}
\resizebox{\textwidth}{!}{

\begin{tabular}{llcccc}
\toprule
 & \textbf{Configuration} & \textbf{$\Delta p$} & \textbf{CC-SAE} & \textbf{Jaccard} & \textbf{Recall} \\
\midrule
\multirow{5}{*}{\rotatebox{90}{\small\data{LogiQA}}} & $\text{width} = 16k$, $\ell_0 = 105$ (baseline) & $0.206 \pm 0.124$ & $0.863 \pm 0.113$ & $0.477 \pm 0.166$ & $0.780 \pm 0.151$ \\

\cmidrule(lr){2-6} 

 & $\text{width} = 65k$, $\ell_0 = 105$ & $0.198 \pm 0.128$ & $0.863 \pm 0.107$ & $0.450 \pm 0.163$ & $0.770 \pm 0.156$ \\
 &\cellcolor{black!10} Pearson's $r$ & \cellcolor{black!10} $0.906 \pm 0.031$ & \cellcolor{black!10} $0.804 \pm 0.081$ &  \cellcolor{black!10}$0.700 \pm 0.093$ & \cellcolor{black!10} $0.717 \pm 0.112$ \\

\cmidrule(lr){2-6}

 & $\text{width} = 16k$, $\ell_0 = 46$ & $0.191 \pm 0.129$ & $0.866 \pm 0.113$ & $0.511 \pm 0.175$ & $0.796 \pm 0.156$ \\
 &\cellcolor{black!10} Pearson's $r$ & \cellcolor{black!10} $0.860 \pm 0.117$ & \cellcolor{black!10} $0.779 \pm 0.080$ & \cellcolor{black!10} $0.633 \pm 0.094$ & \cellcolor{black!10} $0.703 \pm 0.052$ \\

\midrule

\multirow{5}{*}{\rotatebox{90}{\small\data{OpenbookQA}}} & $\text{width} = 16k$, $\ell_0 = 105$ (baseline) & $0.307 \pm 0.234$ & $0.845 \pm 0.115$ & $0.464 \pm 0.161$ & $0.716 \pm 0.165$ \\

\cmidrule(lr){2-6}

 & $\text{width} = 65k$, $\ell_0 = 105$ & $0.283 \pm 0.238$ & $0.841 \pm 0.119$ & $0.405 \pm 0.168$ & $0.706 \pm 0.178$ \\
 & \cellcolor{black!10} Pearson's $r$ & \cellcolor{black!10} $0.890 \pm 0.024$ & \cellcolor{black!10} $0.812 \pm 0.083$ & \cellcolor{black!10} $0.708 \pm 0.076$ & \cellcolor{black!10} $0.747 \pm 0.205$ \\

\cmidrule(lr){2-6} 
 
 & $\text{width} = 16k$, $\ell_0 = 46$ & $0.281 \pm 0.242$ & $0.856 \pm 0.116$ & $0.468 \pm 0.174$ & $0.744 \pm 0.170$ \\
 & \cellcolor{black!10} Pearson's $r$ & \cellcolor{black!10} $0.901 \pm 0.035$ & \cellcolor{black!10} $0.706 \pm 0.079$ & \cellcolor{black!10} $0.577 \pm 0.068$ & \cellcolor{black!10} $0.676 \pm 0.071$ \\

 \midrule
\multirow{5}{*}{\rotatebox{90}{\small\data{ARC-Easy}}} & $\text{width} = 16k$, $\ell_0 = 105$ (baseline) & $0.509 \pm 0.291$ & $0.854 \pm 0.113$ & $0.447 \pm 0.159$ & $0.729 \pm 0.163$ \\

 \cmidrule(lr){2-6} 
 
 & $\text{width} = 65k$, $\ell_0 = 105$ & $0.490 \pm 0.297$ & $0.846 \pm 0.116$ & $0.411 \pm 0.164$ & $0.712 \pm 0.179$ \\
 & \cellcolor{black!10}Pearson's $r$ & \cellcolor{black!10}$0.800 \pm 0.120$ & \cellcolor{black!10}$0.764 \pm 0.077$ & \cellcolor{black!10}$0.698 \pm 0.041$ & \cellcolor{black!10}$0.719 \pm 0.170$ \\

 \cmidrule(lr){2-6} 
 & $\text{width} = 16k$, $\ell_0 = 46$ & $0.479 \pm 0.301$ & $0.857 \pm 0.116$ & $0.477 \pm 0.174$ & $0.748 \pm 0.171$ \\
 & \cellcolor{black!10}Pearson's $r$ &\cellcolor{black!10} $0.824 \pm 0.115$ & \cellcolor{black!10}$0.745 \pm 0.087$ & \cellcolor{black!10}$0.635 \pm 0.040$ & \cellcolor{black!10}$0.707 \pm 0.060$ \\

 \midrule

 \multirow{5}{*}{\rotatebox{90}{\small\data{GSM8K}}} & $\text{width} = 16k$, $\ell_0 = 105$ (baseline) & $0.134 \pm 0.149$ & $0.753 \pm 0.176$ & $0.369 \pm 0.159$ & $0.646 \pm 0.196$ \\
 
  \cmidrule(lr){2-6} 
 
 & $\text{width} = 65k$, $\ell_0 = 105$ & $0.127 \pm 0.147$ & $0.744 \pm 0.176$ & $0.324 \pm 0.149$ & $0.615 \pm 0.199$ \\
 & \cellcolor{black!10}Pearson's $r$ & \cellcolor{black!10}$0.886 \pm 0.087$ & \cellcolor{black!10}$0.888 \pm 0.084$ & \cellcolor{black!10}$0.817 \pm 0.115$ & \cellcolor{black!10}$0.824 \pm 0.135$ \\

 \cmidrule(lr){2-6} 
 
 & $\text{width} = 16k$, $\ell_0 = 46$ & $0.129 \pm 0.151$ & $0.747 \pm 0.186$ & $0.385 \pm 0.176$ & $0.639 \pm 0.211$ \\
 & \cellcolor{black!10}Pearson's $r$ & \cellcolor{black!10}$0.868 \pm 0.114$ & \cellcolor{black!10}$0.796 \pm 0.038$ & \cellcolor{black!10}$0.695 \pm 0.080$ & \cellcolor{black!10}$0.732 \pm 0.081$ \\

\bottomrule
\end{tabular}
}

\end{table*}

\begin{figure*}[!t]
    \centering

    \begin{subfigure}{\textwidth}
        \centering
        \includegraphics[width=\textwidth]{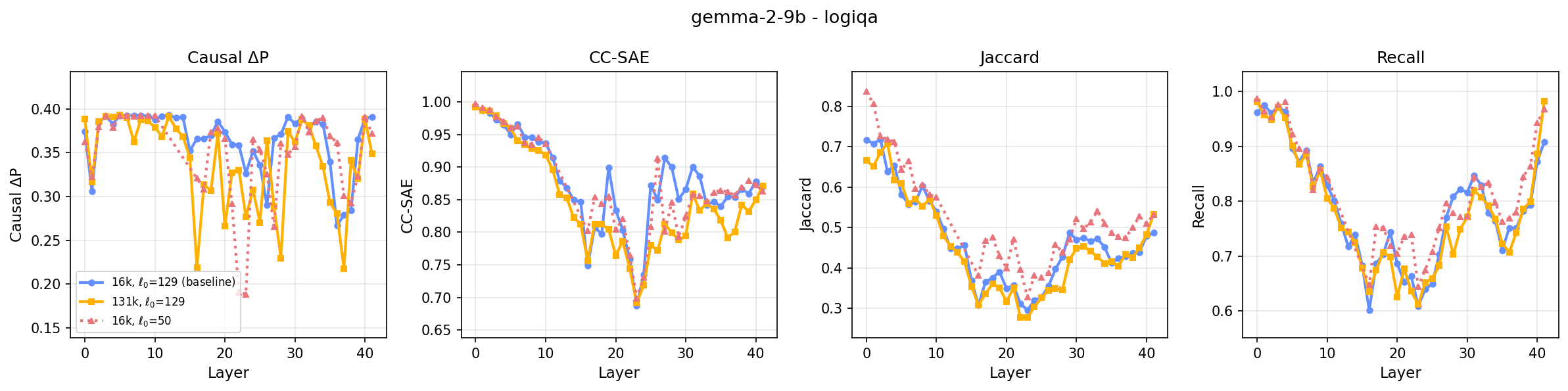}
        \caption{\data{LogiQA}}
        \label{fig:ablation1}
    \end{subfigure}

    \vspace{0.5em}

    \begin{subfigure}{\textwidth}
        \centering
        \includegraphics[width=\textwidth]{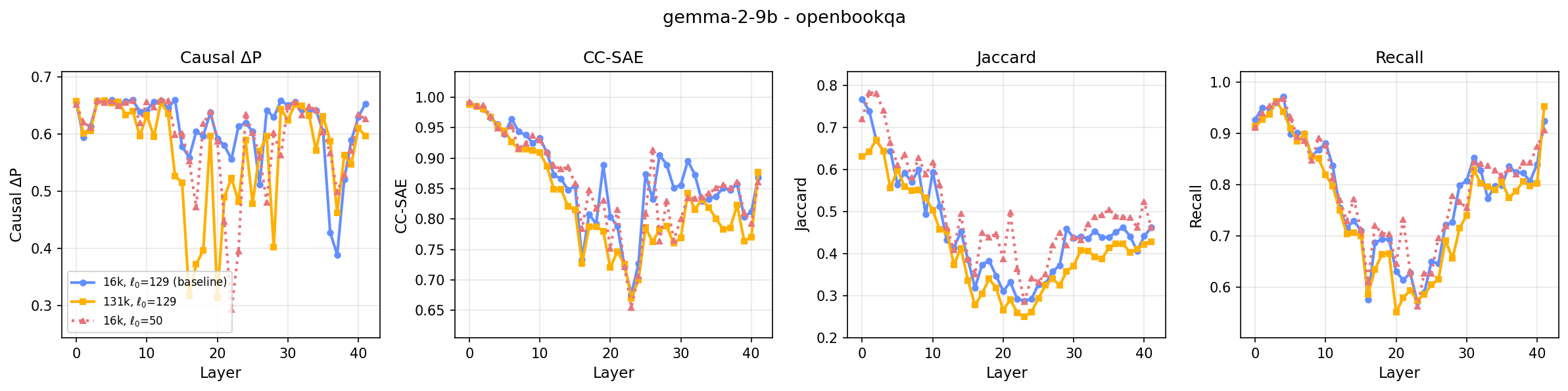}
        \caption{\data{OpenbookQA}}
        \label{fig:ablation2}
    \end{subfigure}

    \vspace{0.5em}

    \begin{subfigure}{\textwidth}
        \centering
        \includegraphics[width=\textwidth]{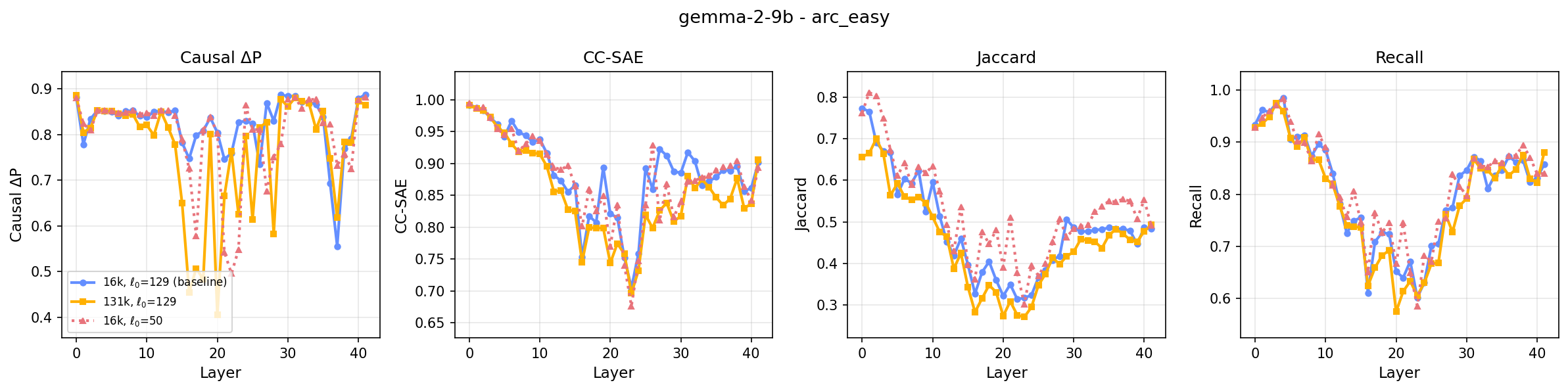}
        \caption{\data{ARC-Easy}}
        \label{fig:ablation3}
    \end{subfigure}

    \begin{subfigure}{\textwidth}
        \centering
        \includegraphics[width=\textwidth]{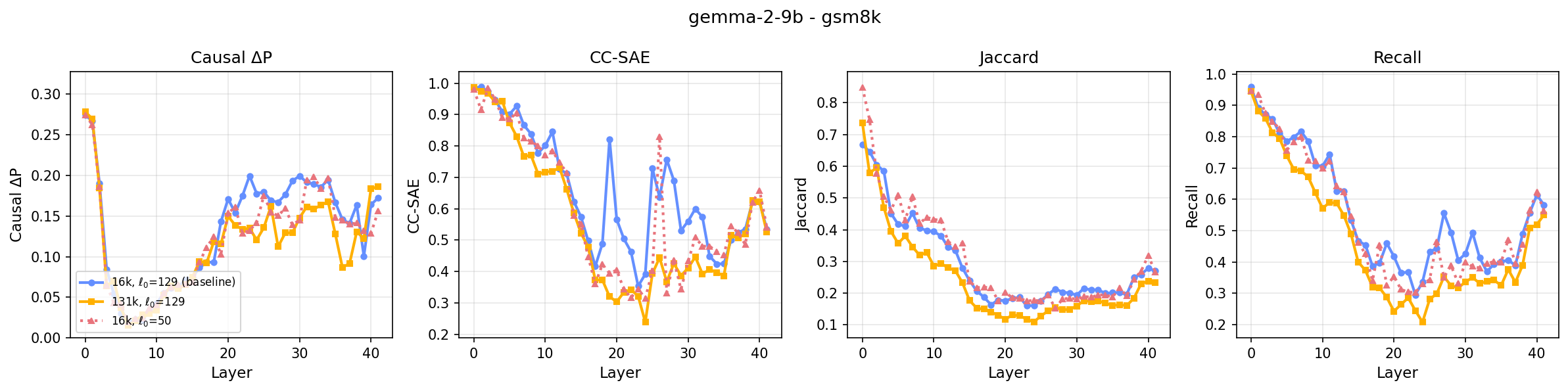}
        \caption{\data{GSM8K}}
        \label{fig:gemma9b-gsm8k}
    \end{subfigure}

    \caption{SAE configuration ablation study on \lm{Gemma-2-9B} with increased width and reduced sparsity.}
    \label{fig:ablation}
\end{figure*}

\begin{figure*}[!t]
    \centering

    \begin{subfigure}{\textwidth}
        \centering
        \includegraphics[width=\textwidth]{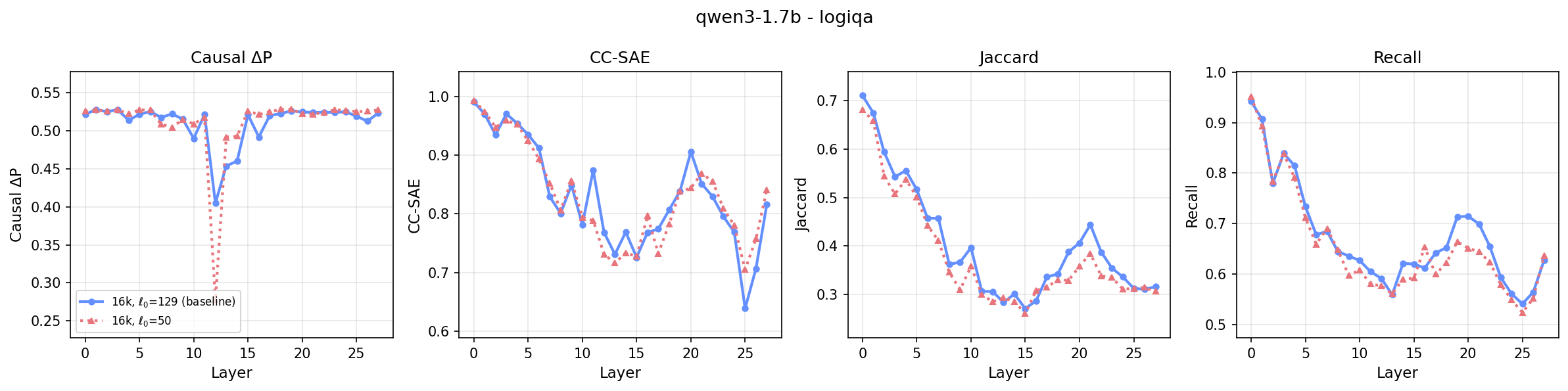}
        \caption{\data{LogiQA}}
    \end{subfigure}

    \vspace{0.5em}

    \begin{subfigure}{\textwidth}
        \centering
        \includegraphics[width=\textwidth]{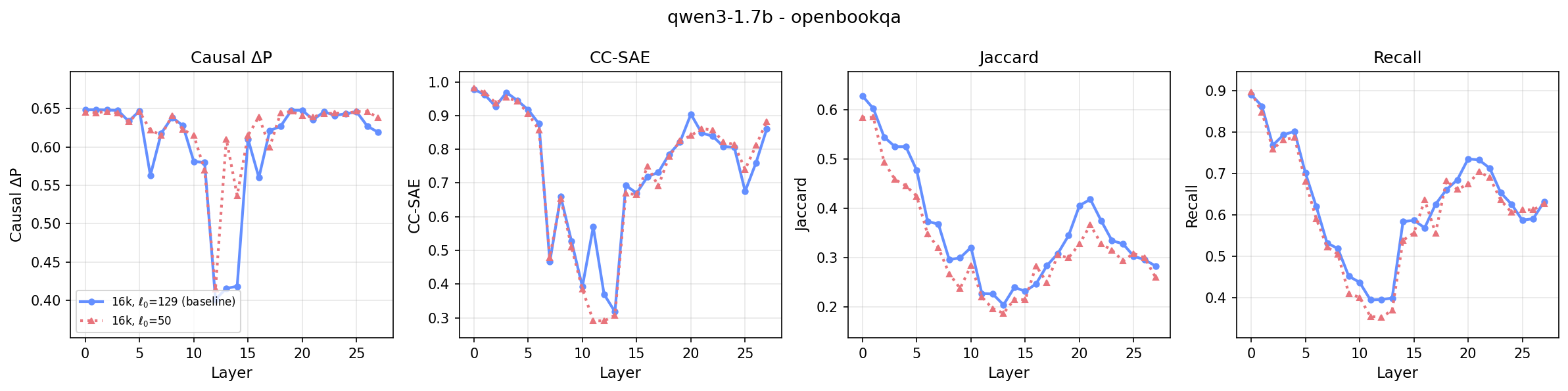}
        \caption{\data{OpenbookQA}}
    \end{subfigure}

     \vspace{0.5em}

    \begin{subfigure}{\textwidth}
        \centering
        \includegraphics[width=\textwidth]{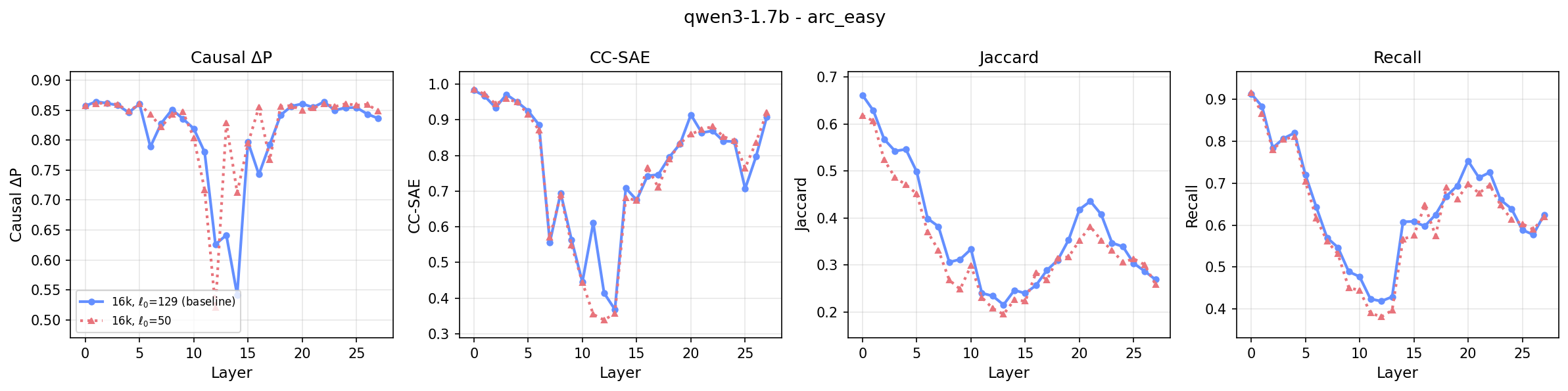}
        \caption{\data{ARC-Easy}}
    \end{subfigure}

    \begin{subfigure}{\textwidth}
        \centering
        \includegraphics[width=\textwidth]{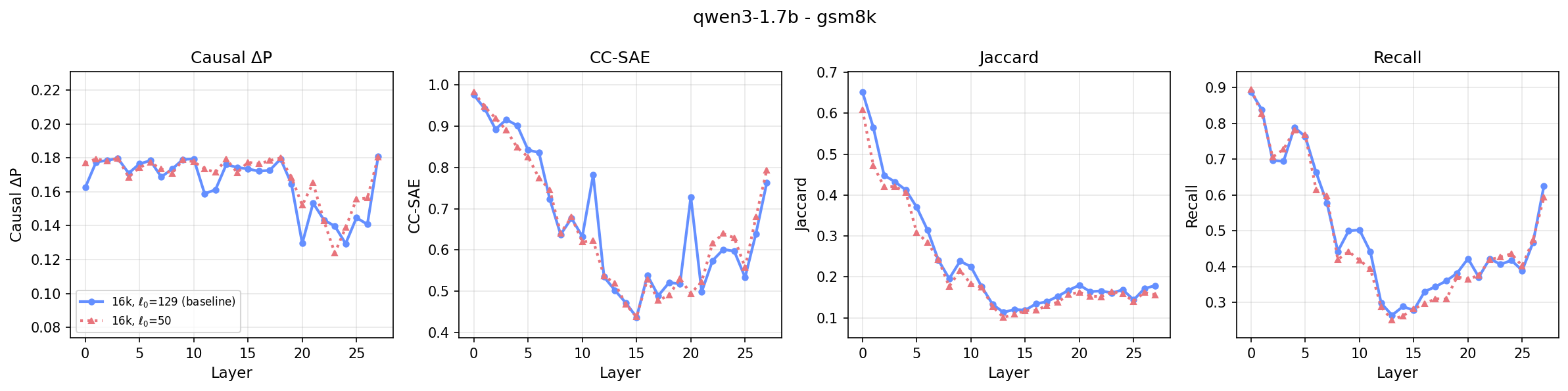}
        \caption{\data{GSM8K}}
    \end{subfigure}

    \caption{Ablation study on \lm{Qwen3-1.7B} with reduced sparsity.}
    \label{fig:ablation-qwen3-1.7b}
\end{figure*}

\begin{figure*}[!t]
    \centering

    \begin{subfigure}{\textwidth}
        \centering
        \includegraphics[width=\textwidth]{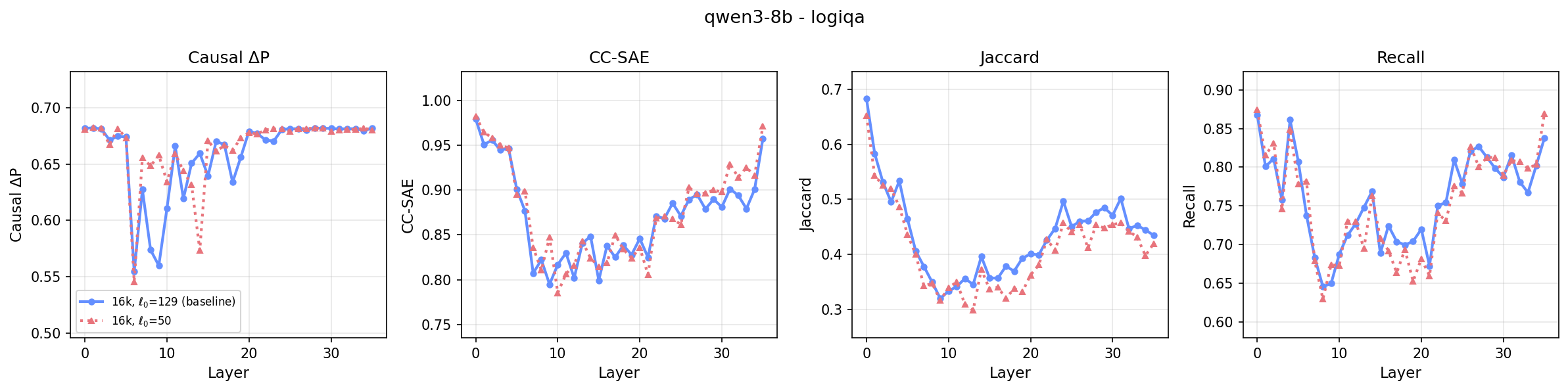}
        \caption{\data{LogiQA}}
    \end{subfigure}

    \vspace{0.5em}

    \begin{subfigure}{\textwidth}
        \centering
        \includegraphics[width=\textwidth]{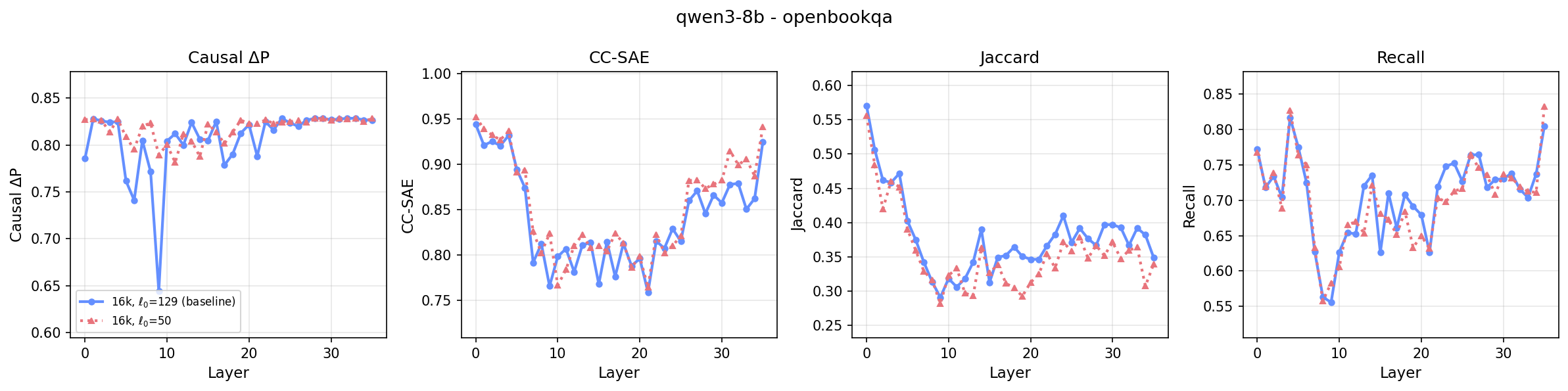}
        \caption{\data{OpenbookQA}}
    \end{subfigure}

    \begin{subfigure}{\textwidth}
        \centering
        \includegraphics[width=\textwidth]{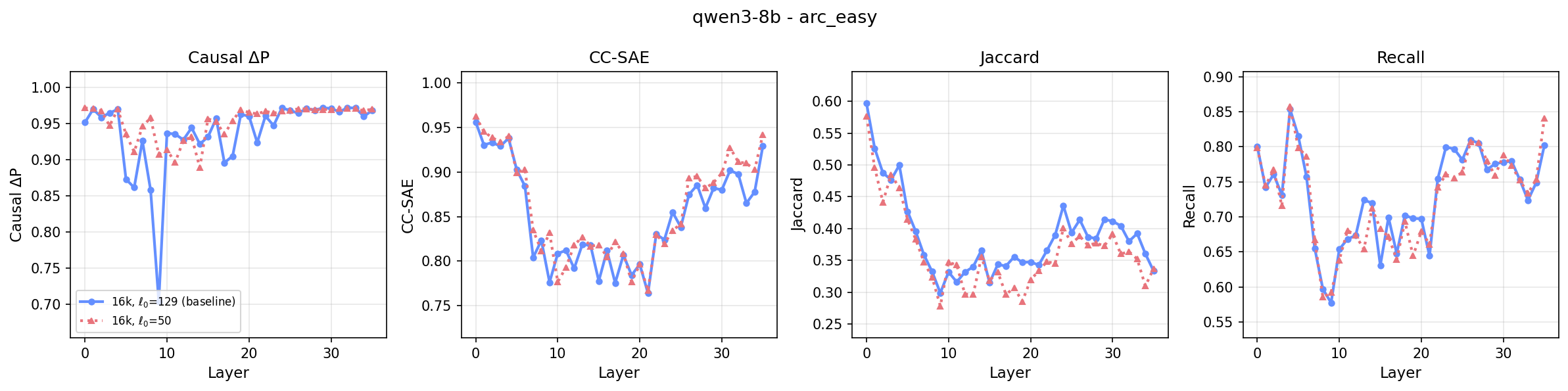}
        \caption{\data{ARC-Easy}}
    \end{subfigure}

    \begin{subfigure}{\textwidth}
        \centering
        \includegraphics[width=\textwidth]{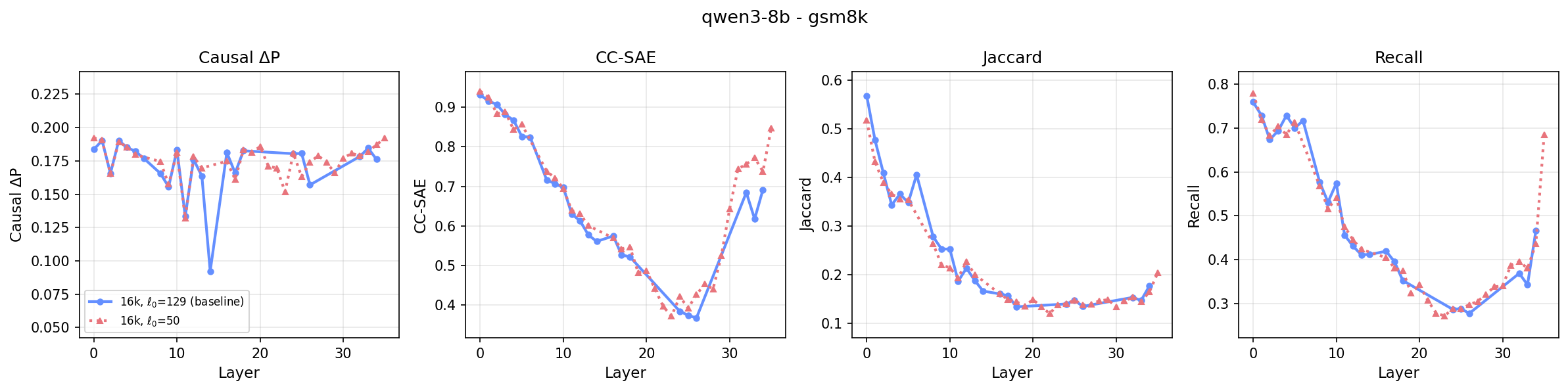}
        \caption{\data{GSM8K}}
    \end{subfigure}

    \caption{Ablation study on \lm{Qwen3-8B} with reduced sparsity.}
    \label{fig:ablation-qwen3-8b}
\end{figure*}

Figure~\ref{fig:ablation-gemma2-2b}--\ref{fig:ablation-qwen3-8b} display layer-wise $\Delta$ across three SAE configurations. Table~\ref{tab:ablation}--\ref{tab:ablation-qwen3-8b} further show the $\Delta$ raw values and Pearson's $r$ correlation values across three SAE configurations.

\begin{table*}[htbp]
\centering
\caption{SAE configuration ablation study results on \lm{Gemma-2-9B}.}
\label{tab:ablation}
    \resizebox{\textwidth}{!}{
\begin{tabular}{llcccc}
\toprule[1.5pt]
& \textbf{Configuration} & \textbf{$\Delta p$} & \textbf{CC-SAE} & \textbf{Jaccard} & \textbf{Recall} \\
\midrule
\multirow{5}{*}{\rotatebox[origin=c]{90}{\small\data{LogiQA}}} & $\text{width} = 16k$, $\ell_0 = 129$ (baseline) & $0.366 \pm 0.034$ & $0.875 \pm 0.069$ & $0.466 \pm 0.112$ & $0.787 \pm 0.102$ \\
\cmidrule(lr){2-6} 

& width=131$k$, $\ell_0 = 129$ & $0.340 \pm 0.050$ & $0.847 \pm 0.076$ & $0.449 \pm 0.115$ & $0.778 \pm 0.105$ \\ 
& \cellcolor{black!10} Pearson's $r$ &  \cellcolor{black!10}$0.880 \pm 0.105$ & \cellcolor{black!10}$0.762 \pm 0.119$ & \cellcolor{black!10}$0.693 \pm 0.105$ & \cellcolor{black!10}$0.761 \pm 0.080$ \\

\cmidrule(lr){2-6} 
& $\text{width} = 16k$, $\ell_0=50$ & $0.350 \pm 0.051$ & $0.872 \pm 0.072$ & $0.525 \pm 0.120$ & $0.819 \pm 0.096$ \\ 
&\cellcolor{black!10} Pearson's $r$ & \cellcolor{black!10} $0.915 \pm 0.095$ & \cellcolor{black!10} $0.756 \pm 0.088$ & \cellcolor{black!10} $0.645 \pm 0.121$ & \cellcolor{black!10} $0.733 \pm 0.071$ \\

  \midrule
  
    \multirow{5}{*}{\rotatebox[origin=c]{90}{\small\data{OpenbookQA}}} & $\text{width} = 16k$, $\ell_0 = 129$ (baseline) & $0.613 \pm 0.059$ & $0.865 \pm 0.073$ & $0.456 \pm 0.120$ & $0.784 \pm 0.113$ \\

    \cmidrule(lr){2-6} 
    
    &  width=131$k$, $\ell_0 = 129$ & $0.567 \pm 0.094$ & $0.829 \pm 0.082$ & $0.418 \pm 0.116$ & $0.761 \pm 0.119$ \\ 
    & \cellcolor{black!10} Pearson's $r$ & \cellcolor{black!10} $0.866 \pm 0.121$ & \cellcolor{black!10} $0.828 \pm 0.065$ & \cellcolor{black!10} $0.703 \pm 0.079$ & \cellcolor{black!10} $0.751 \pm 0.092$ \\

    \cmidrule(lr){2-6} 
    
    & $\text{width} = 16k$, $\ell_0=50$ & $0.595 \pm 0.079$ & $0.856 \pm 0.077$ & $0.498 \pm 0.120$ & $0.796 \pm 0.105$ \\

    & \cellcolor{black!10} Pearson's $r$ & \cellcolor{black!10} $0.914 \pm 0.085$ & \cellcolor{black!10} $0.790 \pm 0.092$ & \cellcolor{black!10} $0.618 \pm 0.116$ & \cellcolor{black!10} $0.699 \pm 0.091$ \\

    \midrule

    \multirow{5}{*}{\rotatebox{90}{\small\data{Arc Easy}}} & width=16$k$, $\ell_0 = 129$ (baseline) & $0.822 \pm 0.209$ & $0.885 \pm 0.078$ & $0.482 \pm 0.134$ & $0.809 \pm 0.119$ \\

 \cmidrule(lr){2-6} 

 & width=131$k$, $\ell_0 = 129$ & $0.765 \pm 0.255$ & $0.855 \pm 0.088$ & $0.449 \pm 0.131$ & $0.792 \pm 0.128$ \\
     
 & \cellcolor{black!10} Pearson's $r$ & \cellcolor{black!10} $0.829 \pm 0.120$ & \cellcolor{black!10} $0.848 \pm 0.066$ & \cellcolor{black!10} $0.667 \pm 0.088$ & \cellcolor{black!10} $0.743 \pm 0.094$ \\

    \cmidrule(lr){2-6} 
 
 & width=16$k$, $\ell_0=50$ & $0.798 \pm 0.227$ & $0.878 \pm 0.083$ & $0.528 \pm 0.140$ & $0.820 \pm 0.117$ \\
 & \cellcolor{black!10} Pearson's $r$ & \cellcolor{black!10} $0.797 \pm 0.187$ & \cellcolor{black!10} $0.829 \pm 0.093$ & \cellcolor{black!10} $0.598 \pm 0.138$ & \cellcolor{black!10} $0.713 \pm 0.109$ \\

\midrule

\multirow{5}{*}{\rotatebox{90}{\small\data{GSM8K}}} & $\text{width} = 16k$, $\ell_0 = 129$ (baseline) & $0.132 \pm 0.188$ & $0.657 \pm 0.219$ & $0.293 \pm 0.178$ & $0.556 \pm 0.222$ \\
    
\cmidrule(lr){2-6} 
    
& $\text{width} = 131k$, $\ell_0 = 129$ & $0.115 \pm 0.177$ & $0.561 \pm 0.256$ & $0.242 \pm 0.181$ & $0.469 \pm 0.245$ \\
& \cellcolor{black!10}Pearson's $r$ & \cellcolor{black!10}$0.837 \pm 0.075$ & \cellcolor{black!10}$0.916 \pm 0.063$ & \cellcolor{black!10}$0.873 \pm 0.089$ & \cellcolor{black!10}$0.889 \pm 0.068$ \\

\cmidrule(lr){2-6} 

& $\text{width} = 16k$, $\ell_0 = 50$ & $0.123 \pm 0.182$ & $0.595 \pm 0.256$ & $0.307 \pm 0.207$ & $0.533 \pm 0.242$ \\
& \cellcolor{black!10}Pearson's $r$ & \cellcolor{black!10}$0.864 \pm 0.069$ & \cellcolor{black!10}$0.934 \pm 0.048$ & \cellcolor{black!10}$0.865 \pm 0.098$ & \cellcolor{black!10}$0.890 \pm 0.067$ \\

\toprule[1.5pt]
\end{tabular}
}

\end{table*}

\begin{table*}[htbp]
\centering
\caption{SAE configuration ablation study results on \lm{Qwen3-1.7B}.}
\label{tab:ablation-qwen3-1.7b}
\resizebox{\textwidth}{!}{
\begin{tabular}{llcccc}
\toprule
 & \textbf{Configuration} & \textbf{$\Delta p$} & \textbf{CC-SAE} & \textbf{Jaccard} & \textbf{Recall} \\
\midrule
\multirow{3}{*}{\rotatebox{90}{\scriptsize\data{LogiQA}}} & $\text{width} = 16k$, $\ell_0 = 129$ (baseline) & $0.511 \pm 0.442$ & $0.832 \pm 0.112$ & $0.404 \pm 0.132$ & $0.673 \pm 0.125$ \\

\cmidrule(lr){2-6} 

 & $\text{width} = 16k$, $\ell_0=50$ & $0.512 \pm 0.441$ & $0.831 \pm 0.110$ & $0.381 \pm 0.126$ & $0.656 \pm 0.130$ \\
 & \cellcolor{black!10} Pearson's $r$ & \cellcolor{black!10} $0.958 \pm 0.075$ & \cellcolor{black!10} $0.607 \pm 0.182$ & \cellcolor{black!10} $0.493 \pm 0.063$ & \cellcolor{black!10} $0.624 \pm 0.070$ \\
\midrule

\multirow{3}{*}{\rotatebox{90}{\scriptsize\data{OpenbookQA}}} & $\text{width} = 16k$, $\ell_0 = 129$ (baseline) & $0.603 \pm 0.452$ & $0.743 \pm 0.196$ & $0.358 \pm 0.125$ & $0.627 \pm 0.148$ \\

\cmidrule(lr){2-6}

 & $\text{width} = 16k$, $\ell_0=50$ & $0.621 \pm 0.445$ & $0.731 \pm 0.216$ & $0.326 \pm 0.114$ & $0.609 \pm 0.155$ \\
 & \cellcolor{black!10} Pearson's $r$ & \cellcolor{black!10} $0.914 \pm 0.107$ & \cellcolor{black!10} $0.604 \pm 0.160$ & \cellcolor{black!10} $0.411 \pm 0.063$ & \cellcolor{black!10} $0.576 \pm 0.065$ \\

\midrule

\multirow{3}{*}{\rotatebox{90}{\scriptsize\data{ARC-Easy}}} & $\text{width} = 16k$, $\ell_0 = 129$ (baseline) & $0.811 \pm 0.366$ & $0.768 \pm 0.186$ & $0.371 \pm 0.132$ & $0.643 \pm 0.145$ \\

 \cmidrule(lr){2-6}

 & $\text{width} = 16k$, $\ell_0=50$ & $0.824 \pm 0.356$ & $0.757 \pm 0.206$ & $0.341 \pm 0.122$ & $0.626 \pm 0.152$ \\
 &\cellcolor{black!10} Pearson's $r$ & \cellcolor{black!10}$0.852 \pm 0.151$ &\cellcolor{black!10} $0.596 \pm 0.204$ &\cellcolor{black!10} $0.415 \pm 0.047$ & \cellcolor{black!10}$0.578 \pm 0.088$ \\

\midrule

\multirow{3}{*}{\rotatebox{90}{\scriptsize\data{GSM8K}}} & $\text{width} = 16k$, $\ell_0 = 129$ (baseline) & $0.165 \pm 0.240$ & $0.670 \pm 0.180$ & $0.240 \pm 0.152$ & $0.497 \pm 0.199$ \\

\cmidrule(lr){2-6} 

& $\text{width} = 16k$, $\ell_0=50$ & $0.169 \pm 0.239$ & $0.657 \pm 0.177$ & $0.220 \pm 0.140$ & $0.479 \pm 0.208$ \\
& \cellcolor{black!10}Pearson's $r$ & \cellcolor{black!10}$0.941 \pm 0.058$ &\cellcolor{black!10} $0.800 \pm 0.145$ & \cellcolor{black!10}$0.673 \pm 0.155$ & \cellcolor{black!10}$0.786 \pm 0.123$ \\

\bottomrule
\end{tabular}
}

\end{table*}


\begin{table*}[htbp]
\centering
\caption{SAE configuration ablation study results on \lm{Qwen3-8B}.}
\label{tab:ablation-qwen3-8b}
\resizebox{\textwidth}{!}{

\begin{tabular}{llcccc}
\toprule
 & \textbf{Configuration} & \textbf{$\Delta p$} & \textbf{CC-SAE} & \textbf{Jaccard} & \textbf{Recall} \\
\midrule
\multirow{3}{*}{\rotatebox{90}{\scriptsize\data{LogiQA}}} & $\text{width} = 16k$, $\ell_0 = 129$ (baseline) & $0.659 \pm 0.396$ & $0.872 \pm 0.060$ & $0.434 \pm 0.092$ & $0.759 \pm 0.081$ \\
\cmidrule(lr){2-6}

 & $\text{width} = 16k$, $\ell_0=50$ & $0.665 \pm 0.392$ & $0.877 \pm 0.065$ & $0.410 \pm 0.091$ & $0.754 \pm 0.089$ \\
 & \cellcolor{black!10} Pearson's $r$ & \cellcolor{black!10} $0.935 \pm 0.091$ & \cellcolor{black!10} $0.464 \pm 0.157$ & \cellcolor{black!10} $0.381 \pm 0.091$ & \cellcolor{black!10} $0.551 \pm 0.101$ \\
\midrule
\multirow{3}{*}{\rotatebox{90}{\scriptsize\data{OpenbookQA}}} & $\text{width} = 16k$, $\ell_0 = 129$ (baseline) & $0.807 \pm 0.334$ & $0.841 \pm 0.063$ & $0.379 \pm 0.074$ & $0.706 \pm 0.081$ \\

\cmidrule(lr){2-6}

 & $\text{width} = 16k$, $\ell_0=50$ & $0.818 \pm 0.327$ & $0.853 \pm 0.064$ & $0.356 \pm 0.073$ & $0.699 \pm 0.082$ \\
 & \cellcolor{black!10} Pearson's $r$ & \cellcolor{black!10} $0.929 \pm 0.085$ & \cellcolor{black!10} $0.722 \pm 0.181$ & \cellcolor{black!10} $0.541 \pm 0.123$ & \cellcolor{black!10} $0.720 \pm 0.176$ \\
\midrule

\multirow{3}{*}{\rotatebox{90}{\scriptsize\data{ARC-Easy}}} & width=16$k$, $\ell_0 = 129$ (baseline) & $0.939 \pm 0.192$ & $0.853 \pm 0.063$ & $0.388 \pm 0.081$ & $0.732 \pm 0.086$ \\

\cmidrule(lr){2-6} 

 & width=16$k$, $\ell_0=50$ & $0.952 \pm 0.166$ & $0.861 \pm 0.064$ & $0.365 \pm 0.079$ & $0.727 \pm 0.089$ \\
 & \cellcolor{black!10}Pearson's $r$ & \cellcolor{black!10} $0.806 \pm 0.166$ & \cellcolor{black!10} $0.778 \pm 0.218$ & \cellcolor{black!10} $0.600 \pm 0.152$ &\cellcolor{black!10} $0.768 \pm 0.187$ \\

 \midrule
\multirow{3}{*}{\rotatebox{90}{\scriptsize\data{GSM8K}}} & $\text{width} = 16k$, $\ell_0 = 129$ (baseline) & $0.175 \pm 0.284$ & $0.689 \pm 0.182$ & $0.266 \pm 0.134$ & $0.520 \pm 0.183$ \\

\cmidrule(lr){2-6} 

& $\text{width} = 16k$, $\ell_0=50$ & $0.175 \pm 0.284$ & $0.636 \pm 0.195$ & $0.209 \pm 0.116$ & $0.450 \pm 0.186$ \\
& \cellcolor{black!10}Pearson's $r$ & \cellcolor{black!10}$0.946 \pm 0.040$ & \cellcolor{black!10}$0.632 \pm 0.121$ & \cellcolor{black!10}$0.483 \pm 0.117$ & \cellcolor{black!10}$0.594 \pm 0.114$ \\

\bottomrule
\end{tabular}
}

\end{table*}

\subsection{$\ell_0$ Variation across Layers}
\lm{Qwen3} and \lm{Llama3.1} models use TopK-ReLU activations, which therefore have the same $\ell_0$ value across layers. However, \lm{Gemma-2} models employ JumpReLU and the $\ell_0$ values vary across layers. As a sanity check, we investigate how $\ell_0$ correlates with $\Delta p$. In our experiments, we consider the $\ell_0$ value at the first layer as the anchor and consistently choose the closest $\ell_0$ from the available $\ell_0$ set at the other layers to largely maintain the $\ell_0$ consistency.\footnote{\url{https://huggingface.co/google/gemma-scope}} 

\begin{figure}[h]
    \centering
    \begin{subfigure}[b]{0.48\textwidth}
        \centering
        \includegraphics[width=\textwidth]{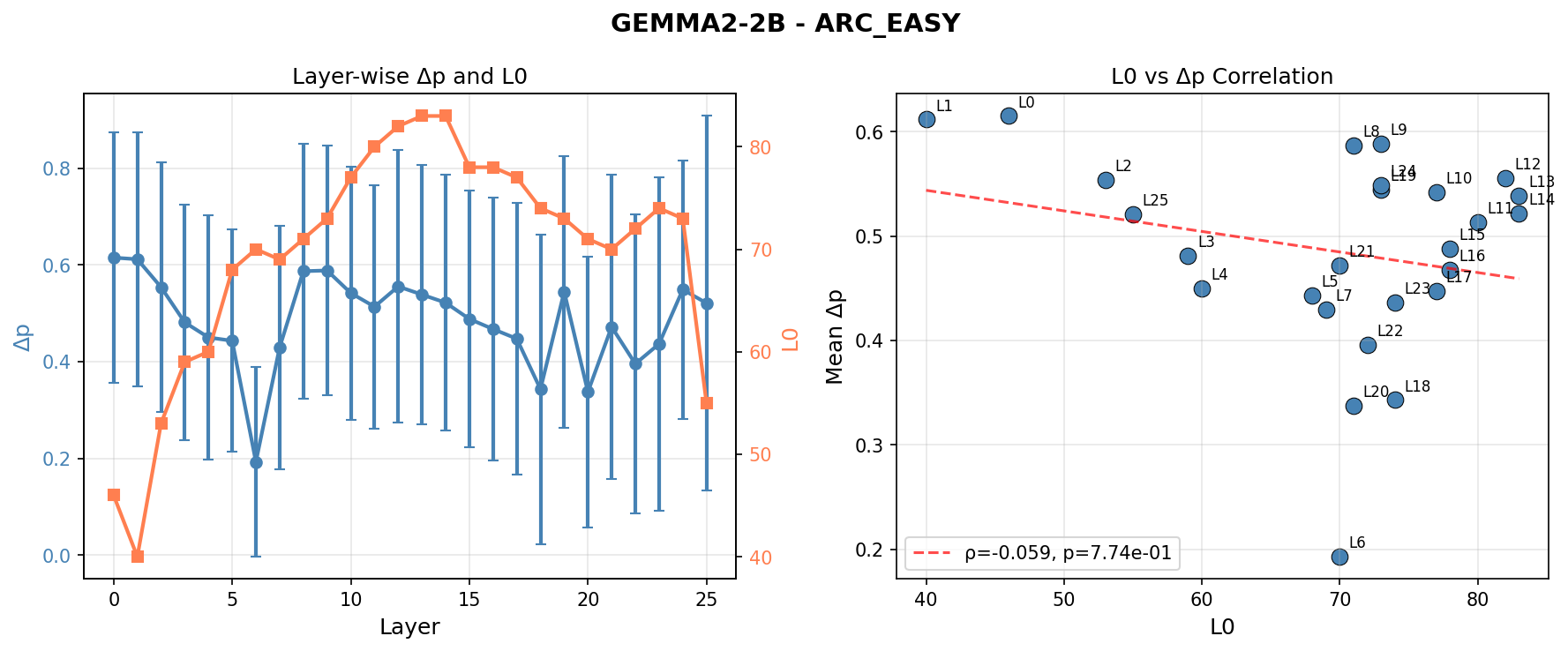}
        \caption{\data{ARC-Easy}}
        \label{fig:corr_arc}
    \end{subfigure}
    \hfill
    \begin{subfigure}[b]{0.48\textwidth}
        \centering
        \includegraphics[width=\textwidth]{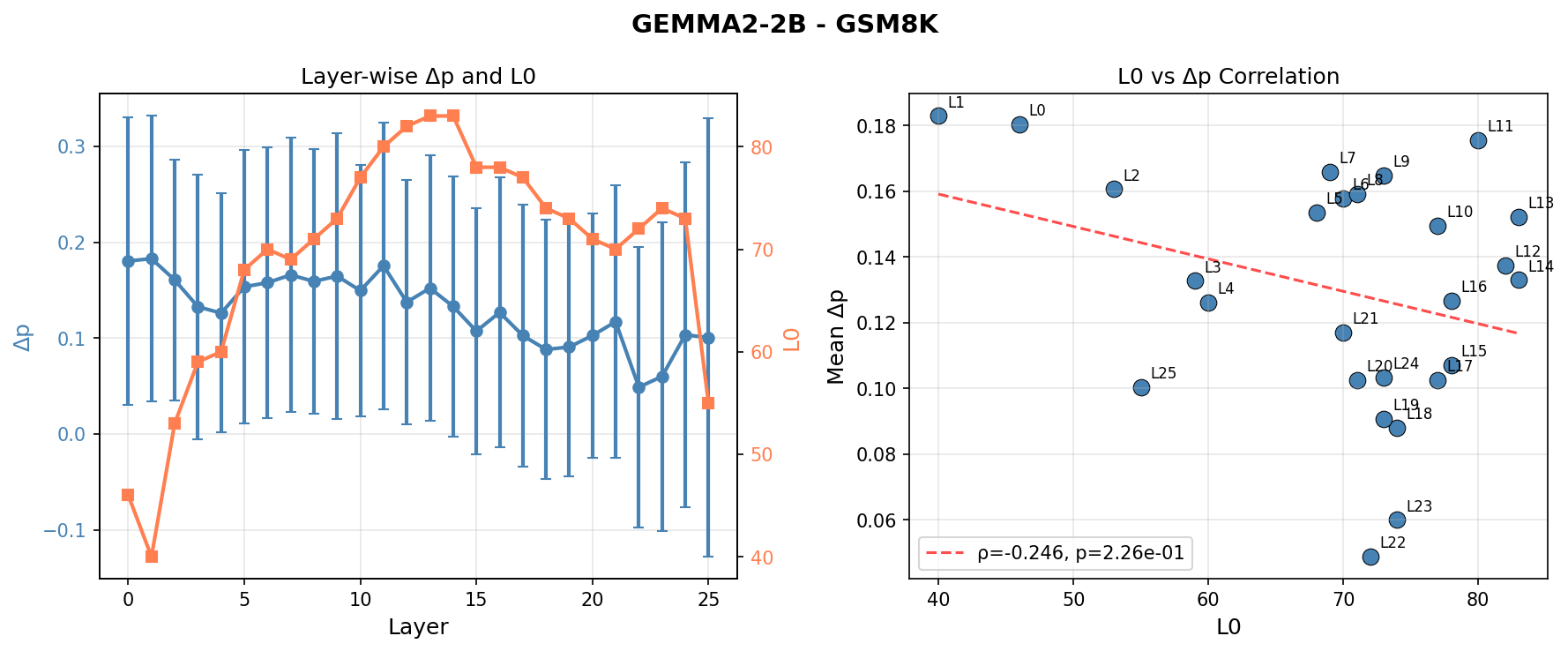}
        \caption{\data{GSM8K}}
        \label{fig:corr_gsm}
    \end{subfigure}

    \vspace{0.4cm} 

    \begin{subfigure}[b]{0.48\textwidth}
        \centering
        \includegraphics[width=\textwidth]{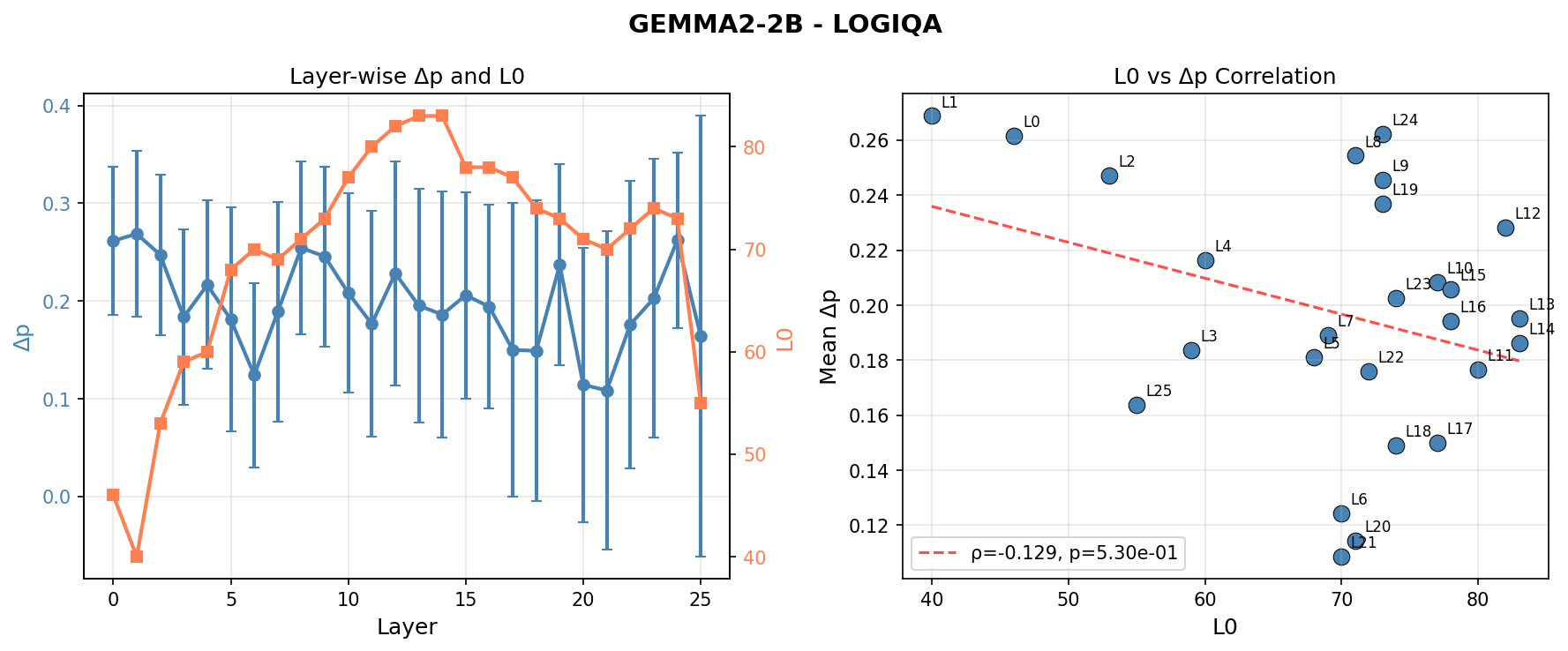}
        \caption{\data{LogiQA}}
        \label{fig:corr_logi}
    \end{subfigure}
    \hfill
    \begin{subfigure}[b]{0.48\textwidth}
        \centering
        \includegraphics[width=\textwidth]{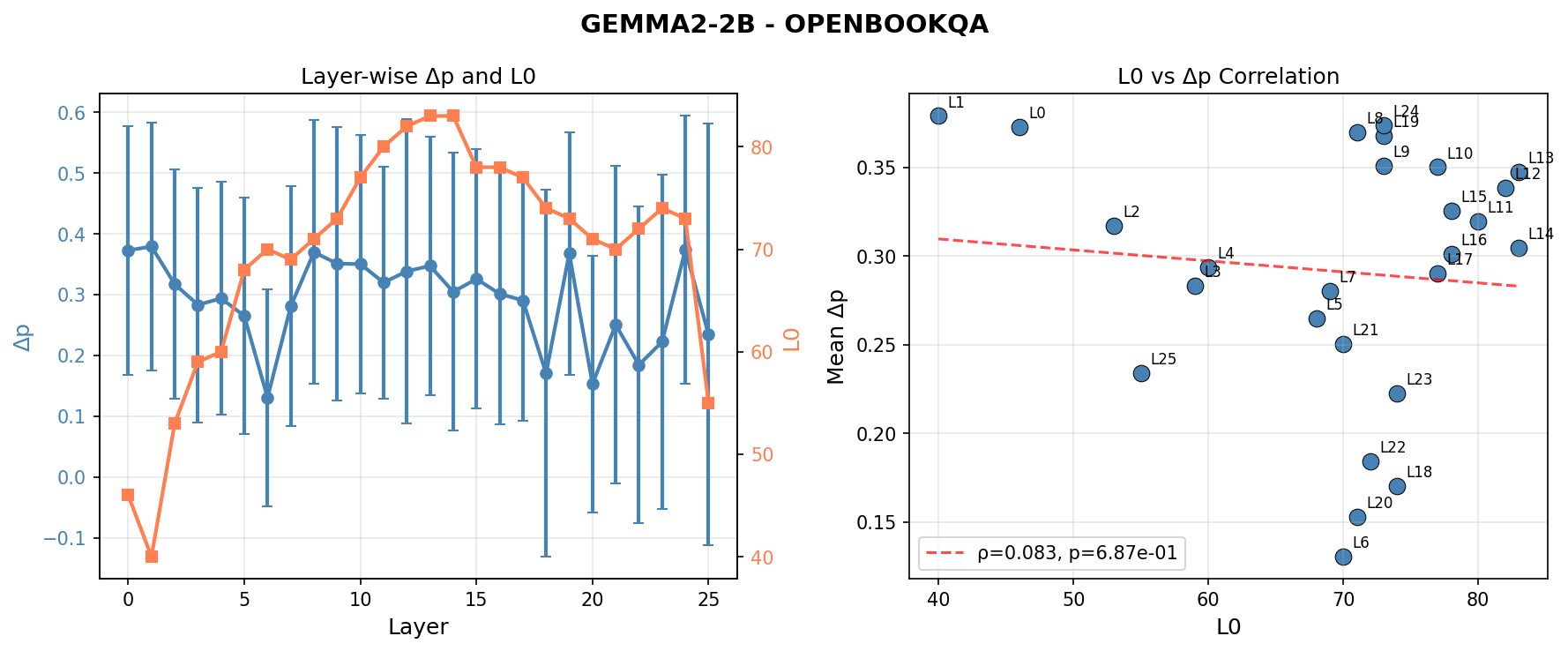}
        \caption{\data{OpenbookQA}}
        \label{fig:corr_open}
    \end{subfigure}

    \caption{$\ell_0$ correlation analysis for \lm{Gemma-2-2B} across four different datasets.}
    \label{fig:gemma2_2b_correlation}
\end{figure}

\begin{figure}[h]
    \centering
    \begin{subfigure}[b]{0.48\textwidth}
        \centering
        \includegraphics[width=\textwidth]{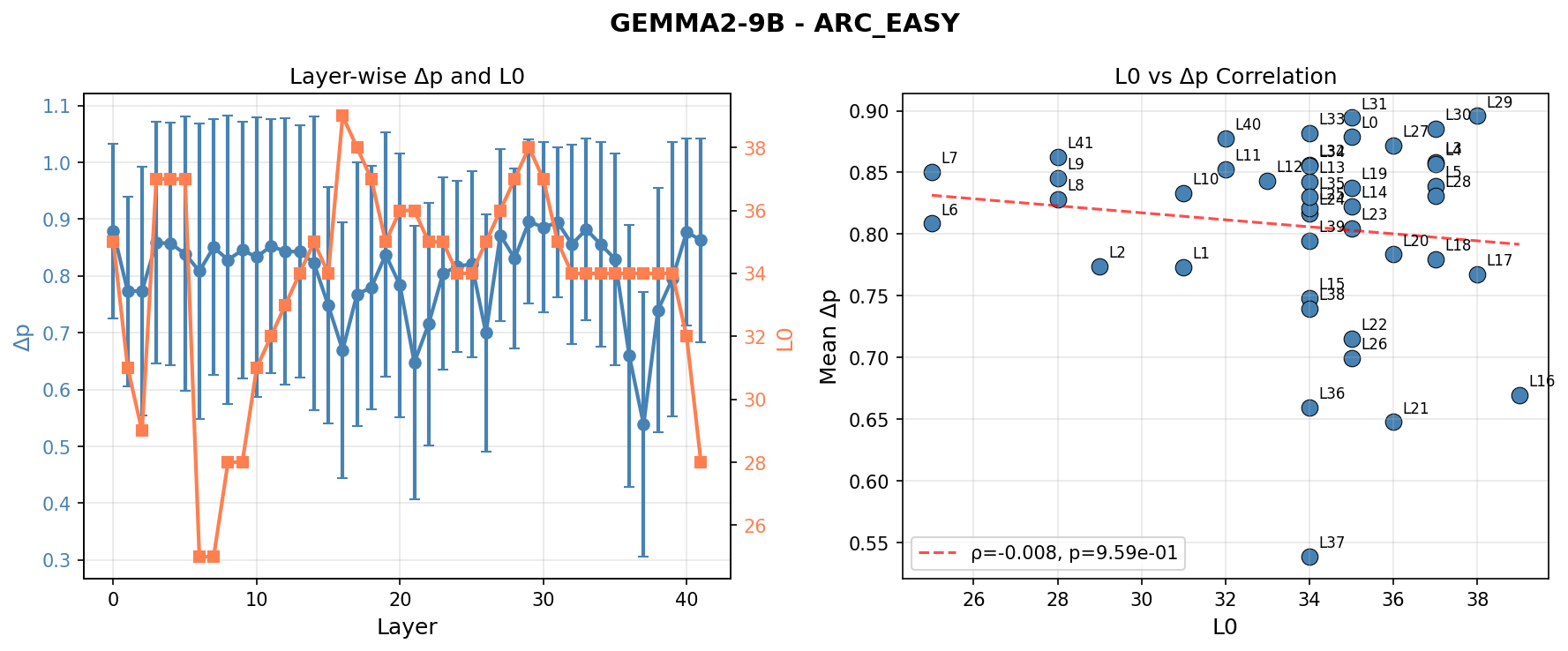}
        \caption{\data{ARC-Easy}}
        \label{fig:corr_arc}
    \end{subfigure}
    \hfill
    \begin{subfigure}[b]{0.48\textwidth}
        \centering
        \includegraphics[width=\textwidth]{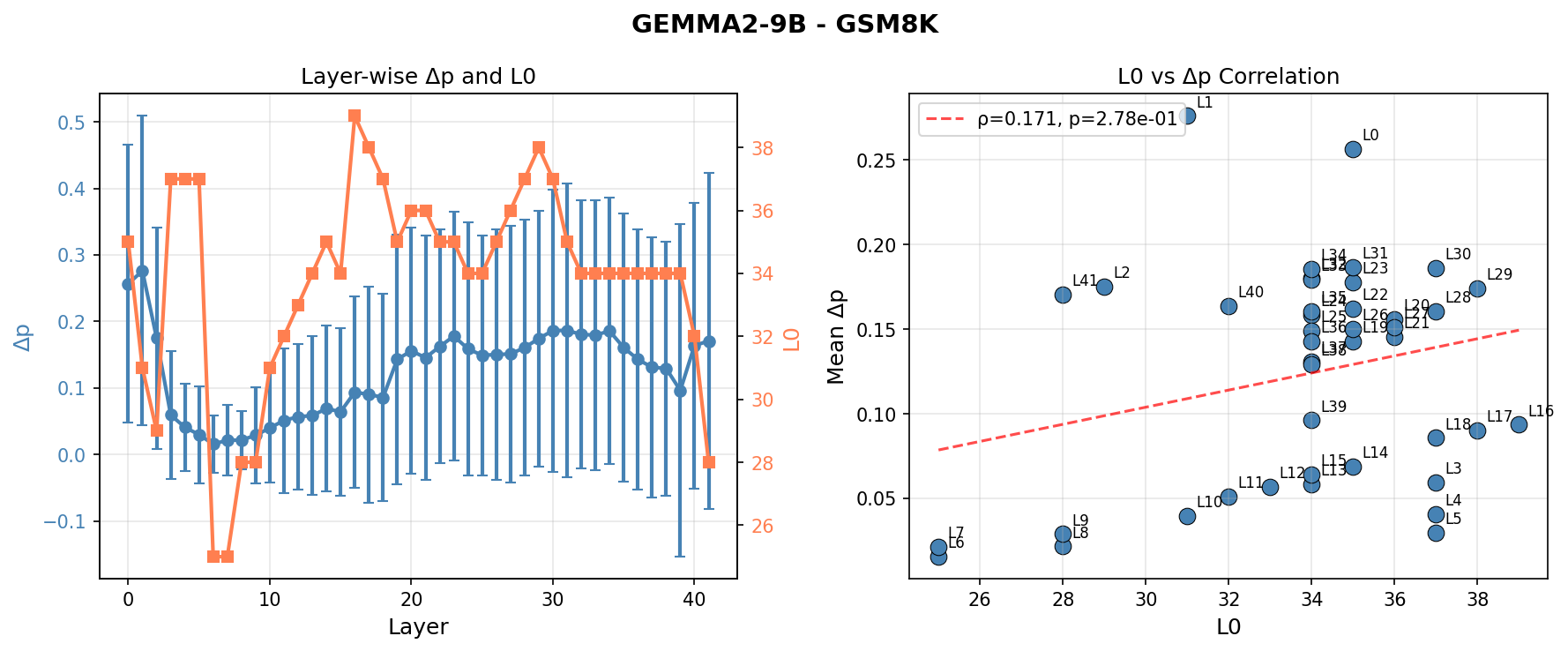}
        \caption{\data{GSM8K}}
        \label{fig:corr_gsm}
    \end{subfigure}

    \vspace{0.4cm} 

    \begin{subfigure}[b]{0.48\textwidth}
        \centering
        \includegraphics[width=\textwidth]{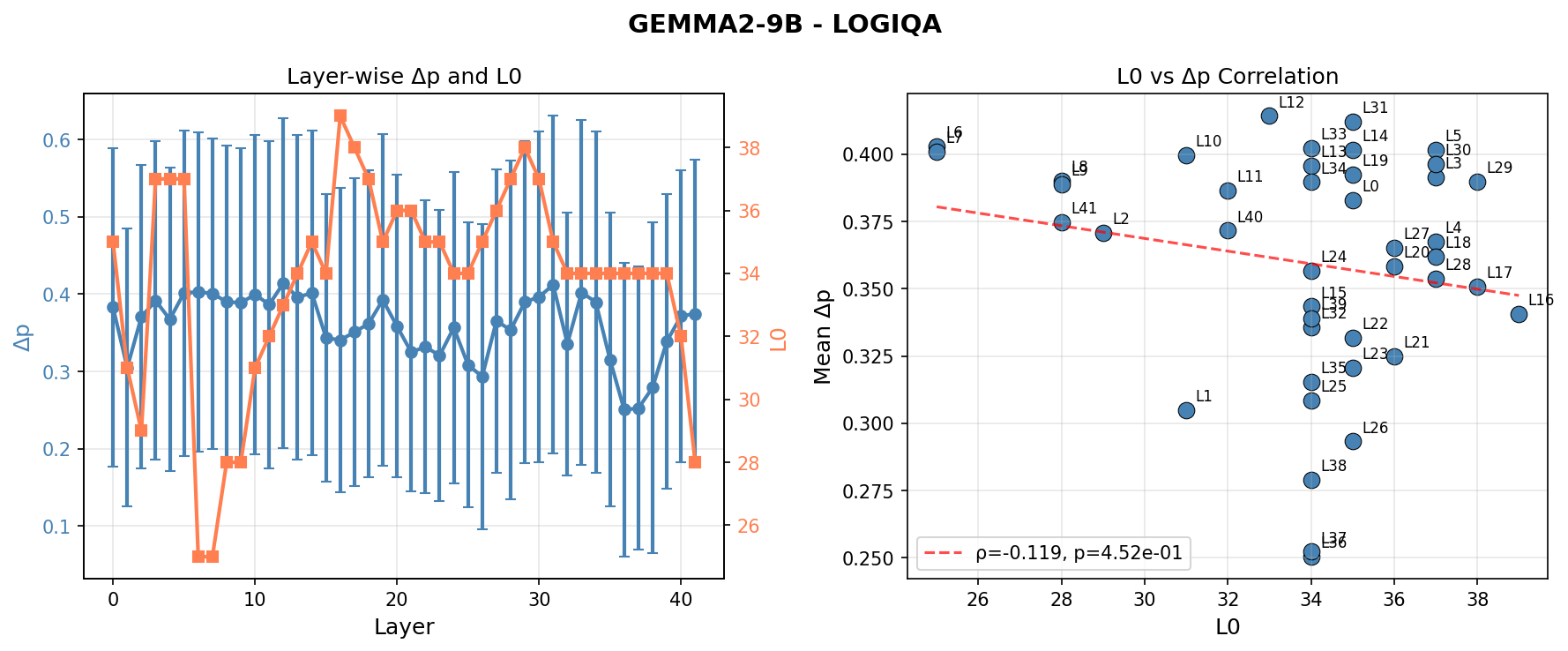}
        \caption{\data{LogiQA}}
        \label{fig:corr_logi}
    \end{subfigure}
    \hfill
    \begin{subfigure}[b]{0.48\textwidth}
        \centering
        \includegraphics[width=\textwidth]{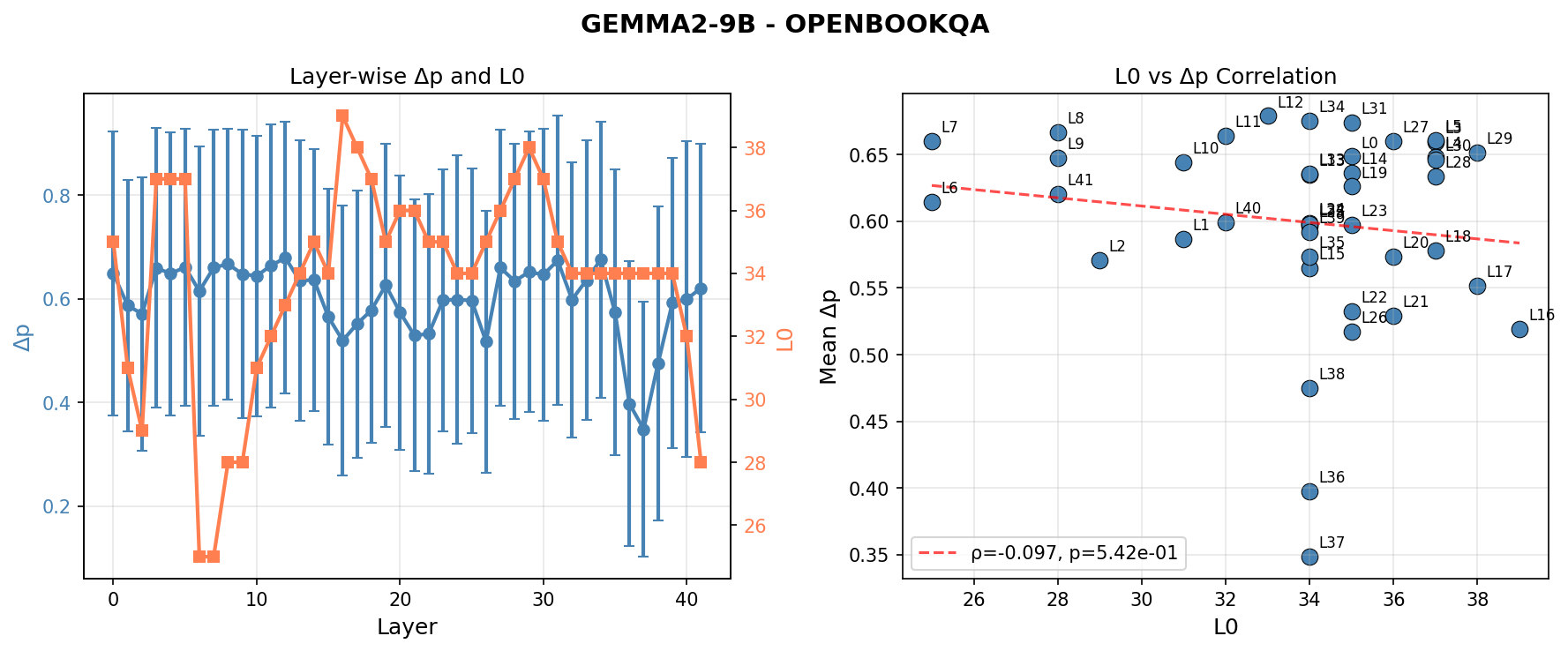}
        \caption{\data{OpenbookQA}}
        \label{fig:corr_open}
    \end{subfigure}

    \caption{$\ell_0$ correlation analysis for \lm{Gemma-2-9B} across four different datasets.}
    \label{fig:gemma2_9b_correlation}
\end{figure}

Figures~\ref{fig:gemma2_2b_correlation} and \ref{fig:gemma2_9b_correlation} demonstrate that the correlation between $\ell_0$ and $\Delta p$ is generally negligible. This indicates that the variation of $\ell_0$ should have minimal impact on the ability to detect shared concepts.

\subsection{SAE Reconstruction}
In order to separate model effects from SAE effects, we examine the SAE reconstruction with \lm{Qwen3-1.7B} on \data{LogiQA} as a sanity check. Specifically, we evaluate SAE reconstruction quality using the Fraction of Variance Unexplained (FVU), defined as:
\begin{equation}
    \text{FVU} = \frac{|x - \hat{x}|^2}{|x - \bar{x}|^2}    
\end{equation}
where $\hat{x} = \text{Dec}(\text{Enc}(x))$ is the SAE reconstruction of residual stream activation $x$. The metric is bounded meaningfully between 0 and 1. An FVU of 0 corresponds to perfect reconstruction, where the SAE recovers the input activations exactly and leaves no residual variance unexplained. An FVU of 1 corresponds to a reconstruction no better than trivially predicting the dataset mean for every input, i.e., the SAE explains none of the variance beyond a constant baseline. 

\begin{figure}[t]
  \centering
  \includegraphics[width=0.7\linewidth]{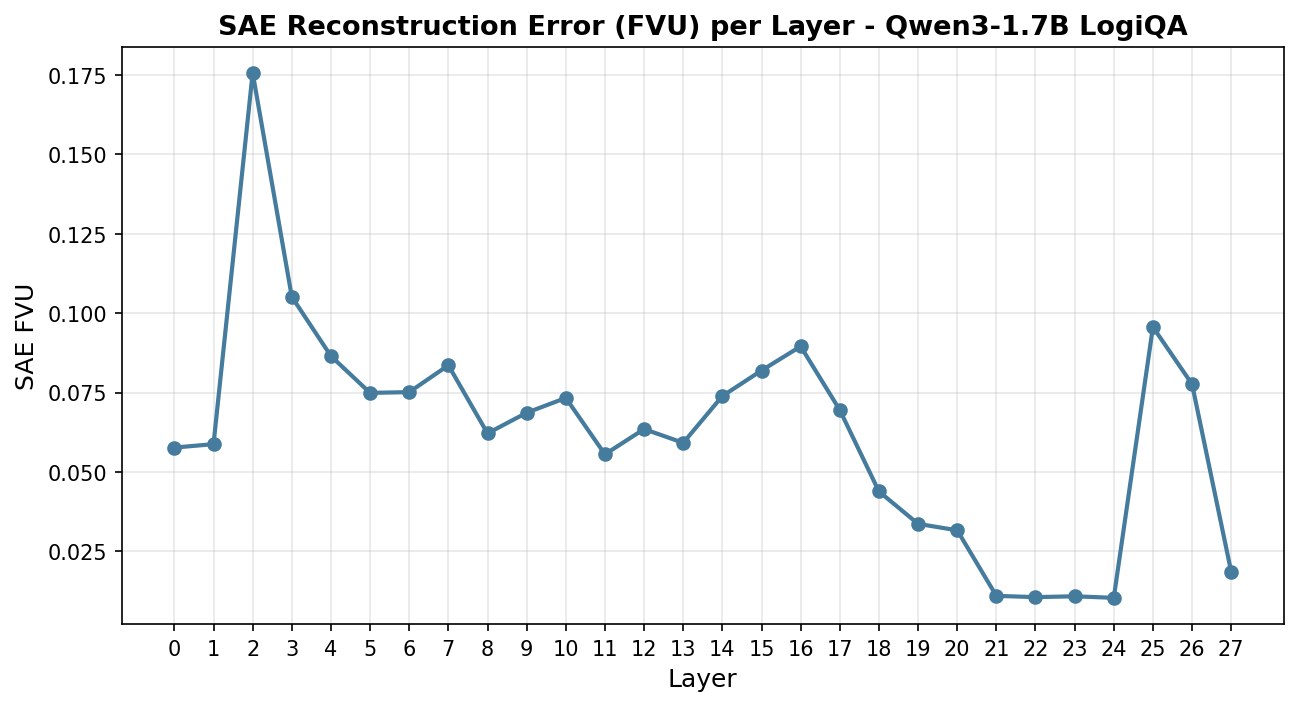}
  \caption{Layer-wise SAE reconstruction quality (FVU) for \lm{Qwen3-1.7B} on \data{LogiQA}. FVU is the ratio of the SAE's reconstruction error to the total variance of the original activations, so lower values indicate better reconstruction.}
  \label{fig:sae-fvu-layerwise}
\end{figure}

Across layers, Figure~\ref{fig:sae-fvu-layerwise} shows that FVU ranges from 0.01 to 0.18, with the strongest reconstruction in late layers (L20--L24, FVU$\approx$ 0.01--0.03) and higher residual error in early-to-middle layers (L2--L7, FVU $\approx$ 0.07---0.18). For most layers FVU falls below 0.1, corresponding to over 90\% of the activation variance being recovered -- a fidelity comparable to or better than that reported for SAEs in prior work \citep{lieberum-etal-2024-gemma, rajamanoharan2024jumpingaheadimprovingreconstruction, gao2025scaling}. This provides a favorable basis for the feature-extraction and causal-intervention experiments that follow.


\section{Shared Concept Transition across Layers}
\label{app:concept_transition}

\subsection{SAE Feature Description Generation}
\label{app:feature_description}

We detail the automated interpretability pipeline used to map the shared concepts $S_\cap$ to natural-language descriptions, following \citet{pmlr-v267-paulo25a} and using \lm{gemini-3-flash-preview}\footnote{\url{https://ai.google.dev/gemini-api/docs/models/gemini-3-flash-preview}} as the explainer. Specifically, for each shared SAE feature we mine its top-activating exemplars from the dataset's question and option texts together with the generated CoT traces. Each exemplar is a $32$-token window centered on the peak-activating token, within which we mark the contiguous high-activation span and quantize its per-token activations to an integer scale from $0$ to $10$. We then prompt the explainer to summarize these exemplars into a single concise description. All case studies use \lm{Gemma-2-9B} with a compact width-$16k$ SAE applied at every layer.

\subsection{Case Studies}
\label{app:concept_case_studies}
Tables~\ref{tab:qual-gemma9b-logiqa} -- \ref{tab:qual-gemma9b-gsm8k} report one qualitative instance each from \data{LogiQA}, \data{OpenbookQA}, \data{ARC-Easy}, and \data{GSM8K}. For every instance, we show the question and its options, the predicted answers under the prediction and CoT passes together with the gold answer, the generated CoT, and, for each layer, the shared SAE features $S_\cap$ ablated in the causal intervention (\S\ref{subsec:causal}), reporting their descriptions, pre-ablation activations, and the per-layer $\Delta p$. $|S_\cap|$ is typically large, so we cannot list every layer or every shared concept in each table. We instead hand-select a few features from $S_\cap$ whose descriptions are topically relevant to the question. 
Across all four datasets, the shared SAE features trace a consistent trajectory: early layers encode SAE features tied to the question topic (e.g., cultural entities in \data{LogiQA}, spatial orientation in \data{OpenbookQA}), while middle and late layers activate more abstract, answer-evaluation concepts. Additionally, $\Delta p$ is largest at layers whose SAE features are semantically relevant to the answer. Additionally, we observe that not all SAE features are verbalized in the CoT, while not all verbalizations correspond to highly activated concepts across layers.

\begin{table*}[t]
\centering
\caption{Qualitative instance from \lm{Gemma-2-9B} on \data{LogiQA}: question, choices, CoT rationale, and prediction under both conditions, together with the shared concepts $S_\cap$ ablated in the causal intervention (\S\ref{subsec:causal}) and their feature descriptions and pre-ablation activation magnitudes.}
\label{tab:qual-gemma9b-logiqa}
\small
\renewcommand{\arraystretch}{1.2}
\arrayrulecolor{purple}
\begin{tcolorbox}[
    colback=lightblue,
    colframe=purple,
    boxrule=0.7pt,
    arc=3pt,
    left=8pt, right=8pt, top=5pt, bottom=6pt,
    title={\lm{Gemma-2-9B} \ \textcolor{purple!40!black}{$\vert$} \ \data{LogiQA}},
    fonttitle=\bfseries,
    colbacktitle=lighpurple,
    coltitle=black,
    boxsep=2pt
]
\textbf{Question.} Many elderly Beijings have never forgotten the unique skills they saw at temple fairs when they were young. Today, these unique skills have more formal names—folk art. However, as the modernization of society accelerates, Chinese folk culture is facing unprecedented crisis. The survival crisis of the city, the continuous change of the urban environment, the rapid diversion and transfer of people's interests and hobbies, and the gradual loss of folk art talents, all of which make the development of folk art face difficulties.
Which of the following can be derived from this text?
\vspace{3pt}
\begin{itemize}[leftmargin=1.4em, itemsep=1.5pt, topsep=3pt]
    \item[\textbf{A:}] Marketization is the way out for folk art
    \item[\textbf{B:}] Folk culture needs rescue protection \goldicon
    \item[\textbf{C:}] Urban construction should highlight cultural characteristics
    \item[\textbf{D:}] The social status of folk art talents should be improved
\end{itemize}
\textbf{Prediction Pass:} B~\okmark \quad
\textbf{CoT Pass:} B~\okmark \quad
\textbf{Gold:}\nobreak\ \mbox{B~\goldicon}
\noindent\textcolor{purple!60!black}{\rule{\linewidth}{0.5pt}}
\textbf{CoT:}~\textit{``The text says that the survival of folk art is facing unprecedented crisis. The text also says that the development of folk art is facing difficulties. So, the answer is B. The text does not say that marketization is the way out for folk art. The text does not say that urban construction should highlight cultural characteristics. The text does not say that the social status of folk art talents should be improved. So, the answer is B. \ldots''}
\noindent\textcolor{purple!60!black}{\rule{\linewidth}{0.5pt}}
\centering
\begin{tabular}{@{}c c c c p{7.5cm}@{}}
\rowcolor{white}
\arrayrulecolor{purple!60!black}
\toprule
\textbf{Layer} & \textbf{$\Delta p$} & \textbf{Feature ID ($S_\cap$)} & \textbf{Act.} & \textbf{Feature Description} \\
\midrule
\rowcolor{lightblue!50}
\arrayrulecolor{purple!30}\cmidrule(lr){1-5}
\rowcolor{lighpurple!50}
\multirow{2}{*}{\textbf{L5}} & \multirow{2}{*}{\textbf{0.67}}
    & \texttt{\#15137} & 1.9 &
Proper nouns and specific cultural or geographic identifiers associated with China, Hong Kong, Taiwan, and Chinese political or social contexts. \\
\rowcolor{lighpurple!50}
 & & \colorbox{yellow!35}{\texttt{\#4649}} & 6.2 & Specialized compound nouns and terminology related to social sciences, economics, and formal academic assessments. \\
 \arrayrulecolor{purple!30}\cmidrule(lr){1-5}
\rowcolor{lightblue!50}
\multirow{2}{*}{\textbf{L8}} & \multirow{2}{*}{\textbf{0.67}}
    & \texttt{\#4149} & 2.9 &
Specific entities, proper nouns, and key subjects within logical reasoning problems or translated Chinese cultural contexts. \\
\rowcolor{lightblue!50}
 & & \colorbox{yellow!35}{\texttt{\#16084}} & 3.0 & Formal and technical terminology related to business, industry trends, and economic development. \\
 
\arrayrulecolor{purple!30}\cmidrule(lr){1-5}
\rowcolor{lighpurple!50}
\multirow{2}{*}{\textbf{L11}} & \multirow{2}{*}{\textbf{0.67}}
    & \texttt{\#10504} & 2.9 &
Pinyin transliterations of Chinese proper nouns, including personal names, locations, and cultural terms. \\
\rowcolor{lighpurple!50}
 & & \texttt{\#5924} & 2.4 & Proper nouns of Chinese or Taiwanese origin and structural formatting elements used in standardized test questions and logical premises. \\
\rowcolor{lightblue!50}
\textbf{L16} & \textbf{0.21} & \texttt{\#6229} & 6.1 &
Abstract nouns and formal terminology associated with systemic processes, institutional mechanisms, and socio-economic concepts in academic or standardized testing contexts. \\
\arrayrulecolor{purple!30}\cmidrule(lr){1-5}
\rowcolor{lighpurple!50}
\multirow{1}{*}{\textbf{L21}} & \multirow{1}{*}{\textbf{0.36}}
    & \texttt{\#7862} & 18.1 &
Formal academic or examination-style text, often characteristic of Chinese-to-English translations in reading comprehension and logical reasoning contexts. \\
\rowcolor{lightblue!50}
\textbf{L32} & \textbf{0.66} & \colorbox{yellow!35}{\texttt{\#8910}} & 15.8 &
Assertions or evaluations regarding the truth, validity, or correctness of a specific statement, theory, or position. \\
\rowcolor{lighpurple!50}
\textbf{L41} & \textbf{0.67} & \colorbox{yellow!35}{\texttt{\#13986}} & 14.8 &
Academic and formal terminology related to cultural heritage, archaeological sites, and scientific or logical observations. \\
\arrayrulecolor{purple}\bottomrule
\end{tabular}
{\small\vspace{2pt}\par\noindent
\textit{\colorbox{yellow!35}{\phantom{X}} marks features whose description semantically overlaps with concepts explicitly verbalized in the CoT above.}}
\end{tcolorbox}

\end{table*}

\begin{table*}[t]
\centering
\small
\caption{Qualitative instance from \lm{Gemma-2-9B} on \data{OpenbookQA}: question, choices, CoT rationale, and prediction under both conditions, together with the shared concepts $S_\cap$ ablated in the causal intervention (\S\ref{subsec:causal}) and their feature descriptions and pre-ablation activation magnitudes.}
\label{tab:qual-gemma9b-openbookqa}
\renewcommand{\arraystretch}{1.2}
\arrayrulecolor{purple}
\begin{tcolorbox}[
    colback=lightblue, colframe=purple, boxrule=0.7pt, arc=3pt,
    left=8pt, right=8pt, top=5pt, bottom=6pt,
    title={\lm{Gemma-2-9B} \ \textcolor{purple!40!black}{$\vert$} \ \data{OpenbookQA}},
    fonttitle=\bfseries, colbacktitle=lighpurple, coltitle=black, boxsep=2pt
]
\textbf{Question.}
The middle of the day usually involves the bright star nearest to the earth to be straight overhead why?
\vspace{3pt}
\begin{itemize}[leftmargin=1.4em, itemsep=1.5pt, topsep=3pt]
    \item[\textbf{A:}] moons gravity
    \item[\textbf{B:}] human planet rotation  \goldicon
    \item[\textbf{C:}] global warming
    \item[\textbf{D:}] moon rotation
\end{itemize}
\vspace{4pt}
\textbf{Prediction Pass:} B~\okmark \quad
\textbf{CoT Pass:} B~\okmark \quad
\textbf{Gold:} B~\goldicon
\vspace{5pt}
\noindent\textcolor{purple!60!black}{\rule{\linewidth}{0.5pt}}
\vspace{4pt}
\textbf{CoT:}~\textit{``The sun is the closest star to the earth. The earth rotates on its axis. The earth rotates from west to east. The sun is in the east in the morning and in the west in the evening. The sun is straight overhead at noon. The sun is straight overhead at noon because the earth rotates from west to east...''}
\vspace{5pt}
\noindent\textcolor{purple!60!black}{\rule{\linewidth}{0.5pt}}
\vspace{4pt}
\centering
\begin{tabular}{@{}c c c c p{7.5cm}@{}}
\rowcolor{white}
\arrayrulecolor{purple!60!black}
\toprule
\textbf{Layer} & \textbf{$\Delta p$} & \textbf{Feature ID ($S_\cap$)} & \textbf{Act.} & \textbf{Feature Description} \\
\midrule
\rowcolor{lightblue!50}
\textbf{L8} & \textbf{0.76} & \colorbox{yellow!35}{\texttt{\#12308}} & 2.3 &
Terms describing spatial orientation, geographical directions, navigation, and physical positioning. \\
\arrayrulecolor{purple!30}\cmidrule(lr){1-5}
\rowcolor{lighpurple!50}
\textbf{L23} & \textbf{0.75} & \colorbox{yellow!35}{\texttt{\#11033}} & 12.7 &
Technical descriptions of physical motion, biological processes, and scientific concepts within educational or multiple-choice contexts. \\
\rowcolor{lightblue!50}
\textbf{L24} & \textbf{0.75} & \colorbox{yellow!35}{\texttt{\#15666}} & 11.3 &
Scientific descriptions of planetary rotation and biological traits, frequently appearing as specific options or answers within a multiple-choice format. \\
\rowcolor{lighpurple!50}
\textbf{L25} & \textbf{0.73} & \colorbox{yellow!35}{\texttt{\#4751}} & 12.2 &
Technical descriptions of physical rotation, directional movement, and specific nouns used as options in multiple-choice questions. \\
\rowcolor{lightblue!50}
\textbf{L26} & \textbf{0.46} & \colorbox{yellow!35}{\texttt{\#1718}} & 18.1 &
Technical terminology and specific answer choices within scientific explanations or multiple-choice questions, particularly regarding planetary rotation and physical properties. \\
\rowcolor{lighpurple!50}
\textbf{L29} & \textbf{0.76} & \colorbox{yellow!35}{\texttt{\#10712}} & 23.6 &
Scientific facts and descriptions concerning animal biology, behaviors, and astronomical or physical processes, frequently appearing in educational or multiple-choice contexts. \\
\rowcolor{lightblue!50}
\textbf{L31} & \textbf{0.76} & \colorbox{yellow!35}{\texttt{\#15964}} & 17.2 &
Factual descriptions and multiple-choice selections concerning biological characteristics, animal types, and physical or astronomical motions. \\
\rowcolor{lighpurple!50}
\textbf{L33} & \textbf{0.75} & \colorbox{yellow!35}{\texttt{\#5895}} & 19.9 &
Scientific and biological terminology describing animal species, ecological behaviors, and physical or astronomical processes. \\
\rowcolor{lightblue!50}
\textbf{L38} & \textbf{0.53} & \texttt{\#1529} & 20.1 &
Key terms and phrases that provide the specific factual or logical completion to a statement or question. \\
\rowcolor{lighpurple!50}
\textbf{L41} & \textbf{0.76} & \colorbox{yellow!35}{\texttt{\#1957}} & 17.4 &
Scientific descriptions and multiple-choice questions regarding celestial bodies, lunar cycles, and astronomical events. \\
\arrayrulecolor{purple}\bottomrule
\end{tabular}
{\small\vspace{2pt}\par\noindent
\textit{\colorbox{yellow!35}{\phantom{X}} marks features whose description semantically overlaps with concepts explicitly verbalized in the CoT above. }}
\end{tcolorbox}

\end{table*}
\begin{table*}[t]
\centering
\small
\caption{Qualitative instance from \lm{Gemma-2-9B} on \data{ARC-Easy}: question, choices, CoT rationale, and prediction under both conditions, together with the shared concepts $S_\cap$ ablated in the causal intervention (\S\ref{subsec:causal}) and their feature descriptions and pre-ablation activation magnitudes.}
\label{tab:qual-gemma9b-arc-easy}
\renewcommand{\arraystretch}{1.2}
\arrayrulecolor{purple}
\begin{tcolorbox}[
    colback=lightblue, colframe=purple, boxrule=0.7pt, arc=3pt,
    left=8pt, right=8pt, top=5pt, bottom=6pt,
    title={\lm{Gemma-2-9B} \ \textcolor{purple!40!black}{$\vert$} \ \data{ARC-Easy}},
    fonttitle=\bfseries, colbacktitle=lighpurple, coltitle=black, boxsep=2pt
]
\textbf{Question.}
A student has just completed a laboratory activity. What is the last action that the student should perform before leaving the lab area?
\vspace{3pt}
\begin{itemize}[leftmargin=1.4em, itemsep=1.5pt, topsep=3pt]
    \item[\textbf{A:}] wash hands \goldicon
    \item[\textbf{B:}] turn off all equipment
    \item[\textbf{C:}] put away all glassware
    \item[\textbf{D:}] wash instruments and table tops
\end{itemize}
\vspace{4pt}
\textbf{Prediction Pass.:} A~\okmark \quad
\textbf{CoT Pass.:} A~\okmark \quad
\textbf{Gold:} A~\goldicon
\vspace{5pt}
\noindent\textcolor{purple!60!black}{\rule{\linewidth}{0.5pt}}
\vspace{4pt}
\textbf{CoT.}~\textit{``The first thing that the student should do is to wash hands. The second thing that the student should do is to turn off all equipment. The third thing that the student should do is to put away all glassware. The fourth thing that the student should do is to wash instruments and table tops. Therefore, the correct answer is (A) wash hands.''}
\vspace{5pt}
\noindent\textcolor{purple!60!black}{\rule{\linewidth}{0.5pt}}
\vspace{4pt}
\centering
\begin{tabular}{@{}c c c c p{7.5cm}@{}}
\rowcolor{white}
\arrayrulecolor{purple!60!black}
\toprule
\textbf{Layer} & \textbf{$\Delta p$} & \textbf{Feature ID ($S_\cap$)} & \textbf{Act.} & \textbf{Feature Description} \\
\midrule
\rowcolor{lightblue!50}
\textbf{L1} & \textbf{0.21} & \colorbox{yellow!35}{\texttt{\#14397}} & 2.2 &
Nouns identifying the participants, safety protocols, and formal designations of scientific activities within an educational laboratory or classroom setting. \\
\rowcolor{lighpurple!50}
 \textbf{L11} & \textbf{0.29} & \colorbox{yellow!35}{\texttt{\#3433}} & 3.3 & Scientific terminology and formal phrasing used in educational assessments, laboratory observations, and multiple-choice questions. \\
\rowcolor{lightblue!50}
\textbf{L15} & \textbf{0.15} & \colorbox{yellow!35}{\texttt{\#8576}} & 3.9 &
Structural elements of multiple-choice science questions, including question framing, student-led scenarios, and the identification of correct answers. \\
\rowcolor{lighpurple!50}
 \textbf{L19} & \textbf{0.29} & \colorbox{yellow!35}{\texttt{\#12737}} & 5.3 &  Technical terminology and vocabulary used in scientific educational materials, standardized test questions, and laboratory contexts.\\
\rowcolor{lightblue!50}
\textbf{L22} & \textbf{0.28} & \colorbox{yellow!35}{\texttt{\#7909}} & 14.7 & Nouns and descriptive phrases identifying physical features, biological structures, or scientific equipment in academic contexts. \\
\rowcolor{lighpurple!50}
 \textbf{L32} & \textbf{0.28} & \colorbox{yellow!35}{\texttt{\#8551}} & 13.3 &  Procedural steps, safety equipment, and experimental methodologies found in scientific laboratory instructions and multiple-choice science assessment questions.\\
\rowcolor{lightblue!50}
\textbf{L38} & \textbf{0.18} & \colorbox{yellow!35}{\texttt{\#6835}} & 12.8 & Key scientific terms and concepts serving as correct answers or critical components within educational multiple-choice questions and explanations. \\
\rowcolor{lighpurple!50}
 \textbf{L40} & \textbf{0.29} & \texttt{\#15047} & 39.5 &  Symbols and notation characters used in scientific formulas, temperature units, and structured multiple-choice options. \\
\arrayrulecolor{purple}\bottomrule
\end{tabular}
{\small\vspace{2pt}\par\noindent
\textit{\colorbox{yellow!35}{\phantom{X}} marks features whose description semantically overlaps with concepts explicitly verbalized in the CoT above.}}
\end{tcolorbox}

\end{table*}
\begin{table*}[t]
\centering
\small
\caption{Qualitative instance from \lm{Gemma-2-9B} on \data{GSM8K}: question, CoT rationale, and prediction under both conditions, together with the shared concepts $S_\cap$ ablated in the causal intervention (\S\ref{subsec:causal}) and their feature descriptions and pre-ablation activation magnitudes.}
\label{tab:qual-gemma9b-gsm8k}
\renewcommand{\arraystretch}{1.2}
\arrayrulecolor{purple}
\begin{tcolorbox}[
    colback=lightblue, colframe=purple, boxrule=0.7pt, arc=3pt,
    left=8pt, right=8pt, top=5pt, bottom=6pt,
    title={\lm{Gemma-2-9B} \ \textcolor{purple!40!black}{$\vert$} \ \data{GSM8K}},
    fonttitle=\bfseries, colbacktitle=lighpurple, coltitle=black, boxsep=2pt
]
\textbf{Question.}
In a dance class of 20 students, 20\% enrolled in contemporary dance, 25\% of the remaining enrolled in jazz dance, and the rest enrolled in hip-hop dance. What percentage of the entire students enrolled in hip-hop dance?
\vspace{3pt}
\vspace{4pt}
\textbf{Prediction Pass.:} 60~\okmark \quad
\textbf{CoT Pass.:} 60~\okmark \quad
\textbf{Gold:} 60~\goldicon
\vspace{5pt}
\noindent\textcolor{purple!60!black}{\rule{\linewidth}{0.5pt}}
\vspace{4pt}
\textbf{CoT:}~\textit{``First, we need to find out how many students enrolled in contemporary dance. 20\% of 20 students = 0.2 * 20 = 4 students. Now, we need to find out how many students are left after 4 students enrolled in contemporary dance. 20 - 4 = 16 students. Next, we need to find out how many students enrolled in jazz dance. 25\% of 16 students = 0.25 * 16 = 4 students. Now, we need to find out how many students are left after 4 students enrolled in jazz dance. 16 - 4 = 12 students. Finally, we need to find out what percentage of the entire students enrolled in hip-hop dance. 12 students out of 20 students = 12/20 = 0.6 = 60\%. Therefore, 60\% of the entire students enrolled in hip-hop dance.''}
\vspace{5pt}
\noindent\textcolor{purple!60!black}{\rule{\linewidth}{0.5pt}}
\vspace{4pt}
\centering
\begin{tabular}{@{}c c c c p{7.5cm}@{}}
\rowcolor{white}
\arrayrulecolor{purple!60!black}
\toprule
\textbf{Layer} & \textbf{$\Delta p$} & \textbf{Feature ID ($S_\cap$)} & \textbf{Act.} & \textbf{Feature Description} \\
\midrule
\rowcolor{lightblue!50}
\textbf{L8} & \textbf{0.00} & \colorbox{yellow!35}{\texttt{\#10482}} & 3.0 &
Mathematical terminology and symbols used to define proportions, percentages, and quantitative relationships within word problems. \\
\rowcolor{lighpurple!50}
 \textbf{L14} & \textbf{0.03} & \colorbox{yellow!35}{\texttt{\#6364}} & 9.8 & Mathematical terminology and symbols used to express probability, percentages, and statistical likelihood within word problems. \\
\rowcolor{lightblue!50}
\textbf{L19} & \textbf{0.01} & \colorbox{yellow!35}{\texttt{\#760}} & 23.2 &
Numerical values and percentages representing proportions, ratios, or fractional parts within mathematical contexts. \\
\rowcolor{lighpurple!50}
 \textbf{L25} & \textbf{0.07} & \colorbox{yellow!35}{\texttt{\#2610}} & 19.5 & Mathematical contexts involving the calculation, phrasing, or conversion of percentages and proportions. \\
\rowcolor{lightblue!50}
\textbf{L28} & \textbf{0.06} & \colorbox{yellow!35}{\texttt{\#10761}} & 24.1 &
Phrases following a fractional or numerical value that specify a portion of a remaining quantity in mathematical word problems. \\
\rowcolor{lighpurple!50}
 \textbf{L31} & \textbf{0.07} & \colorbox{yellow!35}{\texttt{\#16056}} & 40.5 & Transitions and qualifiers used in mathematical word problems to introduce the final, remaining, or resulting quantity in a sequence of calculations. \\
\rowcolor{lightblue!50}
\textbf{L33} & \textbf{0.06} & \colorbox{yellow!35}{\texttt{\#7582}} & 19.9 &
Clauses in mathematical word problems that introduce specific constraints or define the final question to be solved. \\
\rowcolor{lighpurple!50}
 \textbf{L34} & \textbf{0.07} & \colorbox{yellow!35}{\texttt{\#3298}} & 55.5 & The transition point and leading digit of a numerical value representing a secondary quantity or proportion in a mathematical word problem. \\
\arrayrulecolor{purple}\bottomrule
\end{tabular}
{\small\vspace{2pt}\par\noindent
\textit{\colorbox{yellow!35}{\phantom{X}} marks features whose description semantically overlaps with concepts explicitly verbalized in the CoT above.}}
\end{tcolorbox}

\end{table*}

\end{document}